\documentclass[lettersize,journal]{IEEEtran}
\usepackage{amsmath,amsfonts}
\usepackage{algorithmic}
\usepackage{algorithm}
\usepackage{array}
\usepackage{textcomp}
\usepackage{stfloats}
\usepackage{url}
\usepackage{verbatim}
\usepackage{graphicx}
\usepackage{cite}

\usepackage{caption}
\usepackage{hyperref}
\usepackage{alphabeta}
\usepackage{multirow}
\usepackage{booktabs}
\usepackage{subfigure}
\usepackage{lipsum}
\usepackage{amssymb}
\usepackage{epsfig}
\usepackage{rotating}

\begin{document}

\title{Masked Discriminators for Content-Consistent Unpaired Image-to-Image Translation}

\author{Bonifaz Stuhr, Jürgen Brauer, Bernhard Schick, and Jordi Gonzàlez
	
\thanks{This work was supported in part by the Spanish Ministry of Economy and Competitiveness (MINECO) and the European
	Regional Development Fund (ERDF): Grant PID2020-120311RB-I00 funded by MCIN/AEI/ 10.13039/501100011033. \textit{(Corresponding author: Bonifaz Stuhr.)}} 
	
\thanks{Bonifaz Stuhr and Jordi Gonzàlez are with the Department of Computer Science, Autonomous University of Barcelona, 08193 Bellaterra, Barcelona, Spain (e-mail: bonifaz.stuhr@hs-kempten.de; jordi.gonzalez@uab.cat).
	
	Jürgen Brauer is with the Department of Computer Science, University of Applied Sciences Kempten, 87435 Kempten, Germany (e-mail: juergen.brauer@hs-kempten.de).
	
	Bernhard Schick and Bonifaz Stuhr are with the IFM, Junkersstraße 1A, 87734 Benningen, Germany (e-mail: bernhard.schick@hs-kempten.de).}%
	
	\vspace{+4.5pt}
	\begin{minipage}{1\textwidth}
	\begin{minipage}{0.5\textwidth}
		\vspace*{+0.75pt}
		\includegraphics[width=0.5\textwidth]{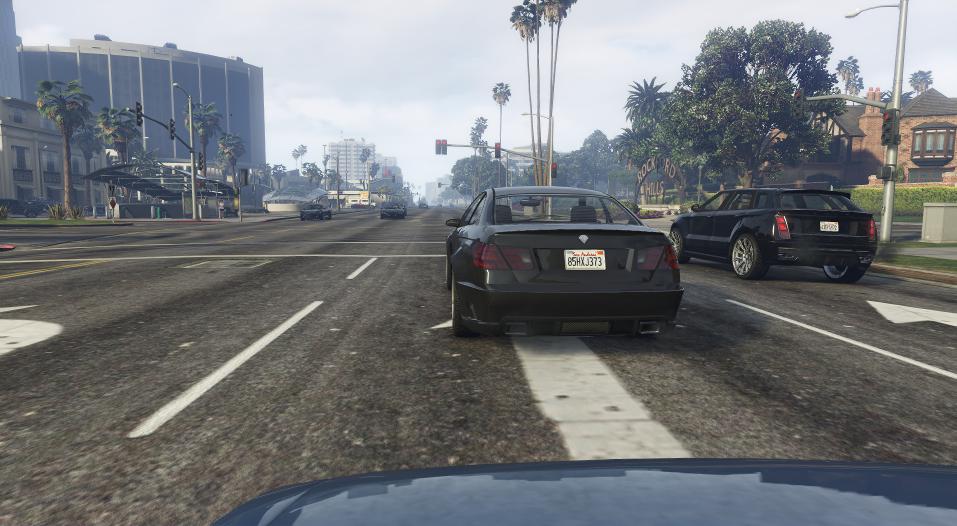}\includegraphics[width=0.5\textwidth]{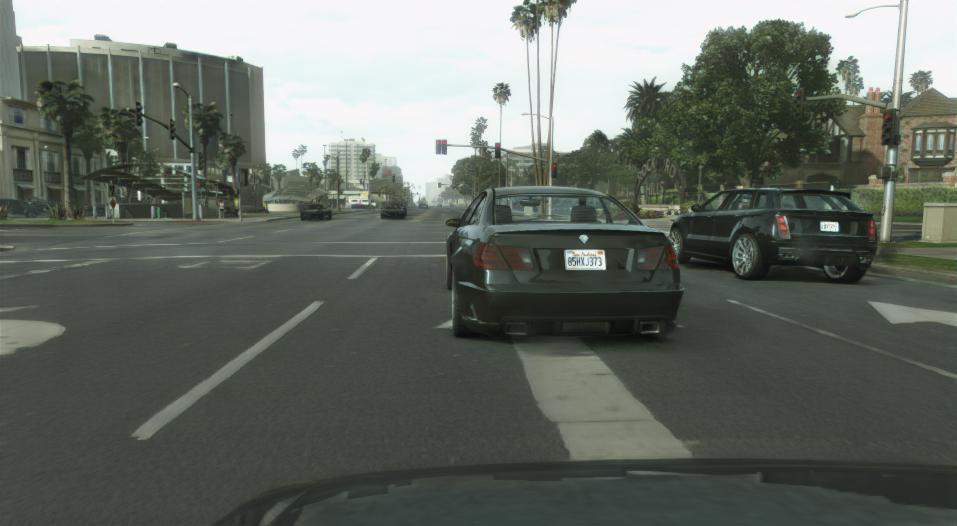}
		\vspace*{-16pt}
		\captionof*{figure}{PFD$\rightarrow$Cityscapes}
		\vspace*{+4.75pt}
		\includegraphics[width=0.5\textwidth]{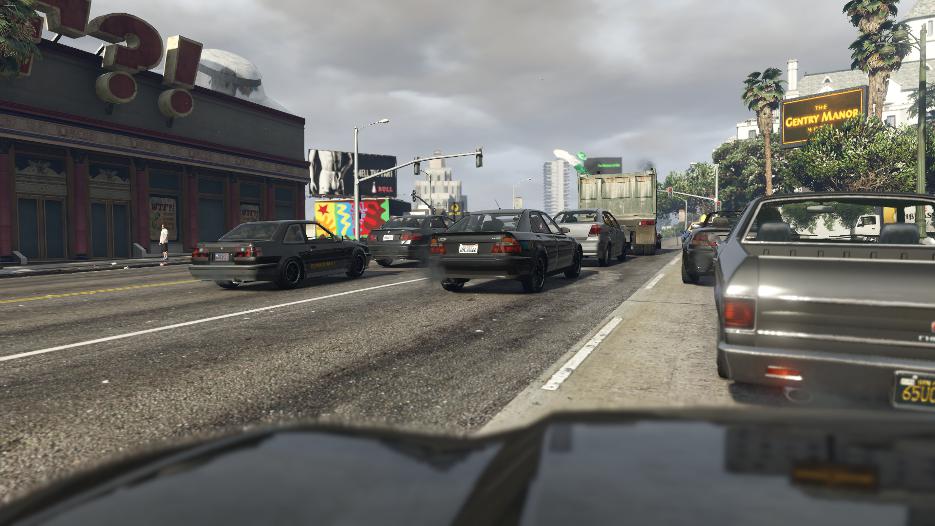}\includegraphics[width=0.5\textwidth]{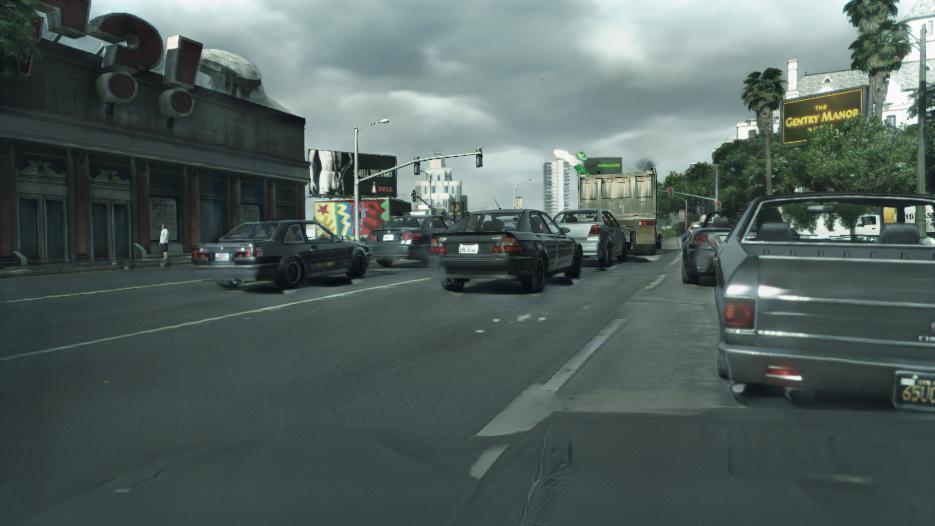}
		\vspace*{-16pt}
		\captionof*{figure}{Viper$\rightarrow$Cityscapes}
	\end{minipage}
	\begin{minipage}{0.5\textwidth}
		\includegraphics[width=0.5\textwidth]{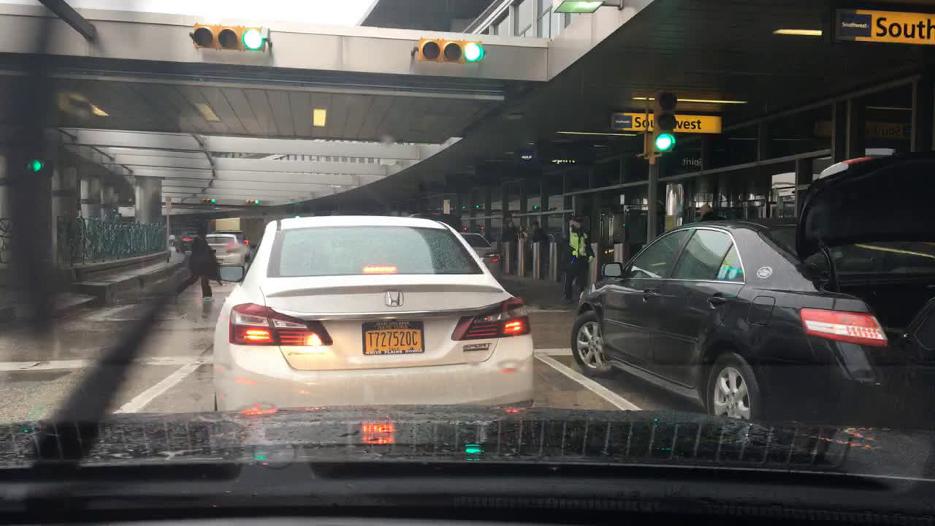}\includegraphics[width=0.5\textwidth]{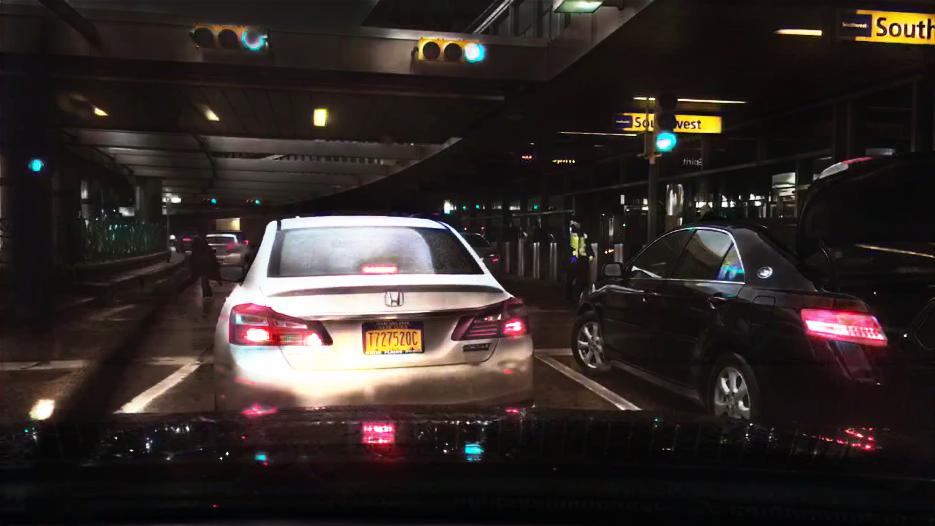}
		\vspace*{-16pt}
		\captionof*{figure}{Day$\rightarrow$Night}
		\vspace*{+4pt}	
		\includegraphics[width=0.5\textwidth]{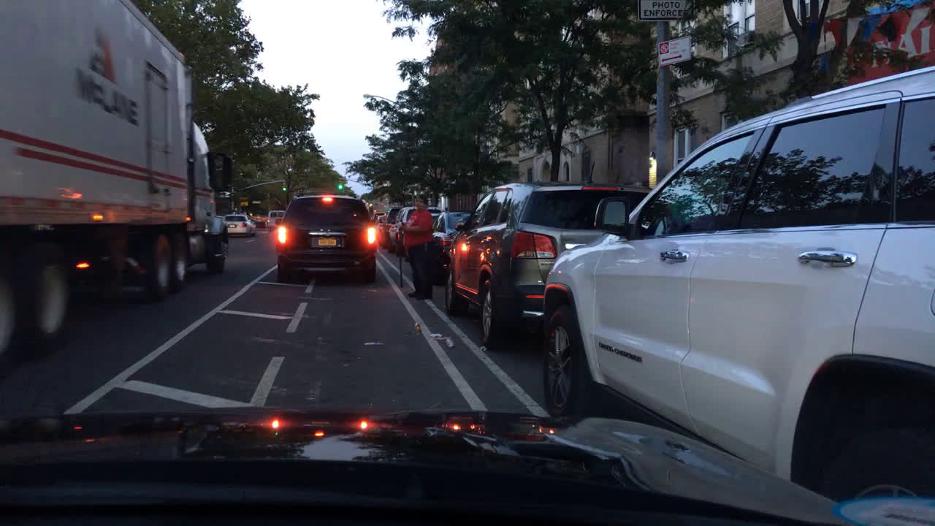}\includegraphics[width=0.5\textwidth]{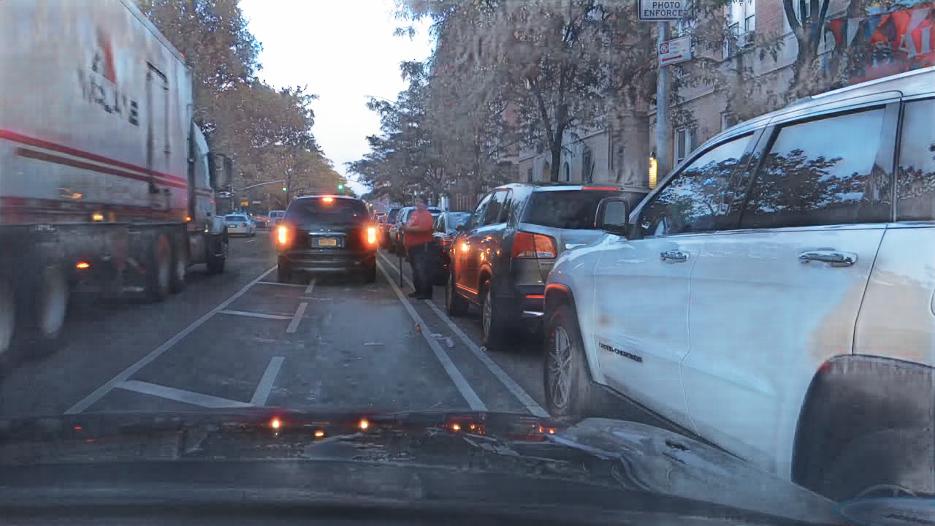}
		\vspace*{-16pt}
		\captionof*{figure}{Clear$\rightarrow$Snowy}
	\end{minipage}
	\vspace{-9pt}
	\setcounter{figure}{0}
	\captionof{figure}{Results of our method.}
	\label{fig:results}\end{minipage}\vspace{-31pt}}


\IEEEpubid{}

\maketitle

\begin{abstract}
A common goal of unpaired image-to-image translation is to preserve content consistency between source images and translated images while mimicking the style of the target domain. Due to biases between the datasets of both domains, many methods suffer from inconsistencies caused by the translation process. Most approaches introduced to mitigate these inconsistencies do not constrain the discriminator, leading to an even more ill-posed training setup. Moreover, none of these approaches is designed for larger crop sizes. In this work, we show that masking the inputs of a global discriminator for both domains with a content-based mask is sufficient to reduce content inconsistencies significantly. However, this strategy leads to artifacts that can be traced back to the masking process. To reduce these artifacts, we introduce a local discriminator that operates on pairs of small crops selected with a similarity sampling strategy. Furthermore, we apply this sampling strategy to sample global input crops from the source and target dataset. In addition, we propose feature-attentive denormalization to selectively incorporate content-based statistics into the generator stream. In our experiments, we show that our method achieves state-of-the-art performance in photorealistic sim-to-real translation and weather translation and also performs well in day-to-night translation. Additionally, we propose the cKVD metric, which builds on the sKVD metric and enables the examination of translation quality at the class or category level. 
\end{abstract}

	
\section{Introduction}
\vspace{+5pt}
\IEEEPARstart{U}{npaired} image-to-image translation aims at transferring images from a source domain to a target domain when no paired examples are given. Recently, this field has attracted increasing interest and has advanced several use cases, such as photorealism \cite{pizzati2021comogan, richter2022enhancing, jia2021semantically}, neural rendering \cite{hao2021gancraft}, domain adaptation \cite{hoffman2018cycada, roy2021trigan}, the translation of seasons or daytime \cite{jiang2020tsit, jeong2021memory, pizzati2021comogan}, and artistic style transfer \cite{huang2017arbitrary, yao2019attention, kim2020u}. Current work has primarily focused on improving translation quality \cite{nederhood2021harnessing,jeong2021memory}, efficiency \cite{liang2021high, shaham2021spatially}, multi-modality \cite{huang2018multimodal,lin2020multimodal}, and content consistency \cite{richter2022enhancing, jia2021semantically}. Due to the ill-posed nature of the unpaired image-to-image translation task and biases between datasets, content consistency is difficult to achieve. To mitigate content inconsistencies, several methods have been proposed that constrain the generator of GANs \cite{zhu2017unpaired,benaim2017one,fu2019geometry,lin2020multimodal,zhang2019harmonic,zhao2020unpaired,liu2017unsupervised,huang2018multimodal,sendik2020crossnet,yang2020phase}. However, only constraining the generator leads to an unfair setup, as biases in the datasets can be detected by the discriminator: The generator tries to achieve contnent consistency by avoiding biases in the output, while the discriminator is still able to detect biases between both datasets and, therefore, forces the generator to include these biases in the output, for example, through hallucinations. Constraining the discriminator \cite{liang2018generative,richter2022enhancing, theiss2022unpaired} or improving the sampling of training pairs \cite{kao2019patch,richter2022enhancing} is currently underexplored, especially for content consistency on a global level, where the discriminator has a global view on larger image crops instead of a local view on small crops. In this work, we propose \textit{masked conditional discriminators}, which operate on masked global crops of the inputs to mitigate content inconsistencies. We combine these discriminators with an efficient sampling strategy based on a pre-trained robust segmentation model to sample similar global crops. Furthermore, we argue that when transferring feature statistics from the content stream of the source image to the \hphantom{g} \clearpage \noindent generator stream, content-unrelated feature statistics from the content stream could affect image quality if the generator is unable to ignore this information since the output image should mimic the target domain. Therefore, we propose a \textit{feature-attentive denormalization (FATE)} block that extends feature-adaptive denormalization (FADE) \cite{jiang2020tsit} with an attention mechanism. This block allows the generator to selectively incorporate statistical features from the content stream into the generator stream. In our experiments, we find that our method achieves state-of-the-art performance on most of the benchmarks shown in \autoref{fig:results}.
\\\\
Our contributions can be summarized as follows:
\begin{itemize}
	\item We propose an efficient sampling strategy that utilizes robust semantic segmentations to sample similar global crops. This reduces biases between both datasets induced by semantic class misalignment. 
	\item We combine this strategy with masked conditional discriminators to achieve content consistency while maintaining a more global field of view.
	\item We extend our method with an unmasked local discriminator. This discriminator operates on local, partially class-aligned patches to minimize the underrepresentation of frequently masked classes and associated artifacts.	
	\item We propose a feature-attentive denormalization (FATE) block, which selectively fuses statistical features from the content stream into the generator stream.
	\item We propose the class-specific Kernel VGG Distance (cKVD) that builds upon the semantically aligned Kernel VGG Distance (sKVD) \cite{richter2022enhancing} and uses robust segmentations to incorporate class-specific content inconsistencies in the perceptual image quality measurement.
	\item In our experiments, we show that our method achieves state-of-the-art performance on photo-realistic sim-to-real transfer and the translation of weather and performs well for daytime translation.
\end{itemize}

\section{Related Work}
\noindent \textbf{Unpaired image-to-image translation.} Following the success of GANs \cite{goodfellow2014generative}, the conditional GAN framework \cite{mirza2014conditional} enables image generation based on an input condition. Pix2Pix \cite{isola2017image} uses images from a source domain as a condition for the generator and discriminator to translate them to a target domain. Since Pix2Pix relies on a regression loss between generated and target images, translation can only be performed between domains where paired images are available. To achieve unpaired image-to-image translation, methods like CycleGAN \cite{zhu2017unpaired}, UNIT \cite{liu2017unsupervised}, and MUNIT \cite{huang2018multimodal} utilize a second GAN to perform the translation in the opposite direction and impose a cycle-consistency constraint or weight-sharing constraint between both GANs. However, these methods require additional parameters for the second GAN, which are used to learn the unpaired translation and are omitted when inferring a one-sided translation. In works such as TSIT \cite{jiang2020tsit} and CUT \cite{park2020contrastive}, these additional parameters are completely omitted at training time by either utilizing a perceptual loss \cite{johnson2016perceptual} between the input image of the generator and the image to be translated or by patchwise contrastive learning. Recently, additional techniques have achieved promising results, like pseudo-labeling \cite{hao2021gancraft} or a conditional discriminator based on segmentations created with a robust segmentation model for both domains \cite{richter2022enhancing}. Furthermore, there are recent efforts to adapt diffusion models to unpaired image-to-image translation \cite{su2022dual,zhao2022egsde,wu2022unifying}.
\\\\
\noindent \textbf{Content consistency in unpaired image-to-image translation.} Due to biases between unpaired datasets, the content of translated samples can not be trivially preserved \cite{richter2022enhancing}. There are ongoing efforts to preserve the content of an image when it is translated to another domain by improving various parts of the training pipeline: Several consistency constraints have been proposed for the generator, which operate directly on the translated image \cite{zhu2017unpaired, benaim2017one, lin2020multimodal}, on a transformation of the translated image \cite{taigman2016unsupervised, zhang2019harmonic, fu2019geometry, yang2020phase, wang2020classes}, or on distributions of multi-modal translated images \cite{zhao2020unpaired}. The use of a perceptual loss \cite{johnson2016perceptual} or LPIPS loss \cite{zhang2018unreasonable} between input images and translated images, as in \cite{jiang2020tsit} and \cite{richter2022enhancing}, can also be considered a consistency constraint between transformed images. In \cite{xie2020self} content consistency is enforced with self-supervised in-domain and cross-domain patch position prediction. There are works that enforce consistency by constraining the latent space of the generator \cite{liu2017unsupervised, huang2018multimodal, sendik2020crossnet}. Semantic scene inconsistencies can be mitigated with a separate segmentation model \cite{yang2020phase, lin2020multimodal}. To avoid inconsistency arising from style transfer, features from the generator stream are masked before AdaIN \cite{huang2017arbitrary,ma2018exemplar}. Another work exploits small perturbations in the input feature space to improve semantic robustness \cite{jia2021semantically}. However, if the datasets of both domains are unbalanced, discriminators can use dataset biases as learning shortcuts, which leads to content inconsistencies. Therefore, only constraining the generator for content consistency still results in an ill-posed unpaired image-to-image translation setup. Constraining discriminators to achieve content consistency is currently underexplored, but recent work has proposed promising directions. There are semantic-aware discriminator architectures \cite{liang2018generative, liu2019learning, richter2022enhancing, hao2021gancraft} that enforce discriminators to base their predictions on semantic classes, or VGG discriminators \cite{richter2022enhancing}, which additionally operate on abstract features of a frozen VGG model instead of the input images. Training discriminators with small patches \cite{richter2022enhancing} is another way to improve content consistency. To mitigate dataset biases during training for the whole model, sampling strategies can be applied to sample similar patches from both domains \cite{kao2019patch, richter2022enhancing}. Furthermore, in \cite{theiss2022unpaired}, a model is trained to generate a hyper-vector mapping between source and target images with an adversarial loss and a cyclic loss for content consistency. In contrast, our work utilizes a robust semantic mask to mask global discriminators with a large field of view, which provide the generator with the gradients of the unmasked regions. This leads to a content-consistent translation while preserving the global context. We combine this discriminator with an efficient sampling method that uses robust semantic segmentations to sample similar crops from both domains.
\\\\
\noindent \textbf{Attention in image-to-image translation.}
Previous work has utilized attention for different parts of the GAN framework. A common technique is to create attention mechanisms that allow the generator or discriminator to focus on important regions of the input \cite{alami2018unsupervised, tang2021attentiongan, yang2019show, kim2020u,zhang2022region} or to capture the relationship between regions of the input(s) \cite{yao2019attention, tang2020dual, hu2022qs}. Other works guide a pixel loss with uncertainty maps computed from attention maps \cite{tang2019multi}, exploit correlations between channel maps with scale-wise channel attention \cite{tang2020dual}, disentangle content and style with diagonal attention \cite{kwon2021diagonal}, or merge features from multiple sources with an attentional block before integrating them into the generator stream \cite{liu2021liquid}. In \cite{lin2021attention}, an attention-based discriminator is introduced to guide the training of the generator with attention maps. Furthermore, ViTs \cite{dosovitskiy2020image} are adapted for unpaired image-to-image translation \cite{torbunov2023uvcgan, zheng2022ittr}, and the computational complexity of their self-attention mechanism is reduced for high-resolution translation \cite{zheng2022ittr}. In contrast, our work proposes an attention mechanism to selectively integrate statistics from the content stream of the source image into the generator stream. This allows the model to focus on statistical features from the content stream that are useful for the target domain.

\section{Method}
\begin{figure*}[t]
	\centering 
	\includegraphics[width=1.0\linewidth]{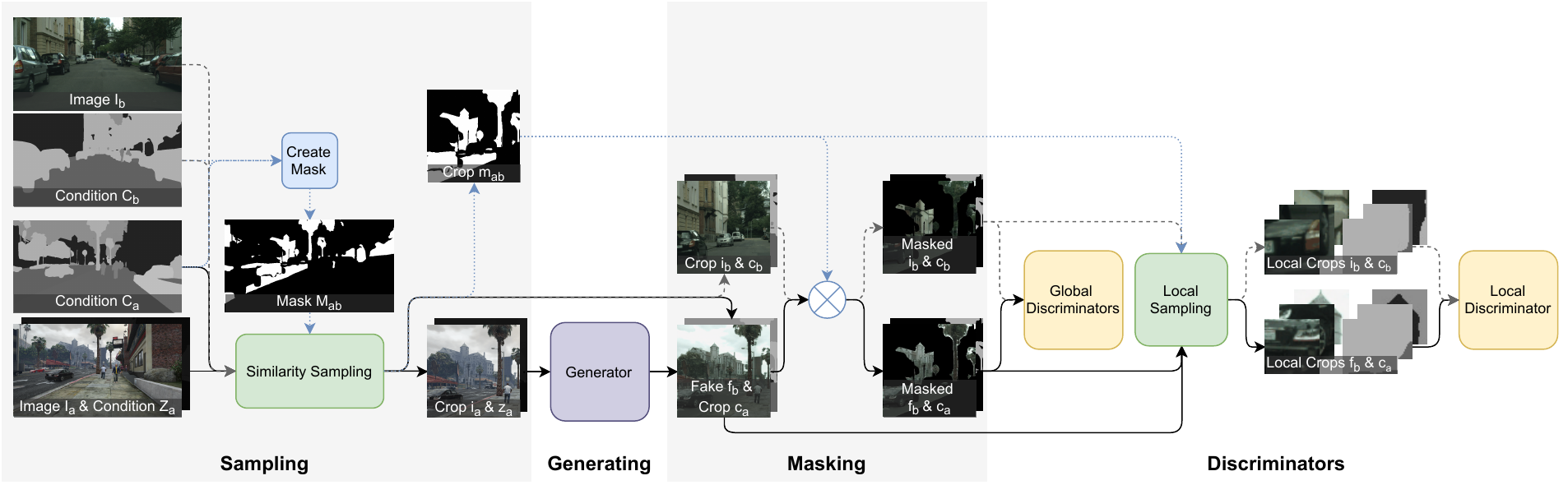}
	\caption{\textbf{Method overview.} In our method, similar image crops from both domains ($i_a$, $i_b$) and their corresponding conditions ($c_a$, $c_b$, $z_a$) are selected via a sampling procedure. In this sampling procedure, a mask $M_{ab}$ is created from the conditions $C_a$ and $C_b$. This mask is used to sample crops from both datasets for which the semantic classes align by at least 50\%. The cropped mask $m_{ab}$ is also used to mask the generated fake image $f_b$, the real images $i_b$, and the corresponding conditions for the global conditional discriminators. Through the mask, these discriminators can only see the parts of the crop where the semantic classes align. To further improve image quality, a local discriminator is introduced that works on a batch of small patches selected from the crop using our sampling technique. This discriminator is not masked and works on patches where the semantic classes do not fully align.}
	\label{fig:MethodOverview}
\end{figure*} 
We propose an end-to-end framework for unpaired image-to-image translation that transfers an image $I_a \in \mathbb{R}^{3\times h\times w}$ from a source domain $a$ to an image $F_b \in \mathbb{R}^{3\times h\times w}$ from a target domain $b$. Our goal is to design a method for content-consistent translations that utilizes a simple masking strategy for the global crops seen by the discriminators. We achieve this by combining an efficient segmentation-based sampling method that samples large crops from the input image with a masked discriminator that operates on these global crops. This is in contrast to EPE \cite{richter2022enhancing}, which achieves content-consistent translation at the local level by sampling small, similar image crops from both domains. To further improve image quality, we use a local discriminator that operates on a batch of small image patches sampled from the global input crops utilizing our sampling method. An overview of our method is shown in \autoref{fig:MethodOverview}.
Furthermore, we propose a feature-attentive denormalization (FATE) block that extends feature-adaptive denormalization (FADE) \cite{jiang2020tsit} with an attention mechanism, allowing the generator to selectively incorporate statistical features from the content stream of the source image into the generator stream.

\subsection{Contend-based Similarity Sampling}
\label{sec:similaritySampling}
To minimize the bias between both datasets in the early stage of our method, we sample similar image crops with an efficient sampling procedure. This procedure uses the one-hot encoded semantic segmentations $C_a \in \mathbb{R}^{d\times h\times w}$ and $C_b \in \mathbb{R}^{d\times h\times w}$ of both domains, where $d$ is the channel dimension of the one-hot encoding. In our case, these segmentations are created with the robust pre-trained MSeg model \cite{lambert2020mseg}. First, a mask $M_{ab} \in \mathbb{R}^{1\times h\times w}$ is computed from the segmentations:
\begin{equation}
	\label{eqn:mask}
	M_{ab} = \operatorname*{max}_d (C_a\circ C_b),
\end{equation}
where $\circ$ denotes the Hadamard product. 
We can now sample semantically aligned image crops $i_a$ and $i_b$ from the images $I_a$ and $I_b$ with the crop $m_{ab}$ from mask $M_{ab}$. Thereby, we calculate the percentage of overlap of semantic classes between both image crops as follows:
\begin{equation}
	\label{eqn:similaritysampling}
	\mathcal{P}_{match}(i_a)=\{i_b\mid\operatorname*{mean}(m_{ab}) > t\} ,
\end{equation}
where $t$ is the similarity sampling threshold. In our case, we sample crops where more than $50\%$ of the semantic classes align ($t > 0.5$). We use this procedure to sample crops $c_a$, $c_b$, and $z_b$ from the discriminator conditions $C_a$, $C_b$, and the generator condition $Z_b$ as well. The cropped mask $m_{ab}$ is also used for our masked conditional discriminator.

\subsection{Contend-based Discriminator Masking}
To train a discriminator with a global field of view that facilitates the usage of global properties of the scene, while simultaneously maintaining content consistency, we mask the discriminator input from both domains with a content-based mask $m_{ab}$. This mask erases all pixels from the discriminator input where the semantic classes do not align. This removes the bias between both datasets caused by the underlying semantic class distribution of the two domains without directly restricting the generator. The objective function of a conditional GAN with a masked discriminator that transfers image crops $i_a$ to domain $b$ can be then defined as follows:
\begin{equation}
	\label{eqn:masked_adv_obj}
	\begin{split}
		&\mathcal{L}_{madv}= ~\mathbb{E}_{i_b,c_b,m_{ab}}[\log D(i_b\circ m_{ab}| c_b\circ m_{ab})] \\ &+ \mathbb{E}_{i_a,z_a,c_a, m_{ab}}[\log (1 - D(G(i_a| z_a)\circ m_{ab}| c_a\circ m_{ab}))].
	\end{split}
\end{equation} 
To ensure that the discriminator does not use the segmentation maps as learning shortcuts, we follow \cite{richter2022enhancing} and create the segmentations of both datasets using a robust segmentation model such as MSeg \cite{lambert2020mseg}. With this setting, we are able to train discriminators with large crop sizes with significantly reduced hallucinations in the translated image.

\subsection{Local Discriminator}
Masking the input of the discriminator may lead to the underrepresentation of some semantic classes. Therefore, we additionally train a local discriminator that operates on a batch of small patches sampled from the global crop. Our local discriminator is not masked but only sees patches where a certain amount of the semantic classes align. In our case, we sample patches with 1/8th the size of the global input crop where more than $50\%$ of the semantic classes align. We use our sampling procedure from Section \ref{sec:similaritySampling} to sample these patches. Using small, partially aligned patches ensures that semantic classes are less underrepresented while maintaining content consistency. 

\subsection{Feature-attentive Denormalization (FATE)}
\begin{figure}[h]
	\centering 
	\includegraphics[width=1.0\linewidth]{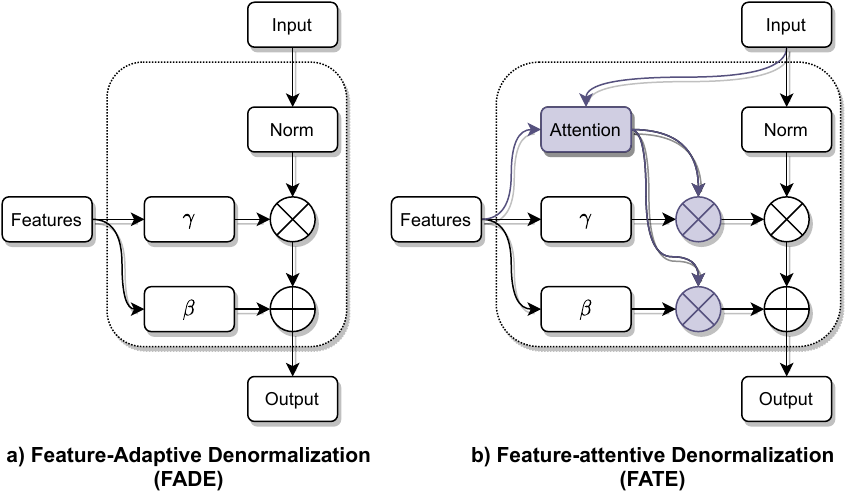}
	\caption{\textbf{FADE and FATE.}}
	\label{fig:FATE}
\end{figure} 
Spatially adaptive denormalization (SPADE) \cite{park2019semantic} fuses resized semantic segmentation maps as content into the generator stream. Feature-adaptive denormalization (FADE) \cite{jiang2020tsit} generalizes SPADE to features learned through a content stream. As shown in \autoref{fig:FATE}, the normalized features $N(h)$ of the generator are modulated with the features $f$ of the content stream using the learned functions $\gamma$ and $\beta$ as follows:
\begin{equation}
	\label{eqn:FADE}
	\operatorname*{FADE}(h,f) = N(h)\circ\gamma(f) + \beta(f) ,
\end{equation}
where $\gamma$ and $\beta$ are one-layer convolutions. This denormalization is applied in several layers of the generator. However, we argue that denormalization with content features is not always appropriate for transferring images to another domain because, as shown in \cite{gatys2016image,li2016combining,li2017demystifying,huang2017arbitrary}, image feature statistics contain not only content information but also style information. When transferring feature statistics from the content stream of the source image to the generator stream, style information from the source image could affect the final image quality if the generator cannot ignore this information since the output image should mimic the style of the target domain. Therefore, we propose an additional attention mechanism to selectively incorporate statistics from the content stream into the generator stream. This allows the model to only fuse the statistical features from the source image into the generator stream that are useful for the target domain. As shown in \autoref{fig:FATE}, this attention mechanism relies on the features of the content stream and the features of the generator stream and attends to the statistics $\gamma$ and $\beta$. With this attention mechanism, we can extend FADE to feature-attentive denormalization (FATE) as follows:
\begin{equation}
	\label{eqn:FATE}
	\operatorname*{FATE}(h,f) = N(h)\circ A(h,f)\circ\gamma(f)+ A(h,f)\circ \beta(f),
\end{equation}
where $A$ is the attention mechanism and $A(h,f)$ is the attention map for the statistics. We use a lightweight two-layer CNN with sigmoid activation in the last layer as the attention mechanism. More details can be found in Appendix \ref{appendix}.

\subsection{Training Objective}
Our training objective consists of three losses: a global masked adversarial loss $\mathcal{L}_{madv}^{global}$, a local adversarial loss $\mathcal{L}_{adv}^{local}$, and the perceptual loss $\mathcal{L}_{perc}$ used in \cite{jiang2020tsit}. We define the final training objective as follows:
\begin{equation}
	\label{eqn:overall_obj_gen}
	\mathcal{L}=\lambda_{madv}^{global}\mathcal{L}_{madv}^{global} + \lambda_{adv}^{local}\mathcal{L}_{adv}^{local} + \lambda_{perc}\mathcal{L}_{perc},
\end{equation}
where we use a hinge loss to formulate the adversarial losses and $\lambda_{madv}^{global}$, $\lambda_{madv}^{local}$, $\lambda_{perc}$ are the corresponding loss weights.

\section{Experiments}
\subsection{Experimental Settings}
\noindent \textbf{Implementation details.} 
Our method is implemented in PyTorch 1.10.0 and trained on an A100 GPU (40 GB) with batch size $1$. For training, we initialize all weights with the Xavier normal distribution \cite{glorot2010understanding} with a gain of $0.02$ and use an Adam optimizer \cite{kingma2014adam} with $\beta_1=0.9$ and $\beta_2=0.999$. The initial learning rates of the generator and discriminators are set to $0.0001$ and halved every $d_e$ epochs. Learning rate decay is stopped after reaching a learning rate of $0.0000125$. We formulate our adversarial objective with a hinge loss \cite{lim2017geometric} and weight the individual parts of our loss function as follows: $\lambda_{madv}^{global}=1.0, \lambda_{madv}^{local}=1.0$, $\lambda_{perc}=1.0$. In addition, we use a gradient penalty on target images \cite{gulrajani2017improved, mescheder2018training} with $\lambda_{rp}=0.03$. The images of both domains are resized and cropped to the same size and randomly flipped before the sampling strategy is applied. In our experiments, we show that we achieve the best performance by cropping global patches of size 352$\times$352. We crop local patches with 1/8th the size of the global crop (i.a., 44$\times$44). The global discriminators are used on two scales. Crops are scaled down by a factor of two for the second scale. We train all our models for $\sim\!400$K iterations. Training a model takes 4-8 days, depending on the dataset, model, and crop size. We report all results as an average across five different runs. We refer to Appendix \ref{appendix} for more details regarding the training and model. Our implementation is publicly available at \href{https://github.com/BonifazStuhr/feamgan}{https://github.com/BonifazStuhr/feamgan}.
\\\\
\noindent \textbf{Memory usage.} Our best model requires $\sim$25 GB of VRAM at training time and performs inference using $\sim$12 GB for an image of size 957$\times$526. Our small model, with a slight performance decrease, runs on consumer graphic cards with $\sim$9 GB of VRAM at training time and performs inference using $\sim$8 GB for an image of size 957$\times$526. 
\\\\
\noindent \textbf{Datasets.} We conduct experiments on four translation tasks across four datasets. For all datasets, we compute semantic segmentations with MSeg \cite{lambert2020mseg}, which we use as a condition for our discriminator and to calculate the discriminator masks.\\
\indent(1) \textit{PFD} \cite{richter2016playing} consists of images of realistic virtual world gameplay. Each frame is annotated with pixel-wise semantic labels, which we use as additional input for our generator. We use the same subset as \cite{richter2022enhancing} to compare with recent work.\\
\indent(2) \textit{Viper} \cite{richter2017playing} consists of sequences of realistic virtual world gameplay. Each frame is annotated with different labels, where we use the pixel-wise semantic segmentations as additional input for our generator. Since Cityscapes does not contain night sequences, we remove them from the dataset.\\
\indent(3) \textit{Cityscapes} \cite{cordts2016cityscapes} consists of sequences of real street scenes from 50 different German cities. We use the sequences of the entire training set to train our models.\\
\indent We use datasets (1-3) for the sim-to-real translation tasks \textit{PFD$\rightarrow$Cityscapes} and \textit{Viper$\rightarrow$Cityscapes}. \\
\indent(4) \textit{BDD100K} \cite{yu2020bdd100k} is a large-scale driving dataset.  We use subsets of the training and validation data for the following translation tasks: \textit{Day$\rightarrow$Night}, \textit{Clear$\rightarrow$Snowy}. 
\\\\
\noindent \textbf{Compared methods.}
We compare our work with the following methods.
\vspace{-2pt}
\begin{itemize}
	\itemsep0em 
	\item Color Transfer (CT) \cite{reinhard2001color} performs color correction by transferring statistical features in lαβ space from the target to the source image.
	\item MUNIT \cite{huang2018multimodal} achieves multimodal translation by recombining the content code of an image with a style code sampled from the style space of the target domain. It is an extension of CycleGAN \cite{zhu2017unpaired} and UNIT \cite{liu2017unsupervised}.
	\item CUT \cite{park2020contrastive} uses a patchwise contrastive loss to achieve one-sided unsupervised image-to-image translation.	
	\item TSIT \cite{jiang2020tsit} achieves one-sided translation by fusing features from the content stream into the generator on multiple scales using FADE and utilizing a perceptual loss between the translated and source images.
	\item QS-Attn \cite{hu2022qs} builds upon CUT \cite{park2020contrastive} with an attention module that selects significant anchors for the contrastive loss instead of features from random locations of the image.
	\item EPE \cite{richter2022enhancing} relies on a variety of gbuffers as input. Techniques such as similarity cropping, utilizing segmentations for both domains generated by a robust segmentation model as input to the conditional discriminators, and small patch training are used to achieve content consistency.
\end{itemize}
Since EPE \cite{richter2022enhancing} provides inferred images of size 957$\times$526 for the \textit{PFD$\rightarrow$Cityscapes} task, comparisons are performed on this resolution. For the \textit{Viper$\rightarrow$Cityscapes}, \textit{Day$\rightarrow$Night}, and \textit{Clear$\rightarrow$Snowy} tasks, we train the models using their official implementations. Furthermore, we retrain models as additional baselines for the \textit{PFD$\rightarrow$Cityscapes} task.
\\\\
\noindent \textbf{Evaluation metrics.} Following prior work \cite{richter2022enhancing}, we use the Fréchet Inception Distance (FID) \cite{heusel2017gans}, the Kernel Inception Distance (KID) \cite{binkowski2018demystifying}, and the semantically aligned Kernel VGG Distance (sKVD) \cite{richter2022enhancing} to evaluate image translation quality quantitatively. The sKVD metric was introduced in \cite{richter2022enhancing} and improved over previous metrics for mismatched layouts in source and target data. In addition, we propose the class-specific Kernel VGG Distance (cKVD), where a robust segmentation model is used before the sKVD calculation to mask input crops by class (or category). Thereby, for each given class, all source and target image crops are filtered using their segmentations by erasing the pixels of all other classes. We select crops where more then $5$\% of the pixels belong to the respective class. Then, the sKVD is calculated class-wise on the filtered crops. Afterward, we can report the cKVD as an average over all classes or separately for each class to achieve a more fine-grained measurement. We follow \cite{richter2022enhancing} and use a crop size of $1/8$ and sample source and target crop pairs with an similarity threshold  of $0.5$ between unmasked source and target segmentation crops. More information on the classes used in the cKVD metric can be found in \autoref{tab:sKVD_class_mapping} of Appendix \ref{appendix}. For the KID, sKVD, and cKVD metrics, we multiply the measurements by $1000$ to improve the readability of results.
\begin{figure*}[h] 
	\centering
	\renewcommand{\thesubfigure}{}
	{\includegraphics[width=0.33\textwidth]{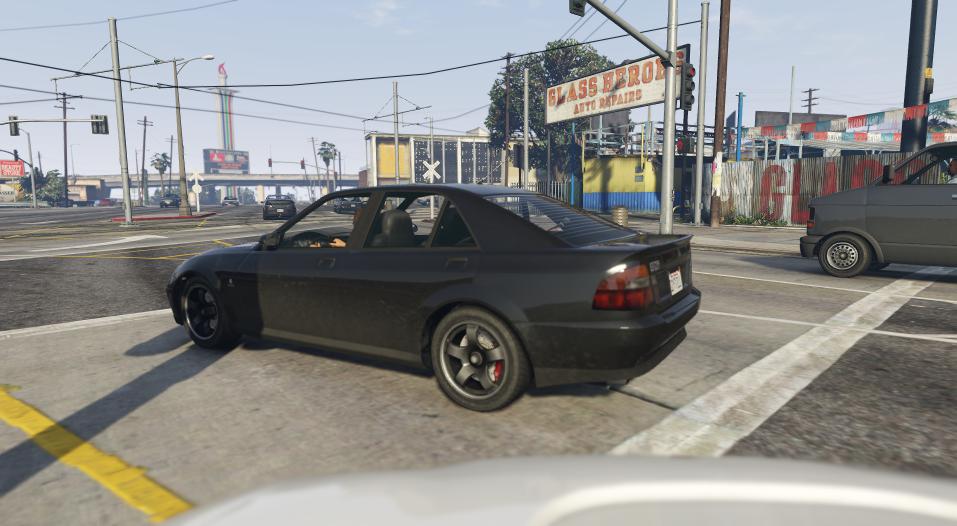}}\hfill
	{\includegraphics[width=0.33\textwidth]{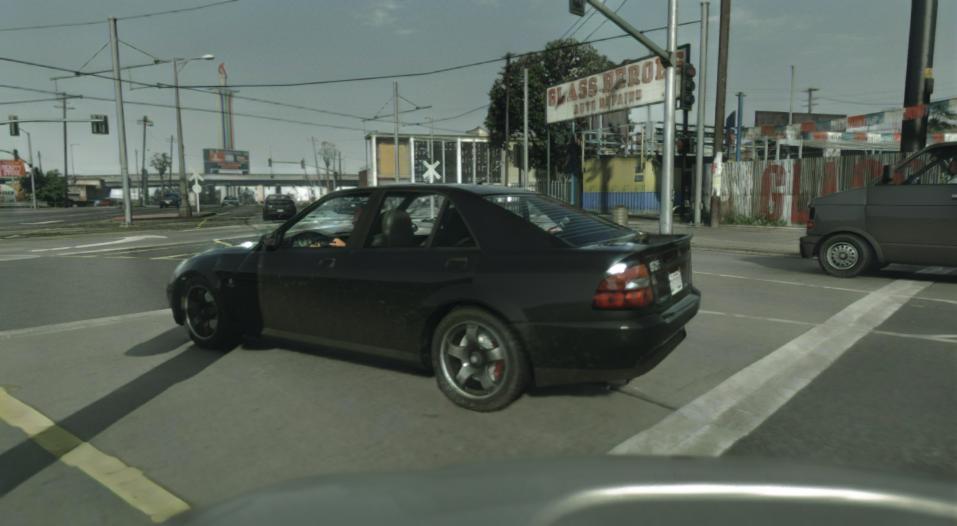}}\hfill
	{\includegraphics[width=0.33\textwidth]{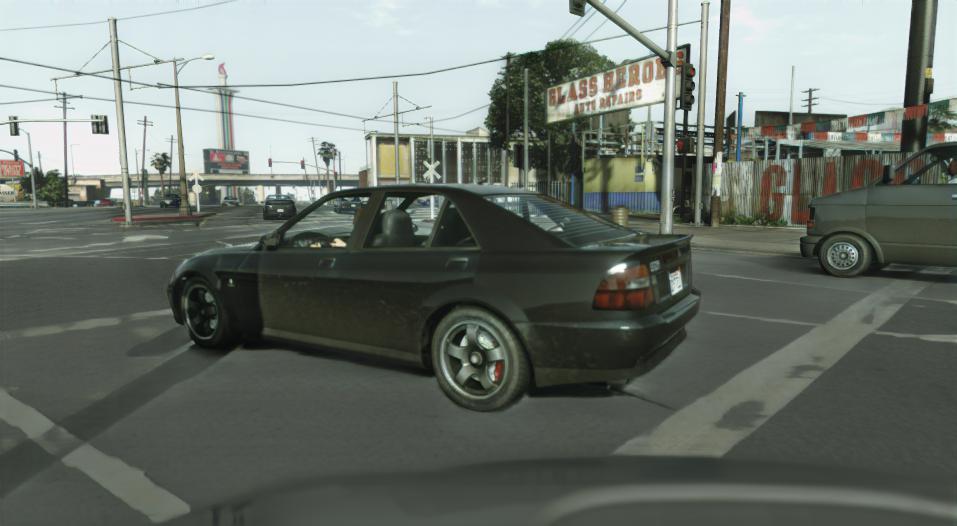}}\hfill\\\vspace{-2pt}
	\subfigure[Input]
	{\includegraphics[width=0.33\textwidth]{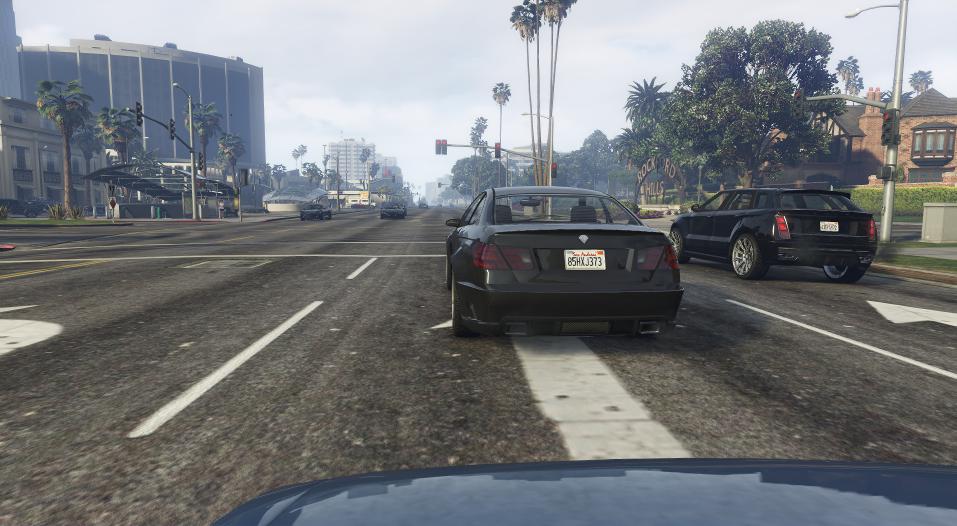}}\hfill
	\subfigure[EPE]
	{\includegraphics[width=0.33\textwidth]{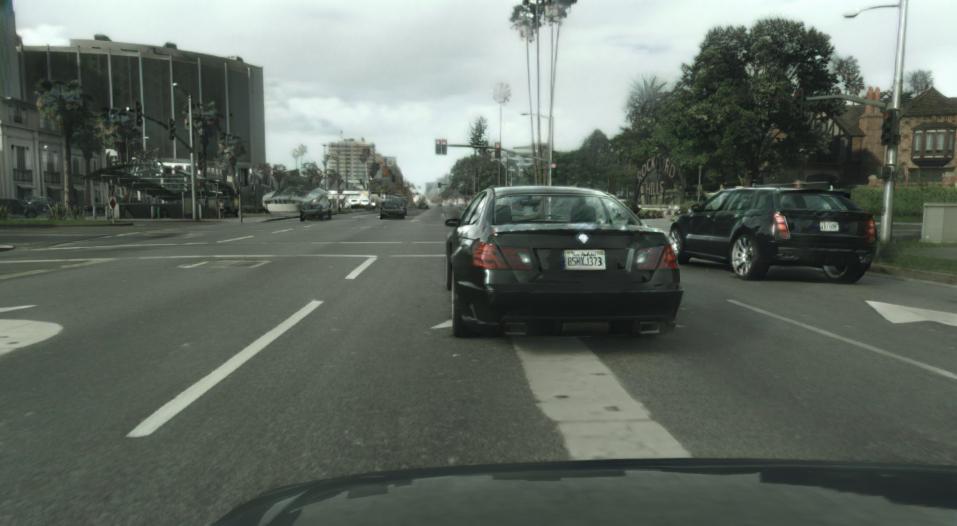}}\hfill
	\subfigure[FeaMGAN (ours)]
	{\includegraphics[width=0.33\textwidth]{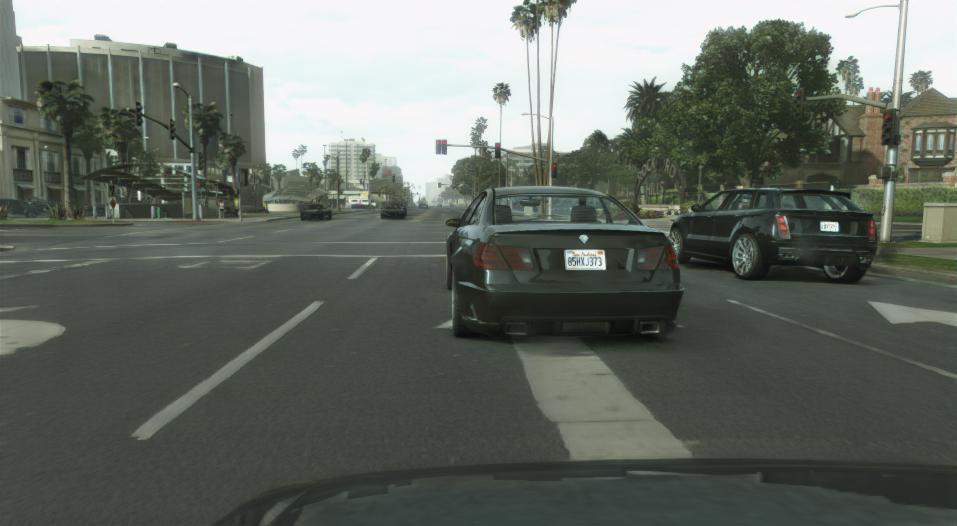}}\hfill
	\subfigure[Generator Input Types]
	{\includegraphics[width=1.0\linewidth]{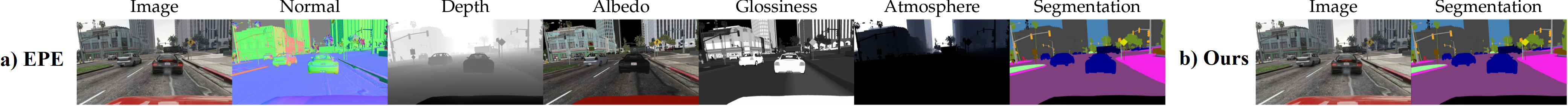}}\\
	\vspace{-6pt}
	\caption{\textbf{Qualitative comparison to EPE.} We compare our method with the provided inferred images of EPE \cite{richter2022enhancing}. }
	\label{fig:qualitative_comparison_epe_baselines}
\end{figure*}
\begin{table*}[h]
	\caption{\textbf{Quantitative comparison to the baselines provided by EPE.} We calculate all metrics on the provided inferred images of EPE and its baselines \cite{richter2022enhancing}.}
	\setlength{\tabcolsep}{3.35pt}
	\begin{center}
		\begin{tabular}{lccccccccccccccc}
			\toprule
			\multirow{2}{*}{Method}&\multirow{2}{*}{FID}&\multirow{2}{*}{KID} &\multirow{2}{*}{sKVD}&\multicolumn{12}{c}{cKVD}\\
			\cmidrule(lr){5-16}
			&    &  &  & AVG& AVG$_{sp}$&		sky& 	ground&	road&	terrain&	vegetation&	building&	roadside-obj.&	person&	vehicle&rest\\
			\midrule	
			ColorTransfer& 84.34 & 88.17 & 16.65 & 36.01 & 33.12& 32.40 & \textbf{12.97} &16.13 & 20.94 & 19.24 & 29.92 & 74.79 & 62.78 & 41.79 & \textbf{49.16}\\
			MUNIT  & 45.00  & 35.05 & 16.51  & 38.57 &34.81 &29.80 & 16.93	& 17.62 & 29.52& 19.29	&\textbf{24.28} & 79.14 & 77.34 & 40.13 & 51.61 \\
			CUT & 47.71 & 42.01   & 18.03   & 35.31 & 33.26 &\textbf{25.96} & 15.32 &	17.87 & \textbf{20.09} & 22.72 & 25.00 & 74.02 & \textbf{60.99} & 41.71	&49.37 \\
			EPE & 44.06 & 33.66  & 13.87  & \textbf{35.22} &\textbf{30.21}&27.14	&13.54	&\textbf{13.56}&24.77&20.77	&26.75&\textbf{50.58} &	83.34&41.29&50.45 \\
			\midrule			
			FeaMGAN-S (ours) & \textbf{43.27} & \textbf{32.59} & \textbf{12.98}  &40.23	&32.69	
			& 38.10	&13.29	&15.34	&26.29	&20.17	&27.32	&61.57&102.65	&42.83	&54.73 \\
			FeaMGAN (ours) & \textbf{40.32}  & \textbf{28.59} & \textbf{12.94} & 40.02  &31.78&46.70	&13.72	&15.60	&23.23	&\textbf{17.69}	&25.57	&66.65	&99.24	&\textbf{39.38}	&52.40 \\	
			\bottomrule
		\end{tabular}
	\end{center}
	\label{tab:quantitative_comparison_epe_baselines}
\end{table*}
\subsection{Comparison to the State of the Art}
We compare our models quantitatively and qualitatively with different baselines. First, we compare our results with EPE and the baselines provided by EPE \cite{richter2022enhancing}. Then, we train our own baselines on the four translation tasks for further comparison. \\\\
\noindent \textbf{Comparison to EPE.}
A set of inferred images is provided for EPE and each of the baselines \cite{richter2022enhancing}. Therefore, we train our models on the same training set and use the inferred images from our best models for this comparison. We select our best models based on scores of various visual metrics and visual inspections of translated images. As shown in \autoref{fig:qualitative_comparison_epe_baselines} a) and b), our model relies solely on segmentation maps as additional input compared to EPE, which uses a variety of gbuffers. In addition, our model is trained with significantly fewer steps ($\sim\!400$K iterations) compared to EPE and the baselines ($1$M iterations). As shown in \autoref{tab:quantitative_comparison_epe_baselines}, our model outperforms the baselines and EPE in all commonly used metrics (FID and KID) and the sKVD metric. More surprisingly, our small model, which can be trained on consumer GPUs, outperforms all baselines and EPE as well. 

\begin{figure*}[h] 
	\renewcommand{\thesubfigure}{}
	\begin{center}
		{\includegraphics[width=0.248\textwidth]{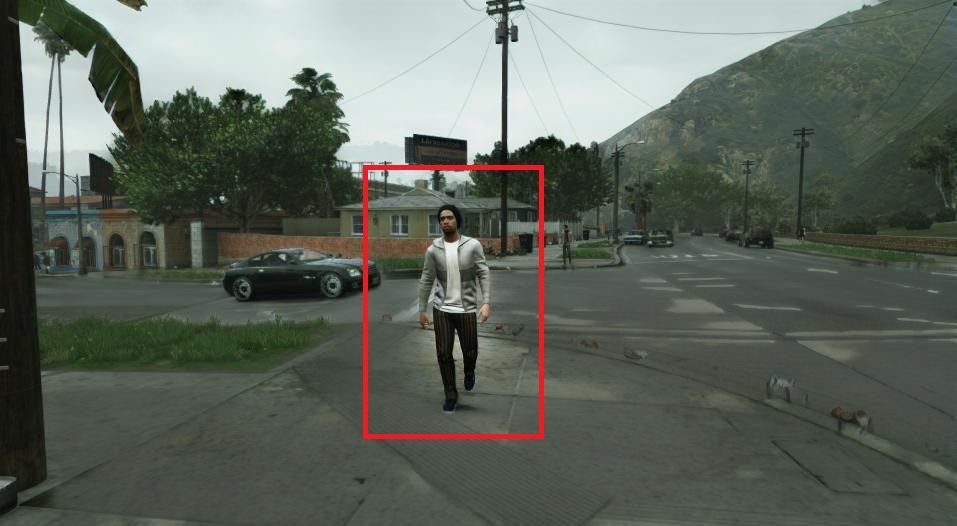}}\hfill
		{\includegraphics[width=0.248\textwidth]{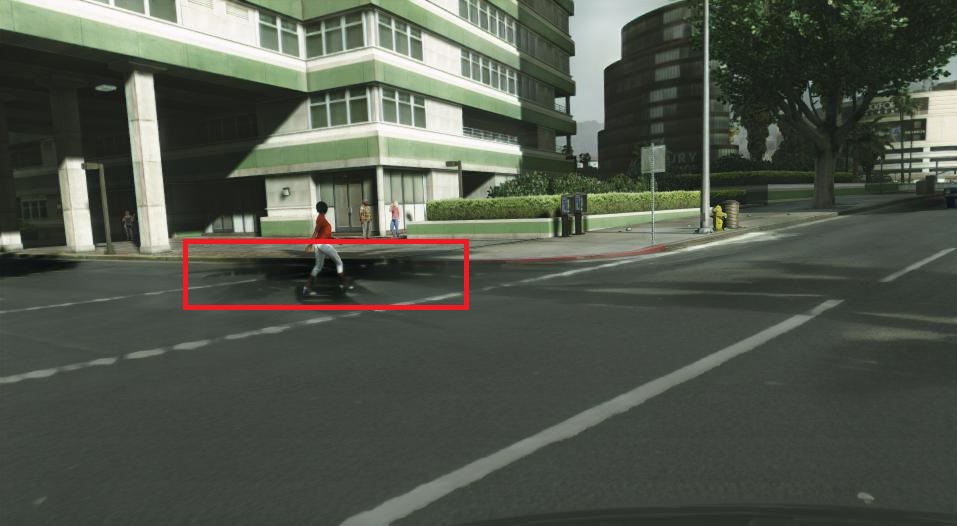}}\hfill
		{\includegraphics[width=0.248\textwidth]{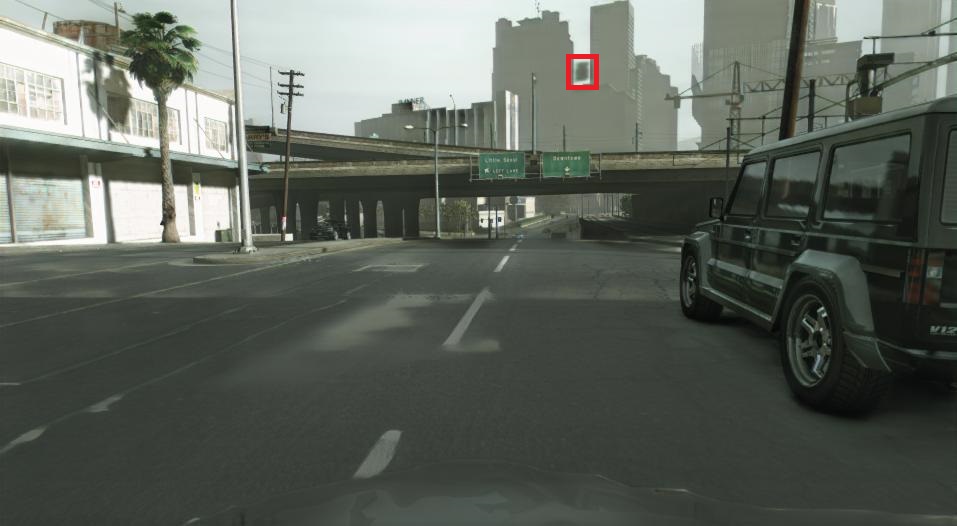}}\hfill
		{\includegraphics[width=0.248\textwidth]{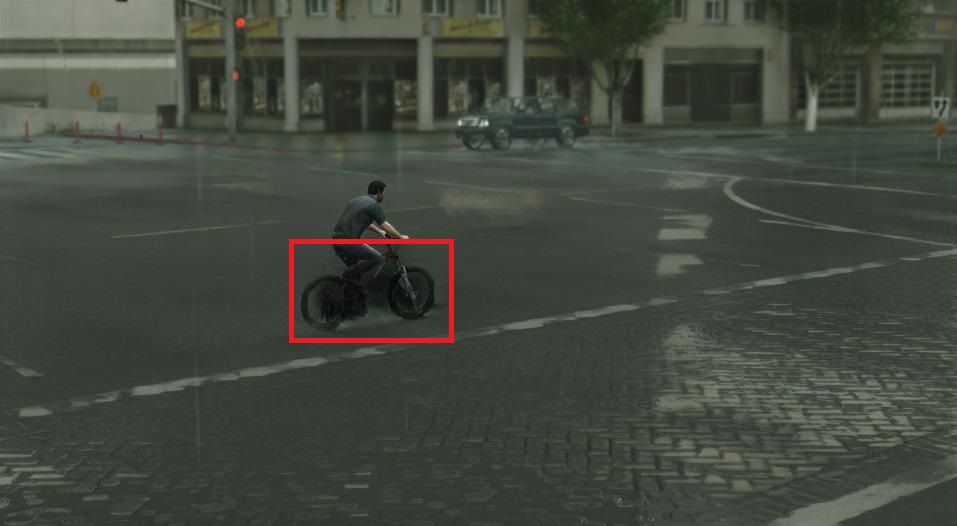}}\hfill \\ \vspace{1.4pt}
		{\includegraphics[width=0.248\textwidth]{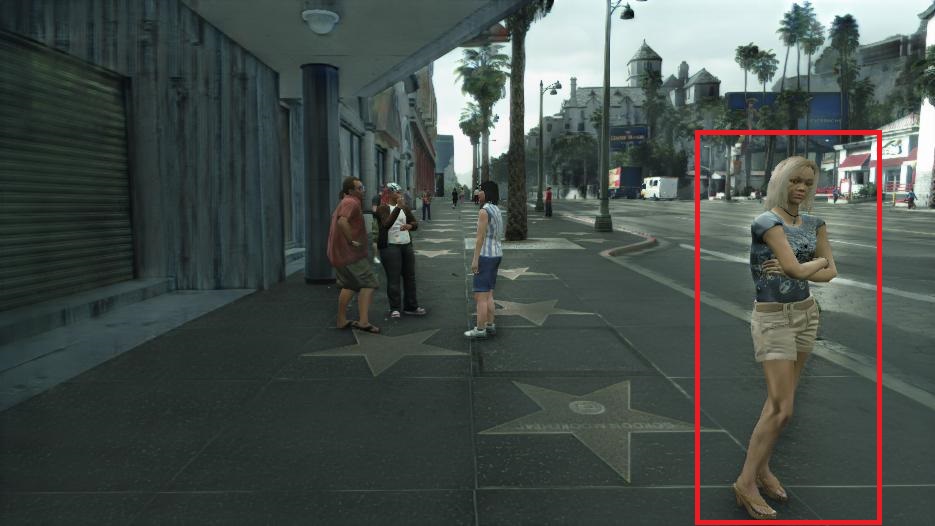}}\hfill
		{\includegraphics[width=0.248\textwidth]{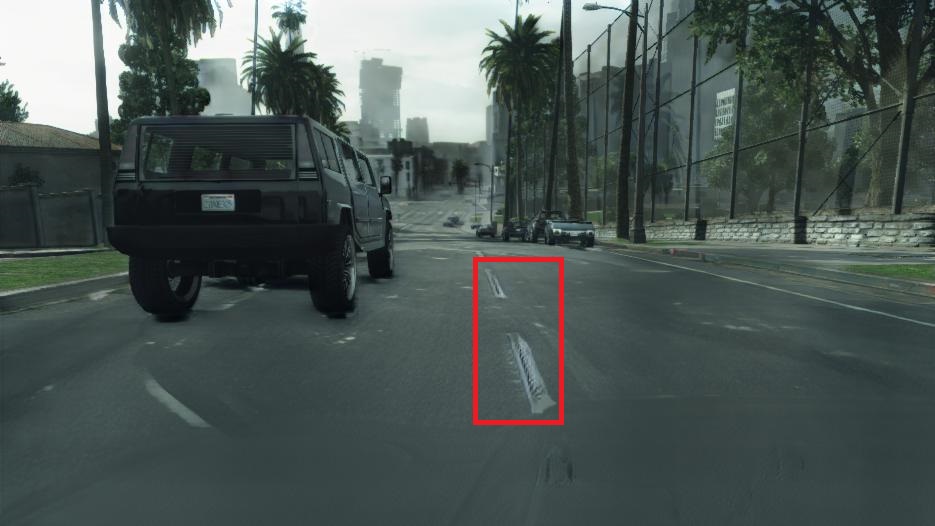}}\hfill
		{\includegraphics[width=0.248\textwidth]{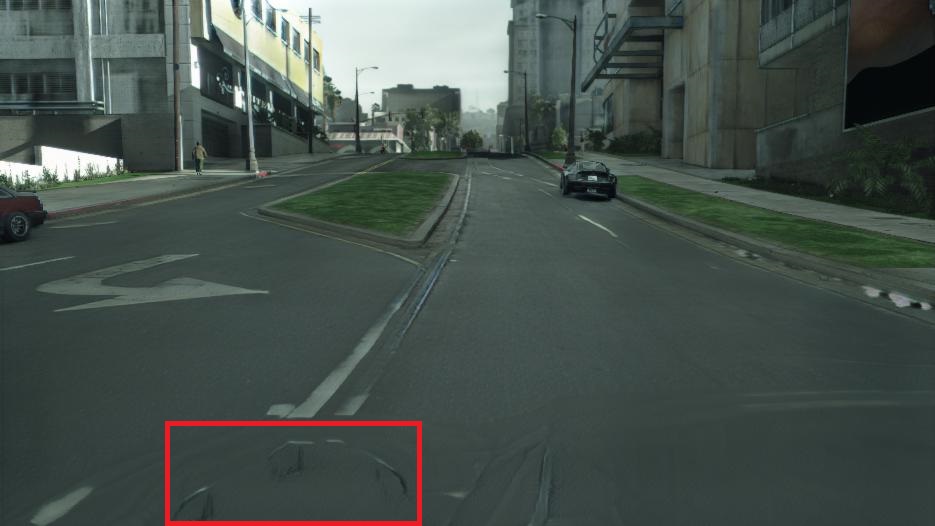}}\hfill
		{\includegraphics[width=0.248\textwidth]{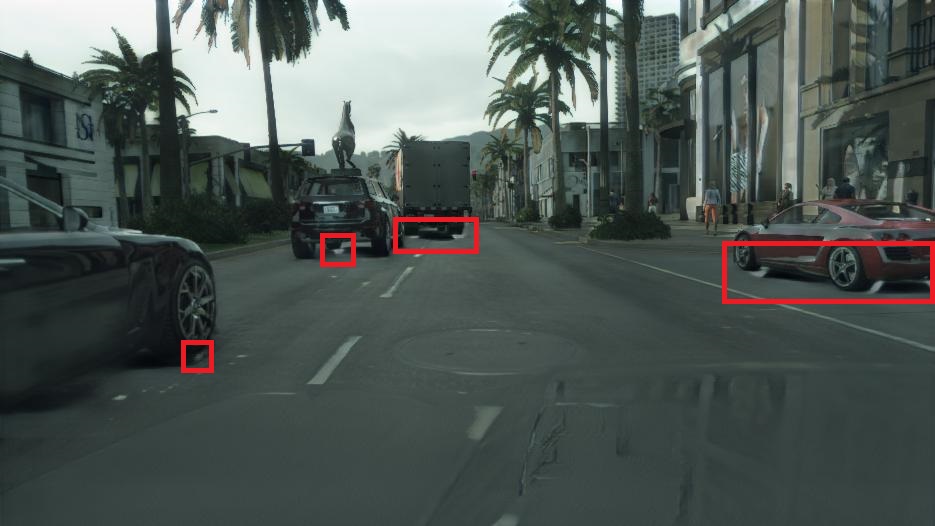}}\hfill \\ \vspace{1.1pt}
		{\includegraphics[width=0.248\textwidth]{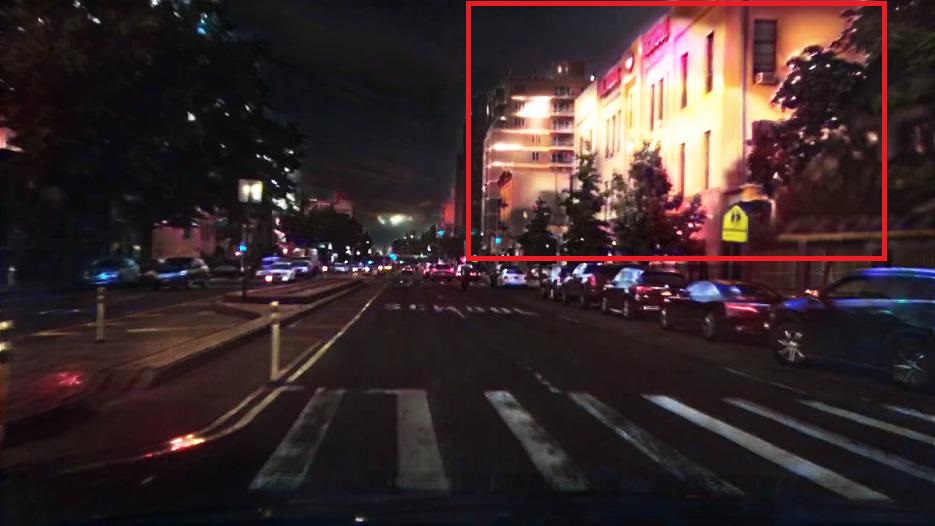}}\hfill
		{\includegraphics[width=0.248\textwidth]{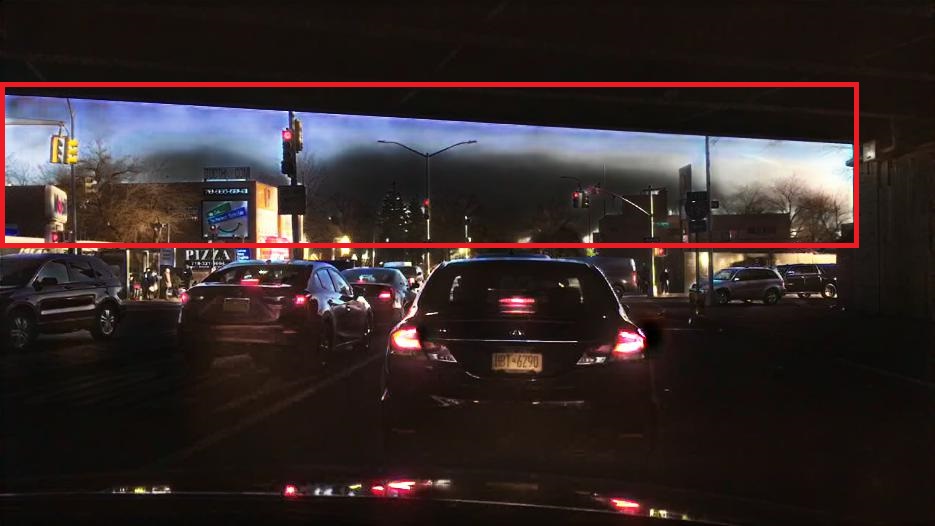}}\hfill
		{\includegraphics[width=0.248\textwidth]{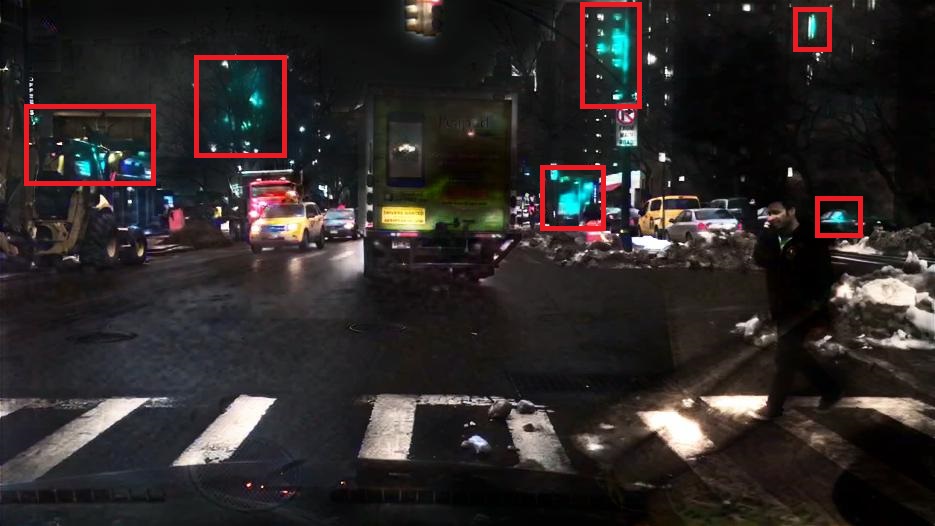}}\hfill 
		{\includegraphics[width=0.248\textwidth]{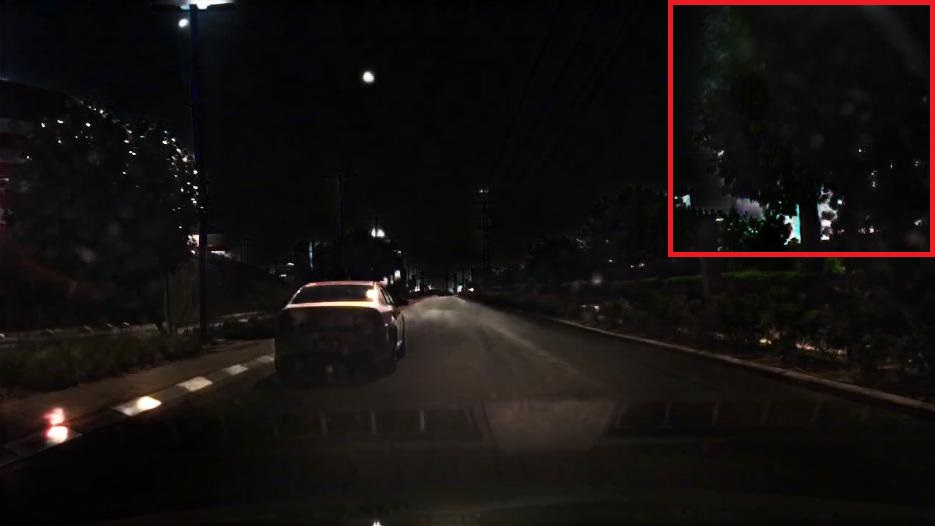}}\hfill  \\
		\vspace{-3.9pt}
		\subfigure[Glowing objects]
		{\includegraphics[width=0.248\textwidth]{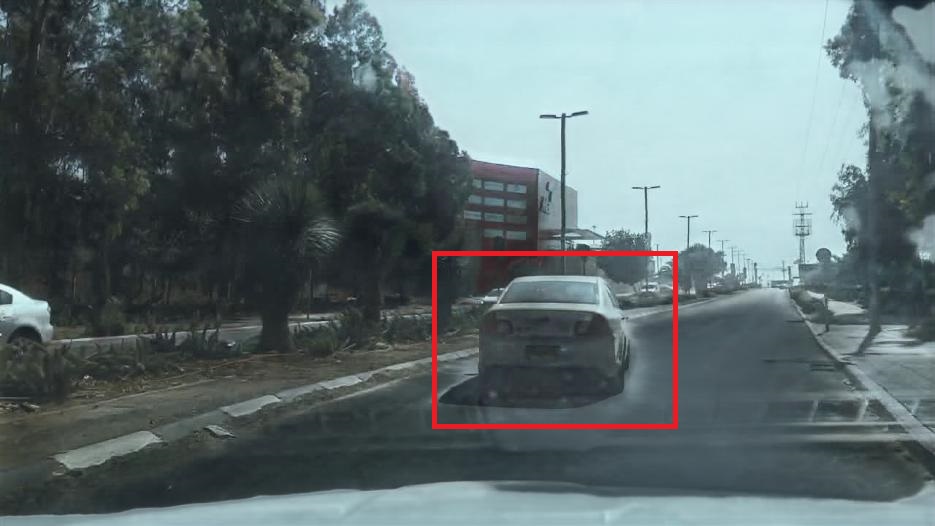}}\hfill
		\subfigure[Intra-class inconsistencies]
		{\includegraphics[width=0.248\textwidth]{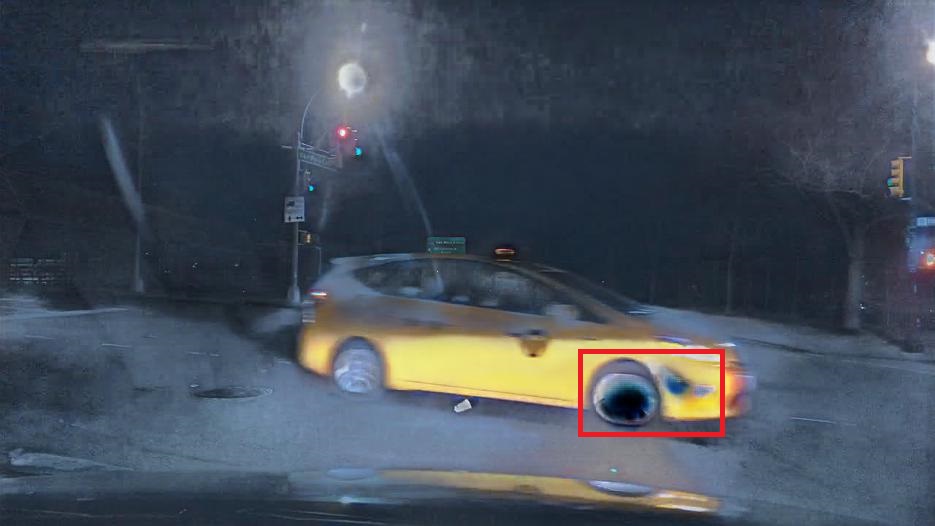}}\hfill
		\subfigure[Minor hallucinations]
		{\includegraphics[width=0.248\textwidth]{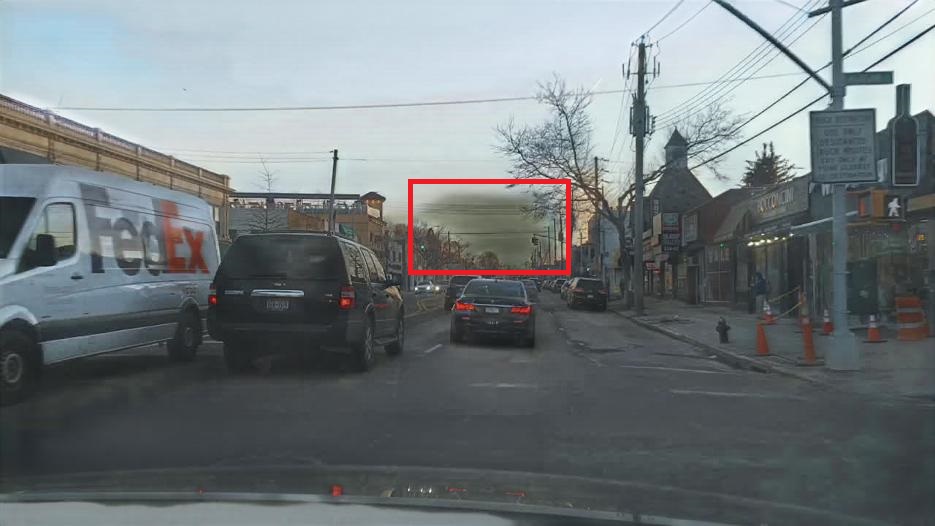}}\hfill
		\subfigure[Class boundary artifiacts]
		{\includegraphics[width=0.248\textwidth]{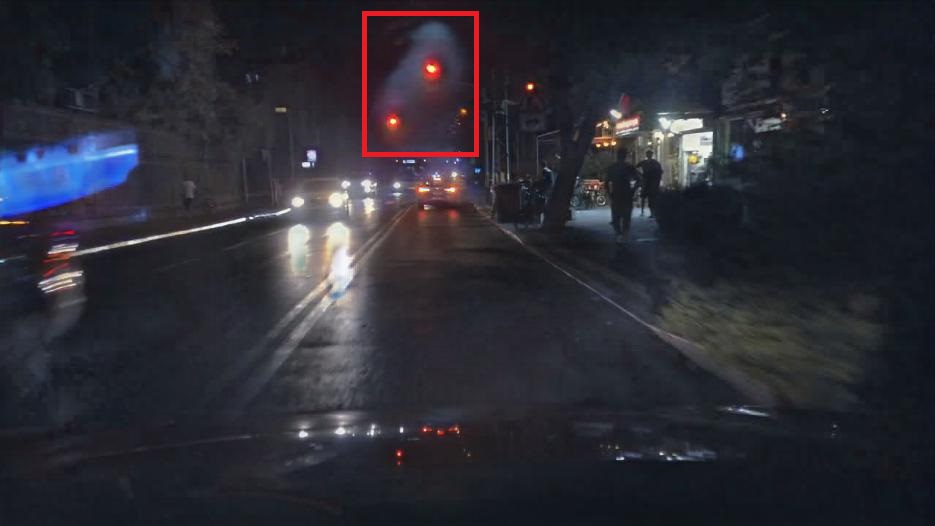}}\hfill
	\end{center}
	\vspace{-8pt}
	\caption{\textbf{Limitations.}}
	\label{fig:limitations}
\end{figure*}

However, our cKVD metric shows that our models have difficulty with the person and sky classes. Therefore, the average cKVD values are high and become low when we remove both classes from the average calculation (AVG$_{sp}$). A possible reason for the weaker performance on the person class is our masking procedure. Since the masking procedure requires overlapping samples in both domains, the person class is not seen frequently during training. This can lead to inconsistencies (a glow) around the person class, as seen in \autoref{fig:limitations} of our limitations. The masking procedure also leads to a drop in performance in the sky class, as seen in \autoref{tab:quantitative_ablation} of our ablation study.
\begin{figure*}[h] 
	\renewcommand{\thesubfigure}{}	
	\begin{center}	
		{\scriptsize PFD$\rightarrow$Cityscapes}\hfill\\\vspace{1pt}
		{\includegraphics[width=0.165\textwidth]{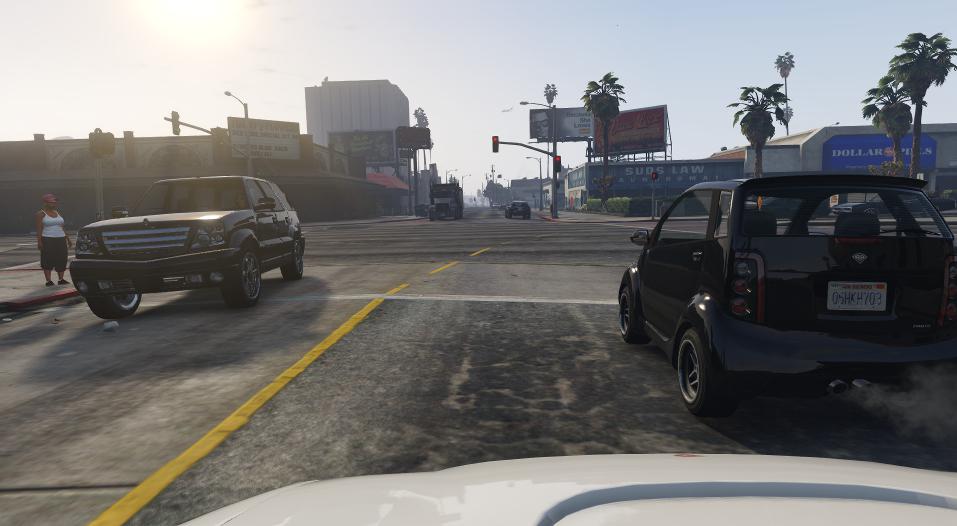}}\hfill
		{\includegraphics[width=0.165\textwidth]{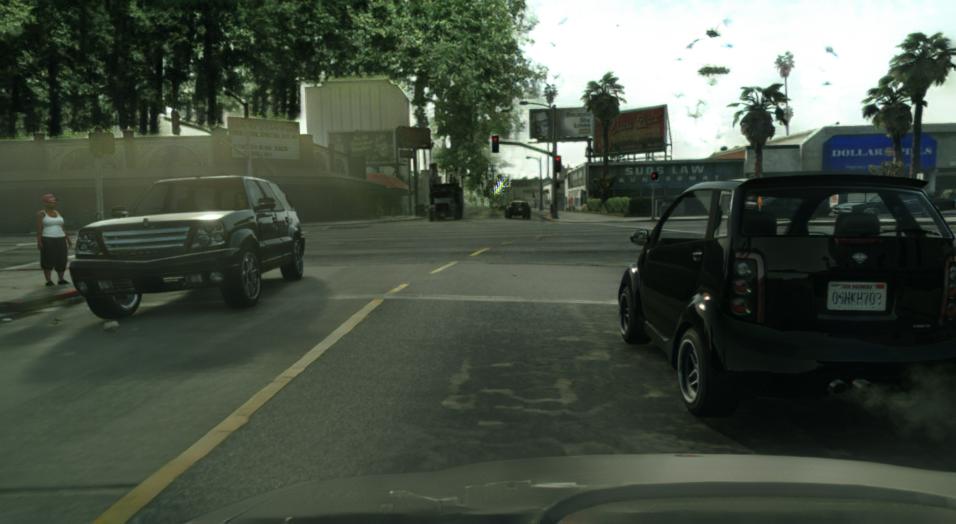}}\hfill
		{\includegraphics[width=0.165\textwidth]{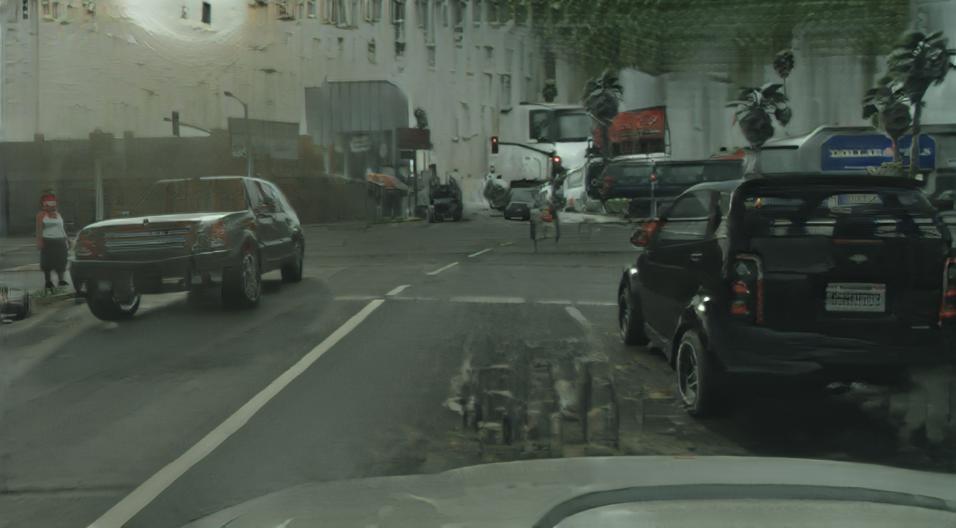}}\hfill
		{\includegraphics[width=0.165\textwidth]{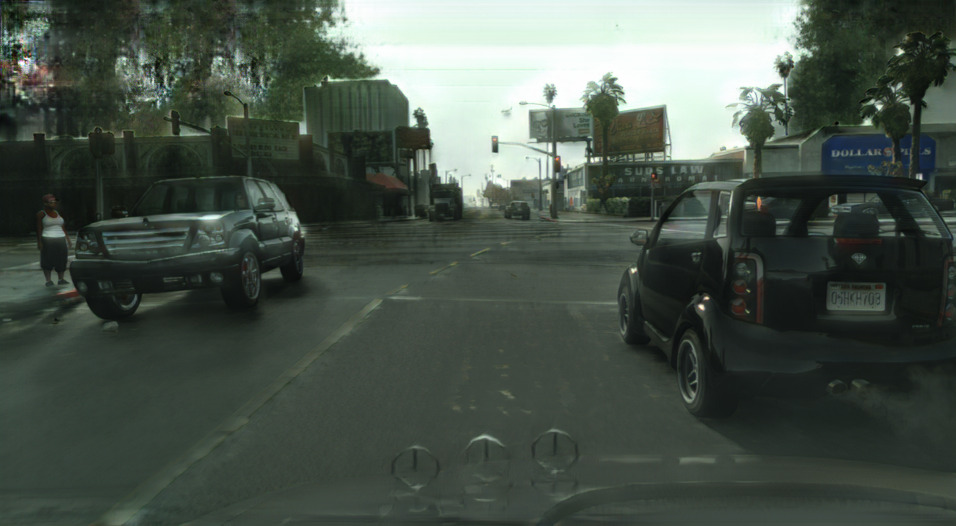}}\hfill
		{\includegraphics[width=0.165\textwidth]{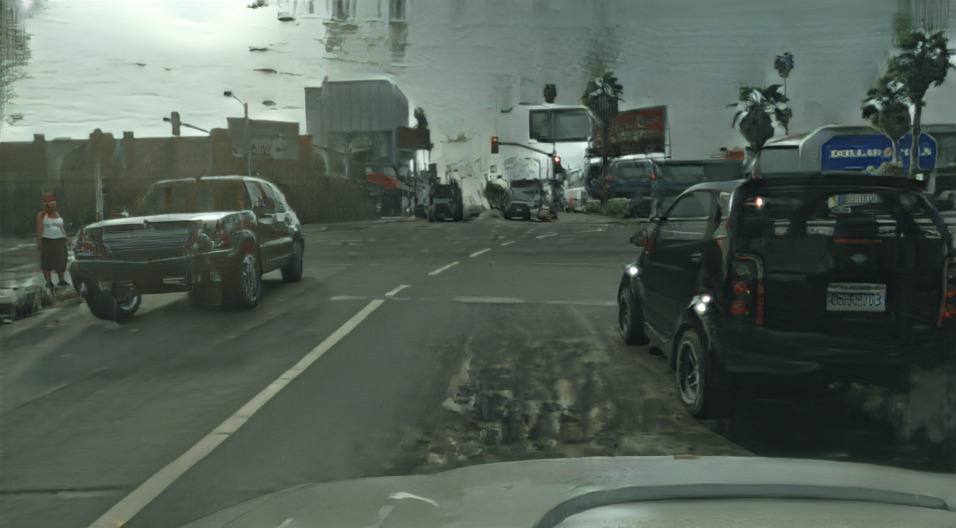}}\hfill
		{\includegraphics[width=0.165\textwidth]{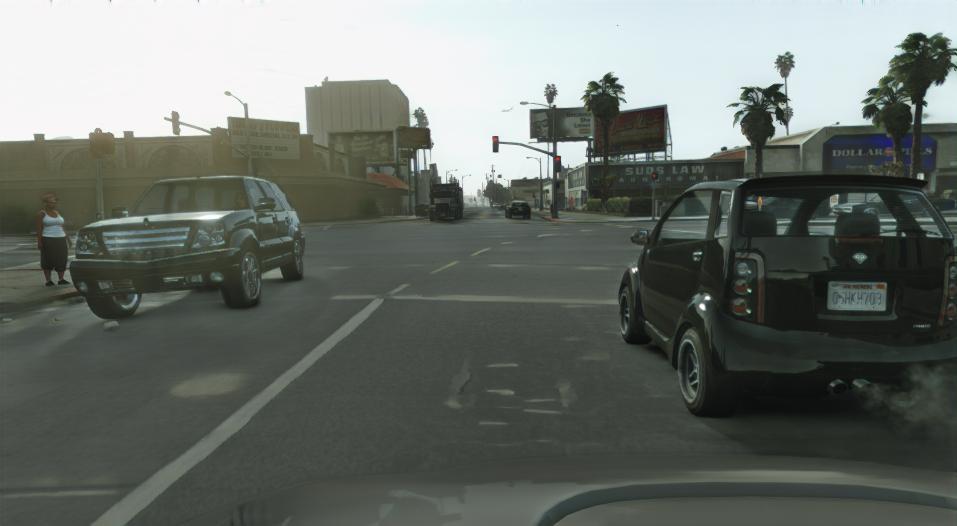}}\hfill \\ \vspace{-2.2pt}
		{\scriptsize Viper$\rightarrow$Cityscapes} \hfill\\\vspace{1pt}
		{\includegraphics[width=0.165\textwidth]{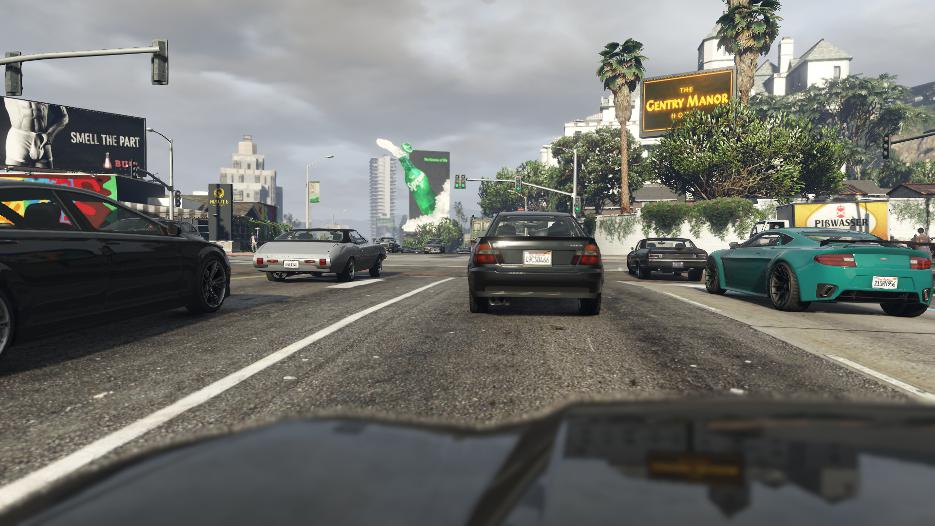}}\hfill
		{\includegraphics[width=0.165\textwidth]{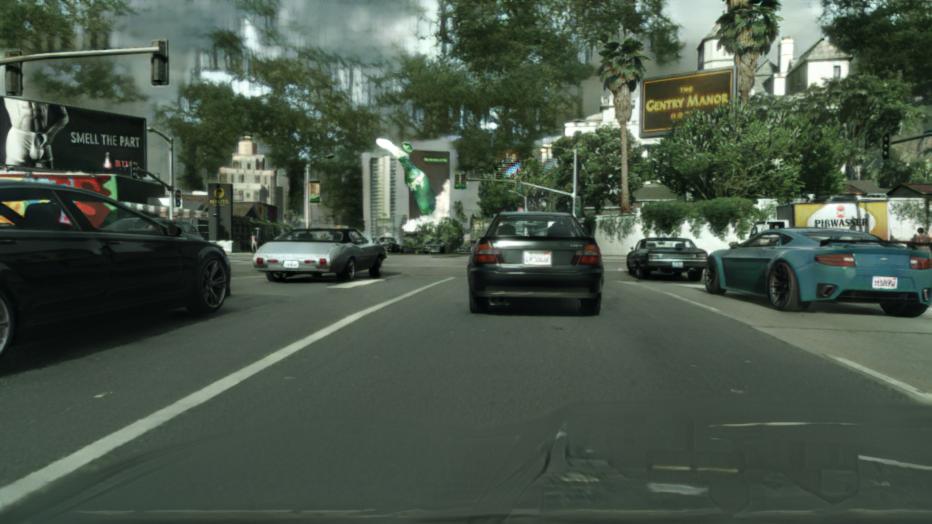}}\hfill
		{\includegraphics[width=0.165\textwidth]{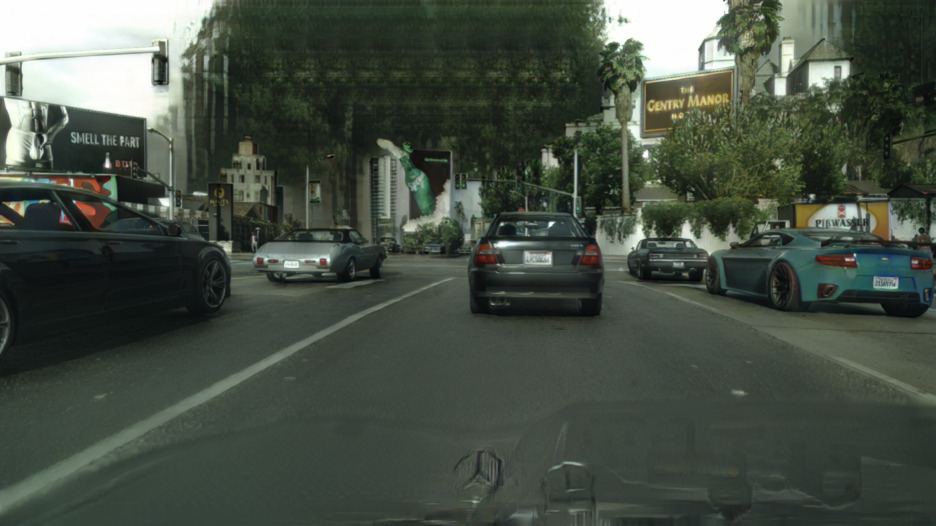}}\hfill
		{\includegraphics[width=0.165\textwidth]{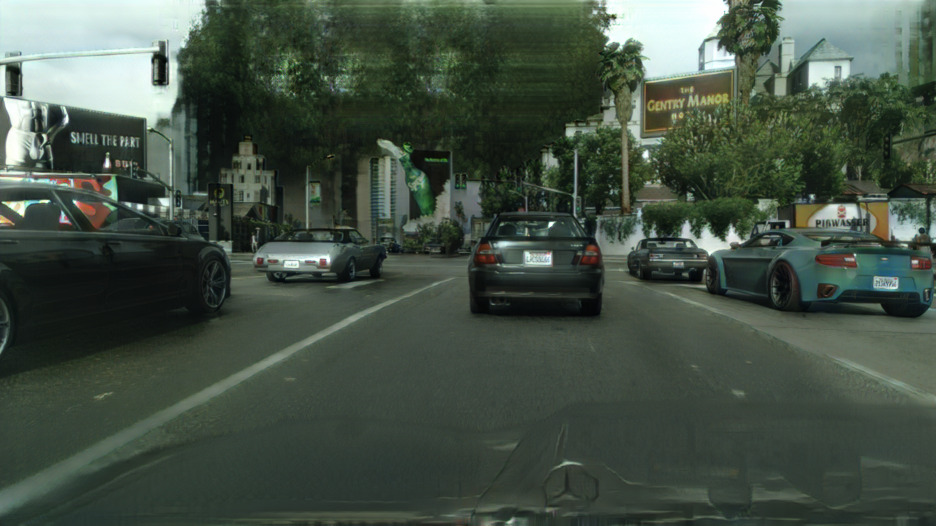}}\hfill
		{\includegraphics[width=0.165\textwidth]{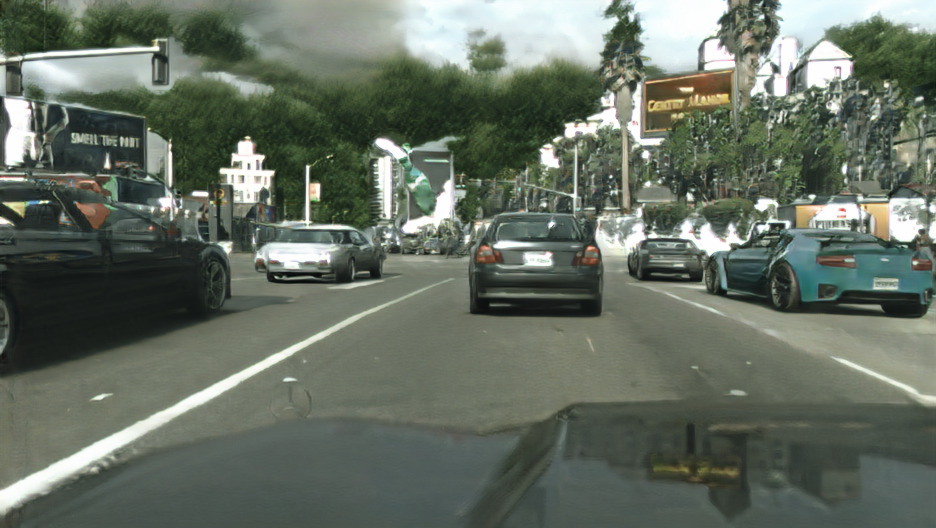}}\hfill
		{\includegraphics[width=0.165\textwidth]{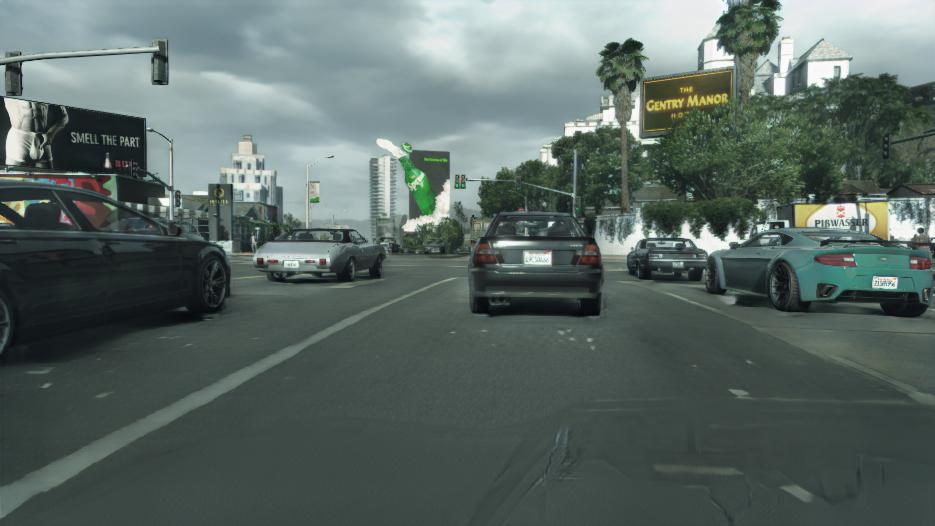}}\hfill \\ \vspace{-2.2pt}
		{\scriptsize Day$\rightarrow$Night} \hfill\\\vspace{1pt}
		{\includegraphics[width=0.165\textwidth]{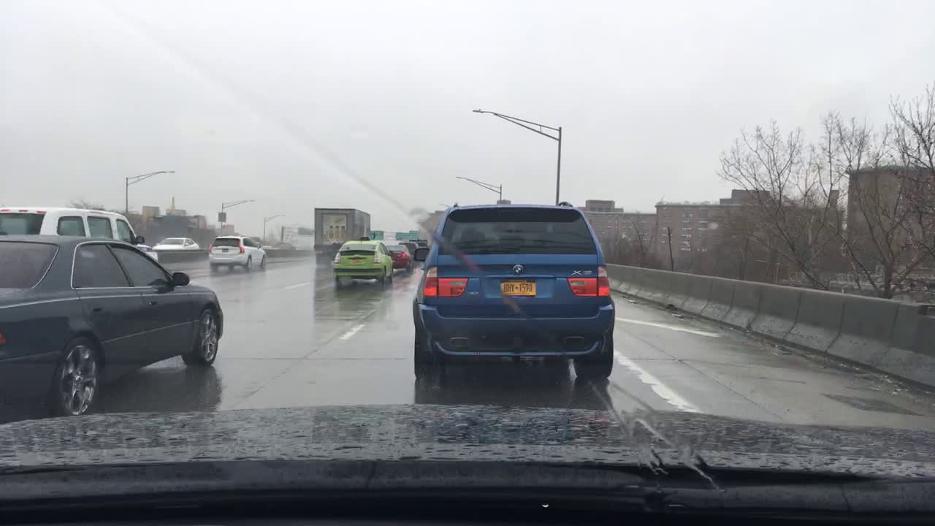}}\hfill
		{\includegraphics[width=0.165\textwidth]{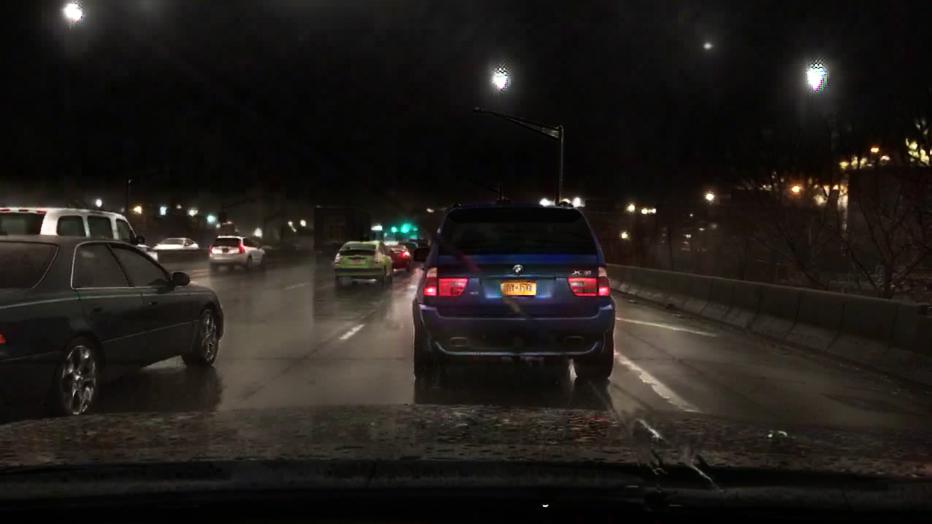}}\hfill
		{\includegraphics[width=0.165\textwidth]{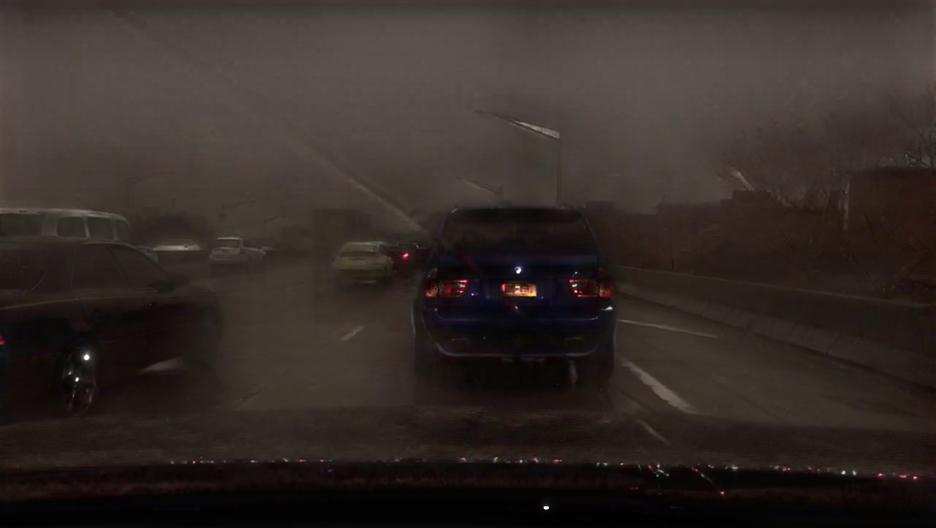}}\hfill
		{\includegraphics[width=0.165\textwidth]{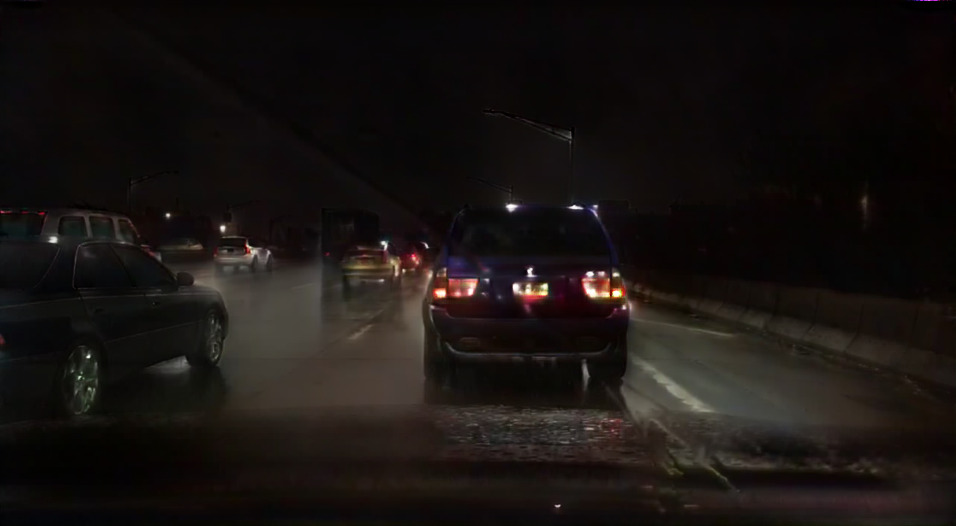}}\hfill
		{\includegraphics[width=0.165\textwidth]{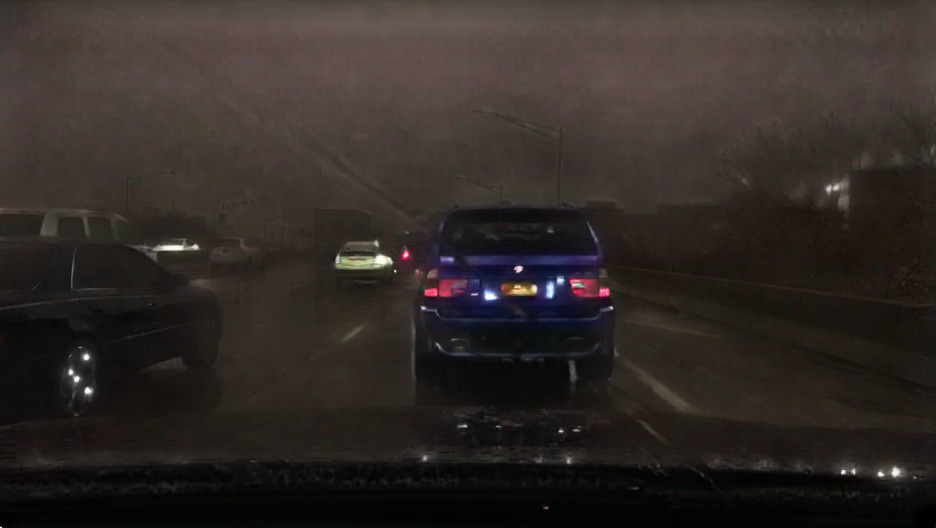}}\hfill
		{\includegraphics[width=0.165\textwidth]{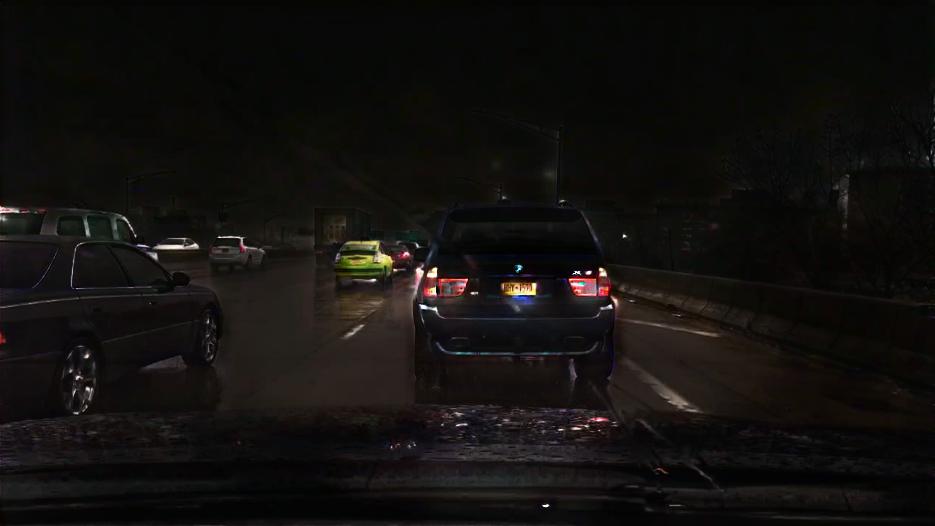}}\hfill \\ \vspace{-2.2pt}
		{\scriptsize Clear$\rightarrow$Snowy} \hfill\\\vspace{-3pt}
		\subfigure[Input]
		{\includegraphics[width=0.165\textwidth]{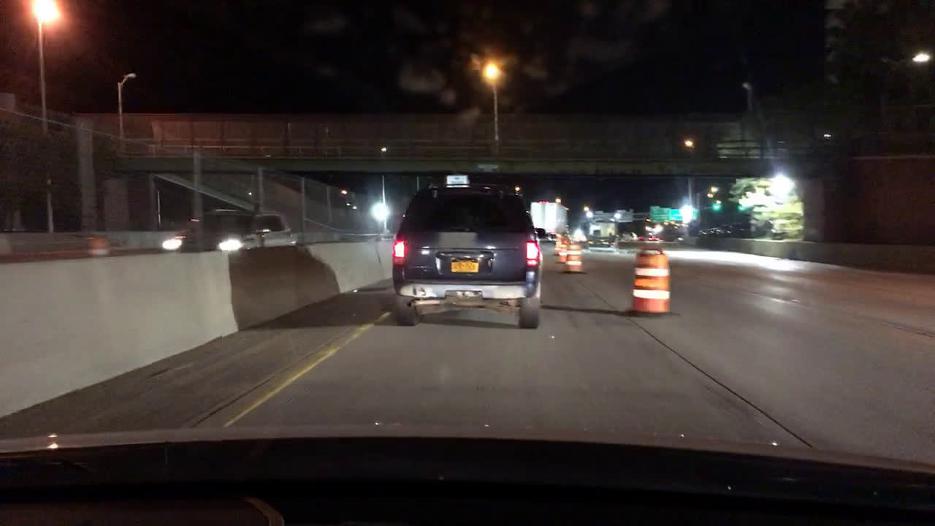}}\hfill
		\subfigure[MUNIT]
		{\includegraphics[width=0.165\textwidth]{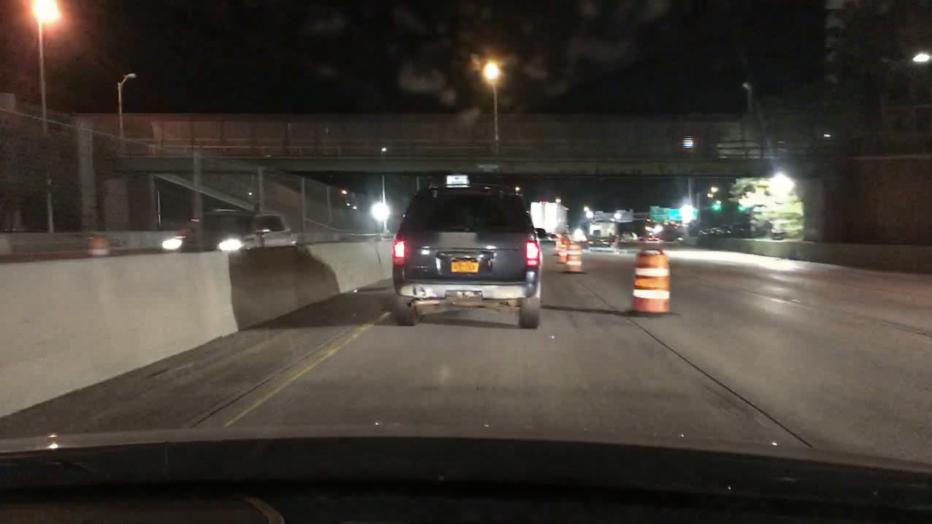}}\hfill
		\subfigure[CUT]
		{\includegraphics[width=0.165\textwidth]{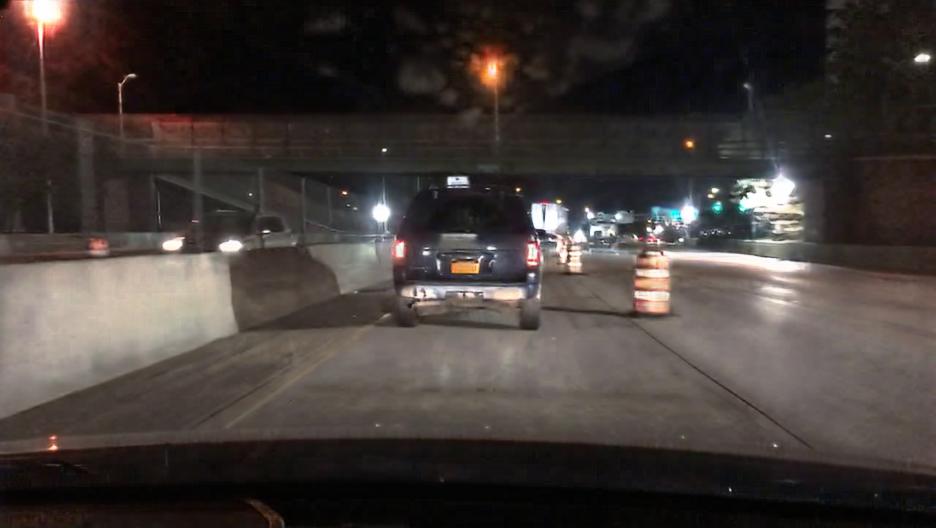}}\hfill
		\subfigure[TSIT]
		{\includegraphics[width=0.165\textwidth]{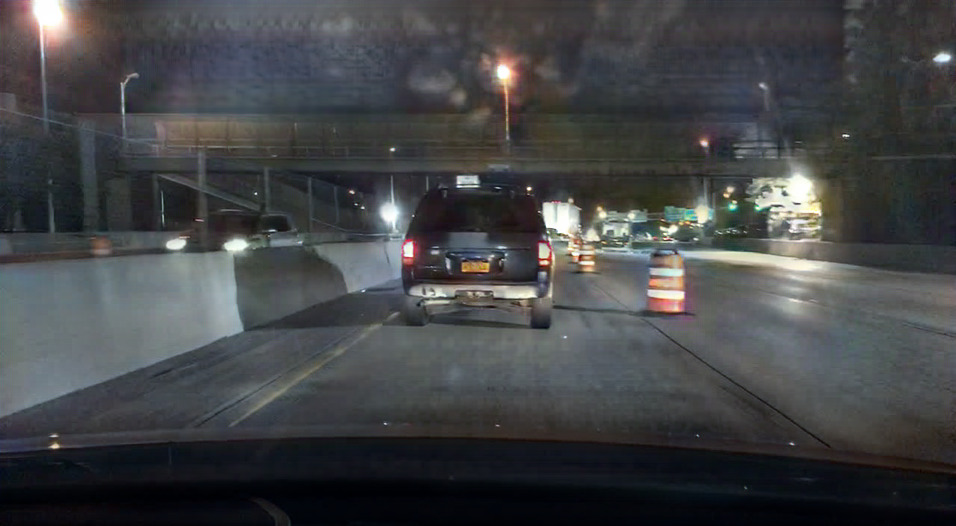}}\hfill
		\subfigure[QS-Attn]
		{\includegraphics[width=0.165\textwidth]{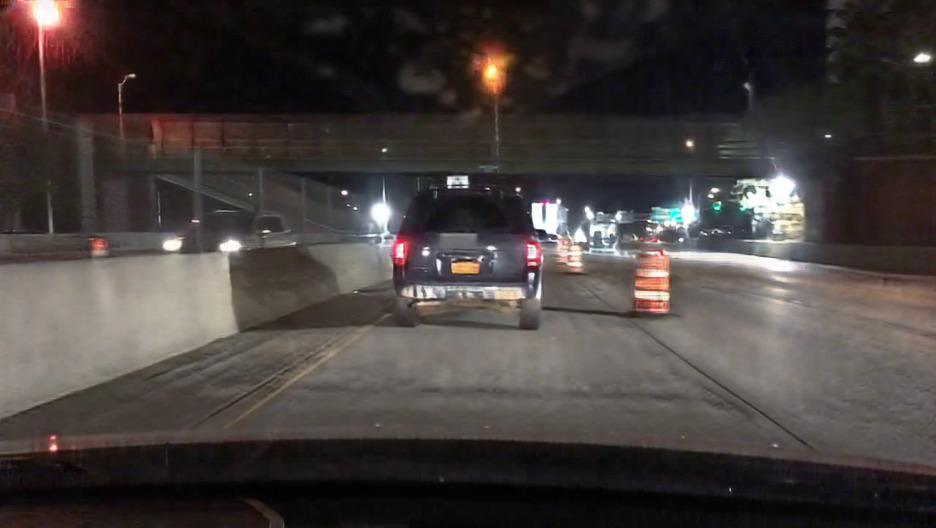}}\hfill
		\subfigure[FeaMGAN (ours)]
		{\includegraphics[width=0.165\textwidth]{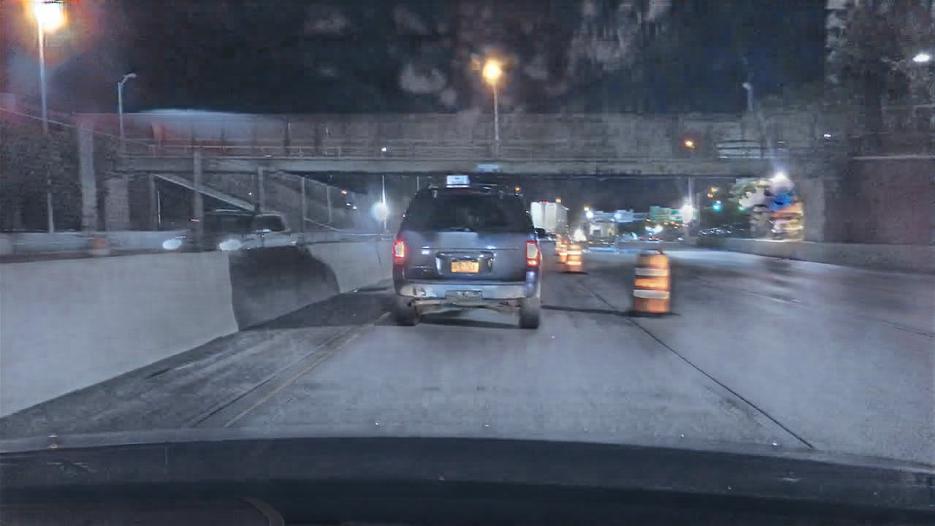}}\hfill
	\end{center}\vspace{-8pt}
	\caption{\textbf{Qualitative comparison to prior work.} Models were trained using their official implementations. Randomly sampled results can be found in \autoref{fig:qualitative_comparison_additional_random} of Appendix \ref{appendix}.}
	\label{fig:qualitative_comparison}
\end{figure*}
\begin{table*}[h]
	\caption{\textbf{Quantitative comparison to prior work.} Models were trained using their official implementations. Results are reported as the average across five runs. We refer to \autoref{tab:quantitative_comparison_extended} of Appendix \ref{appendix} for an extended version of this table.}
	\setlength{\tabcolsep}{3.9pt}
	\begin{center}
		\begin{tabular}{lcccccccccccccccc}
			\toprule
			\multirow{2}{*}{Method}&\multicolumn{4}{c}{PFD$\rightarrow$Cityscapes}&\multicolumn{4}{c}{Viper$\rightarrow$Cityscapes}&\multicolumn{4}{c}{Day$\rightarrow$Night}&\multicolumn{4}{c}{Clear$\rightarrow$Snowy}\\
			\cmidrule(lr){2-5}
			\cmidrule(lr){6-9}
			\cmidrule(lr){10-13}
			\cmidrule(lr){14-17}
			& FID  & KID & sKVD& cKVD& FID  & KID & sKVD& cKVD& FID  & KID & sKVD& cKVD& FID  & KID & sKVD& cKVD\\
			\midrule
			Color Transfer
			& 91.01  & 94.82  & 18.16 & 50.87	
			& 89.30  & 83.51  & 20.20 & 51.23
			& 125.90 & 140.60 & 32.58 & 56.52
			& 46.85  & 19.44  & 14.91 & 42.89 \\				
			MUNIT
			& 40.36 & 29.98 & 14.99 & 43.24
			& 47.96 & 30.35 & 14.14 & 59.62
			& 42.53 & 31.83 & 15.02 & 50.83
			& \textbf{44.74} & 17.48 & 11.65 & 48.10 \\
			CUT
			& 49.55	& 44.25	& 16.85 & \textbf{37.53}
			& 60.35	& 49.48	& 16.80 & 51.02
			& \textbf{34.36} & \textbf{20.54}& 10.16 & 53.55
			& 46.03	& 15.70	& 14.71 & 43.91 \\					
			TSIT
			& \textbf{38.70} & \textbf{28.70} & \textbf{10.80} & 42.35
			& \textbf{45.26} & \textbf{28.40} &\textbf{8.47} & 50.03
			& 54.96 & 33.21 & 12.71 & 57.91
			& 79.28 & 40.02 & 12.97 & 41.52 \\	
			QS-Attn
			& 49.41 & 42.87 & 14.01 & 38.57
			& 55.62 & 39.31 & 12.99 & 63.22
			& 46.67 & 21.47 &  \textbf{7.58} & 52.02	
			& 60.91 & 18.85 & 14.19 & 44.00 \\	
			\midrule
			FeaMGAN-S (ours) 
			& 45.16	& 34.93 & 13.87 & 40.50
			& 52.79	& 35.92 & 14.34 & \textbf{45.38}
			& 70.40	& 51.30 & 14.68 & \textbf{46.66}
			& 57.93	& 16.24 & 11.88 & \textbf{38.28} \\	
			FeaMGAN (ours)
			& 46.12 & 36.56 & 13.69 & 41.19
			& 51.56 & 34.63 & 14.01 & \textbf{47.21}
			& 66.39 & 46.96 & 13.14 & \textbf{46.88} 
			& 56.78 & \textbf{14.77} & \textbf{11.36}  & 41.72 \\
			\bottomrule	
	
		\end{tabular}
	\end{center}
	\label{tab:quantitative_comparison}
\end{table*}
As shown in the first row of \autoref{fig:qualitative_comparison_epe_baselines} and the results of Figures \ref{fig:qualitative_comparison_epe_additional} and \ref{fig:qualitative_comparison_epe_additional_random} of Appendix \ref{appendix}, our model translates larger structures, such as lane markings, more consistently, but fails to preserve some in-class characteristics from the source dataset. This is evident, for example, in the structure of translated streets and the corresponding cKVD value (road). As shown in the second row and Appendix \ref{appendix}, EPE achieves visually superior modeling of the reflective properties of materials (e.g., the car) but suffers from inconsistencies (erased objects) regarding the vegetation, which can be seen in the palm trees and the corresponding cKVD value (vegetation). The superior modeling of reflective properties can be attributed to the availability of gbuffers (i.a., glossines) in EPE's input. 

By surpassing EPE in all commonly used quantitative metrics while maintaining content consistency, we are able to show that our model improves overall quantitative translation performance. However, our method has specific drawbacks that we discussed with the help of the cKVD metric and visual comparisons. \\

\noindent \textbf{Comparisons to retrained baselines.}
We find that retraining the baselines with their original training setup for the PFD$\rightarrow$Cityscapes task significantly improves their performance on commonly used metrics compared to the baselines provided by EPE, as can be seen in \autoref{tab:quantitative_comparison}. However, as shown in \autoref{fig:qualitative_comparison} and the random results of \autoref{fig:qualitative_comparison_additional_random} of Appendix \ref{appendix}, content-consistency problems remain. This indicates again that simply relying on commonly used metrics does not provide a complete picture if content consistency is taken into account. When qualitatively comparing our model to the baselines for the PFD$\rightarrow$Cityscapes and Viper$\rightarrow$Cityscapes tasks in \autoref{fig:qualitative_comparison}, we observe that our method significantly reduces content inconsistencies. However, a limitation of our masking strategy are class boundary artifacts, which are particularly evident in the Day$\rightarrow$Night translation task (\autoref{fig:limitations}). Since masking allows our method to focus on specific classes, we achieve state-of-the-art performance for the Clear$\rightarrow$Snowy translation task.
\subsection{Ablation Study}
\begin{figure*}[h] 
	\renewcommand{\thesubfigure}{}
	\begin{center}
		{\includegraphics[width=0.198\textwidth]{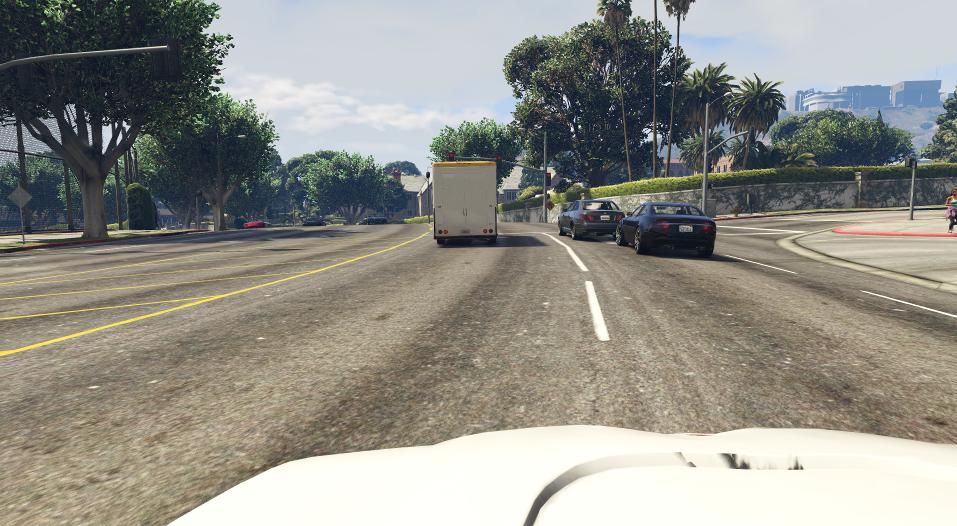}}\hfill
		{\includegraphics[width=0.198\textwidth]{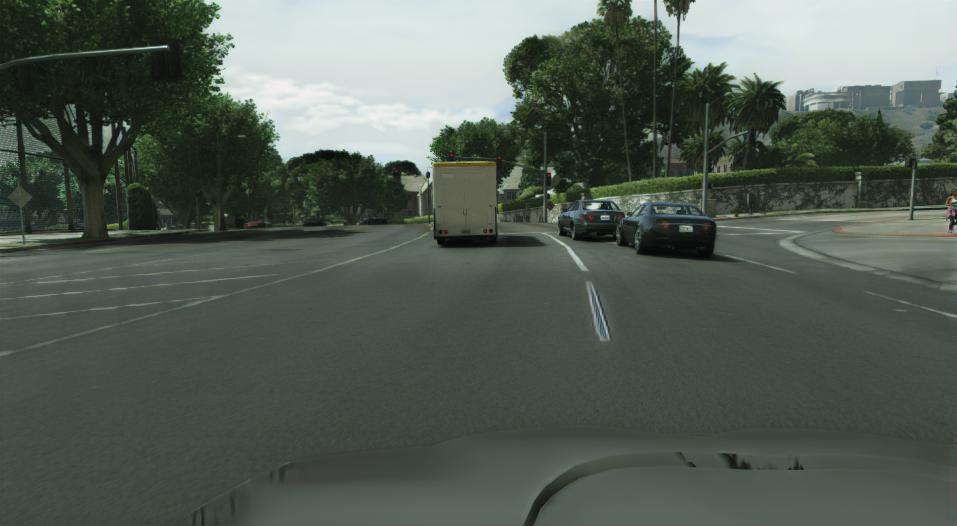}}\hfill
		{\includegraphics[width=0.198\textwidth]{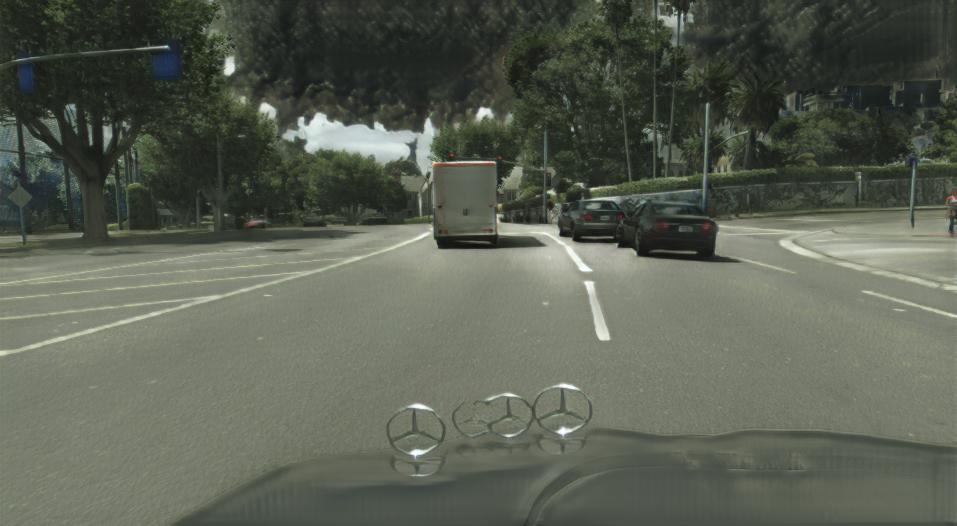}}\hfill
		{\includegraphics[width=0.198\textwidth]{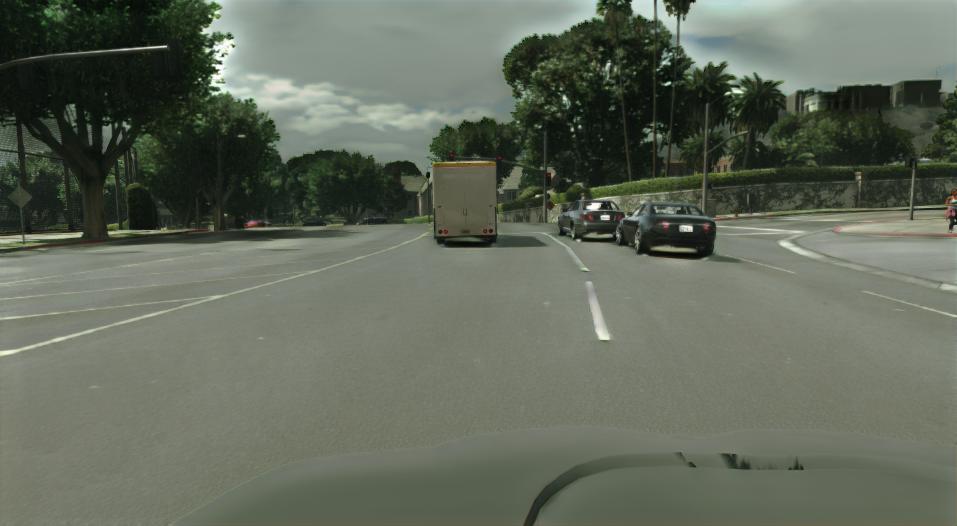}}\hfill
		{\includegraphics[width=0.198\textwidth]{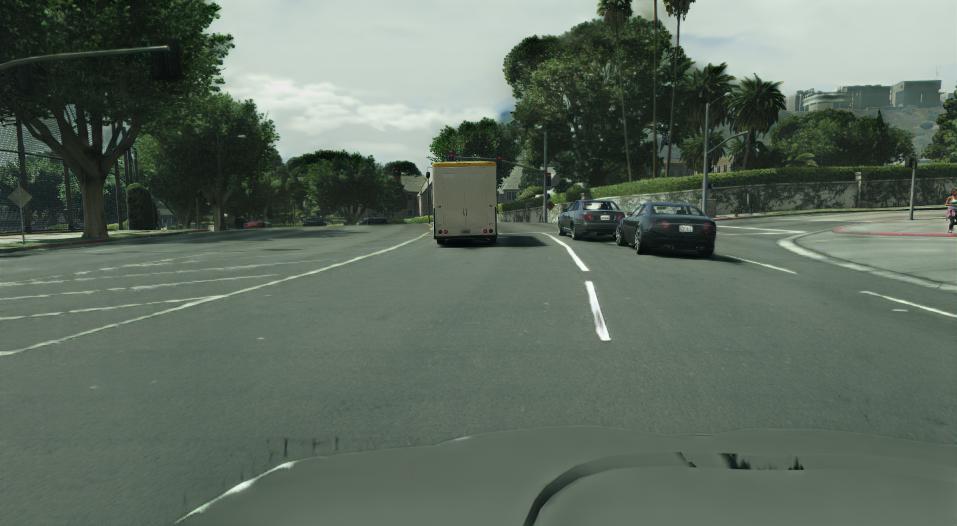}}\hfill \\ \vspace{1.33pt}
		{\includegraphics[width=0.198\textwidth]{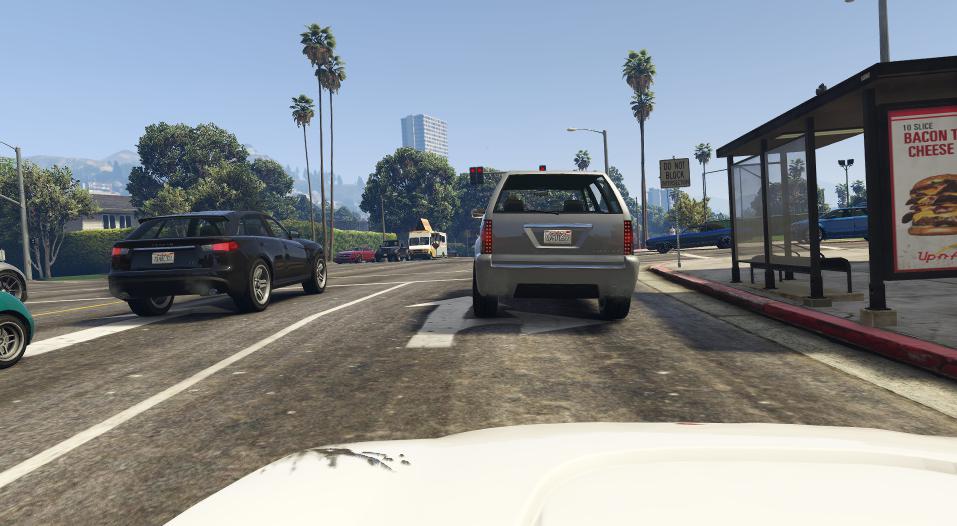}}\hfill
		{\includegraphics[width=0.198\textwidth]{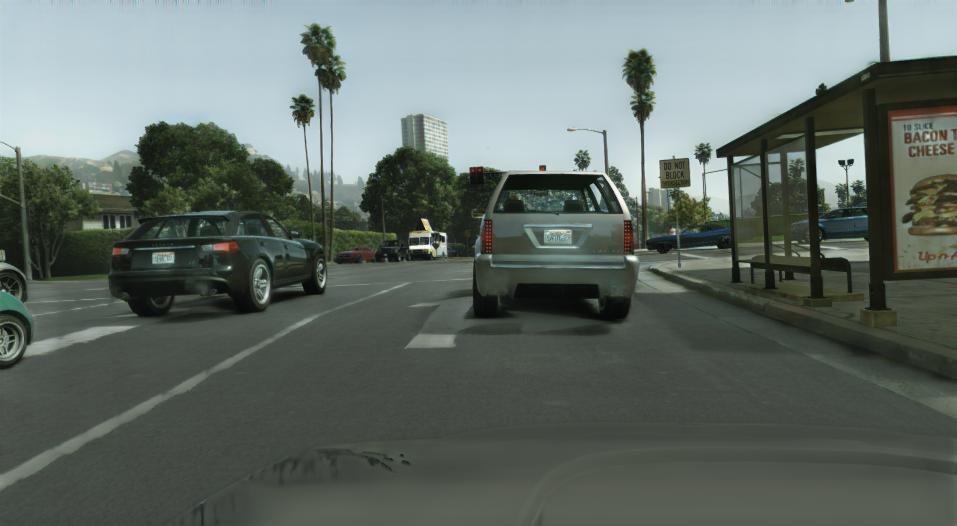}}\hfill
		{\includegraphics[width=0.198\textwidth]{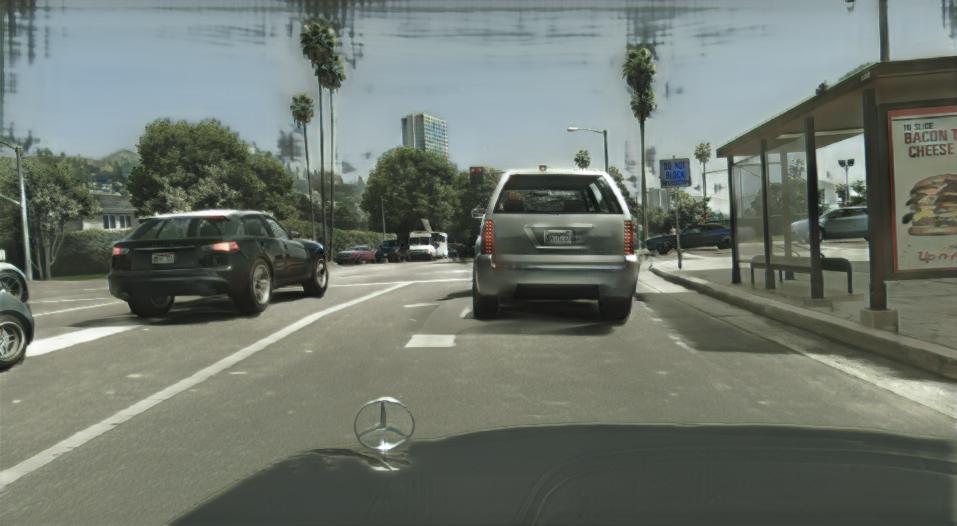}}\hfill
		{\includegraphics[width=0.198\textwidth]{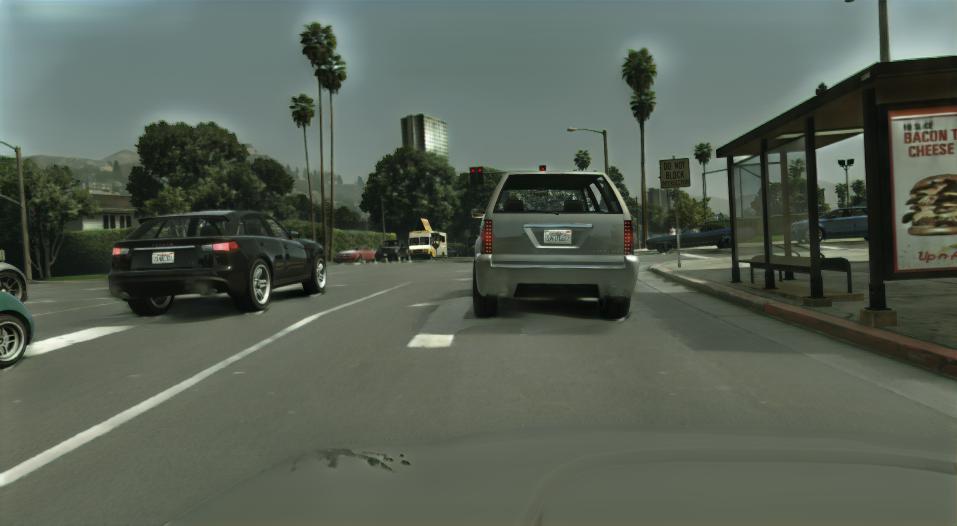}}\hfill
		{\includegraphics[width=0.198\textwidth]{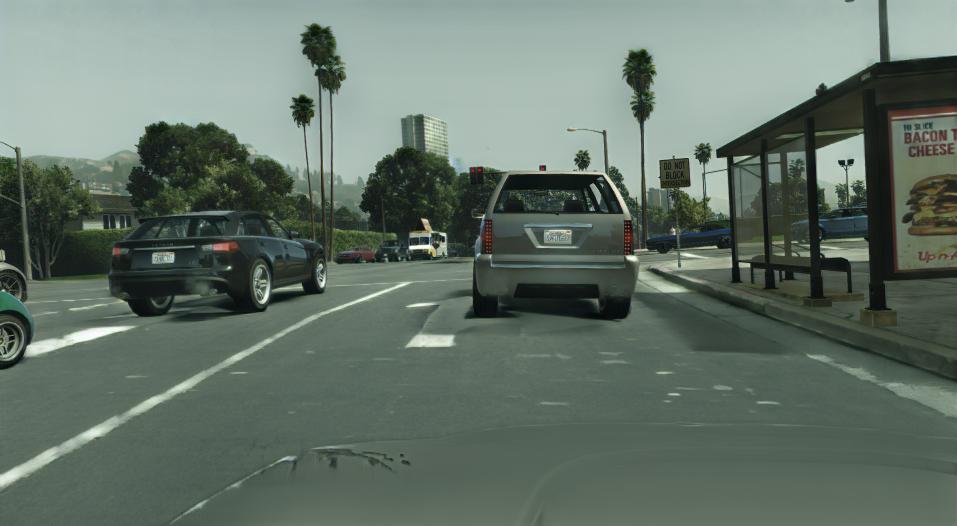}}\hfill \\ 
		\vspace{-3.66pt}
		\subfigure[Input]
		{\includegraphics[width=0.198\textwidth]{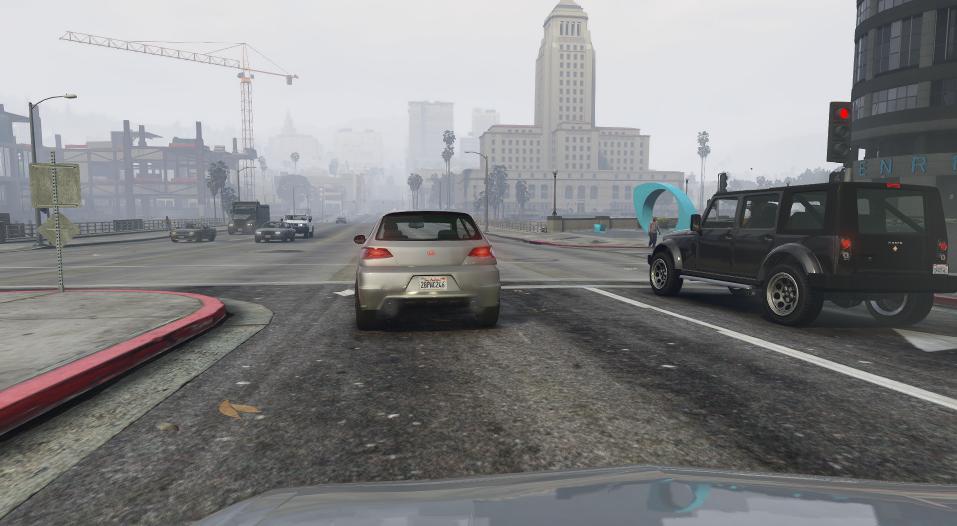}}\hfill
		\subfigure[Full]
		{\includegraphics[width=0.198\textwidth]{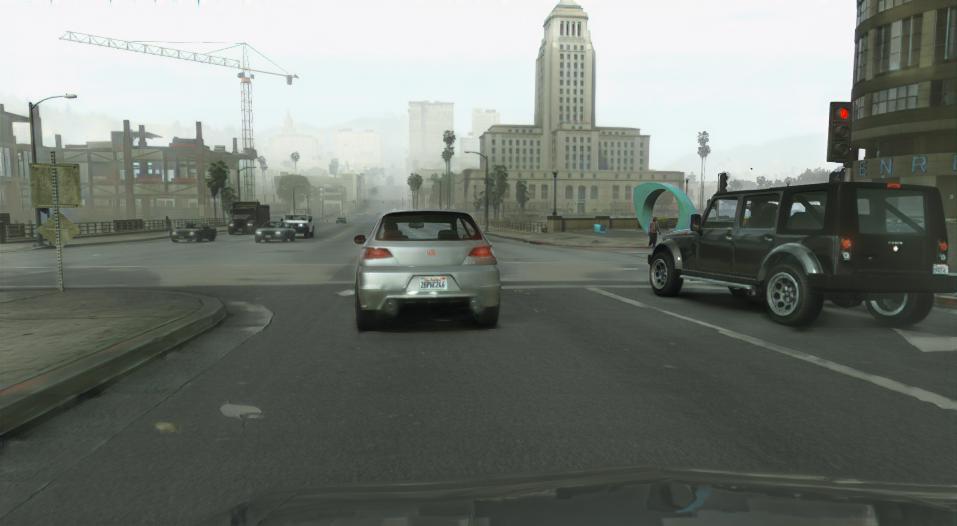}}\hfill
		\subfigure[w/o Dis. Mask]
		{\includegraphics[width=0.198\textwidth]{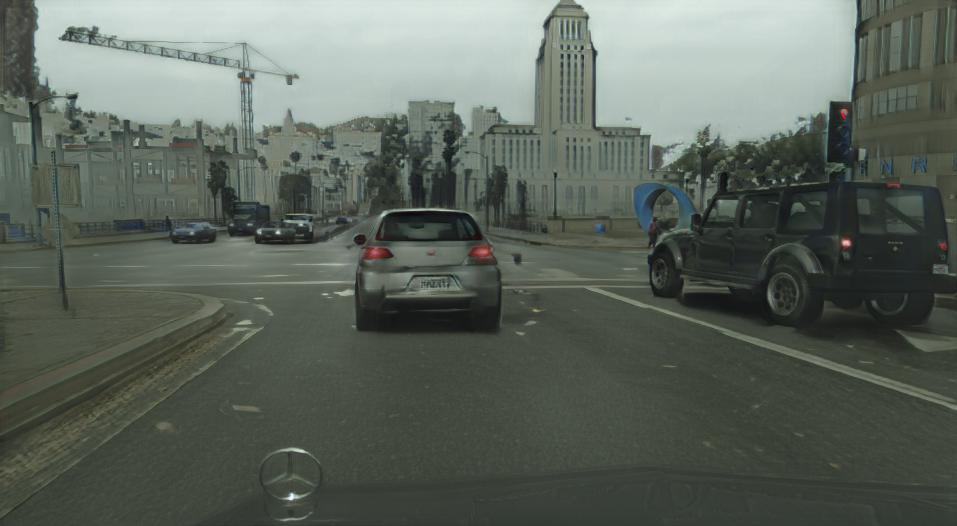}}\hfill
		\subfigure[w/o Local Dis.]
		{\includegraphics[width=0.198\textwidth]{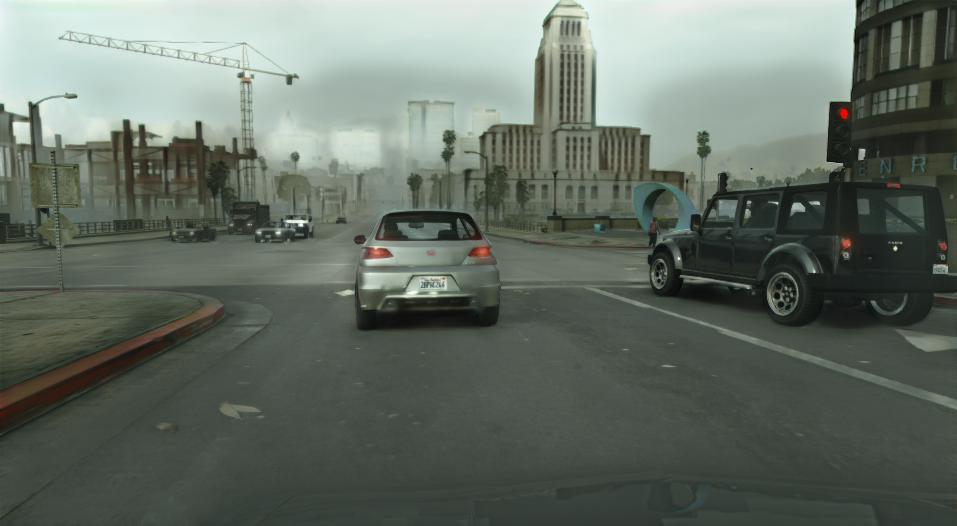}}\hfill
		\subfigure[w/ FADE w/o FATE ]
		{\includegraphics[width=0.198\textwidth]{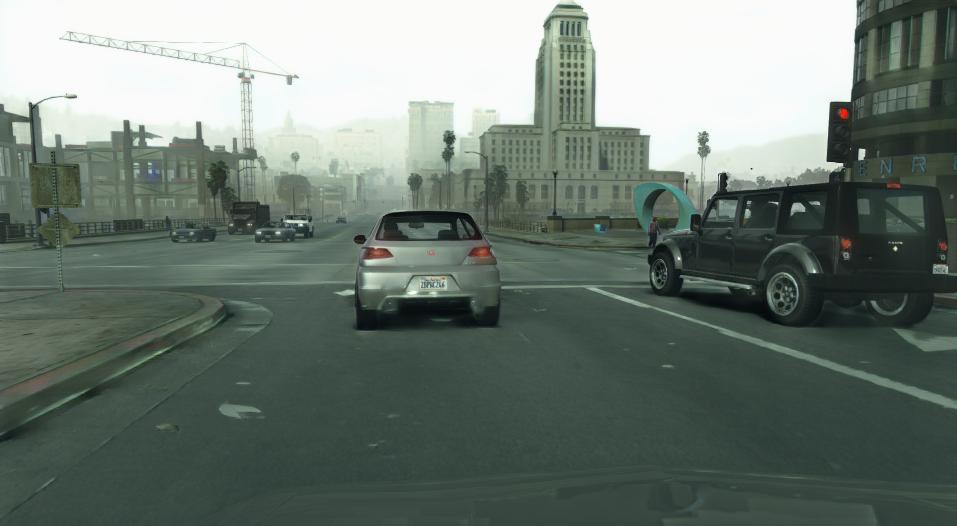}}\hfill 
	\end{center}
	\vspace{-8pt}
	\caption{\textbf{Qualitative ablations.} Results are selected from the best model. Randomly sampled results can be found in \autoref{fig:qualitative_ablations_additional_random} of Appendix \ref{appendix}.}
	\label{fig:qualitative_ablations}
\end{figure*}
\begin{table*}[h]
	\caption{\textbf{Quantitative evaluation for ablation study.} Results are reported as the average across five runs. We refer to \autoref{tab:quantitative_ablation_extended} of Appendix \ref{appendix} for an extended version of this table.}
	\setlength{\tabcolsep}{4.3pt}
	\begin{center}
		\begin{tabular}{lcccccccccccccc}
			\toprule
			\multirow{2}{*}{Method}&\multirow{2}{*}{FID}&\multirow{2}{*}{KID} &\multirow{2}{*}{sKVD}&\multicolumn{11}{c}{cKVD}\\
			\cmidrule(lr){5-15}
			&    &  &  & AVG& 		sky& 	ground&	road&	terrain&	vegetation&	building&	roadside-obj.&	person&	vehicle&	rest	\\
			\midrule	
			FeaMGan (Full)& 46.12 &	36.56 & 13.69 & 41.19 &42.69	&14.97	&17.35	&26.51	&\textbf{20.25}	&26.34	&64.64	&102.23	&\textbf{42.38}	&54.52 \\
			w/o Dis. Mask  & \textbf{37.10} & \textbf{25.88} & 14.73 &\textbf{39.65}	&\textbf{26.70}	&15.81	&16.65	&31.02	&22.97	&\textbf{25.39}	&67.01	&\textbf{93.78}	&44.23	&\textbf{52.91}
			\\
			w/ FADE w/o FATE & 45.46 & 35.73 & \textbf{13.17} &40.90	&41.49	&13.78	&16.78	&25.30	&20.58	&27.21	&\textbf{63.12}	&104.43	&42.44	&53.83\\
			w/ Random Crop & 47.88	&38.48&	13.37	& 40.18	&39.88	&\textbf{12.90}	&\textbf{14.65}	&\textbf{25.09}	&21.89	&27.32	&64.32	&98.81	&43.08	&53.86 \\
			w/ VGG Crop& 51.23&	42.46&	13.56 & 40.62	&40.32	&13.38	&15.67	&26.47	&21.09	&27.28	&65.23	&99.61	&43.19	&53.94 \\
			\midrule
			w/o Local Dis. &   &  &  &  &  &  &   &  &  &  &    &  &  &    \\
			- w/ 256$\times$256 Crop& 48.57 & 38.89 & \textbf{12.89} & 41.26 &	42.31 &	13.57	& 15.98 &	\textbf{25.28} &	22.18	& \textbf{26.56} &	\textbf{61.13} &	107.48 & 42.44&	55.62\\
			- w/ 352$\times$352 Crop& 47.26 & 37.75 & 14.38 & 39.30 &34.44	&\textbf{13.09}	&15.84& 25.83	&21.50	&27.20	&61.24	&98.25	&42.24	&53.38\\
			- w/ 464$\times$464 Crop& \textbf{46.61} & \textbf{37.25} & 15.04 &\textbf{38.62}	&\textbf{31.60}	&13.13	&\textbf{15.38}	&27.06	&22.23	&29.67	&63.38	&\textbf{87.51}	&44.41	&51.77\\
			- w/ 512$\times$512 Crop& 55.89	& 49.12 & 15.94 & 39.35	&36.48 &14.68 & 16.06	&26.87	&\textbf{19.61}	&27.37	&62.40	&98.90	&\textbf{40.32}	&\textbf{50.86}
			\\	
			\bottomrule
		\end{tabular}
	\end{center}
	\label{tab:quantitative_ablation}
\end{table*}
\begin{figure*}[h] 
	\renewcommand{\thesubfigure}{}
	\begin{center}
		{\includegraphics[width=0.198\textwidth]{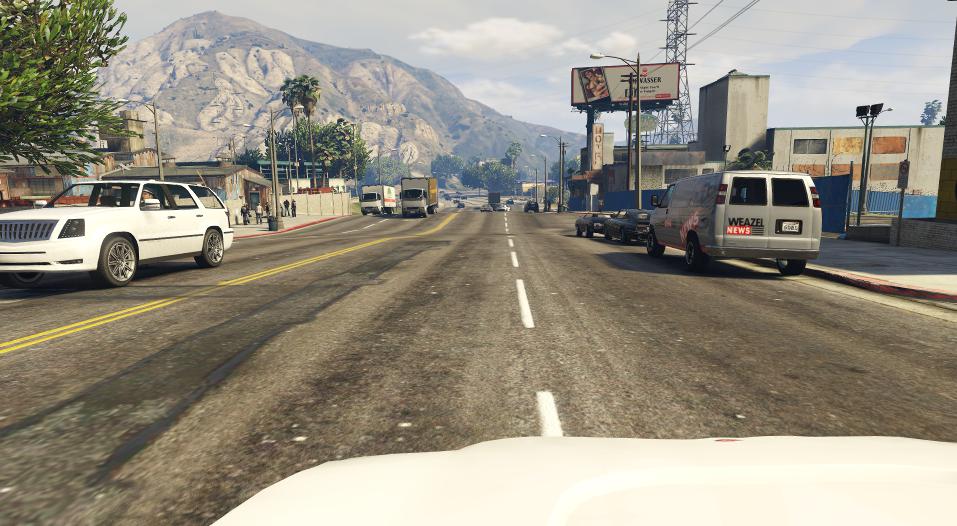}}\hfill
		{\includegraphics[width=0.198\textwidth]{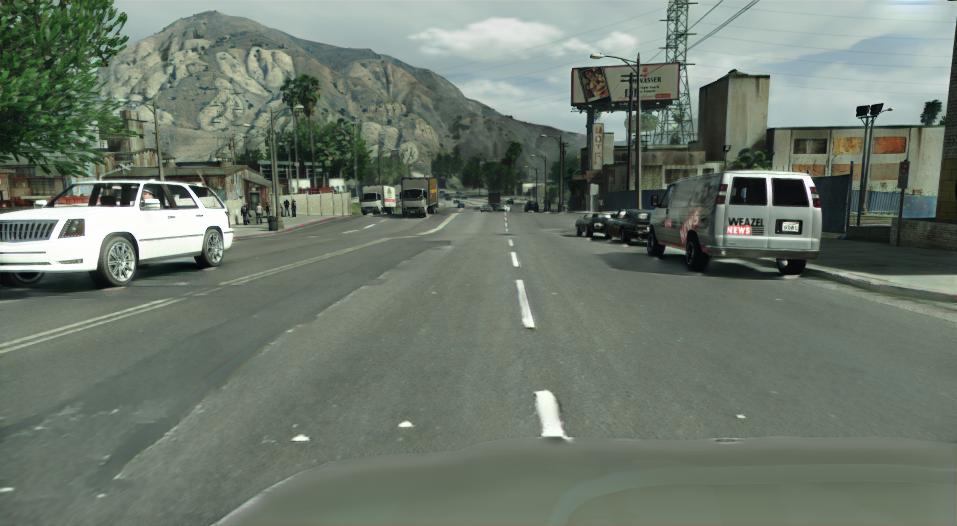}}\hfill
		{\includegraphics[width=0.198\textwidth]{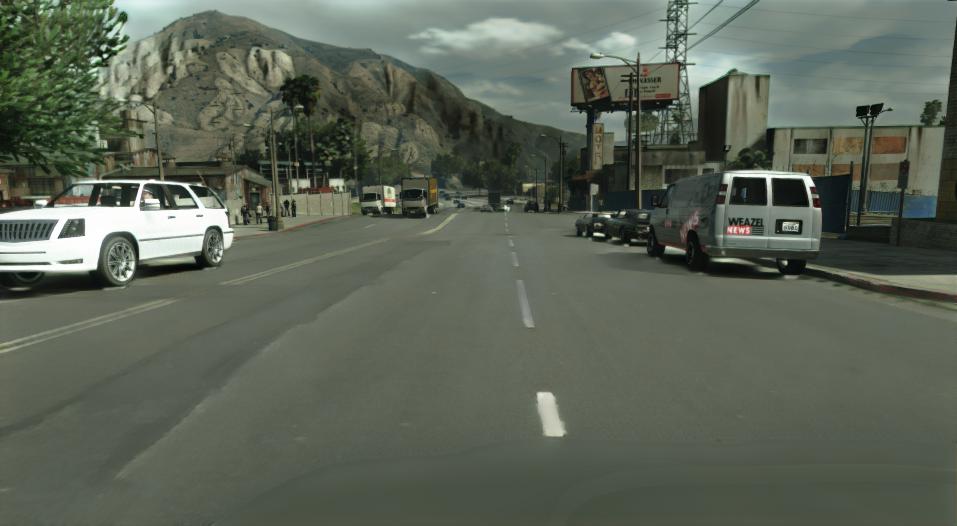}}\hfill
		{\includegraphics[width=0.198\textwidth]{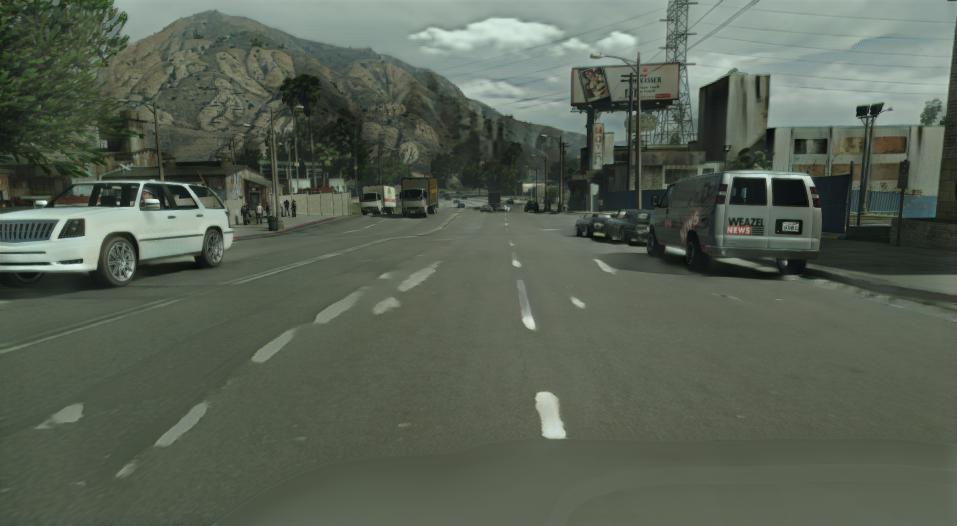}}\hfill
		{\includegraphics[width=0.198\textwidth]{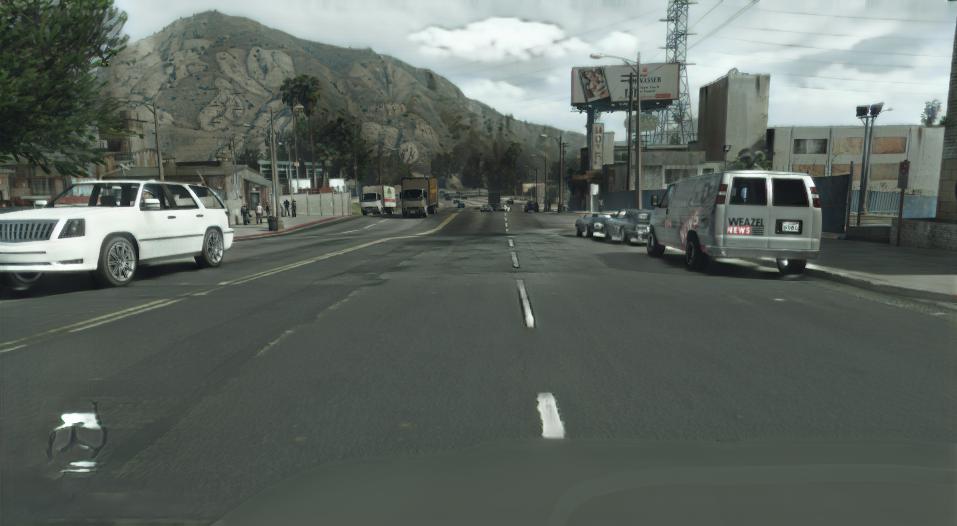}}\hfill \\ \vspace{1.33pt}
		{\includegraphics[width=0.198\textwidth]{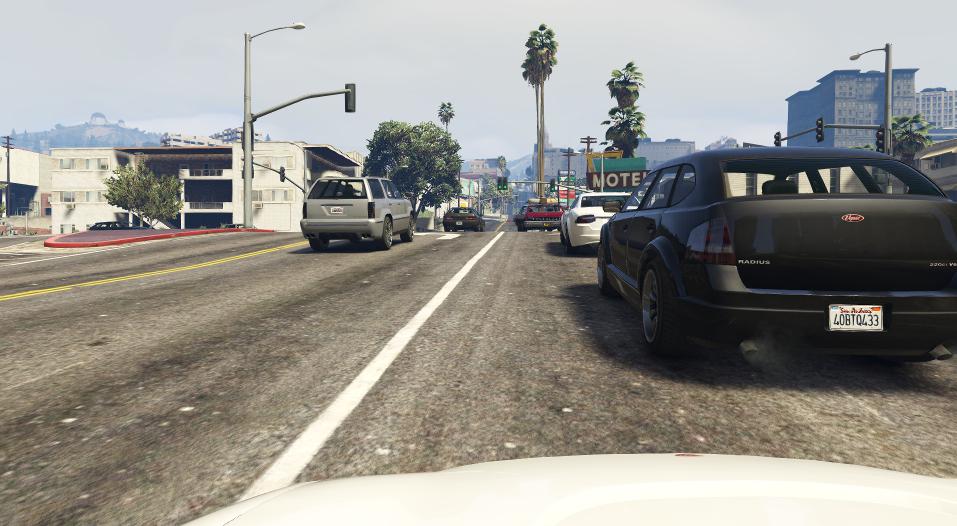}}\hfill 
		{\includegraphics[width=0.198\textwidth]{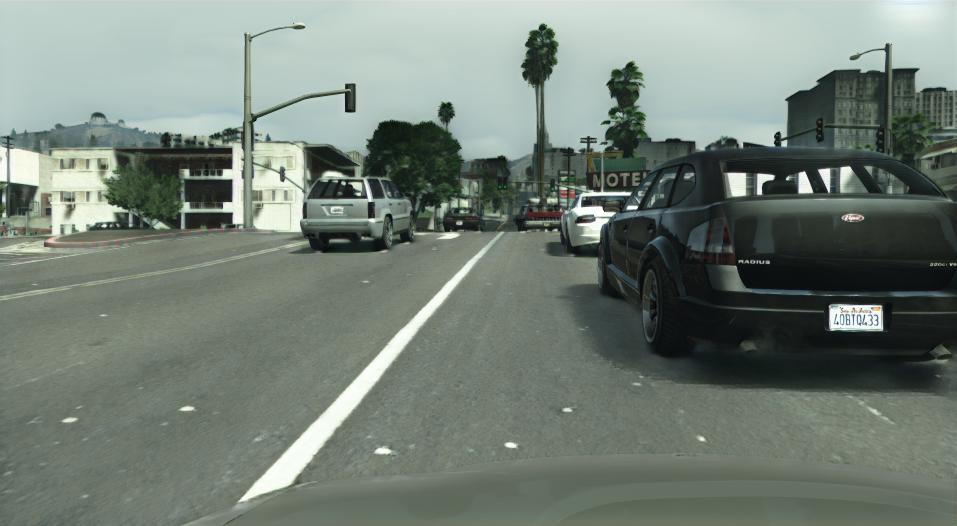}}\hfill
		{\includegraphics[width=0.198\textwidth]{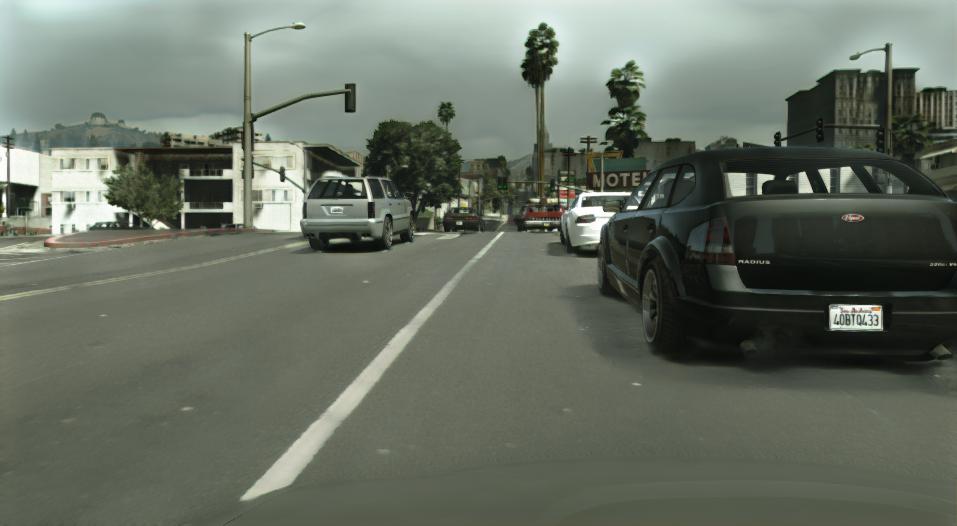}}\hfill
		{\includegraphics[width=0.198\textwidth]{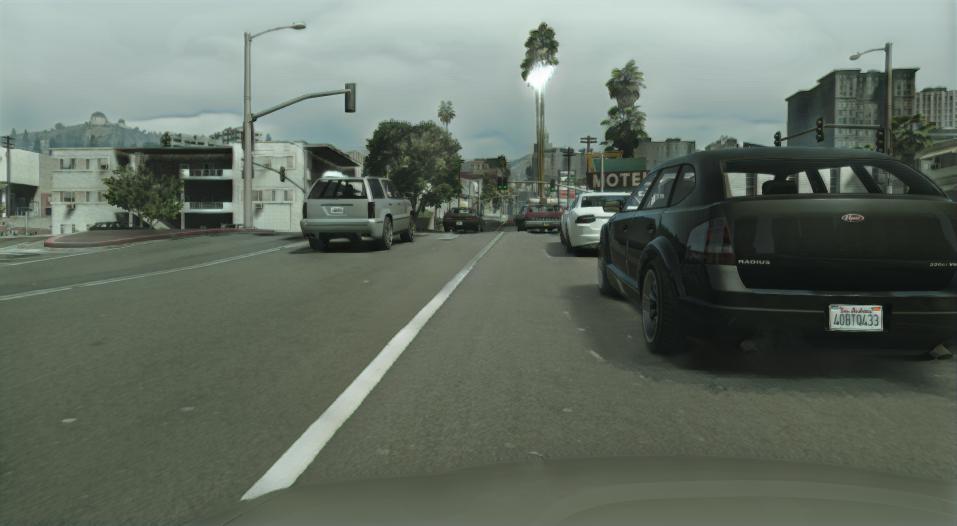}}\hfill
		{\includegraphics[width=0.198\textwidth]{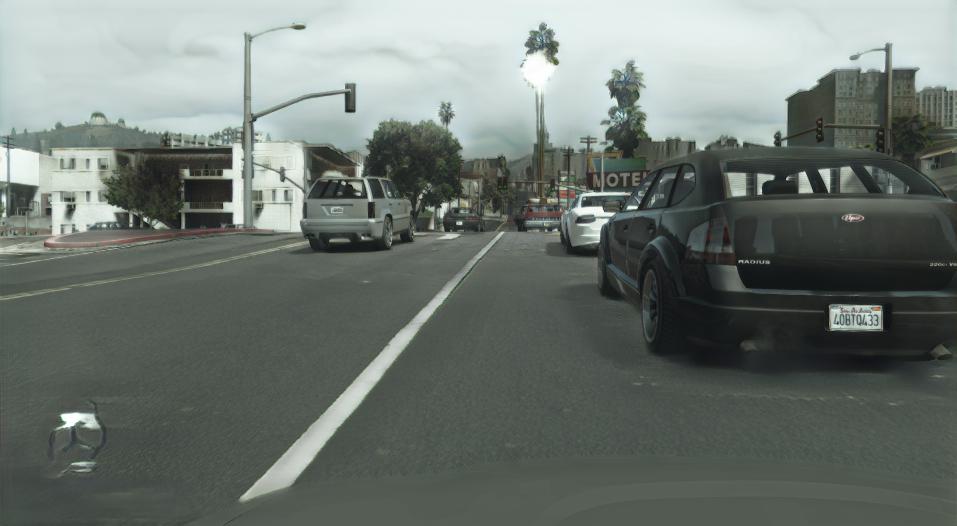}}\hfill \\ 
		\vspace{-3.66pt}
		\subfigure[Input]
		{\includegraphics[width=0.198\textwidth]{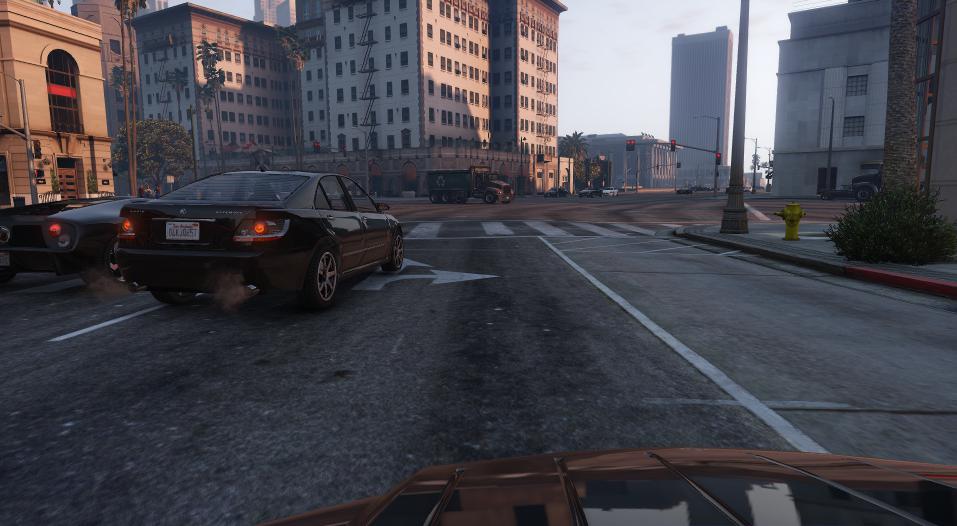}}\hfill
		\subfigure[252$\times$252]
		{\includegraphics[width=0.198\textwidth]{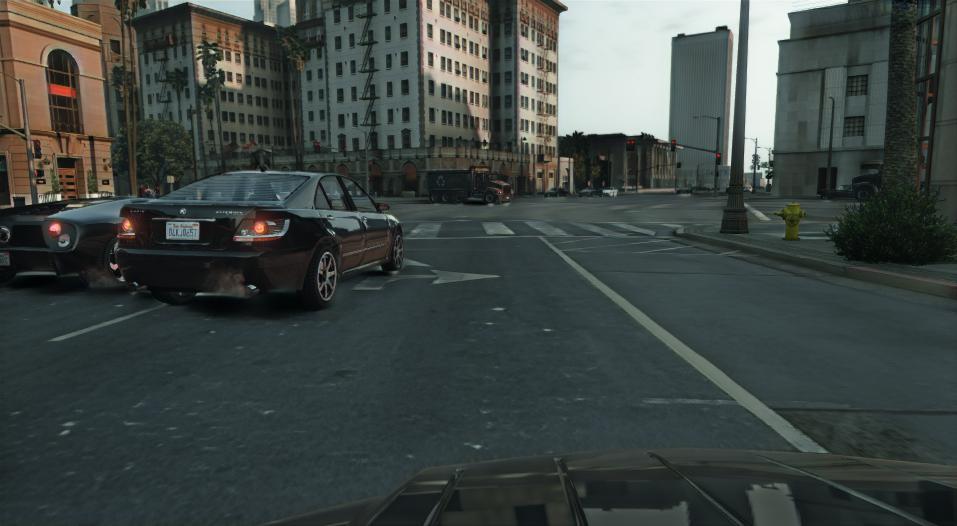}}\hfill
		\subfigure[352$\times$352]
		{\includegraphics[width=0.198\textwidth]{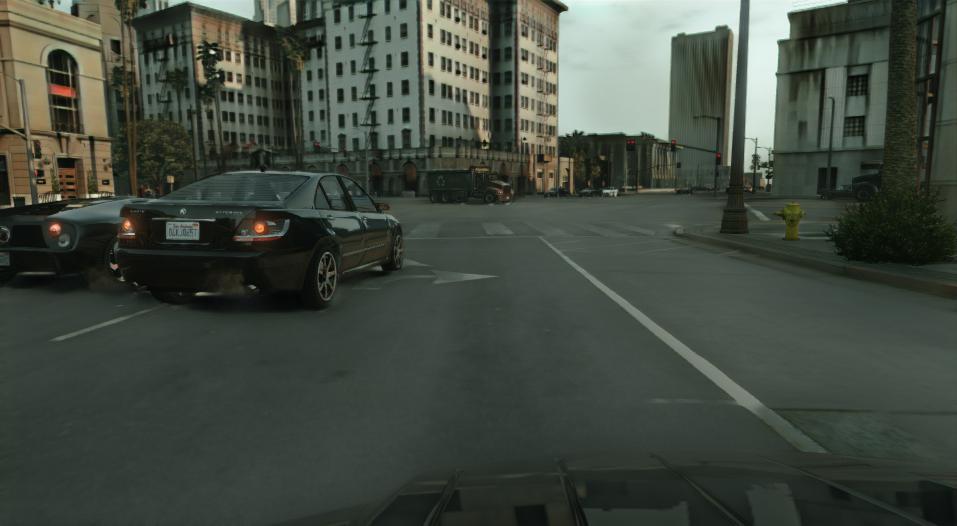}}\hfill
		\subfigure[464$\times$464]
		{\includegraphics[width=0.198\textwidth]{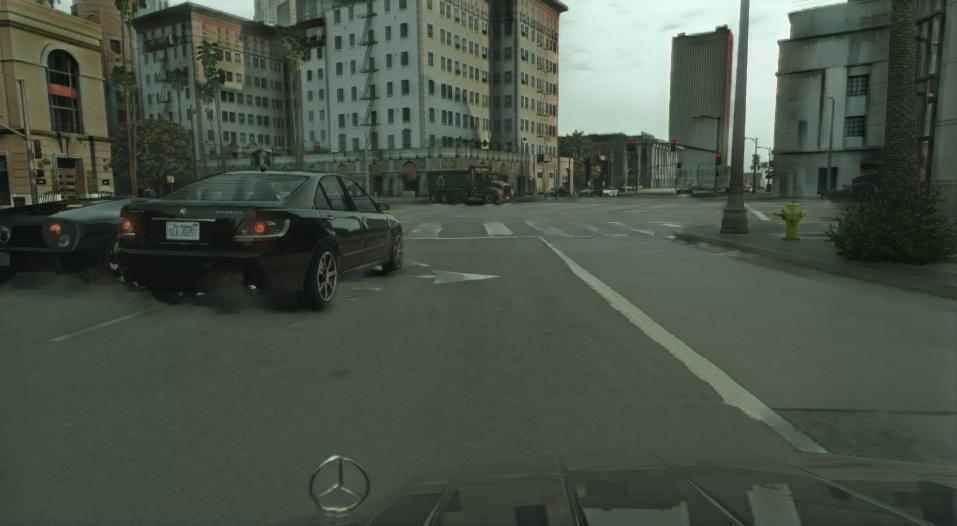}}\hfill
		\subfigure[512$\times$512]
		{\includegraphics[width=0.198\textwidth]{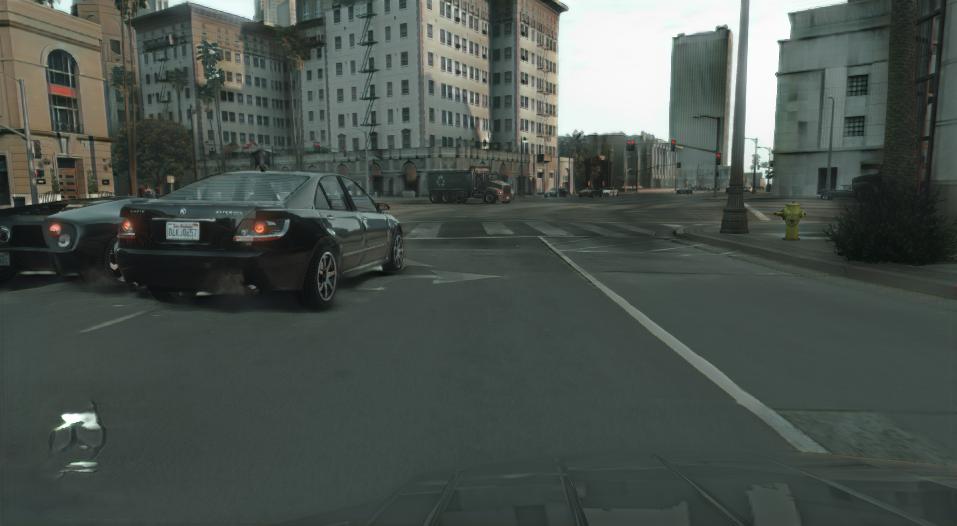}}\hfill 
	\end{center}
	\vspace{-8pt}
	\caption{\textbf{Qualitative ablation of crop sizes.} For each crop size, results are selected from the best model. Randomly sampled results can be found in \autoref{fig:qualitative_ablation_crop_size_additional_random} of Appendix \ref{appendix}.}
	\label{fig:qualitative_ablation_crop_size}
\end{figure*}
\begin{figure*}[h]
	\centering 
	\includegraphics[width=\linewidth]{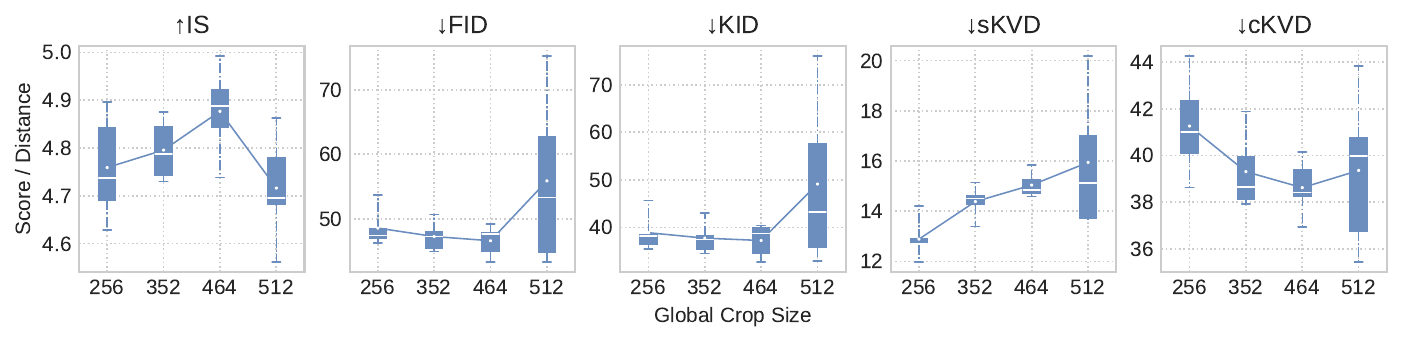} 
	\vspace{-16pt}
	\caption{\textbf{Quantitative ablation of crop sizes.}}
	\label{fig:quantitative_ablation_crop_size}
\end{figure*} 
\noindent \textbf{Effectiveness of masked discriminator.}
As shown in \autoref{fig:qualitative_ablations} and the random samples in \autoref{fig:qualitative_ablations_additional_random} of Appendix \ref{appendix}, our masking strategy for the discriminator positively impacts content consistency. Without masking, inconsistencies occur that correlate with biases between the class distributions of the source and target domains. As shown in \cite{richter2022enhancing}, the distributions of certain classes in the spatial image dimension vary greatly between the PFD dataset and the Cityscapes dataset. For example, trees in Cityscapes appear more frequently in the top half of the image, resulting in hallucinated trees when the images are translated without accounting for biases. In the first and second row of \autoref{fig:qualitative_ablations}, we show that our masking strategy (Full) prevents these inconsistencies in contrast to our model trained without masking (w/o Dis. Mask). However, as shown in \autoref{tab:quantitative_ablation}, this comes with a quantitative tradeoff in performance on commonly used metrics.\\

\noindent \textbf{Effectiveness of local discriminator.}
We compare our model trained with a local discriminator (Full) to the model trained without a local discriminator (w/o Local Dis. 352x352). As shown in \autoref{fig:qualitative_ablations}, the local discriminator leads to an increase in quantitative performance. Furthermore, we show the qualitative effects of the local discriminator in \autoref{fig:qualitative_ablations}, where we observe a decrease in glowing objects and a significant decrease of erased objects in the translation. An example of a glowing object is the palm tree in row two of \autoref{fig:qualitative_ablations}. An example of erased objects are the missing houses in the background of the images from row three. In addition, small inconsistencies near object boundaries are reduced, as shown by the randomly sampled results in \autoref{fig:qualitative_ablations_additional_random} of Appendix \ref{appendix} (e.g., the wheels of the car in row one and three). Overall, we can conclude that local discriminators can reduce local inconsistencies, which might arise from the robust but not flawless segmentation maps used for masking. \\

\noindent \textbf{Effectiveness of segmentation-based sampling.}
We compare our segmentation-based sampling method with random sampling and sampling based on VGG features. For the sampling strategy based on VGG features, we follow EPE \cite{richter2022enhancing} to calculate scores for 352$\times$352 crops of the input images. Crops with a similarity score higher than $0.5$ are selected for training. As shown in \autoref{tab:quantitative_ablation}, our segmentation-based sampling strategy (Full) slightly outperforms the other sampling strategies in overall translation performance.\\

\noindent \textbf{Effectiveness of FATE.}
\begin{figure*}[h] 
	\renewcommand{\thesubfigure}{}
	\begin{center}
		{\includegraphics[width=0.164\textwidth]{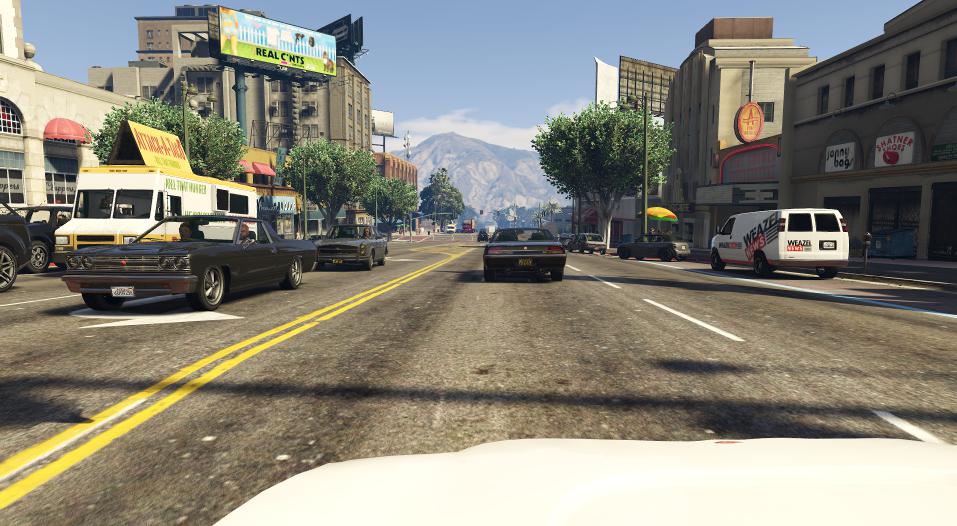}}\hfill
		{\includegraphics[width=0.164\textwidth]{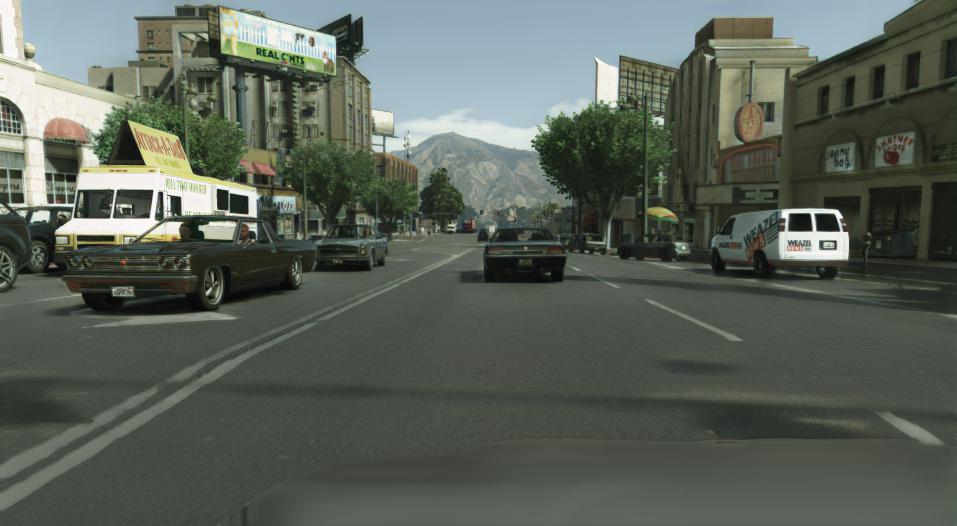}}\hfill
		{\includegraphics[width=0.164\textwidth]{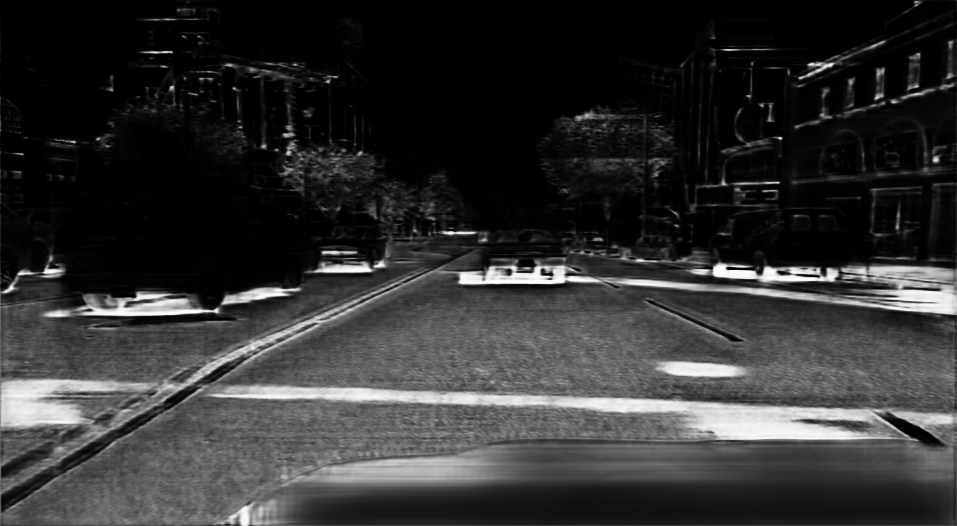}}\hfill
		{\includegraphics[width=0.164\textwidth]{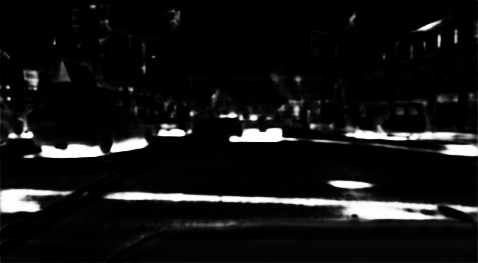}}\hfill
		{\includegraphics[width=0.164\textwidth]{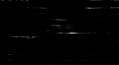}}\hfill
		{\includegraphics[width=0.164\textwidth]{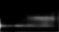}}\hfill \\ \vspace{1.33pt}
		{\includegraphics[width=0.164\textwidth]{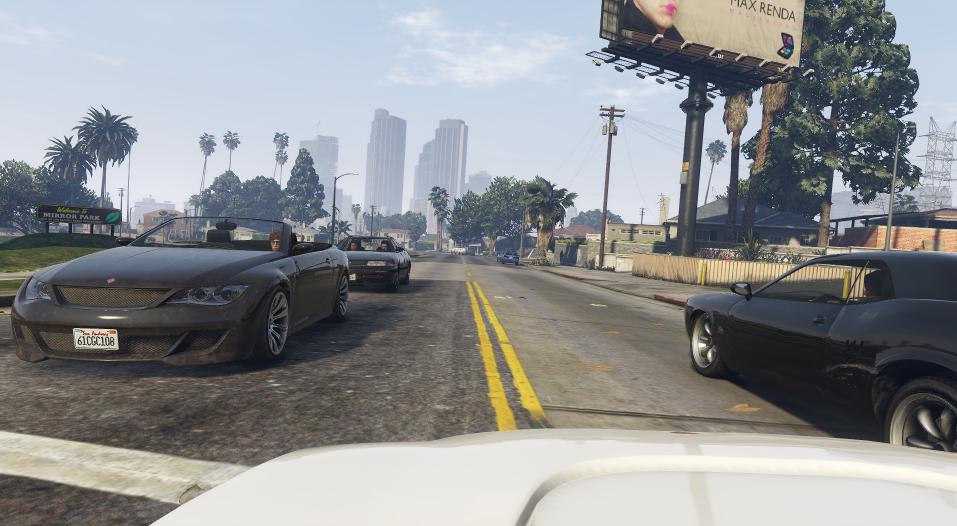}}\hfill
		{\includegraphics[width=0.164\textwidth]{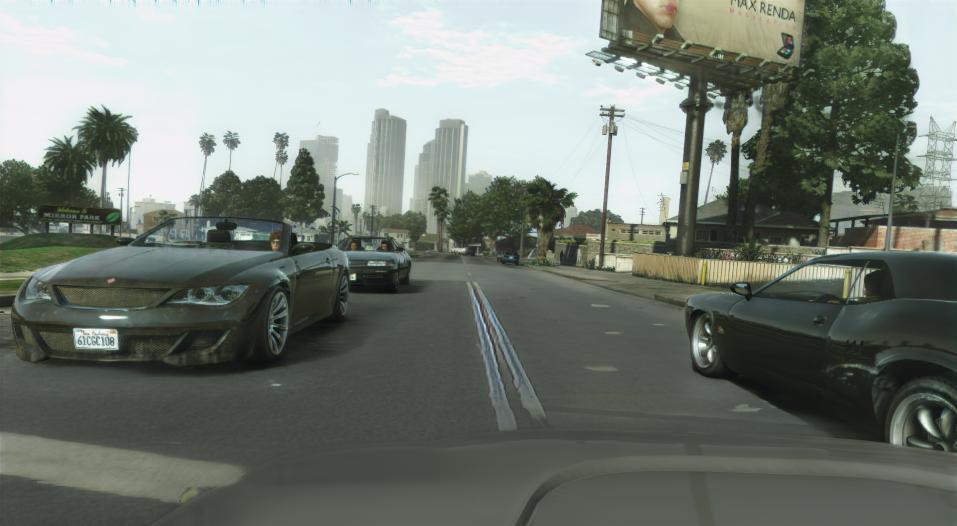}}\hfill
		{\includegraphics[width=0.164\textwidth]{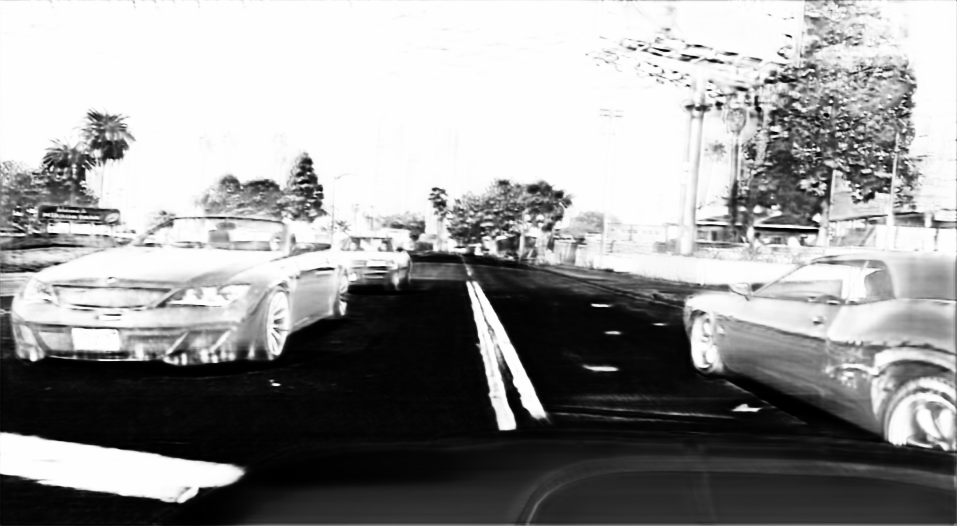}}\hfill
		{\includegraphics[width=0.164\textwidth]{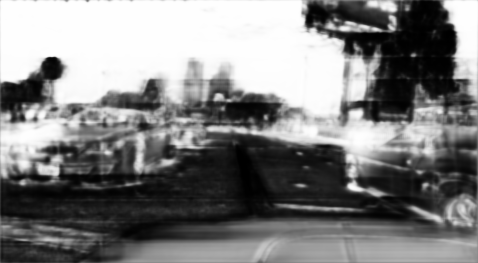}}\hfill
		{\includegraphics[width=0.164\textwidth]{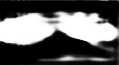}}\hfill
		{\includegraphics[width=0.164\textwidth]{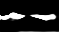}}\hfill \\ 
		\vspace{-3.66pt}
		\subfigure[957$\times$526 Input]
		{\includegraphics[width=0.164\textwidth]{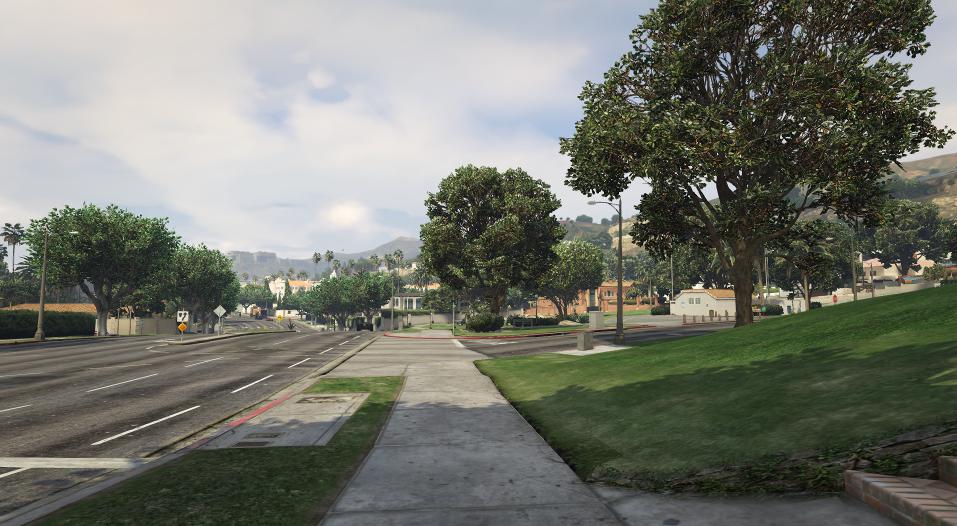}}\hfill
		\subfigure[957$\times$526 Output]
		{\includegraphics[width=0.164\textwidth]{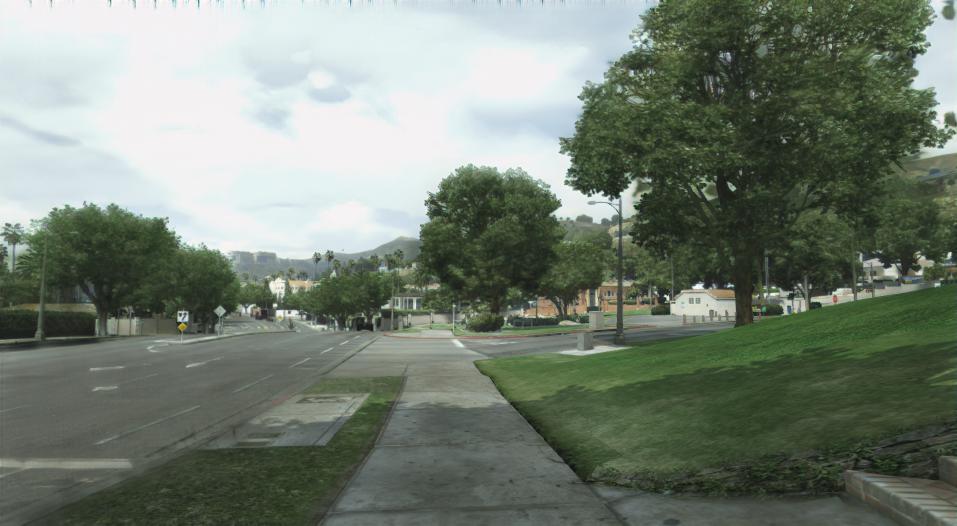}}\hfill
		\subfigure[957$\times$526 Layer]
		{\includegraphics[width=0.164\textwidth]{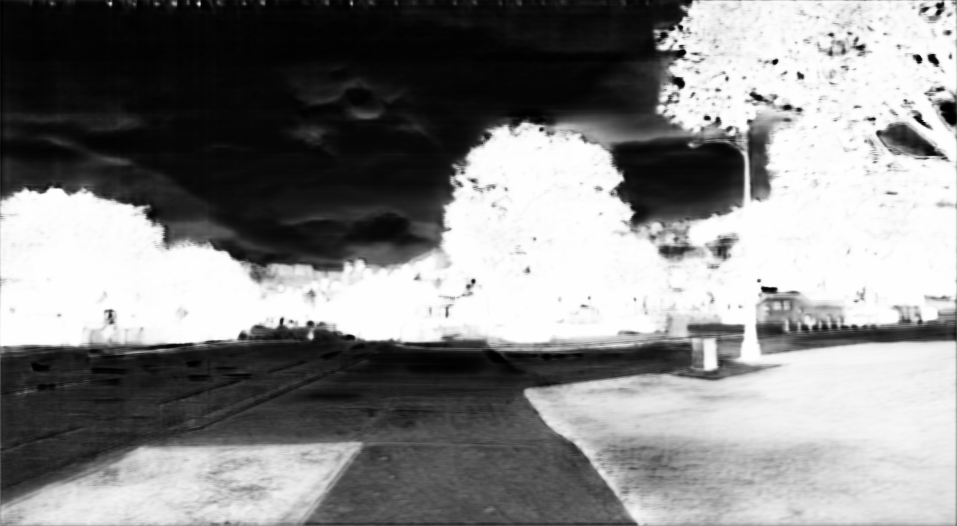}}\hfill
		\subfigure[478$\times$263 Layer]
		{\includegraphics[width=0.164\textwidth]{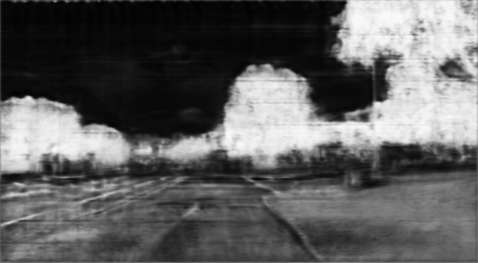}}\hfill
		\subfigure[119$\times$65 Layer]
		{\includegraphics[width=0.164\textwidth]{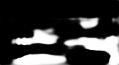}}\hfill
		\subfigure[59$\times$32 Layer]
		{\includegraphics[width=0.164\textwidth]{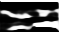}}\hfill 
	\end{center}
	\vspace{-8pt}
	\caption{\textbf{FATE attention maps.} Results are selected from the best model.}
	\label{fig:FATE_attention_maps}
\end{figure*}
For each spatial point ("pixel") in the input feature map, our feature-attentive denormalization block selects the features in the feature dimension to be incorporated into the output stream of the generator by denormalization. We show the attention values of our feature-attentive denormalization block in \autoref{fig:FATE_attention_maps} by visualizing all attention values for a single feature across the entire feature map. Since a single feature represents a property of the input, a spatial pattern should emerge. This is expected especially in earlier layers, where the spatiality of the convolutional model's feature map is best preserved. As shown in \autoref{fig:FATE_attention_maps}, our attention mechanism learns to attend to features that correlate with a property. Examples are the shadows of a scene (row 1), cars and their lighting (row 2), and vegetation (row 3). In addition, we find increasingly more white feature maps in deeper layers. This can be interpreted positively as an indication that the learned content (source) features in deeper layers are important for the translation task and that more shallow content features of earlier layers are increasingly ignored. However, this can also be interpreted negatively and could indicate that our simple attention mechanism is not able to separate deeper features properly.

Comparing FATE to FADE, we find that FATE leads to a subtle increase in training instability, resulting in slightly worse average performance over the five runs per model. However, FATE also leads to our best models. Therefore, we select the FATE block as the standard configuration for our model. The deviation from the average values for all runs can be found in \autoref{tab:quantitative_ablation_extended} of Appendix \ref{appendix}. The slight increased instability suggests that the attention mechanism of FATE can be further improved. \\

\noindent \textbf{Effect of global crop size.}
We successively increase the global crop size of the generator and discriminators from 256$\times$256 to 512$\times$512 and examine the effects on translation performance. As shown in \autoref{fig:qualitative_ablations}, increasing the global crop size results in a better approximation of the target domain style. However, increasing the global crop size also leads to an increasing number of artifacts in the translated image. In \autoref{fig:quantitative_ablation_crop_size}, we report the score of various metrics with respect to the global crop size. The commonly used metrics for measuring translation quality (IS, FID, and KID) show that translation quality increases steadily up to a global crop size of 464$\times$464, after which the results become unstable. The cKVD metric also shows an increase in average performance up to a crop size of 464$\times$464, mainly because translation quality for the underrepresented person class increases. This is intuitive since a larger crop size leads to a more frequent appearance of underrepresented classes during training. Furthermore, the sKVD metric shows a steady decline in consistency as the global crop size increases. Therefore, we choose a tradeoff between approximation of the target domain style, artifacts, and computational cost, and select 352$\times$352 as the global crop size for our model.\\

\begin{figure*}[h] 
	\renewcommand{\thesubfigure}{}
	\begin{center}
		{\includegraphics[width=0.249\textwidth]{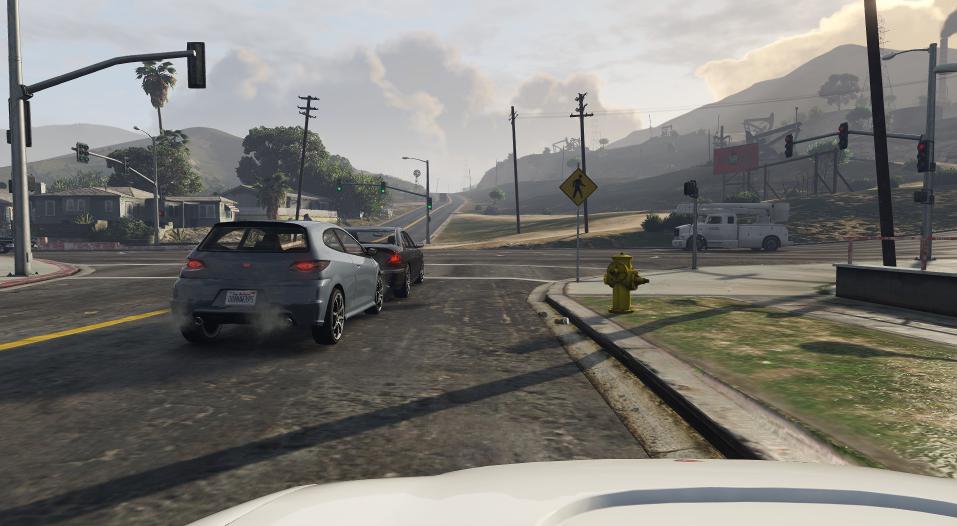}{\includegraphics[width=0.249\textwidth]{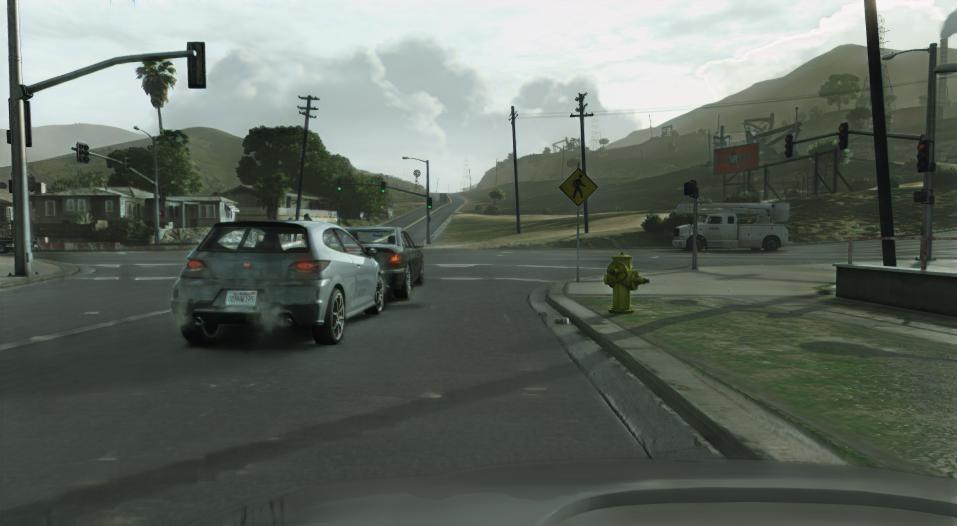}}}\hfill
		{\includegraphics[width=0.249\textwidth]{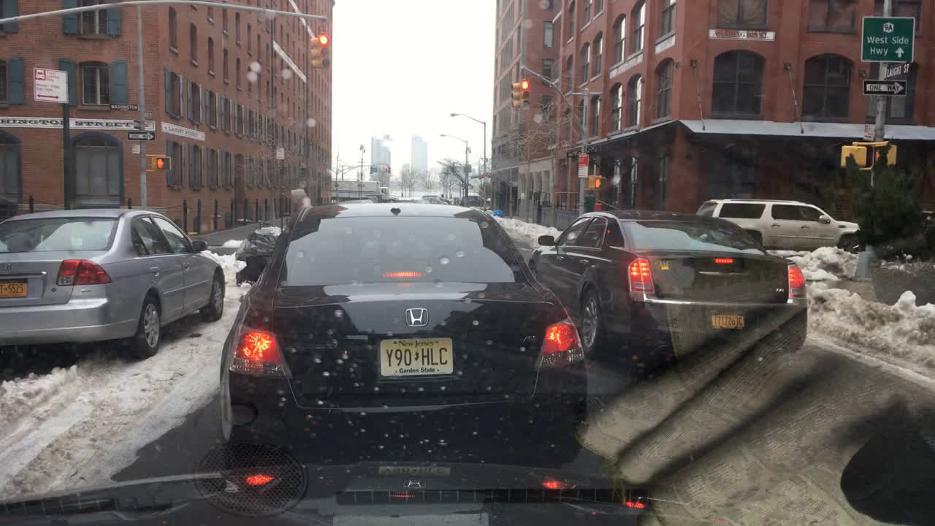}{\includegraphics[width=0.249\textwidth]{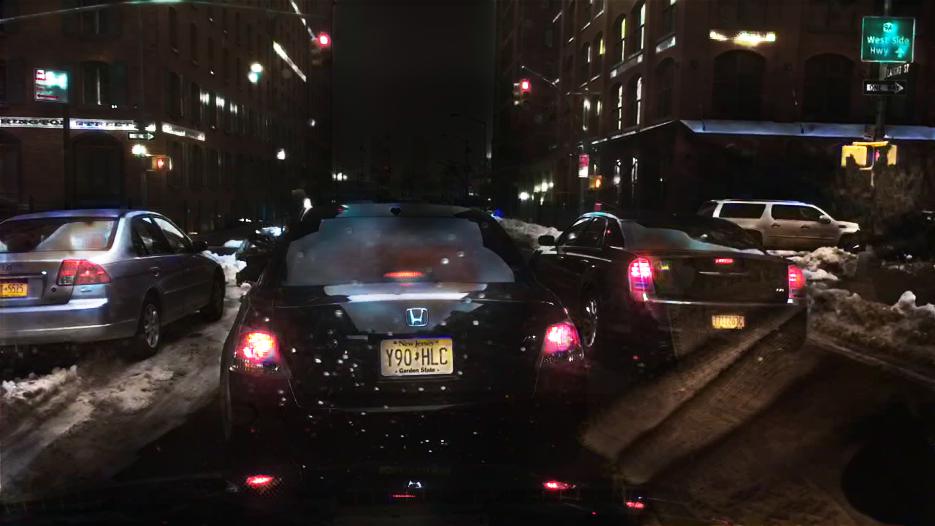}}}\hfill\\  \vspace{2pt}
		{\includegraphics[width=0.249\textwidth]{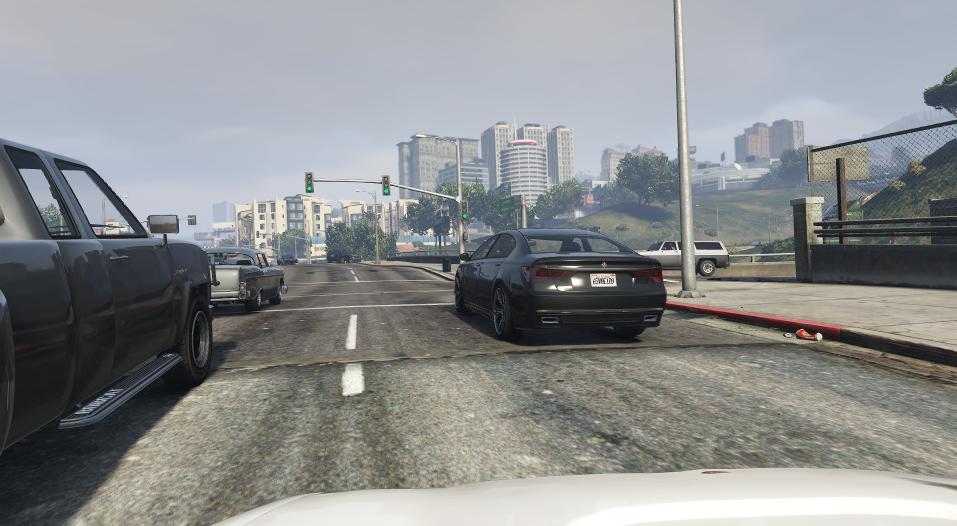}{\includegraphics[width=0.249\textwidth]{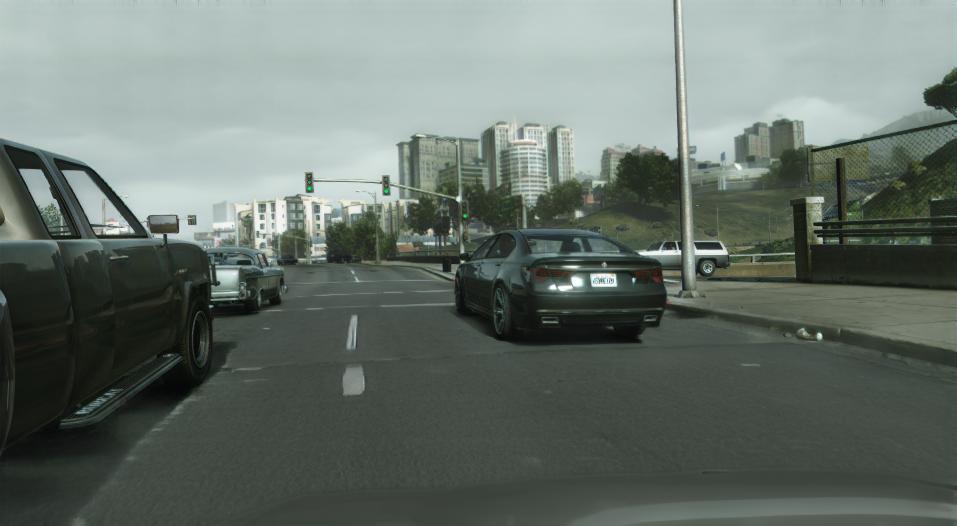}}}\hfill
		{\includegraphics[width=0.249\textwidth]{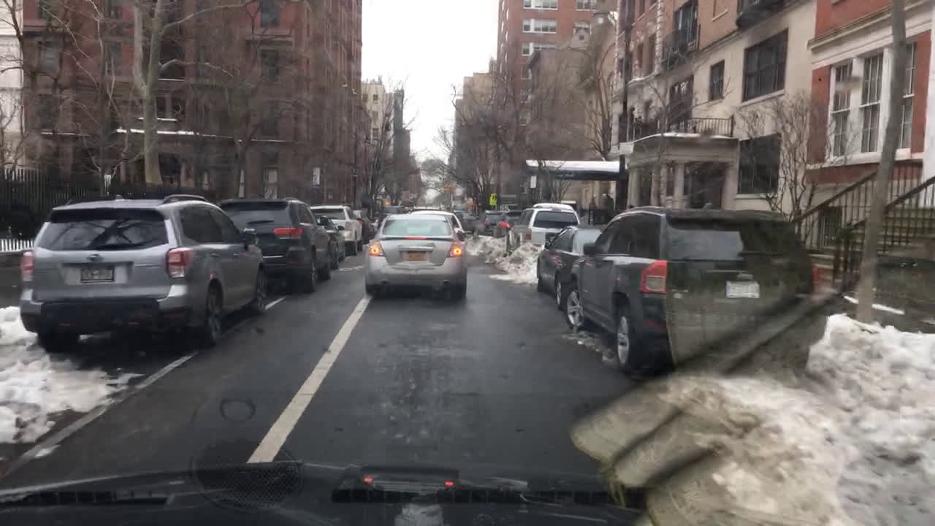}{\includegraphics[width=0.249\textwidth]{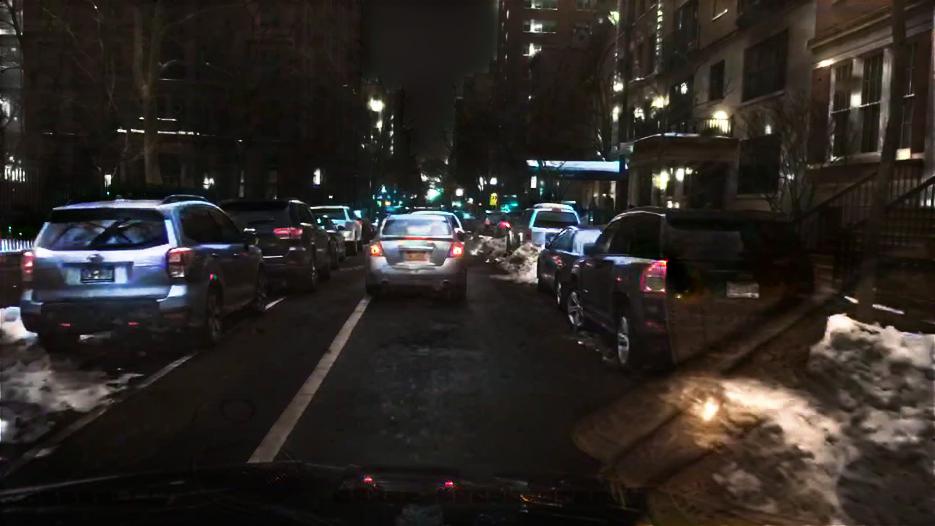}}}\hfill \\  \vspace{2pt}
		{\includegraphics[width=0.249\textwidth]{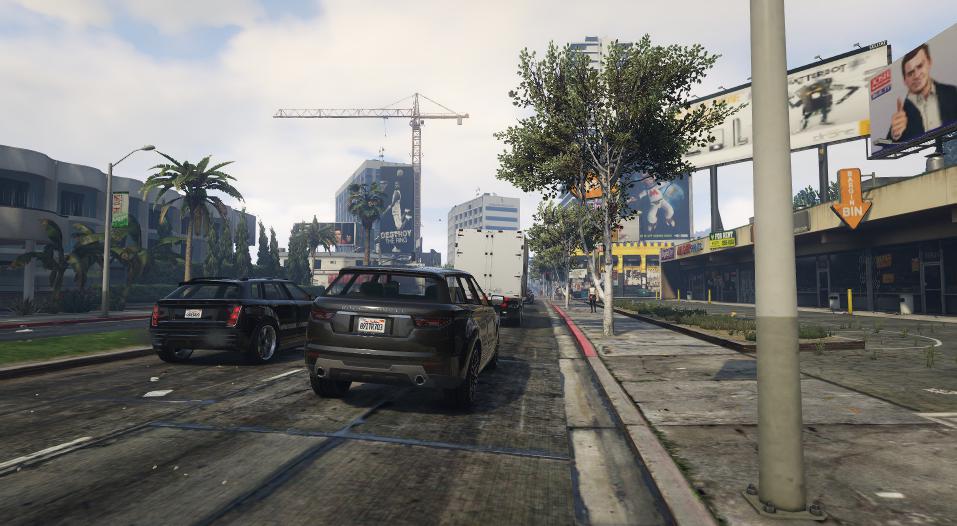}{\includegraphics[width=0.249\textwidth]{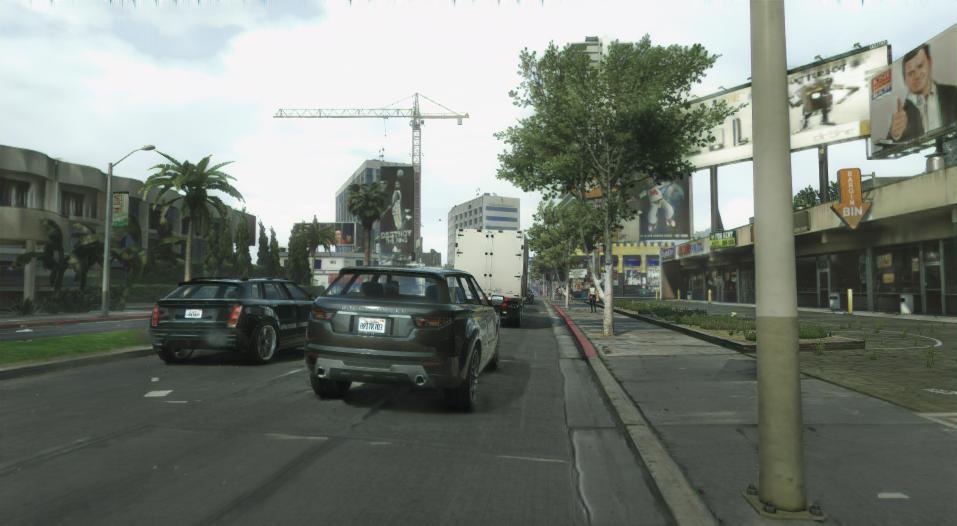}}}\hfill
		{\includegraphics[width=0.249\textwidth]{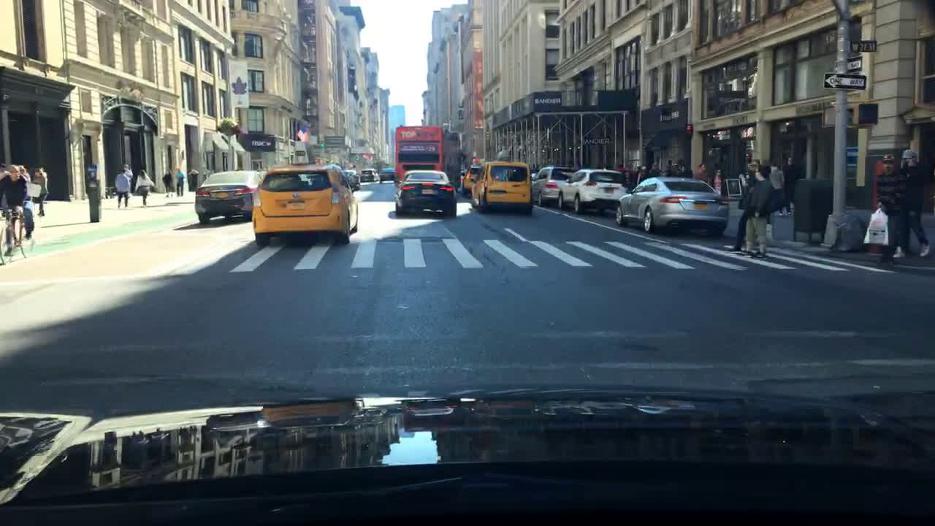}{\includegraphics[width=0.249\textwidth]{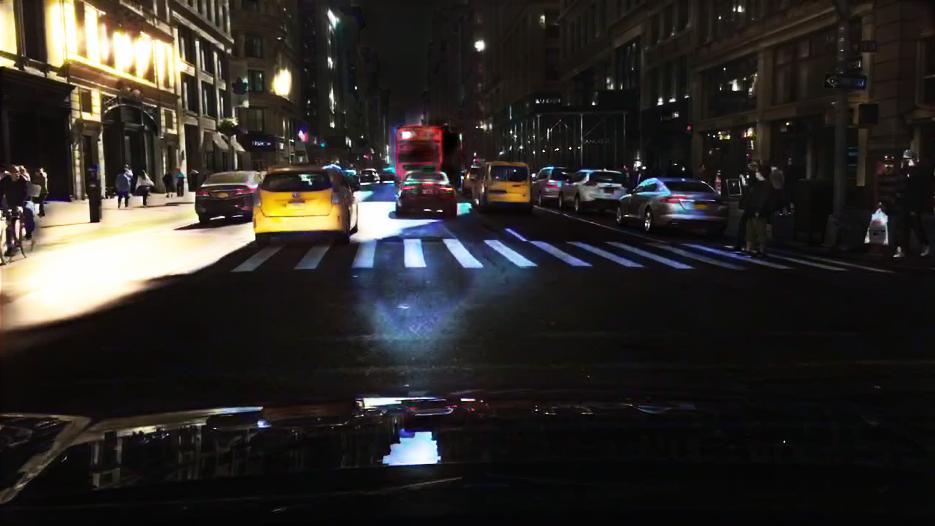}}}\hfill\\ 
		\vspace{-3pt}
		\subfigure[PFD$\rightarrow$Cityscapes]
		{\includegraphics[width=0.249\textwidth]{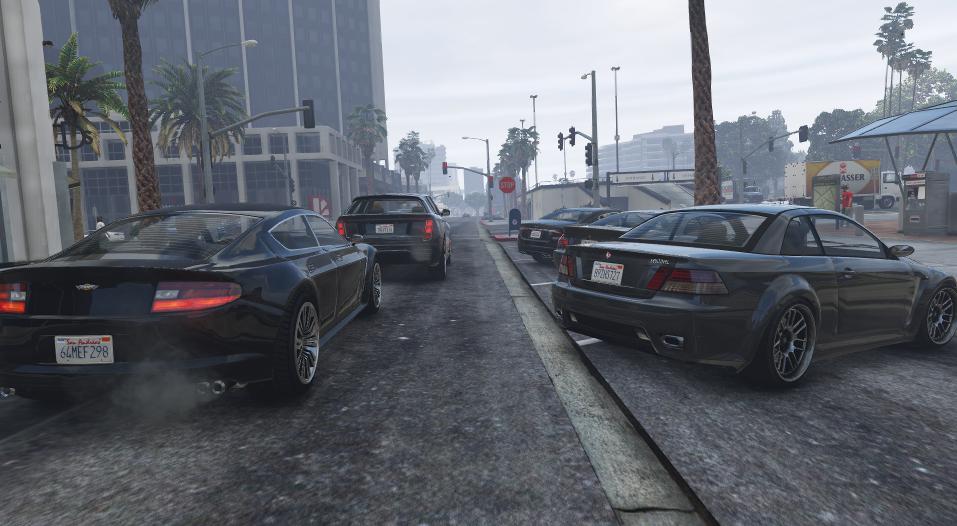}{\includegraphics[width=0.249\textwidth]{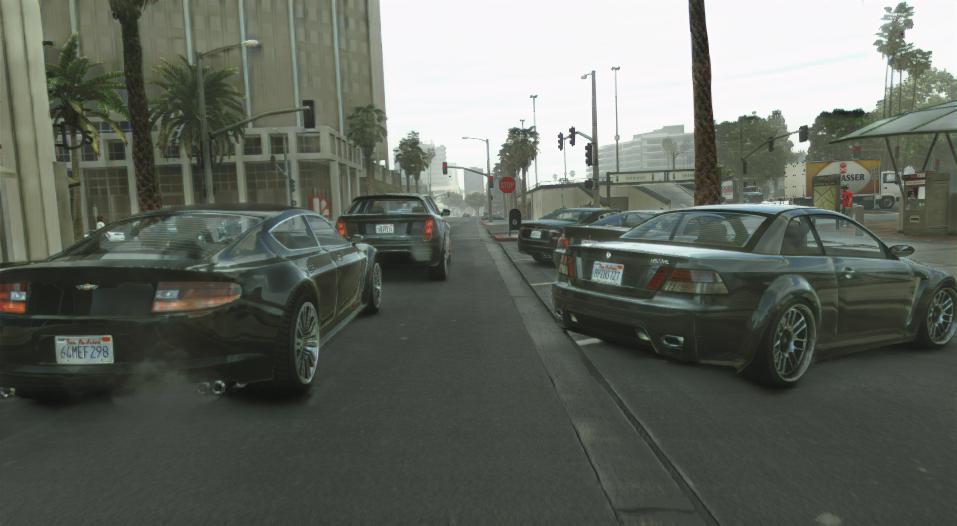}}}\hfill
		\subfigure[Day$\rightarrow$Night]
		{\includegraphics[width=0.249\textwidth]{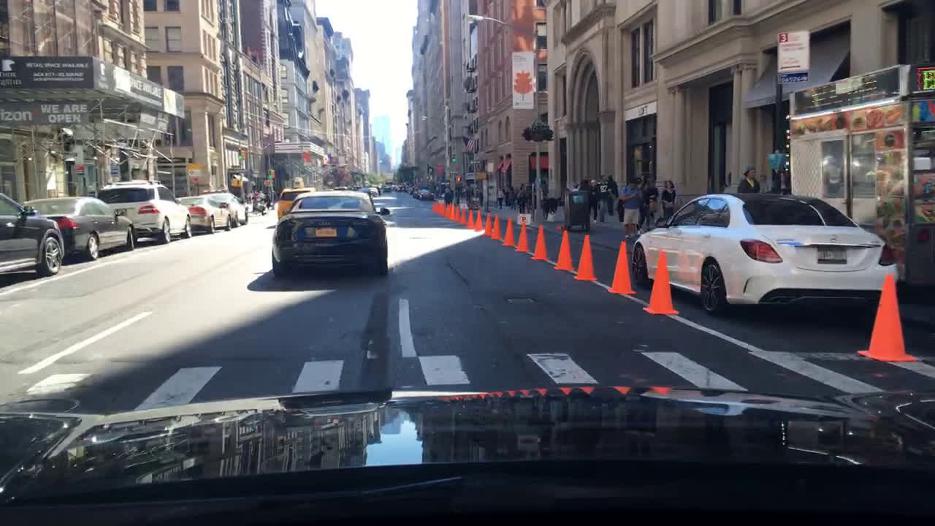}{\includegraphics[width=0.249\textwidth]{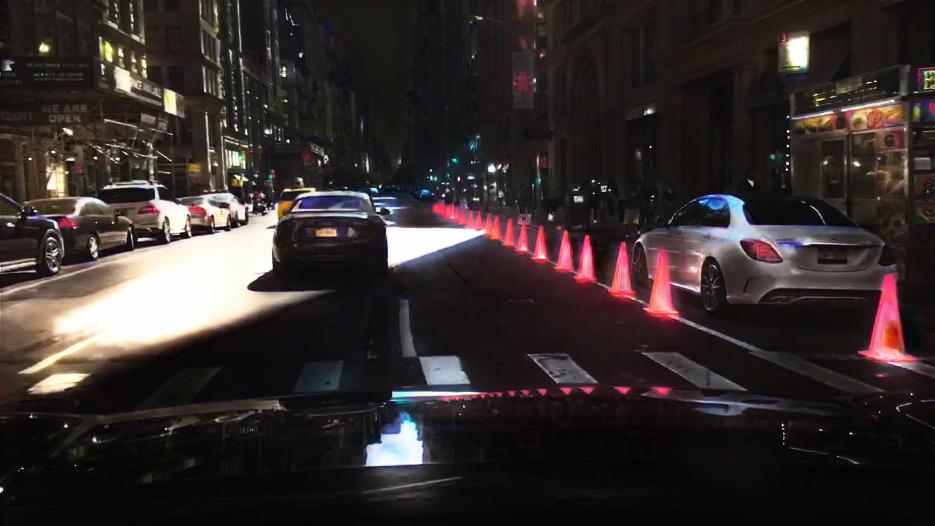}}}\hfill\\ 
		\vspace{+5pt}
		
		{\includegraphics[width=0.249\textwidth]{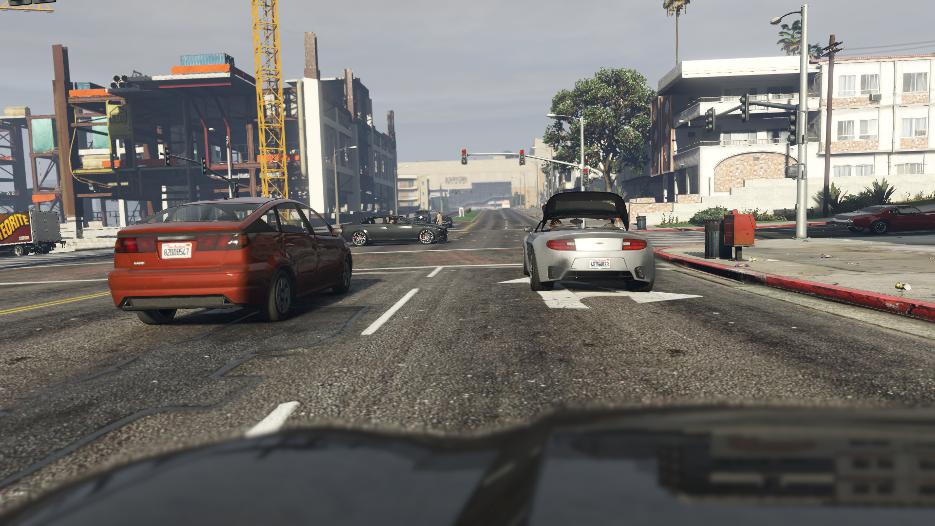}{\includegraphics[width=0.249\textwidth]{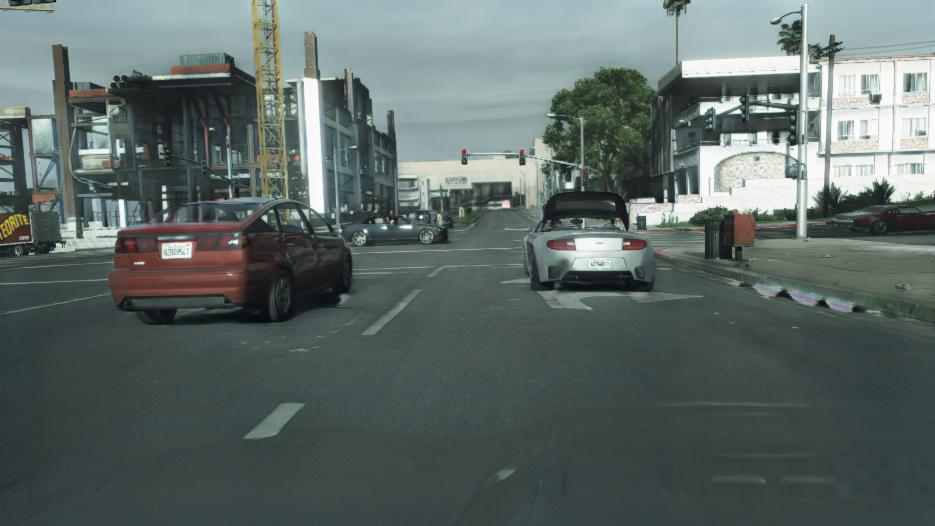}}}\hfill
		{\includegraphics[width=0.249\textwidth]{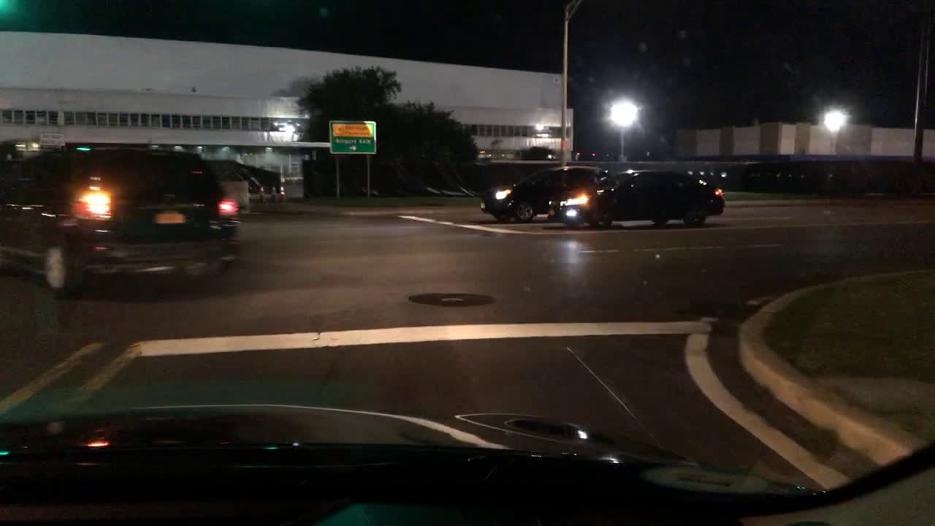}{\includegraphics[width=0.249\textwidth]{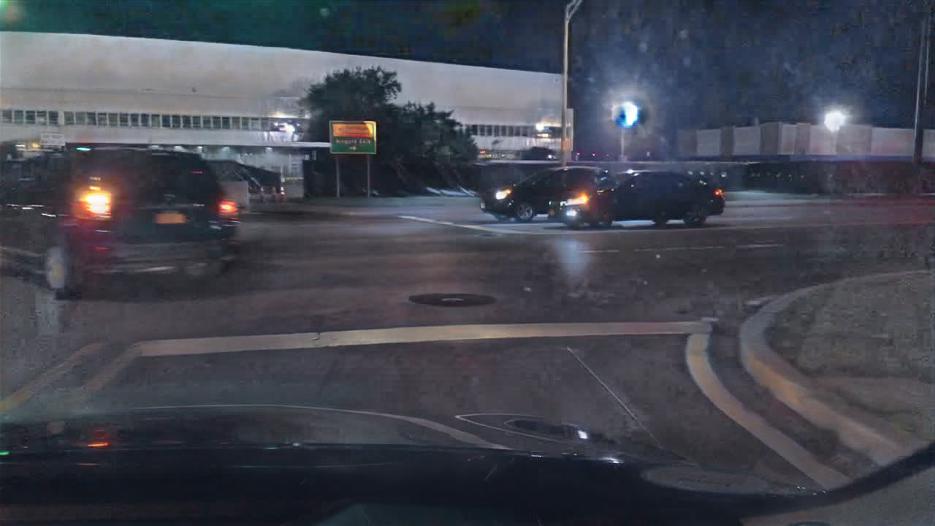}}}\hfill\\  \vspace{2pt}
		{\includegraphics[width=0.249\textwidth]{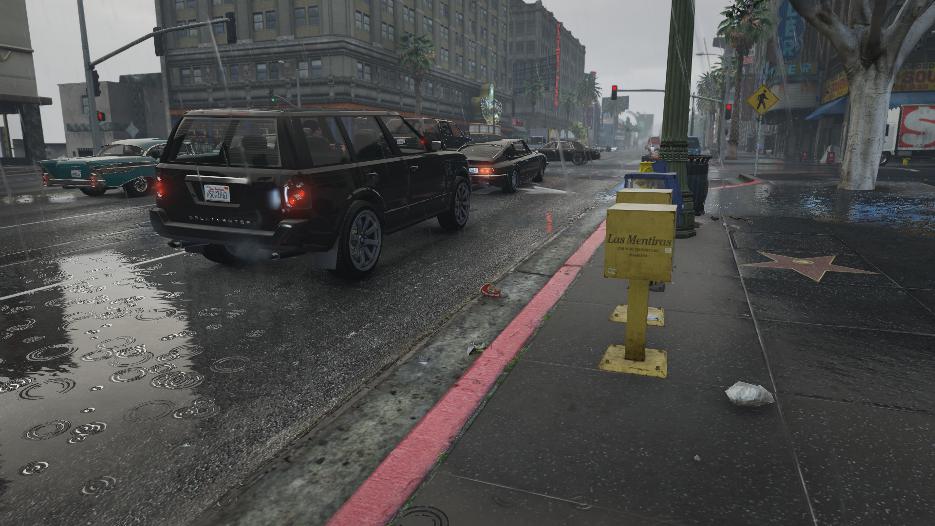}{\includegraphics[width=0.249\textwidth]{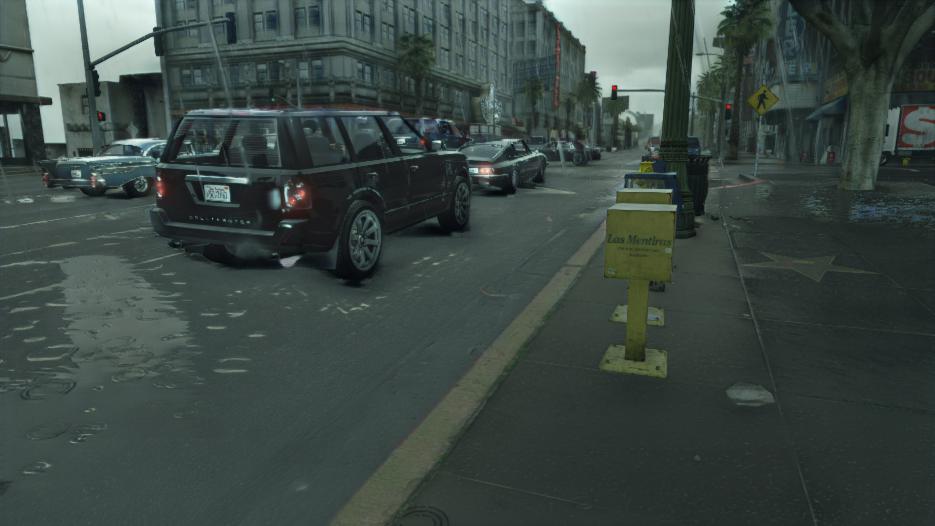}}}\hfill
		{\includegraphics[width=0.249\textwidth]{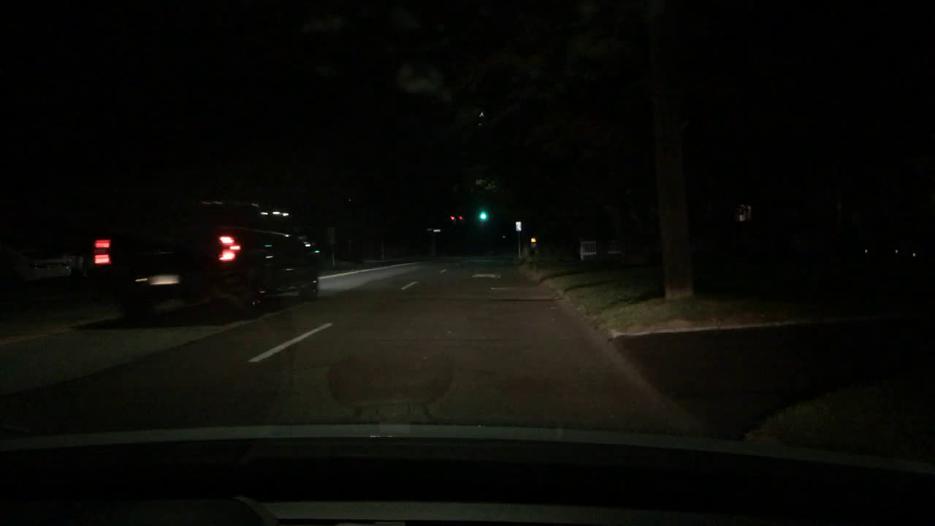}{\includegraphics[width=0.249\textwidth]{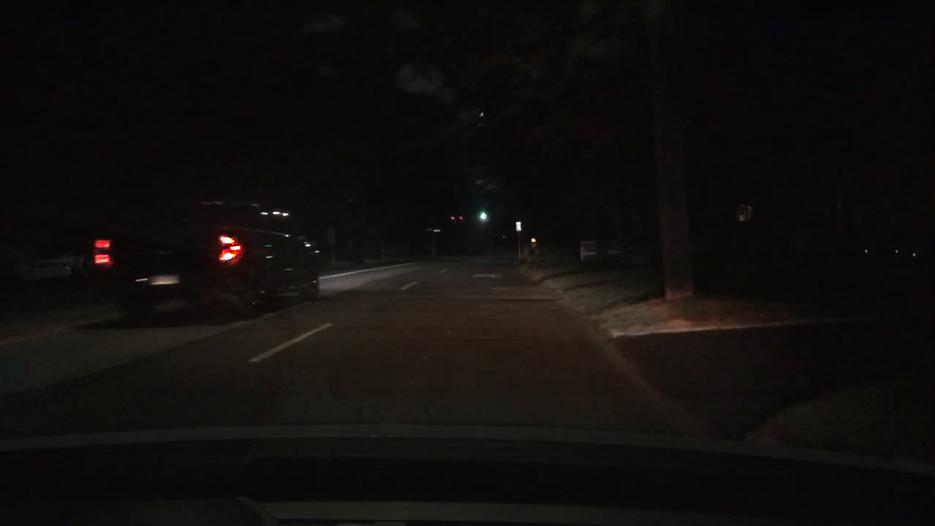}}}\hfill \\  \vspace{2pt}
		{\includegraphics[width=0.249\textwidth]{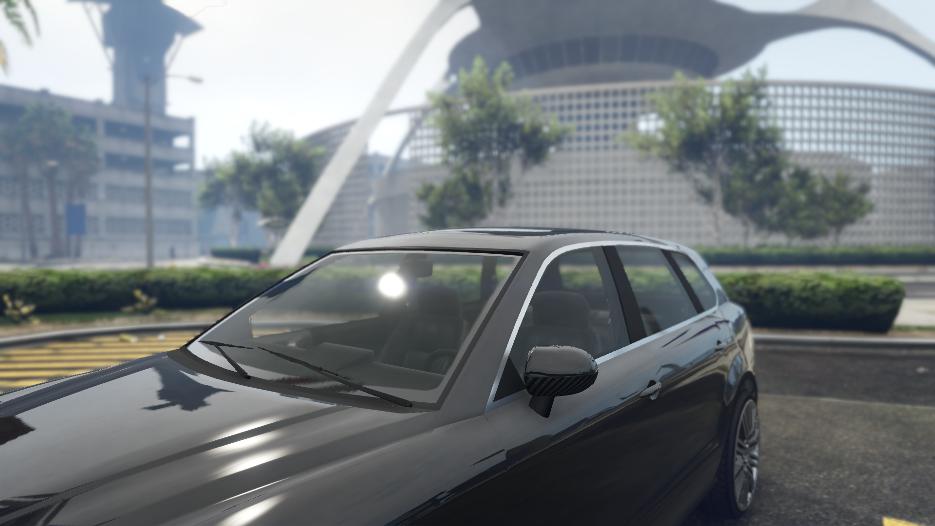}{\includegraphics[width=0.249\textwidth]{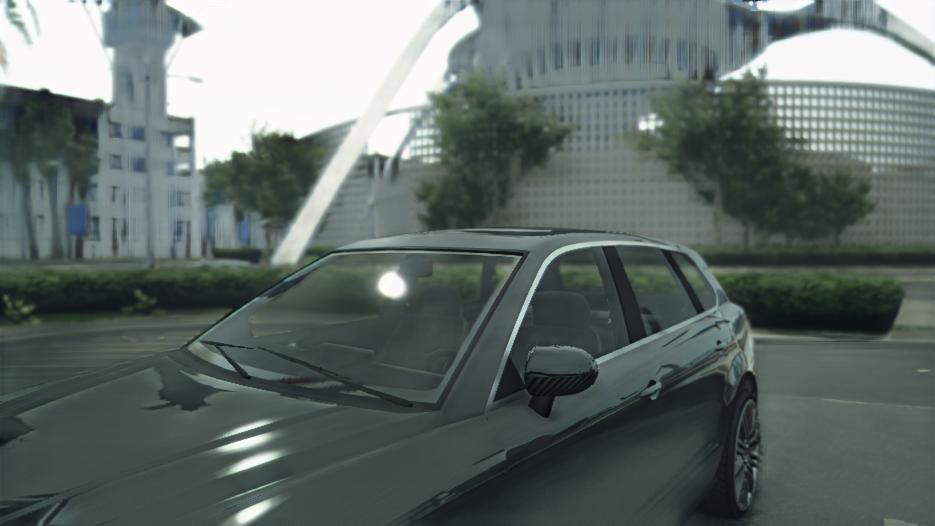}}}\hfill
		{\includegraphics[width=0.249\textwidth]{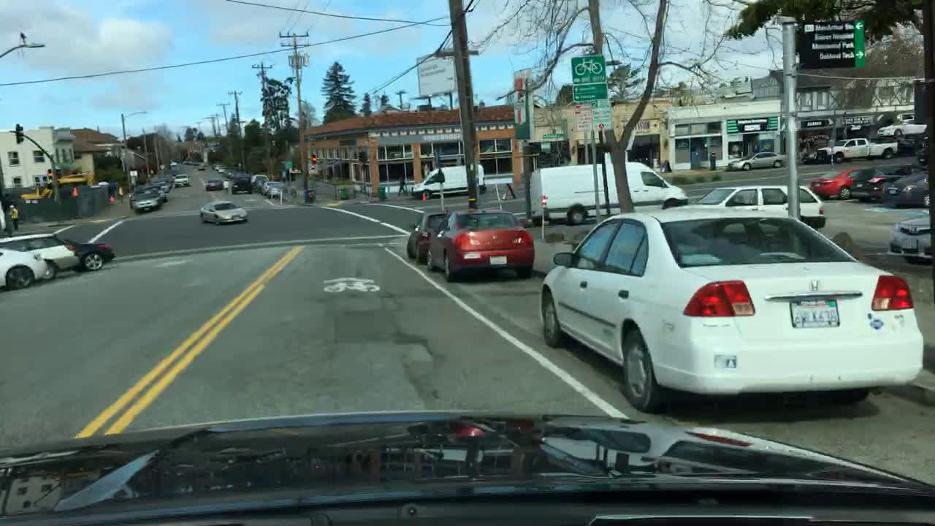}{\includegraphics[width=0.249\textwidth]{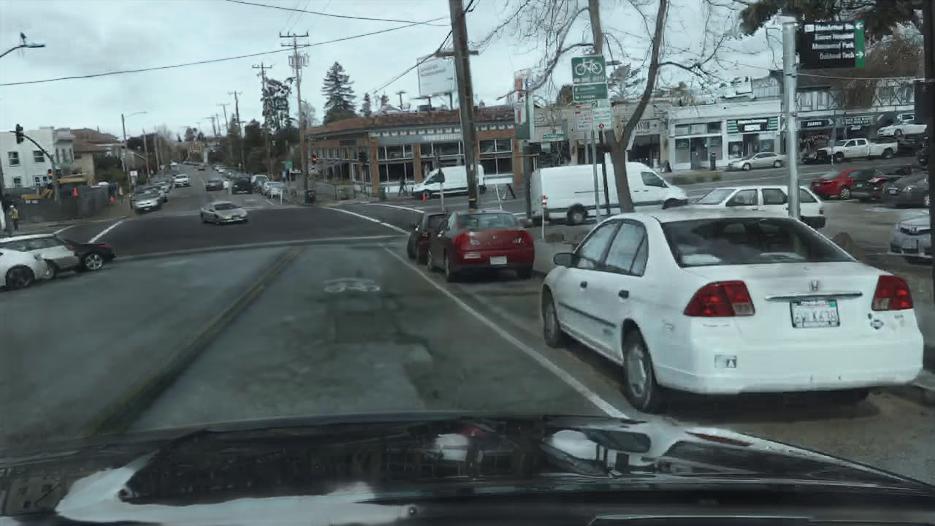}}}\hfill\\ 
		\vspace{-3pt}
		\subfigure[Viper$\rightarrow$Cityscapes]
		{\includegraphics[width=0.249\textwidth]{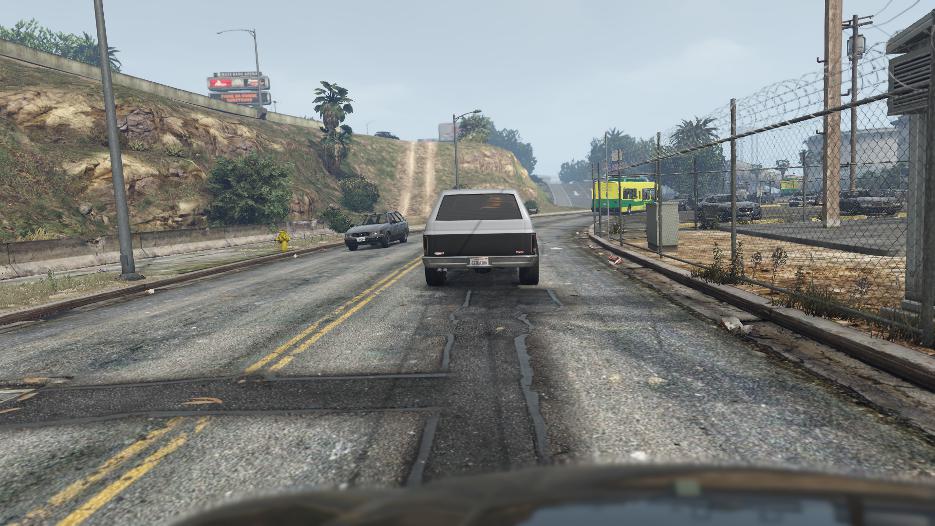}{\includegraphics[width=0.249\textwidth]{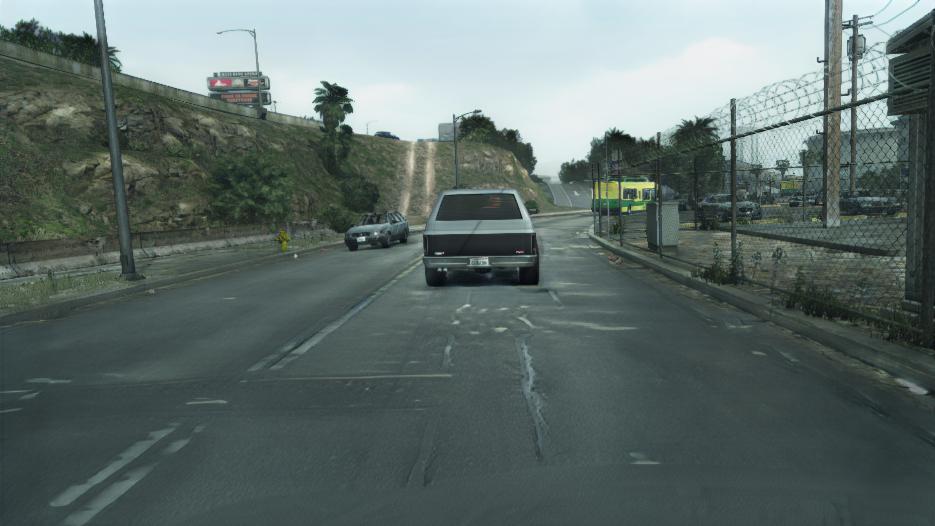}}}\hfill
		\subfigure[Clear$\rightarrow$Snowy]
		{\includegraphics[width=0.249\textwidth]{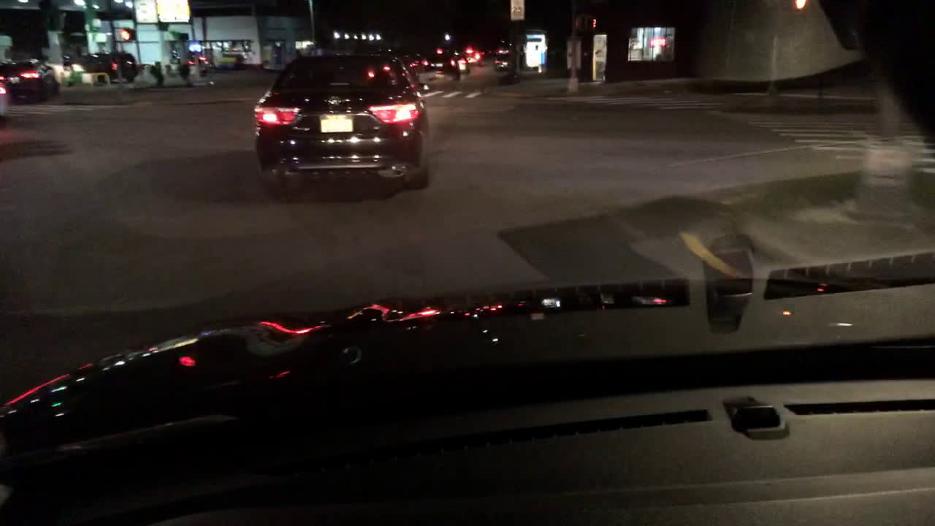}{\includegraphics[width=0.249\textwidth]{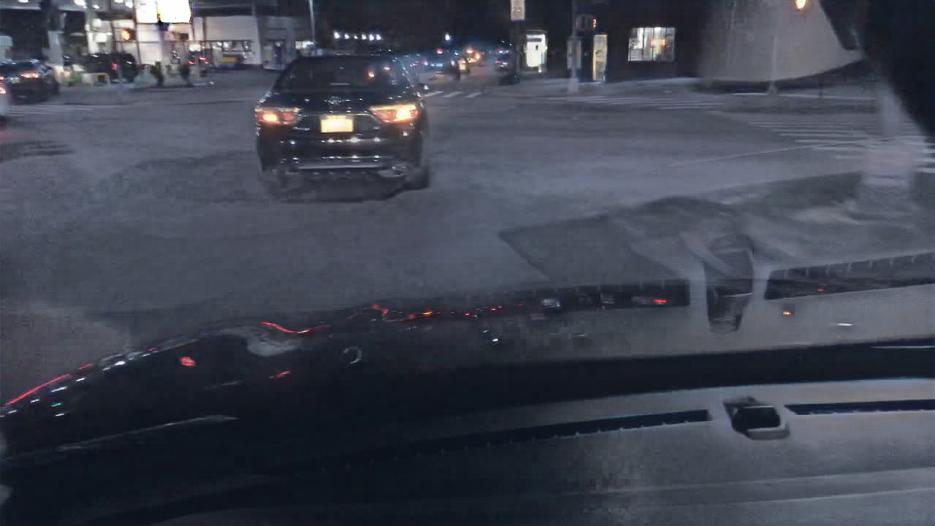}}}\hfill
	\end{center}
	\vspace{-8pt}
	\caption{\textbf{Additional qualitative results.}}
	\label{fig:qualitatative_additional_results}
\end{figure*}

\section{Conclusion}
In this work, we have shown that content-based masking of the discriminator is sufficient to significantly reduce content inconsistencies that arise in unpaired image-to-image translation. Furthermore, artifacts caused by the masking procedure can be significantly reduced by introducing a local discriminator that utilizes a segmentation-based similarity sampling technique. Moreover, our similarity sampling technique leads to a further increase in performance when applied to global input crops. We have also shown that our feature-based denormalization block is able to attend to specific content features, such as features of shadows, but can slightly increase training instability. In addition, we have proposed the cKVD metric to examine translation quality at the class or category level. In our experiments, we have found that these techniques lead to state-of-the-art performance on photo-realistic sim-to-real transfer and the translation of weather. Although our method performs well in Day$\rightarrow$Night translation, the remaining limitations of our approach are especially evident in this task.
\\\\
\textbf{Limitations.} We remark on limitations regarding the dataset, sampling, method, and implementation. Probably the most significant limitations are the complex public datasets currently available and in use, as they are not specifically designed for unpaired translation. Collection strategies and datasets that mitigate biases between source and target domains would be beneficial. Furthermore, our sampling strategy only works on an image basis and could be extended across the entire dataset to sample more significant pairs for training. Although our method works for large crops, there is still a crop size limit that must be taken into account when tuning the hyperparameters. In addition, our method for mitigating content inconsistencies depends on the segmentation model. In theory, the number of classes could be used to control how fine-grained the content consistency should be, which leads to flexibility but allows for errors depending on the segmentation quality. This can result in artifacts such as glowing objects, as shown in \autoref{fig:limitations}. Intra-class inconsistencies that may arise from intra-class biases ignored by the loss, such as small textures, represent another problem. Intra-class inconsistencies are currently underexplored in unpaired image-to-image translation and are an interesting direction for future research. Finally, we would like to point out that the efficiency of our implementation could be further improved. Apart from these limitations, our method achieves state-of-the-art performance in complex translation tasks while mitigating inconsistencies through a masking strategy that works by applying few tricks. Simple masking strategies have proven to be very successful in other fields. Therefore, we believe that masking strategies for unpaired image-to-image translation represent a promising direction for further research.
\\\\
\textbf{Ethical and responsible use.} Considering the limitations of current methods, unpaired image-to-image translation methods should be trained and tested with care, especially for safety-critical domains like autonomous driving. A major concern is that it is often unclear or untested whether the transferred content can still be considered consistent for subsequent tasks in the target domain. Even though measures exist for content-consistent translation, they do not allow for the explainability of what exactly is being transferred and changed by the model on a fine-grained level. With our proposed cKVD metric we contribute to this field by allowing class-specific translation measurements - a direction that we hope is the right one. However, even if the content is categorically consistent at a high (class) level, subcategories (like parts of textures) may still be interchanged. At a lower level, content consistency and style consistency are intertwined (e.g., a yellow stop sign). Another privacy and security question is whether translation methods are (or will) be able to (indirectly) project sensitive information from the target domain to the translated images (e.g., exchange faces from simulation with faces of existing persons during the translation). A controllable (class-level and in-class-level) consistency method could help to resolve such issues.

\section*{Acknowledgments}
The authors would like to sincerely thank all reviewers for their helpful feedback, which contributed to the quality of this paper. The authors would also like to sincerely thank Markus Klenk for proofreading this work. \clearpage

\bibliographystyle{IEEEtran}
\bibliography{IEEEabrv,bib}

\clearpage

{\appendix
\label{appendix}
\section*{The FeaMGAN Architecture}
	
\noindent \textbf{Generator}. As shown in \autoref{fig:FeaMGenerator}, our generator consists of a content stream encoder, a content stream, a generator stream encoder, and a generator stream. The content stream encoder shown in \autoref{fig:ContentStreamEncoder} is utilized to create the initial features of the source image and condition. These initial features are the input to the content stream, which creates features for multiple levels with residual blocks. The statistics of these features are then integrated into the generator at multiple levels utilizing the residual FATE blocks shown in \autoref{fig:FATEResBlk}. The generator stream utilizes the encoder shown in \autoref{fig:GeneratorStreamEncoder} to create the initial latent from which the target image is generated. To further enforce content consistency, we do not use a variational autoencoder to obtain an deterministic latent. In addition, we found that utilizing additional residual blocks in the last layers of the generator stream improves performance, likely due to further refinement of the preceding upsampled features. We use spectral instance normalization for the residual blocks in the content stream and spectral batch normalization for the residual blocks in the generator stream. The convolutional layers in the generator stream encoder have the following numbers of filters: $[256,512,1024]$. The residual blocks in the generator have the following numbers of filters: $[1024,1024,1024,512,256,128,64,64,64,64]$. The numbers of filters of the convolutional layers in the content streams encoder are $[64,64]$. The numbers of filters in the content stream match those of the output of the preceding residual block in the generator stream at the respective level: $[64,128,256,512,1024,1024,1024,1024]$. For all residual blocks, we use $3\times3$ convolutions and $1\times1$ convolutions for the skip connections. $\gamma$ and $\beta$ in the FATE and FADE blocks are created with $3\times3$ convolutions. Throughout the generator, we use a padding of $1$ for the convolutions - we only downsample with strides and downsampling layers. We utilize the "nearest" upsampling and downsampling from Pytorch. For our small model, we halve the number of filters.
\begin{figure*}[t]
	\begin{minipage}{1.0\textwidth}
		\centering 
		\includegraphics[width=0.9\linewidth]{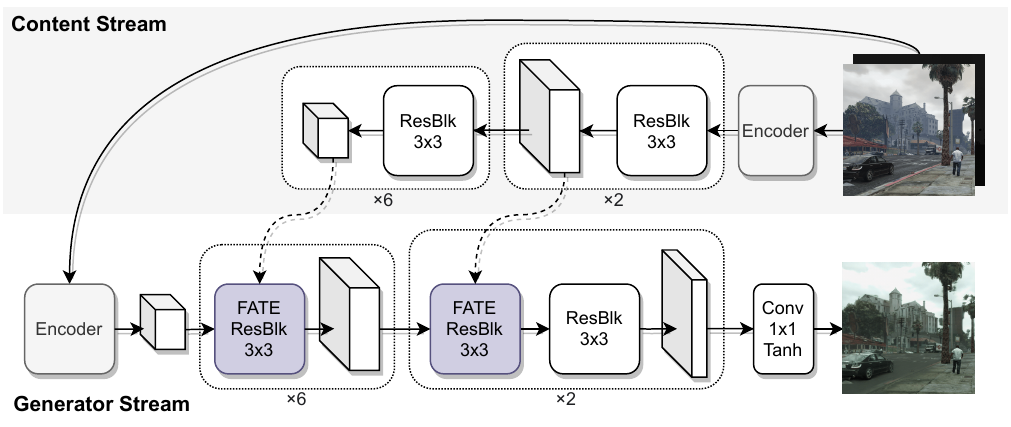}
		\caption{\textbf{Generator architecture}. Arrows with dashed lines indicate connections at multiple levels between the two streams.}
		\label{fig:FeaMGenerator}
		\vspace{15pt}
	\end{minipage}

	\begin{minipage}{0.5\textwidth}
		\centering 
		\includegraphics[width=0.9\linewidth]{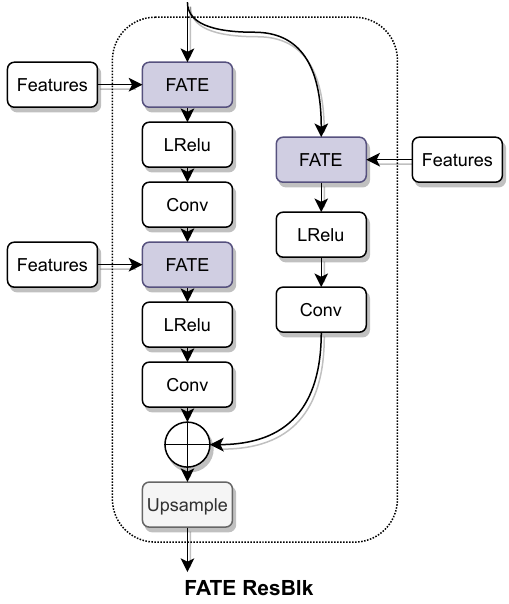}
		\caption{The FATE residual block used in the generator stream.}
		\label{fig:FATEResBlk}
	\end{minipage}
	\begin{minipage}{0.5\textwidth}	
		\centering 
		\includegraphics[width=0.6\linewidth]{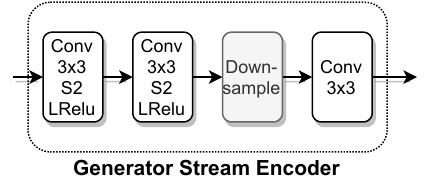}
		\caption{The generator stream encoder used to encode the input image and condition for the generator stream.}
		\label{fig:GeneratorStreamEncoder}
		\vspace{7pt}
		\includegraphics[width=0.3\linewidth]{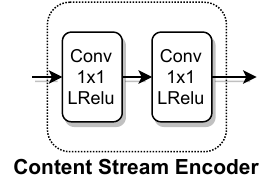}
		\caption{The content stream encoder used to encode the input image and condition for the content stream.}
		\label{fig:ContentStreamEncoder}
		\vspace{7pt}
		\includegraphics[width=0.5\linewidth]{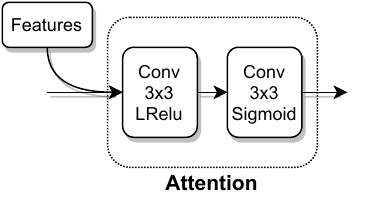}
		\caption{The attention module used in the FATE block to attend to the statistics of the features.}
		\label{fig:Attention}
	\end{minipage}
\end{figure*} \\\\
\noindent \textbf{Discriminator}. As shown in \autoref{fig:FeaMDiscriminator}, our discriminator consists of downsampling, upsampling, and prediction components. First, the input images of the source or target domain are downsampled via $5$ stride $2$ convolutions. We transform the output feature map of the last $4$ downsampling convolutions with $1\times1$ convolutions. The last transformed feature map is used as input for the upsampling components, while the other transformed feature maps are added to the feature maps of the upsampling component for the receptive level. Then we utilize the feature maps of the $3$ upsampling levels to create the final prediction on $3$ levels. Thereby, we first apply a convolutional layer on the upsampled features. This convolution is followed by two convolutional layers: One is used to create the prediction feature map of depth $1$, and the feature map of the other convolutional layer is multiplied by the segmentation map. The resulting segmentation feature map is then collapsed into depth $1$ by adding the depth dimensions together. At last, the collapsed segmentation feature map is added to the prediction feature map to produce the final prediction. In this way, the discriminator is encouraged to produce class-specific predictions. We use spectral instance normalization for all convolutional layers. The $3\times3$ convolutions of the downsampling component have the following numbers of filters: $[64,128,256,512,512]$. The $1\times1$ convolutions of the downsampling component have the following numbers of filters: $[256,256,256,256]$. The first convolutions in the prediction component have the following numbers of filters: $[128,128,128]$. The convolutions that are multiplied by the downsampled segmentation maps have the following numbers of filters: $[128,128,128]$. The convolution used to create the downsampled segmentation map has $128$ filters. The convolutions to create the predictions have the following numbers of filters: $[1,1,1]$. Throughout the discriminator, we use a padding of $1$ for the convolutions - we only downsample with strides and downsampling layers. We utilize the "bilinear" upsampling and downsampling from Pytorch. For our small model, we halve the number of filters.
\begin{figure*}[t]
	\centering 
	\includegraphics[width=0.8\linewidth]{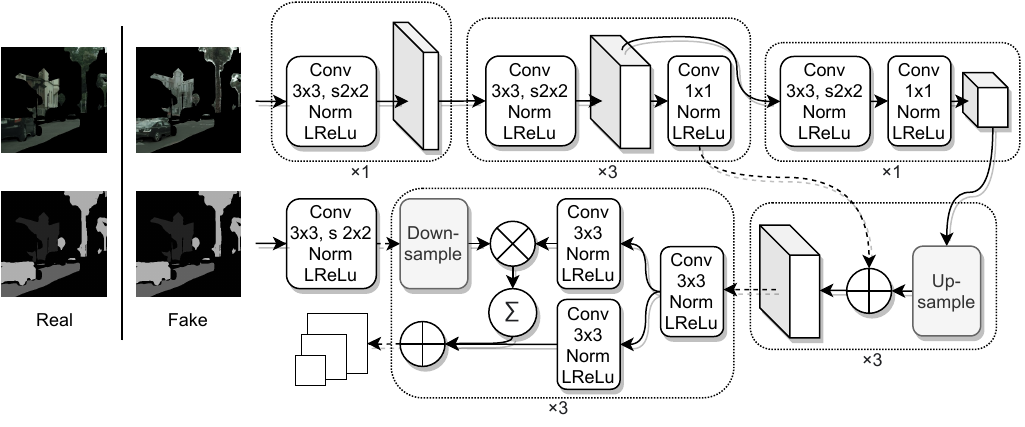}
	\caption{\textbf{Discriminator architecture}. Arrows with dashed lines indicate connections at multiple levels between the two components.}
	\label{fig:FeaMDiscriminator}
\end{figure*}
\section*{Additional Dataset Details}
\begin{table}
	\caption{\textbf{cKVD class mapping.}}
	\begin{center}
		\begin{tabular}{ll}
			\toprule
			cKVD Class & MSeg-Id(Name)\\   
			\midrule	
			sky & 142(sky) \\ [+3pt] 
			\multirow{2}*{ground}  & 94(gravel), 95(platform), 97(railroad), \\
			& 100(pavement-merged), 101(ground) \\ [+3pt] 
			road & 98(road) \\ [+3pt] 
			terrain & 102(terrain) \\ [+3pt] 
			vegetation & 174(vegetation) \\ [+3pt] 
			\multirow{2}*{building} & 31(tunnel), 32(bridge), 33(building-parent), \\
			&  35(building), 36(ceiling-merged) \\ 			[+3pt] 			
			\multirow{4}*{roadside-obj.}  & 130(streetlight), 131(road\_barrier), 132(mailbox), \\
			& 133(cctv\_camera), 134(junction\_box), 135(traffic\_sign), \\
			& 136(traffic\_light), 137(fire\_hydrant), 138(parking\_meter), \\
			& 139(bench), 140(bike\_rack), 141(billboard) \\ 	[+3pt] 
			\multirow{2}*{person} & 125(person), 126(rider\_other), 127(bicyclist), \\
			& 128(motorcyclist) \\[+3pt] 
			\multirow{3}*{vehicle}  &  175(bicycle), 176(car), 177(autorickshaw), \\
			& 178(motorcycle), 180(bus), 181(train), \\
			& 182(truck), 183(trailer), 185(slow\_wheeled\_object) \\ [+3pt] 
			rest & all other MSeg classes \\ 
			\bottomrule	
		\end{tabular}
	\end{center}
	\label{tab:sKVD_class_mapping}
	\vspace{-13pt}
\end{table}
\begin{table*}
	\caption{\textbf{Additional details of the used datasets.}}
	\begin{center}
		\begin{tabular}{lcclcccc}
			\toprule
			Dataset & Resolution & fps & Used Train/Val Data & Task & Input Resolution & Input Cropping  \\   
			\midrule	
			PFD \cite{richter2016playing} & 1914$\times$1052& - & all images & \textit{PFD$\rightarrow$Cityscapes} & 957$\times$526 &  -  \\  [+3pt] 
			Viper \cite{richter2017playing} & 1920$\times$1080 & $\sim$15 & all train/val data, but no night sequences& \textit{Viper$\rightarrow$Cityscapes} & 935$\times$526 & - \\ 	[+3pt] 
			\multirow{2}*{Cityscapes \cite{cordts2016cityscapes}} & \multirow{2}*{2048$\times$1024} & \multirow{2}*{17} & \multirow{2}*{all sequences of the train/val data} & \textit{PFD$\rightarrow$Cityscapes}  & 1.052$\times$526 &  957$\times$526  \\  
			&  &  &  &  \textit{Viper$\rightarrow$Cityscapes}  & 1.052$\times$526 & 935$\times$526 \\  [+3pt] 
			\multirow{2}*{BDD100K \cite{yu2020bdd100k}} & \multirow{2}*{1280$\times$720} & \multirow{2}*{30} & train: first 100k, val: first 40k & \textit{Day$\rightarrow$Night}& \multirow{2}*{935$\times$526} & \multirow{2}*{-} \\ 
			& &  & train: first 50k, val: first 40k   & \textit{Clear$\rightarrow$Snowy} &  &  \\ 
			\bottomrule	
		\end{tabular}
	\end{center}
	\label{tab:datasets_details}
\end{table*}
\begin{table*}
	\caption{\textbf{Additional training details.}}
	\begin{center}
		\begin{tabular}{lcccc}
			\toprule
			Task & Epochs & Schedule & Decay & Local Discriminator Batch Size  \\   
			\midrule	
			\textit{PFD$\rightarrow$Cityscapes} & $20$ & half learning rate stepwise, learning rate $\geq 0.0000125$ & after each $3$rd epoch  & 32\\  
			\textit{Viper$\rightarrow$Cityscapes} & $5$ & half learning rate stepwise, learning rate $\geq 0.0000125$ & after each epoch  & 32\\    
			\textit{Day$\rightarrow$Night} & $5$ & half learning rate stepwise, learning rate $\geq 0.0000125$ & after each epoch  & 32\\    
			\textit{Clear$\rightarrow$Snowy} & $10$ & half learning rate stepwise, learning rate $\geq 0.0000125$ & after each epoch & 32\\    
			\bottomrule	
		\end{tabular}
	\end{center} 
	\label{tab:training_details}
\end{table*}
In \autoref{tab:datasets_details}, we show additional details about the used datasets. Since we compare our method i.a. to the transferred PFD \cite{richter2016playing} images provided by EPE \cite{richter2022enhancing}, we use a base image height of $526$ throughout our experiments. The aspect ratio is preserved when resizing the input images. To match the input sizes of the images of both domains, we apply cropping to the image with the larger width if the image sizes of both domains do not align. The resulting images are randomly flipped before the sampling strategy is applied. 
	
\section*{Additional Training Details}
In \autoref{tab:training_details}, we show additional details about the hyperparameters used for training the four translation tasks. No tuning was performed for other translation tasks then PFD$\rightarrow$Cityscapes besides adapting the learning rate schedule for the dataset lengths of these tasks.

\section*{Additional Results}
We show additional results of our experiments in Figures \ref{fig:qualitative_comparison_epe_additional}, \ref{fig:qualitative_comparison_epe_additional_random}, \ref{fig:qualitative_comparison_additional_random}, \ref{fig:qualitative_ablation_crop_size_additional_random}, and \ref{fig:qualitative_ablations_additional_random}. In \autoref{tab:quantitative_comparison_extended}, we report additional results from our cKVD metric and the stability of all results over five runs. Furthermore, we report the stability of all results from the ablation study in \autoref{tab:quantitative_ablation_extended}. We note that the results for most baselines and for our method show non-negligible deviations in many tasks.

\begin{figure*}[h] 
	\centering
	\renewcommand{\thesubfigure}{}
	{\includegraphics[width=0.33\textwidth]{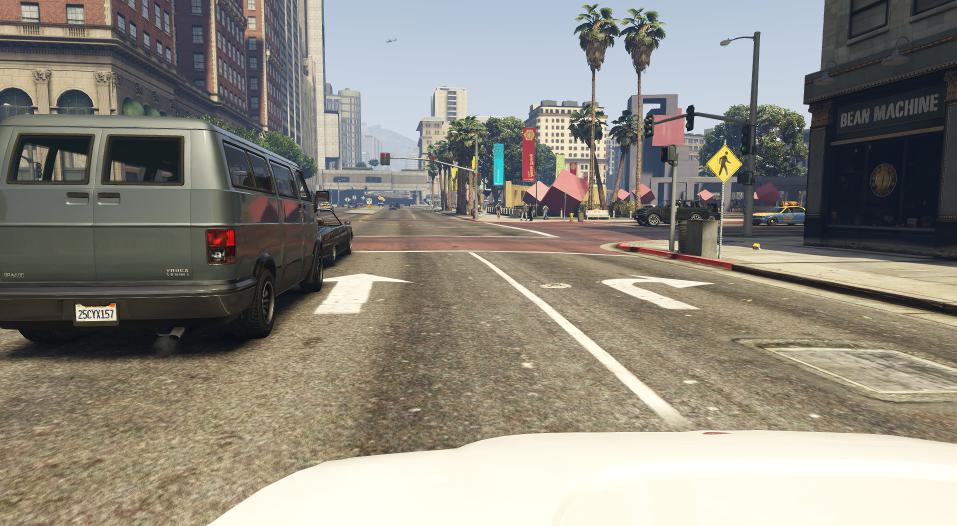}}\hfill
	{\includegraphics[width=0.33\textwidth]{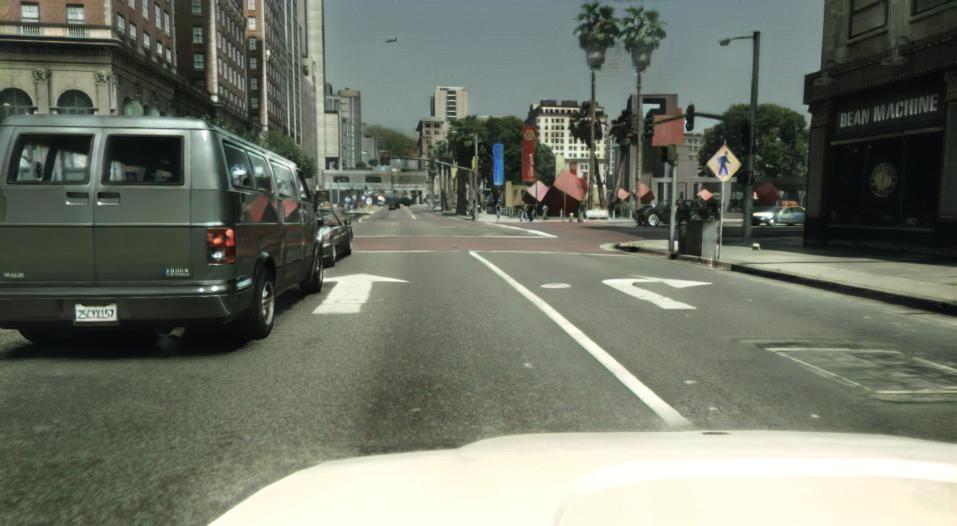}}\hfill
	{\includegraphics[width=0.33\textwidth]{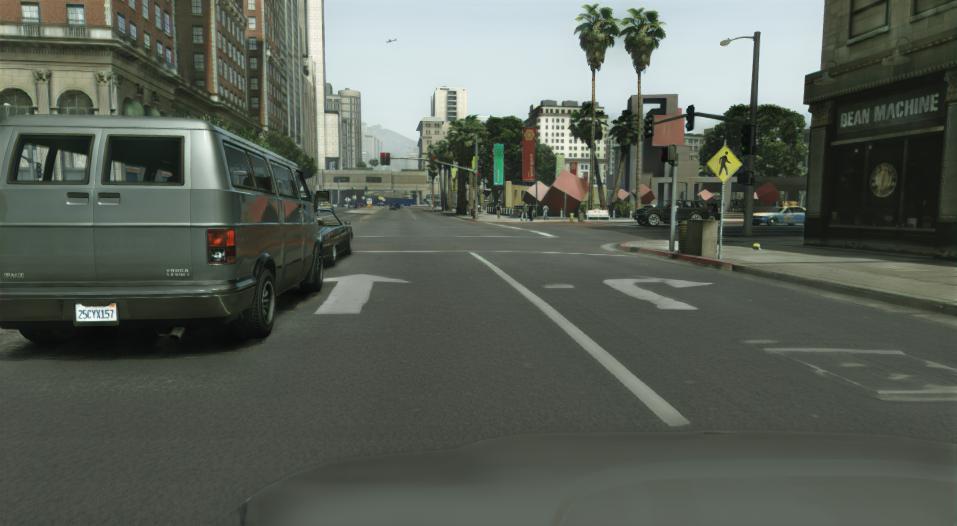}}\hfill\\\vspace{2.5pt}
	{\includegraphics[width=0.33\textwidth]{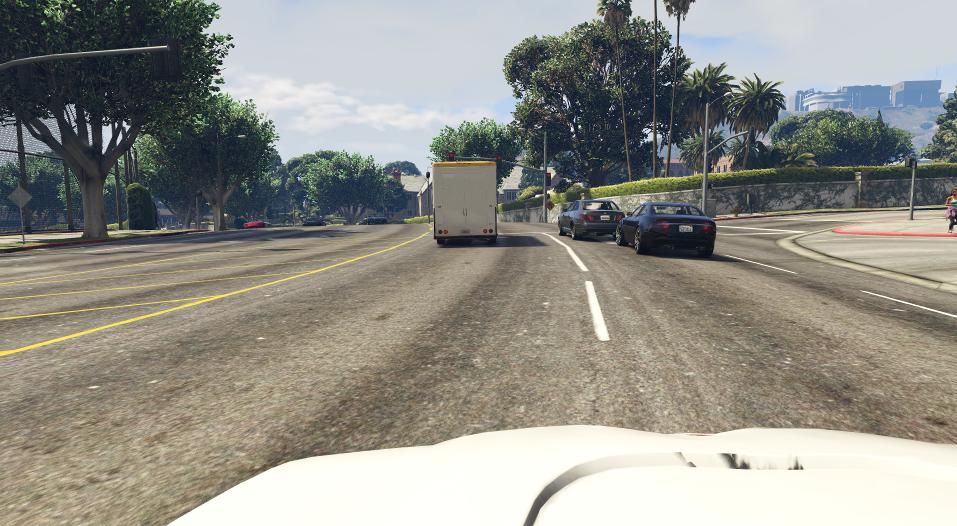}}\hfill
	{\includegraphics[width=0.33\textwidth]{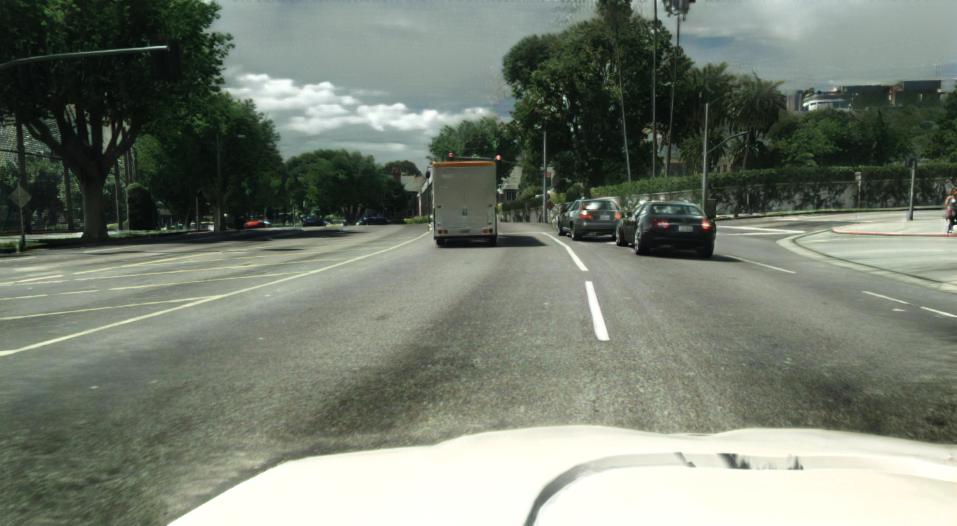}}\hfill
	{\includegraphics[width=0.33\textwidth]{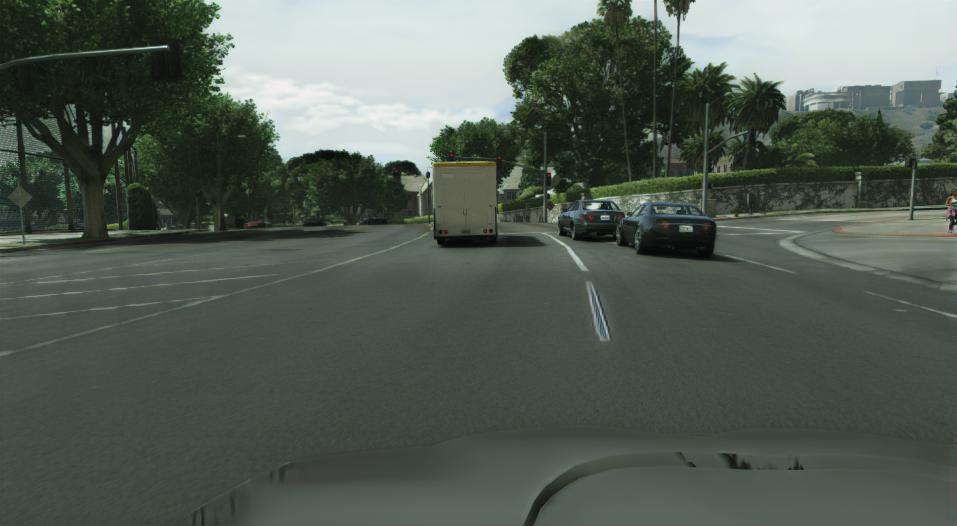}}\hfill\\\vspace{2.5pt}
	{\includegraphics[width=0.33\textwidth]{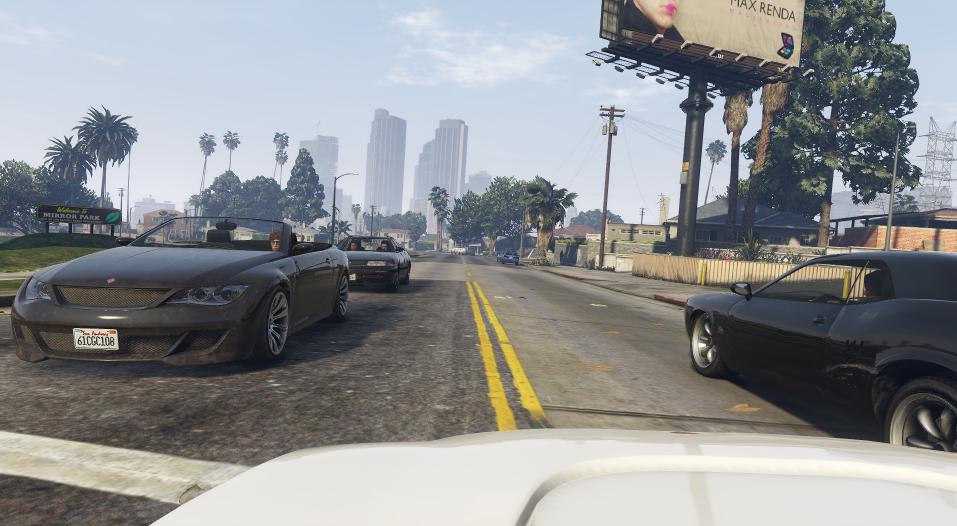}}\hfill
	{\includegraphics[width=0.33\textwidth]{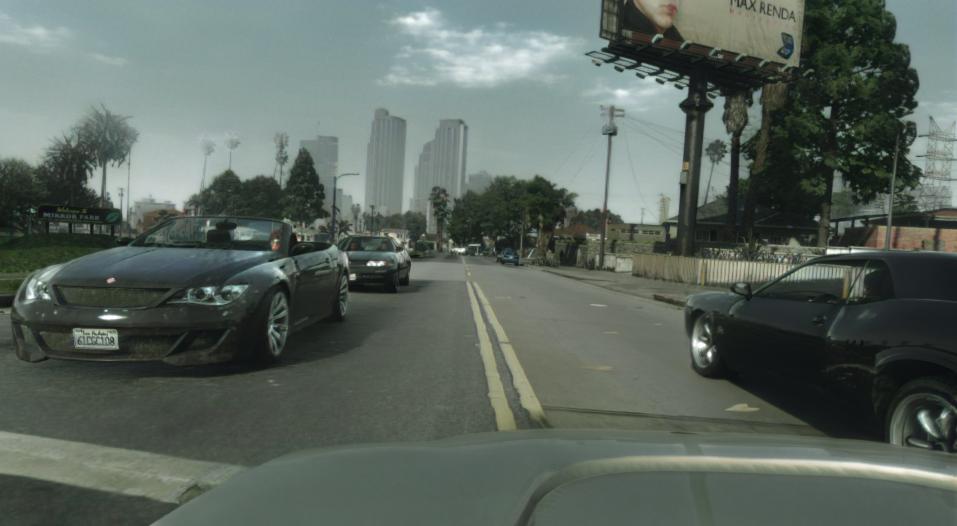}}\hfill
	{\includegraphics[width=0.33\textwidth]{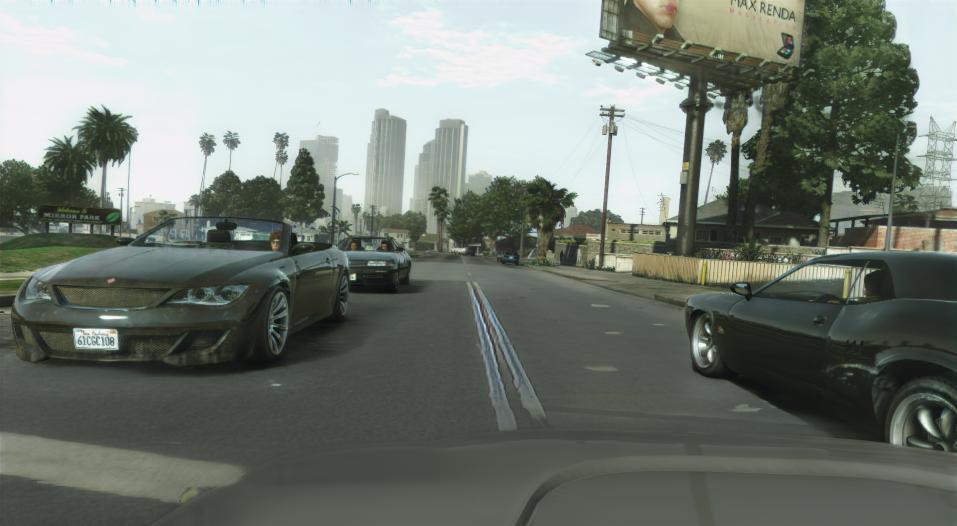}}\hfill\\\vspace{2.5pt}
	{\includegraphics[width=0.33\textwidth]{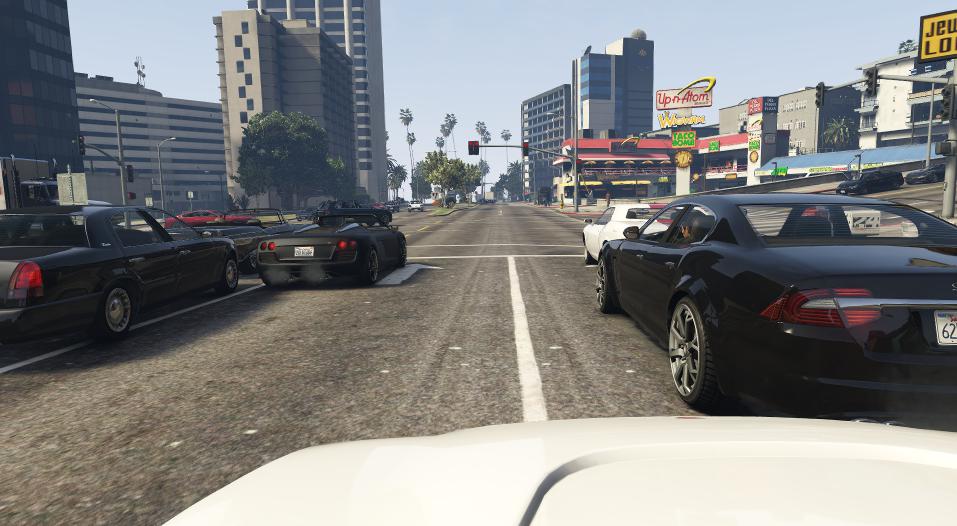}}\hfill
	{\includegraphics[width=0.33\textwidth]{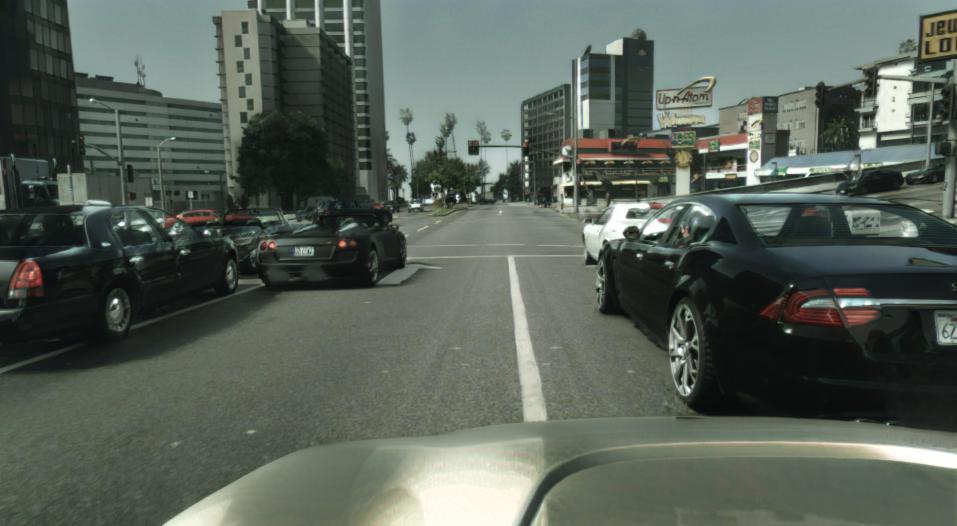}}\hfill
	{\includegraphics[width=0.33\textwidth]{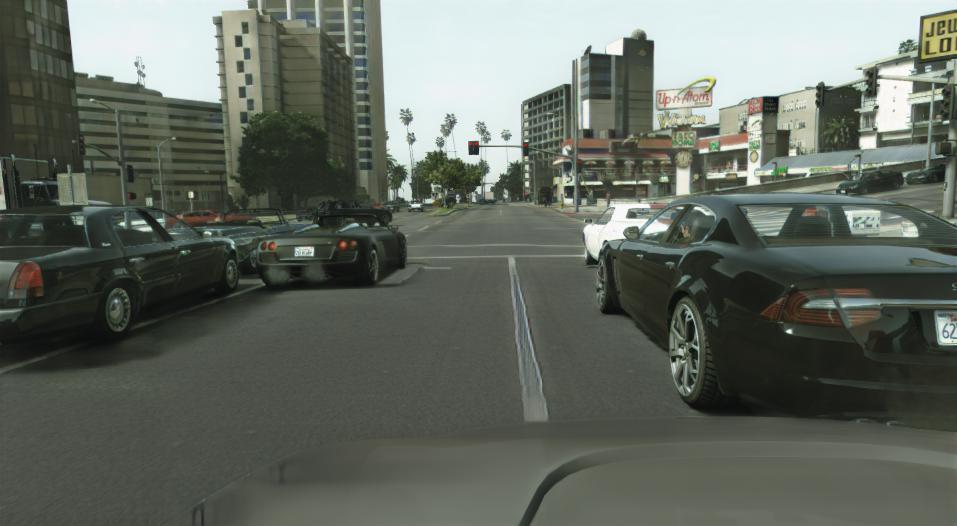}}\hfill\\\vspace{2.5pt}
	{\includegraphics[width=0.33\textwidth]{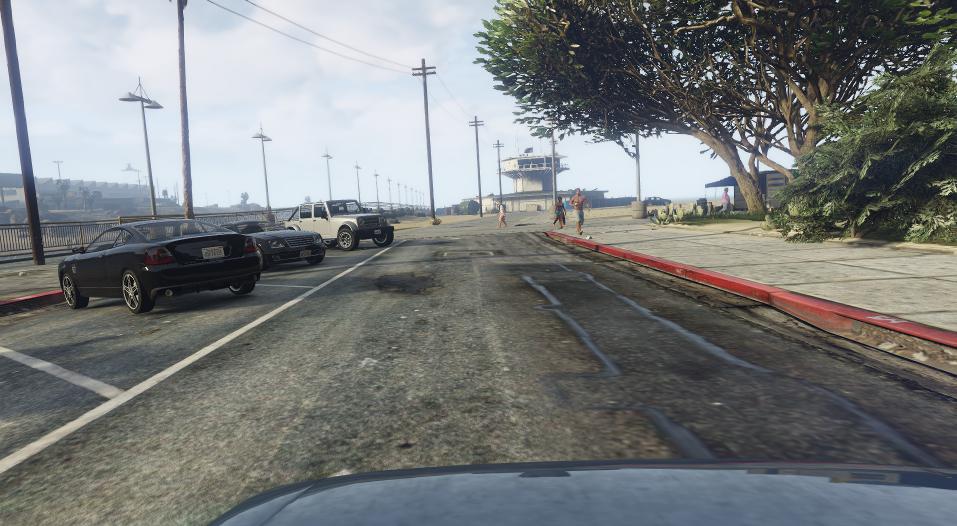}}\hfill
	{\includegraphics[width=0.33\textwidth]{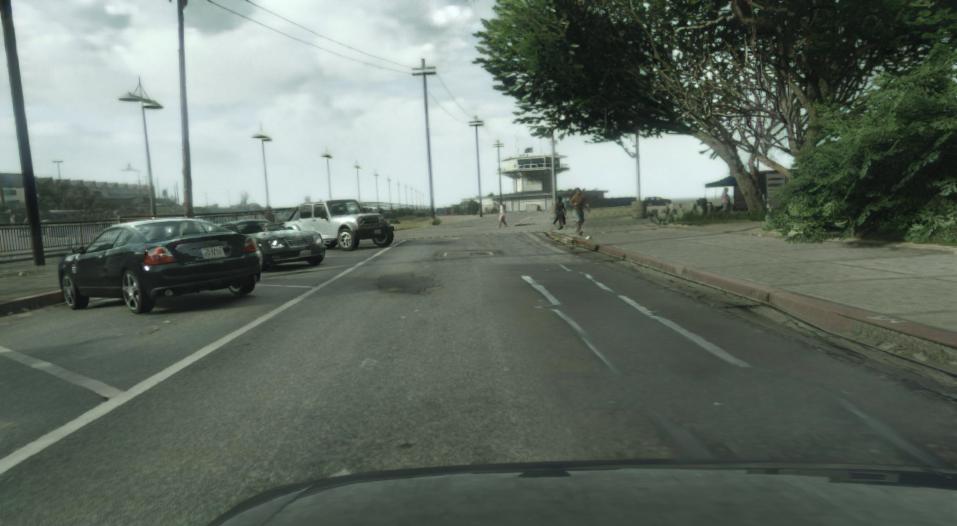}}\hfill
	{\includegraphics[width=0.33\textwidth]{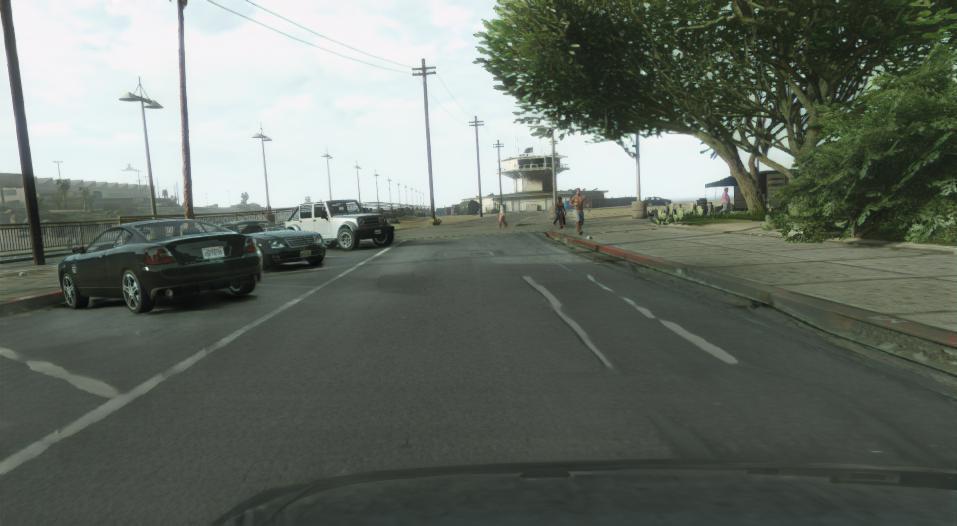}}\hfill\\\vspace{-2.5pt}
	\subfigure[Input]
	{\includegraphics[width=0.33\textwidth]{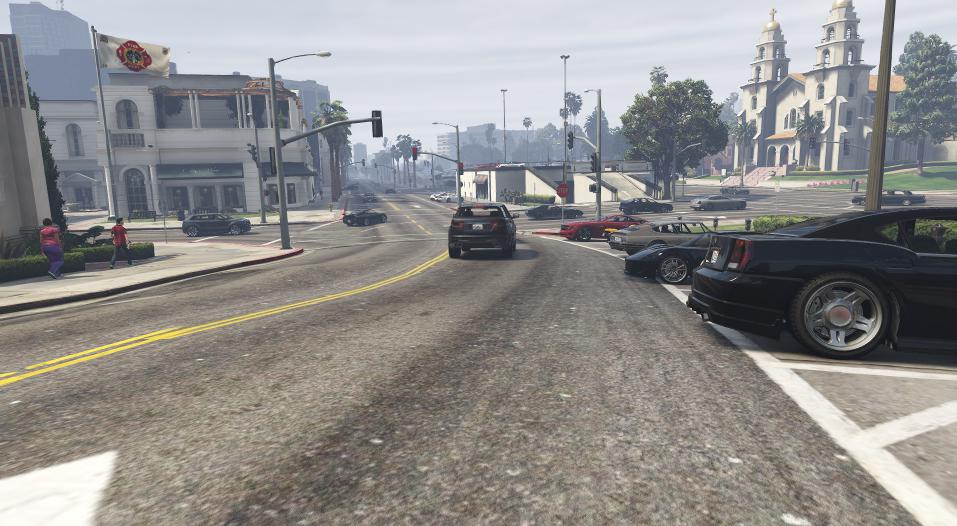}}\hfill
	\subfigure[EPE]
	{\includegraphics[width=0.33\textwidth]{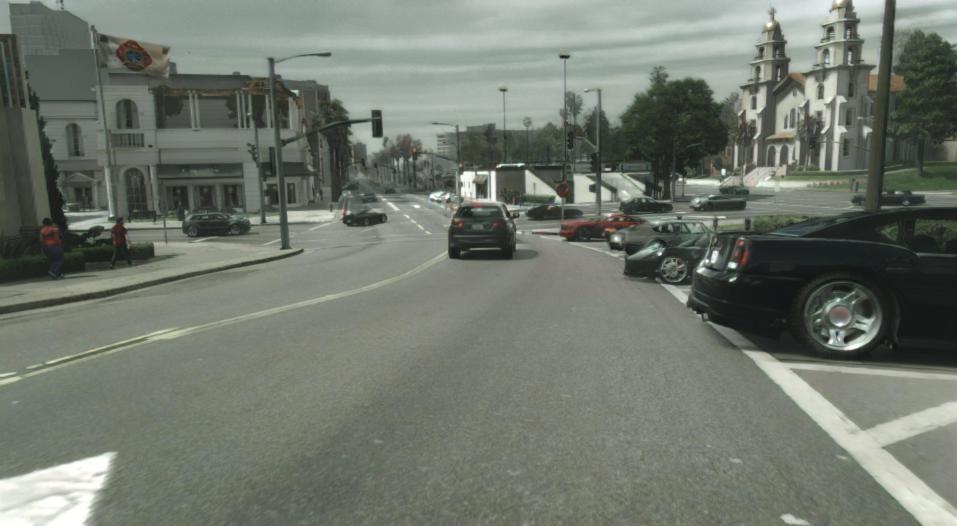}}\hfill
	\subfigure[FeaMGAN (ours)]
	{\includegraphics[width=0.33\textwidth]{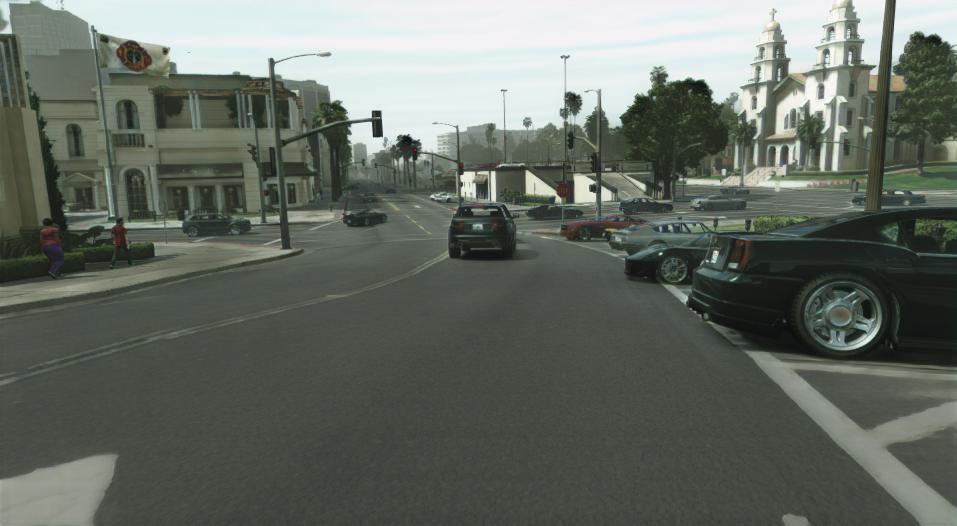}}\hfill 
	\vspace{-2pt}
	\caption{\textbf{Qualitative comparison to EPE.} We compare our method with the provided inferred images of EPE \cite{richter2022enhancing}.}
	\label{fig:qualitative_comparison_epe_additional}
\end{figure*}

\begin{figure*}[h] 
	\centering
	\renewcommand{\thesubfigure}{}
	{\includegraphics[width=0.33\textwidth]{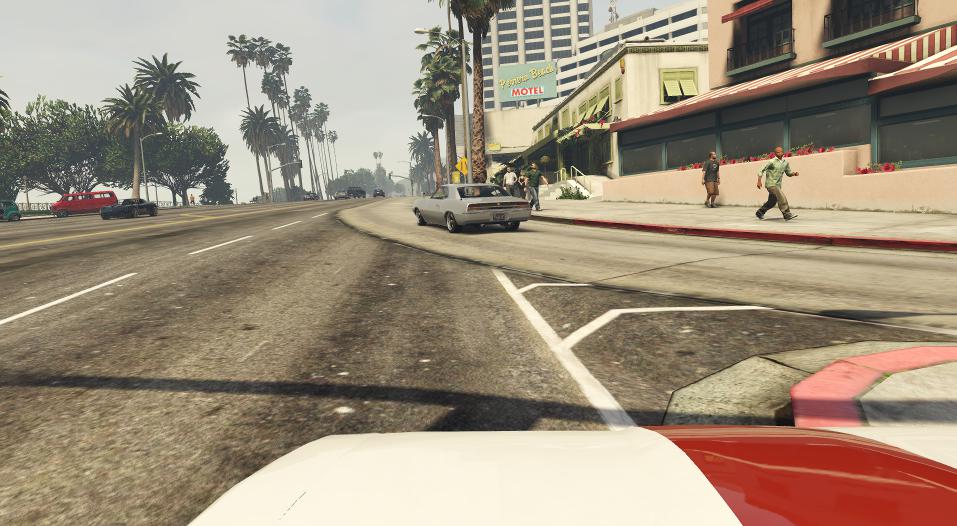}}\hfill
	{\includegraphics[width=0.33\textwidth]{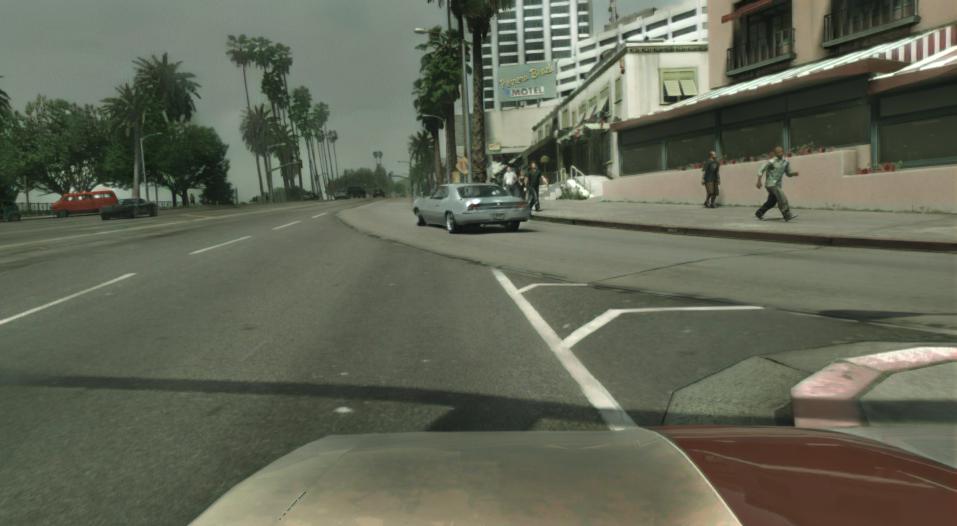}}\hfill
	{\includegraphics[width=0.33\textwidth]{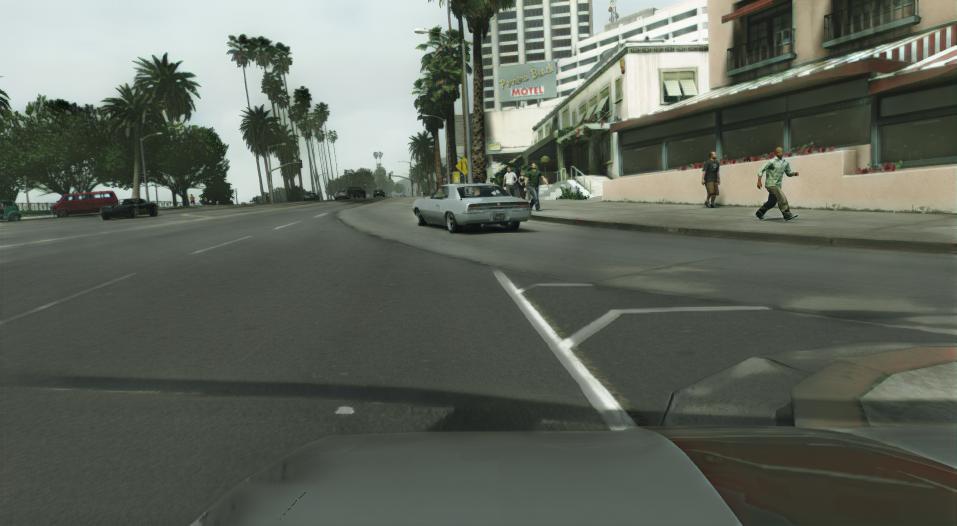}}\hfill\\\vspace{2.5pt}
	{\includegraphics[width=0.33\textwidth]{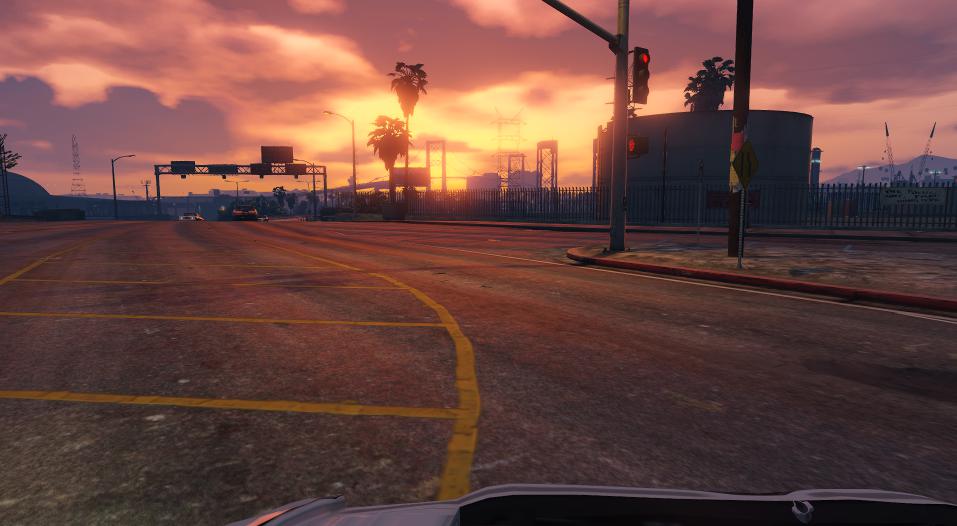}}\hfill
	{\includegraphics[width=0.33\textwidth]{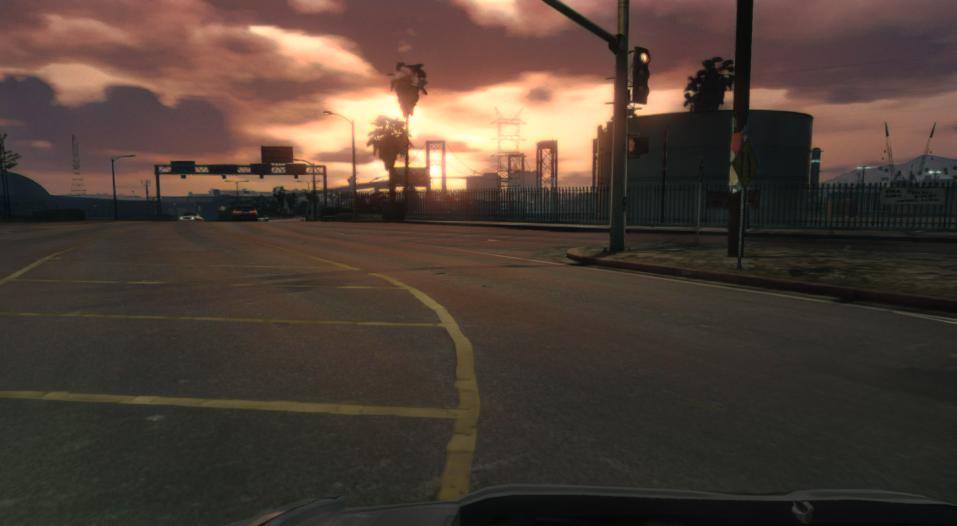}}\hfill
	{\includegraphics[width=0.33\textwidth]{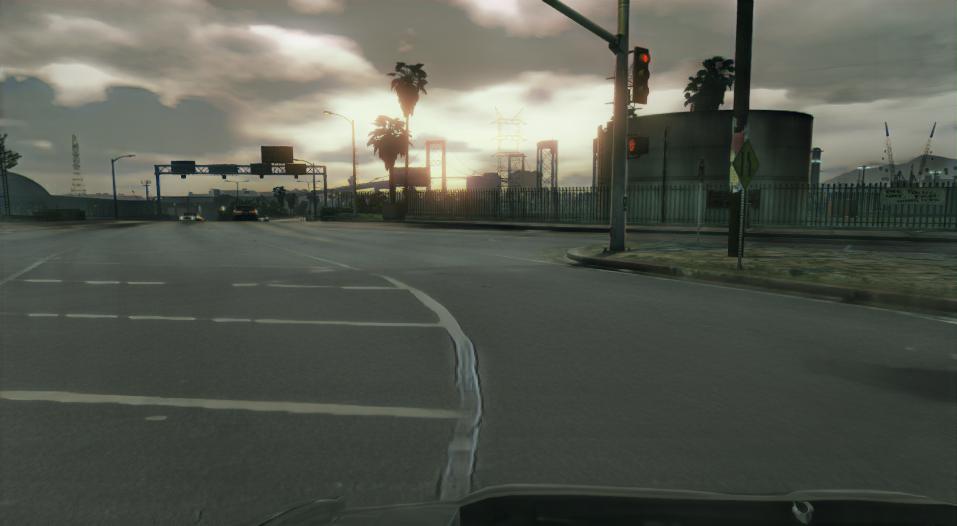}}\hfill\\\vspace{2.5pt}
	{\includegraphics[width=0.33\textwidth]{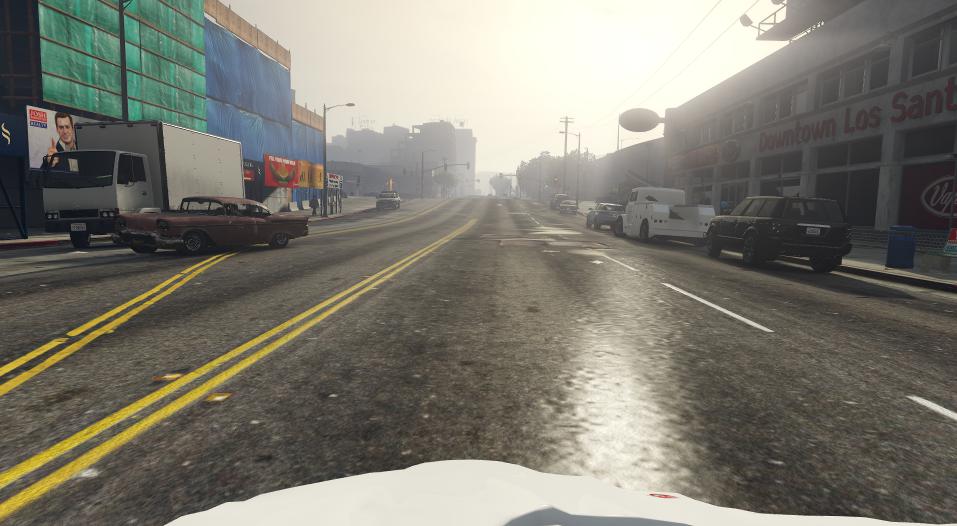}}\hfill
	{\includegraphics[width=0.33\textwidth]{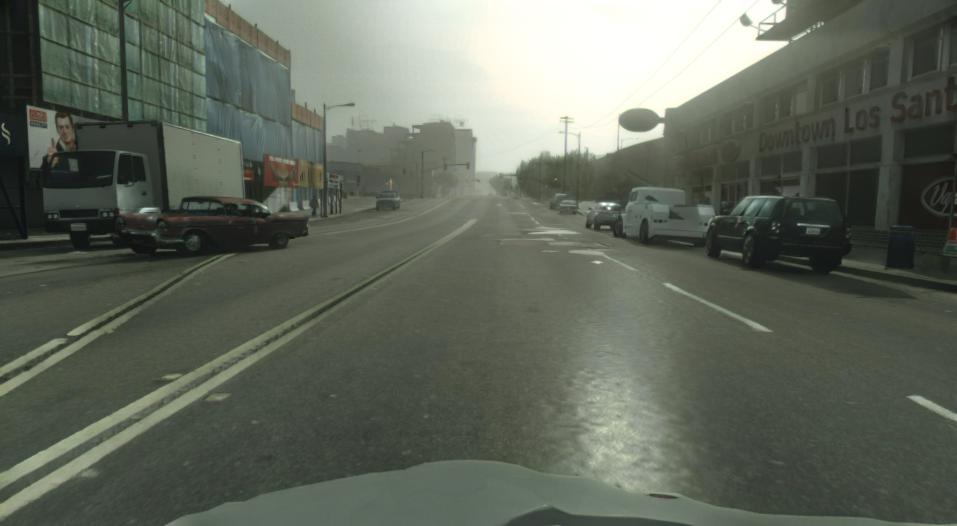}}\hfill
	{\includegraphics[width=0.33\textwidth]{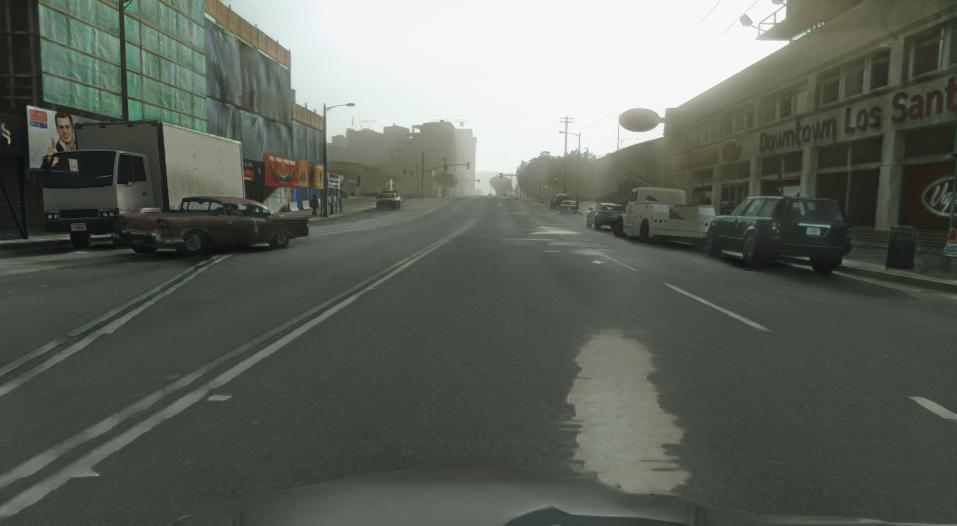}}\hfill\\\vspace{2.5pt}
	{\includegraphics[width=0.33\textwidth]{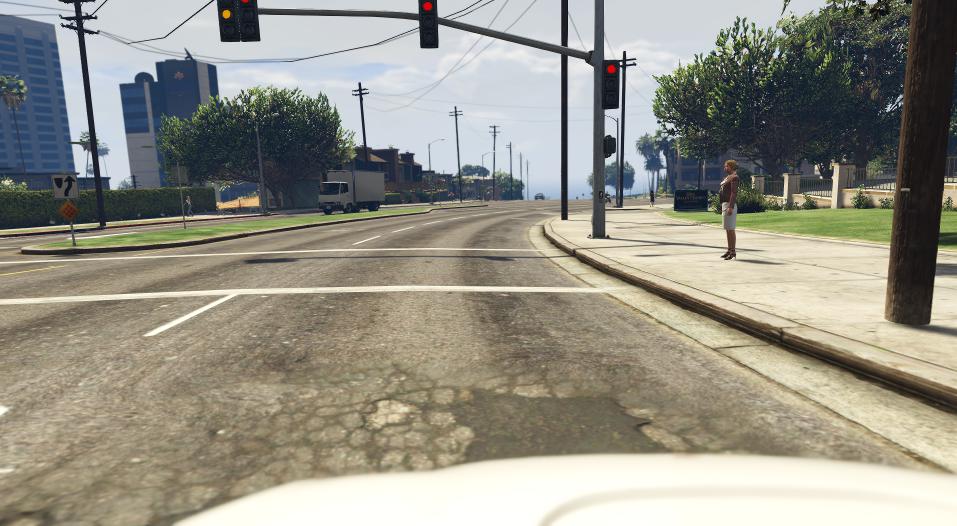}}\hfill
	{\includegraphics[width=0.33\textwidth]{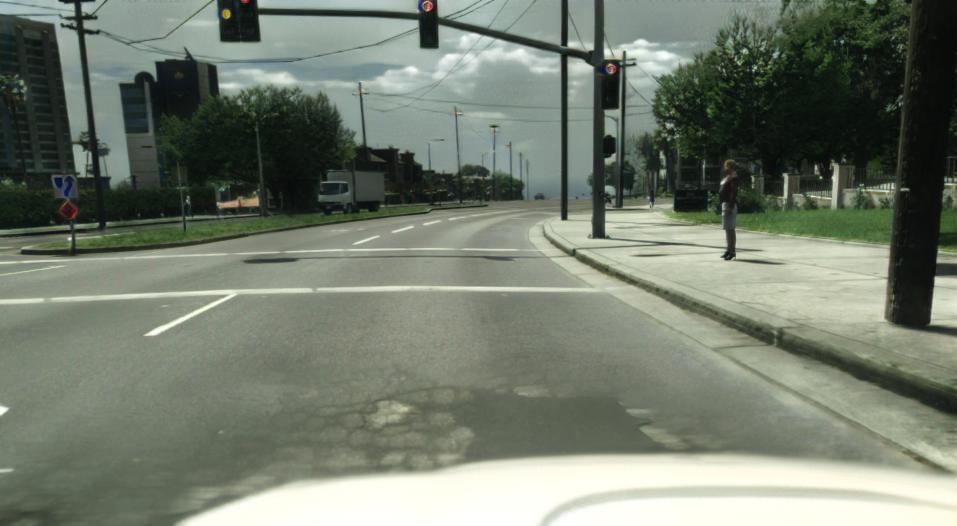}}\hfill
	{\includegraphics[width=0.33\textwidth]{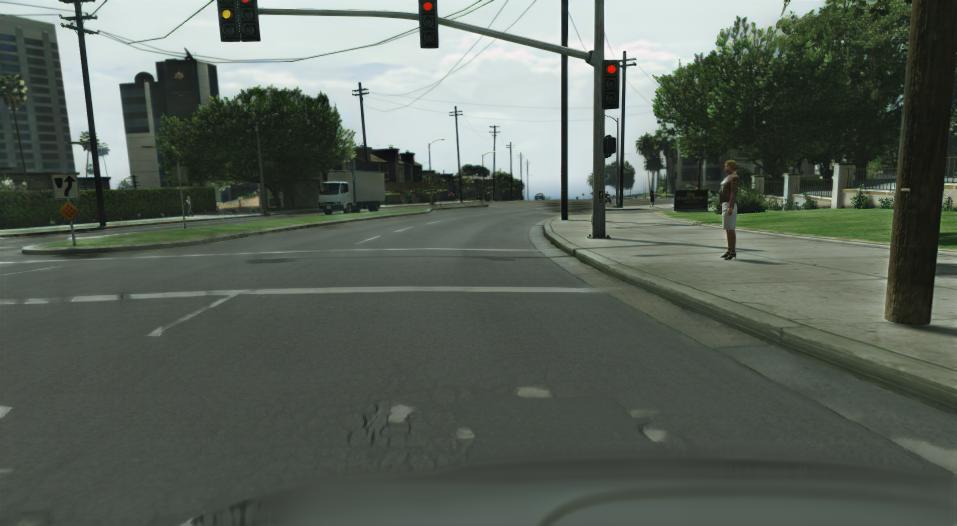}}\hfill\\\vspace{2.5pt}
	{\includegraphics[width=0.33\textwidth]{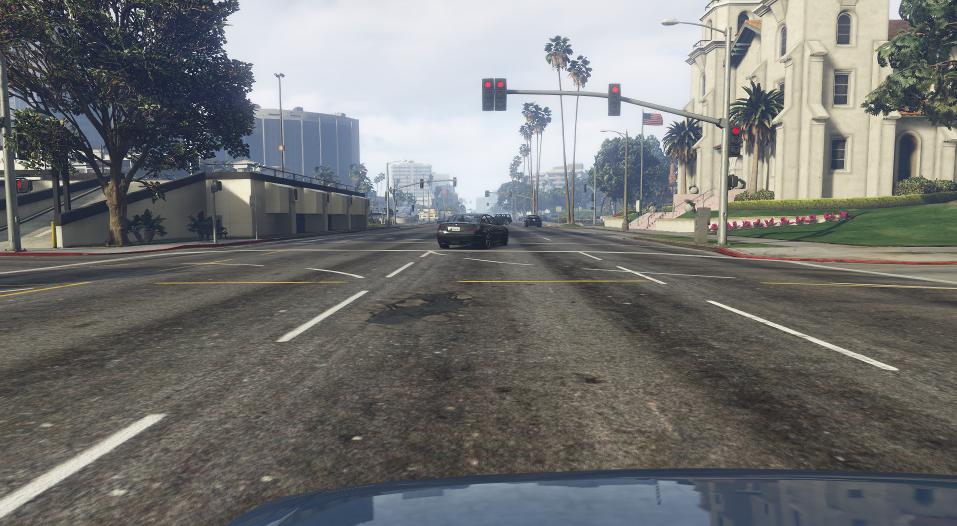}}\hfill
	{\includegraphics[width=0.33\textwidth]{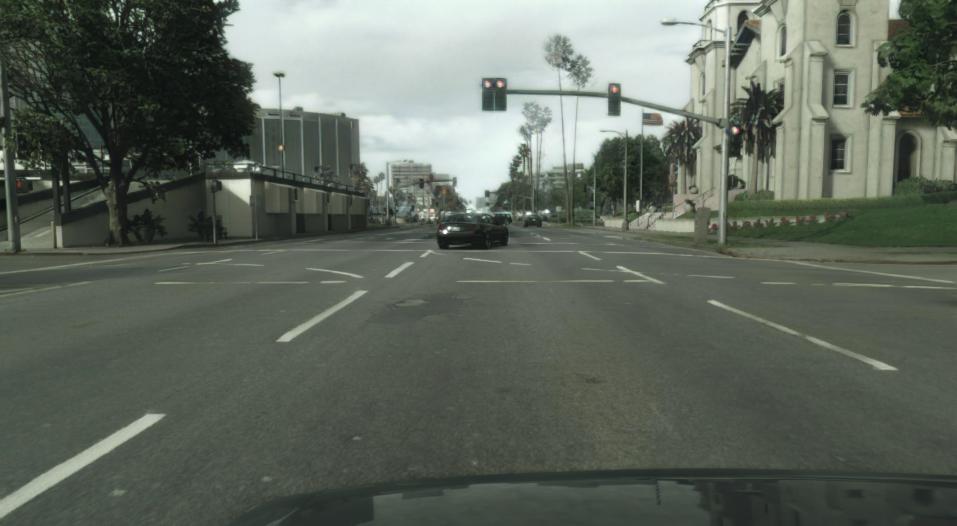}}\hfill
	{\includegraphics[width=0.33\textwidth]{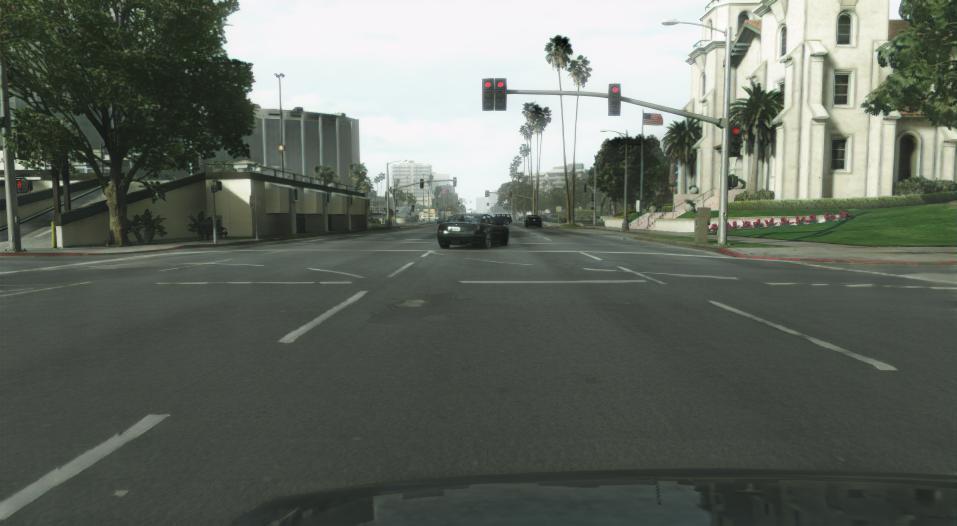}}\hfill\\\vspace{-2.5pt}
	\subfigure[Input]
	{\includegraphics[width=0.33\textwidth]{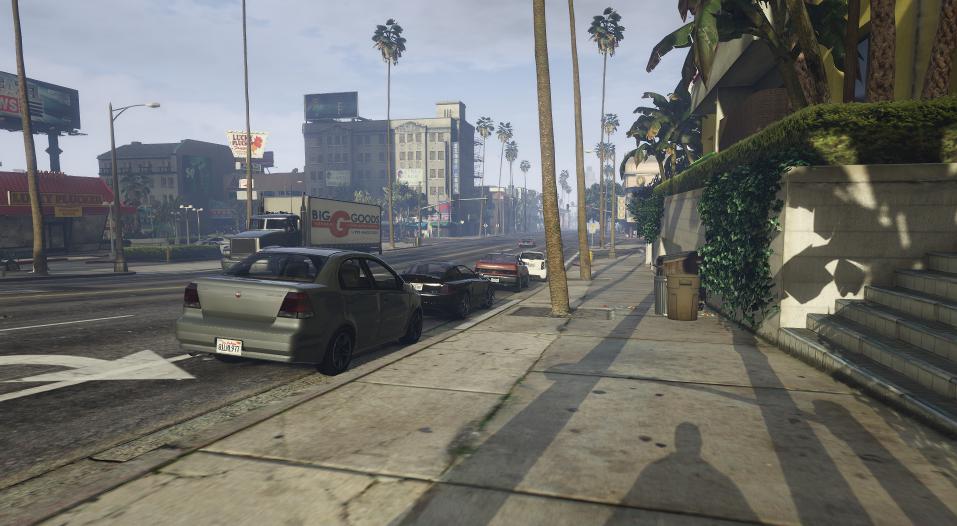}}\hfill
	\subfigure[EPE]
	{\includegraphics[width=0.33\textwidth]{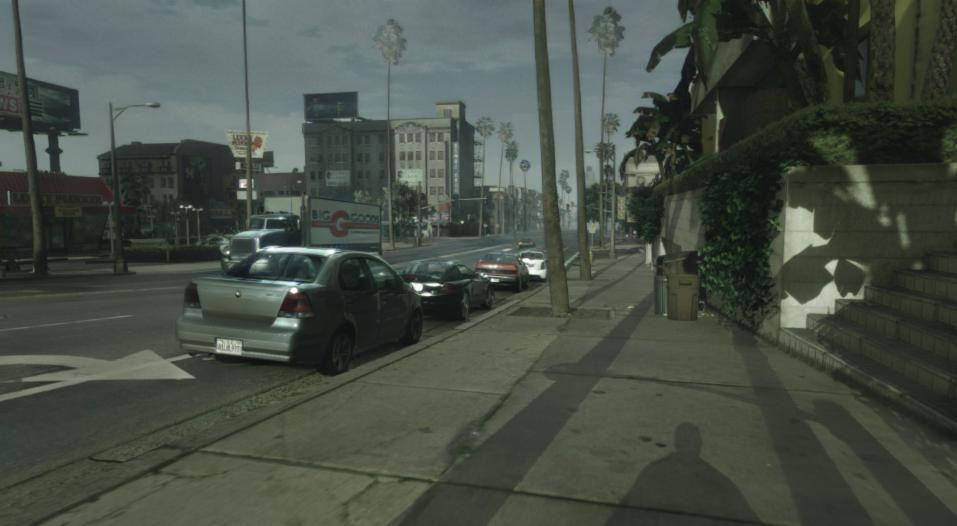}}\hfill
	\subfigure[FeaMGAN (ours)]
	{\includegraphics[width=0.33\textwidth]{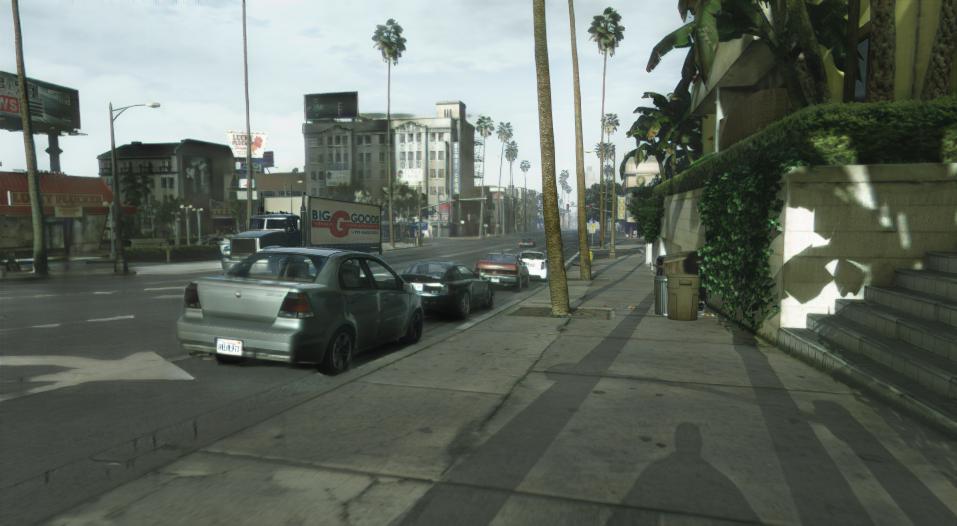}}\hfill 
	\vspace{-2pt}
	\caption{\textbf{Qualitative comparison to EPE.} We compare our method with the provided inferred images of EPE \cite{richter2022enhancing}. Results are randomly sampled from the best model.}
	\label{fig:qualitative_comparison_epe_additional_random}
\end{figure*}

\begin{figure*}[h] 
	\renewcommand{\thesubfigure}{}
	\begin{center}
		{\scriptsize PFD$\rightarrow$Cityscapes} \hfill\\\vspace{1pt}
		{\includegraphics[width=0.165\textwidth]{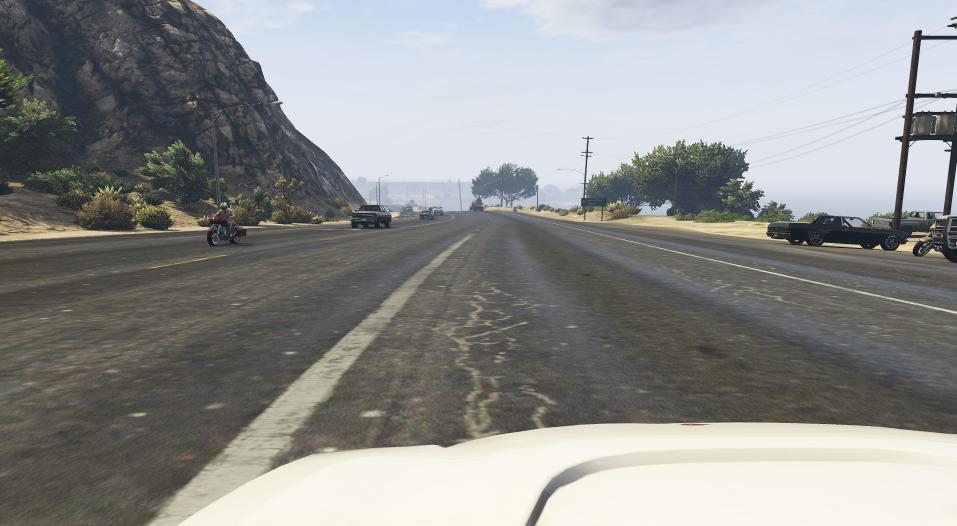}}\hfill
		{\includegraphics[width=0.165\textwidth]{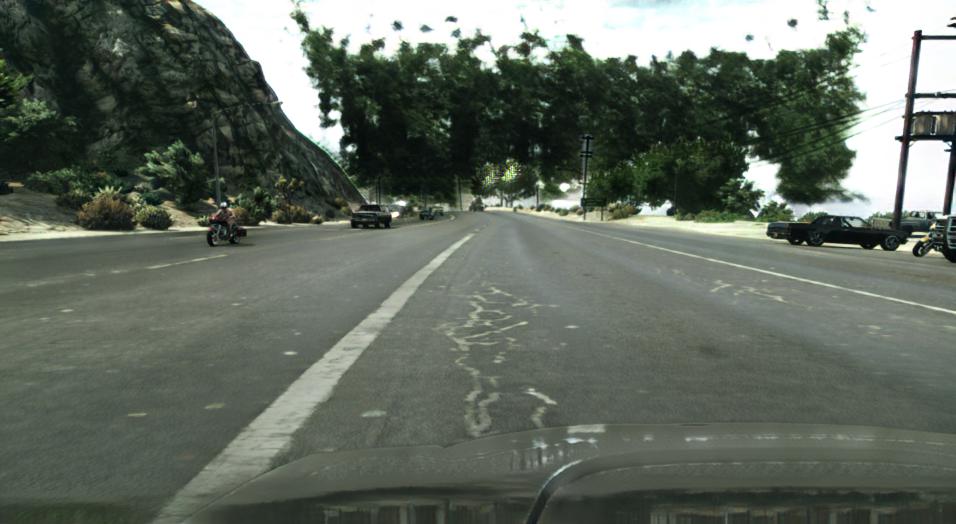}}\hfill
		{\includegraphics[width=0.165\textwidth]{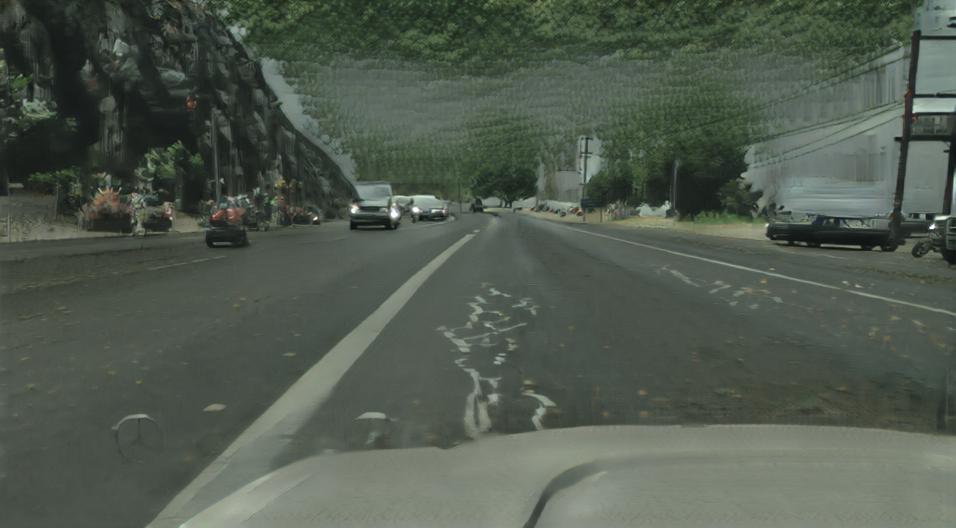}}\hfill
		{\includegraphics[width=0.165\textwidth]{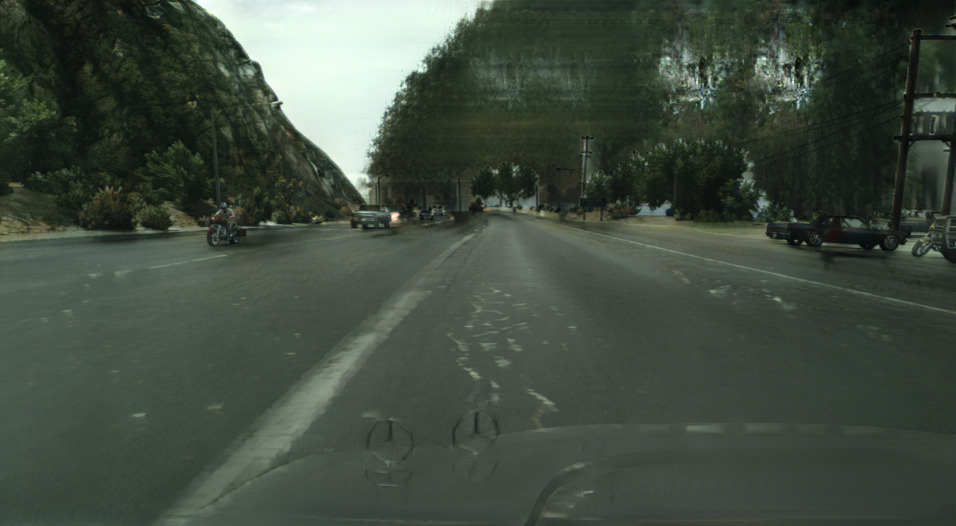}}\hfill
		{\includegraphics[width=0.165\textwidth]{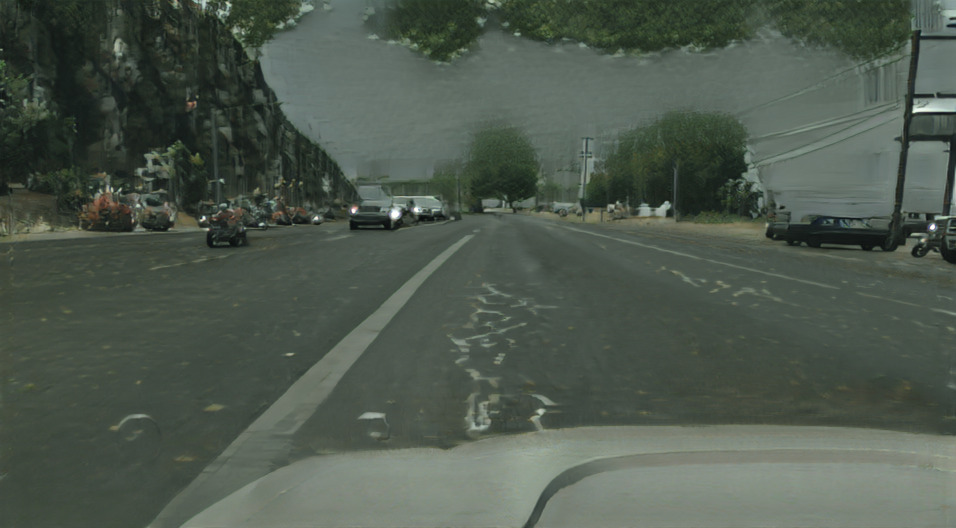}}\hfill
		{\includegraphics[width=0.165\textwidth]{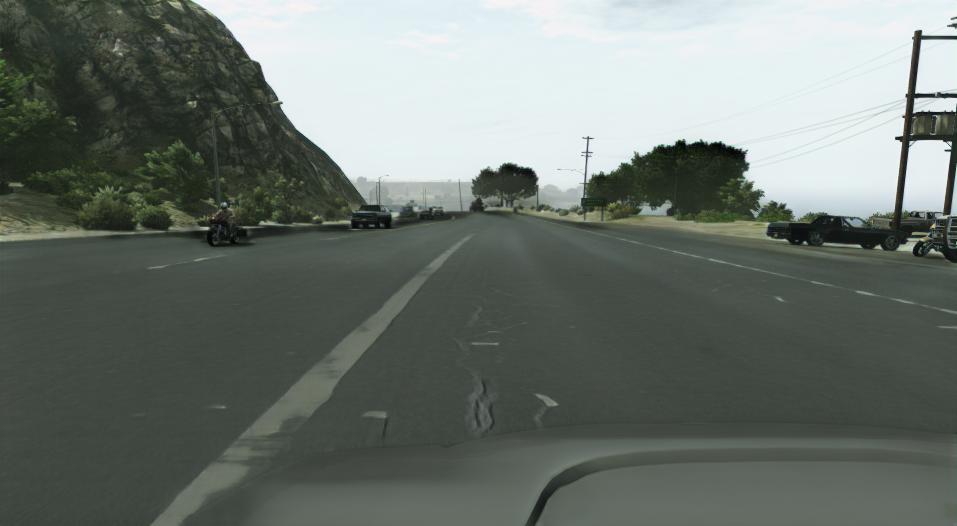}}\hfill\\\vspace{1pt}
		{\includegraphics[width=0.165\textwidth]{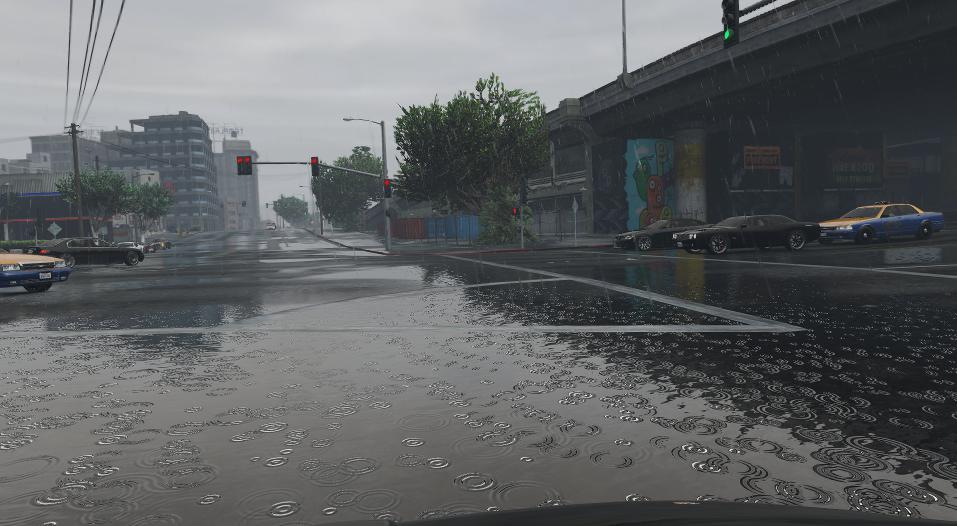}}\hfill
		{\includegraphics[width=0.165\textwidth]{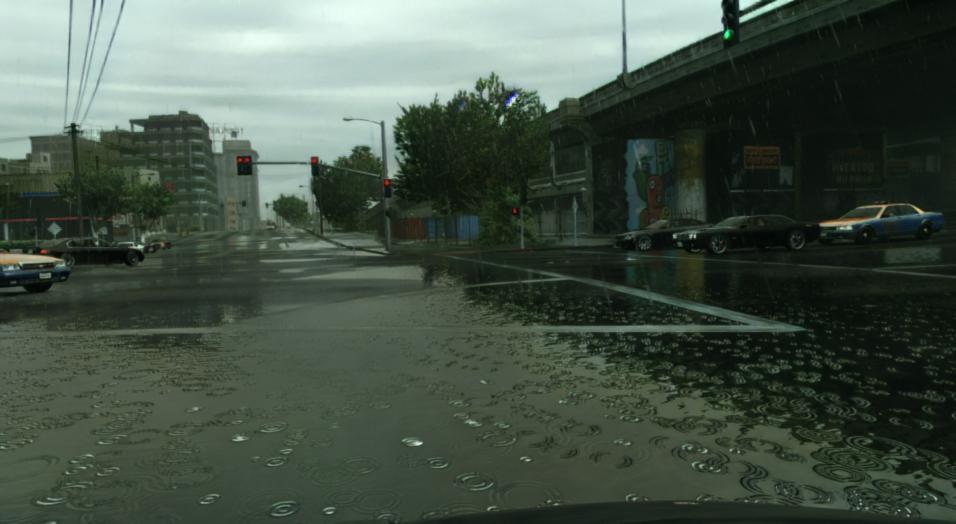}}\hfill
		{\includegraphics[width=0.165\textwidth]{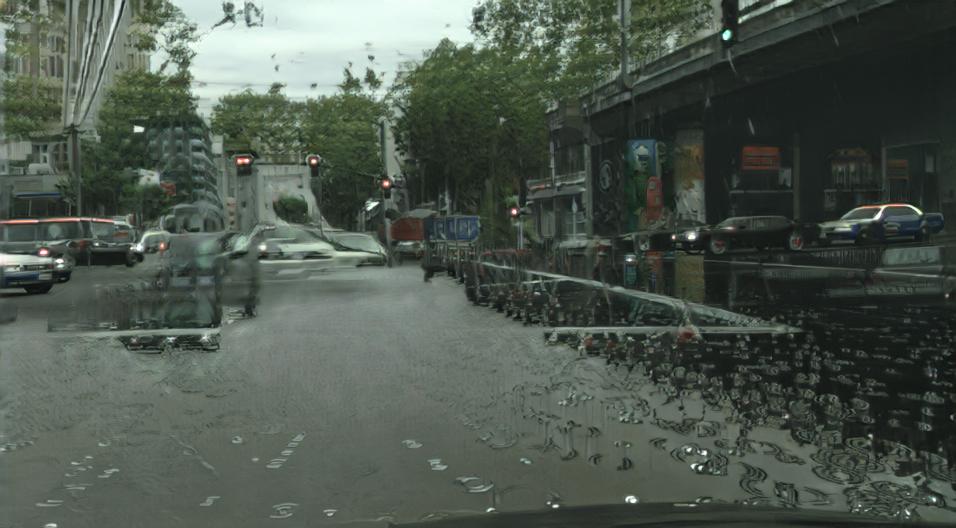}}\hfill
		{\includegraphics[width=0.165\textwidth]{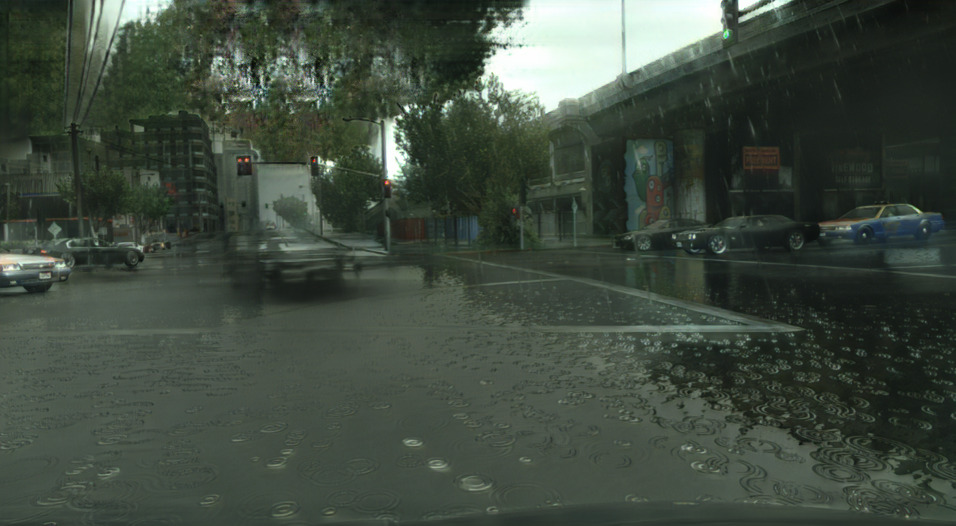}}\hfill
		{\includegraphics[width=0.165\textwidth]{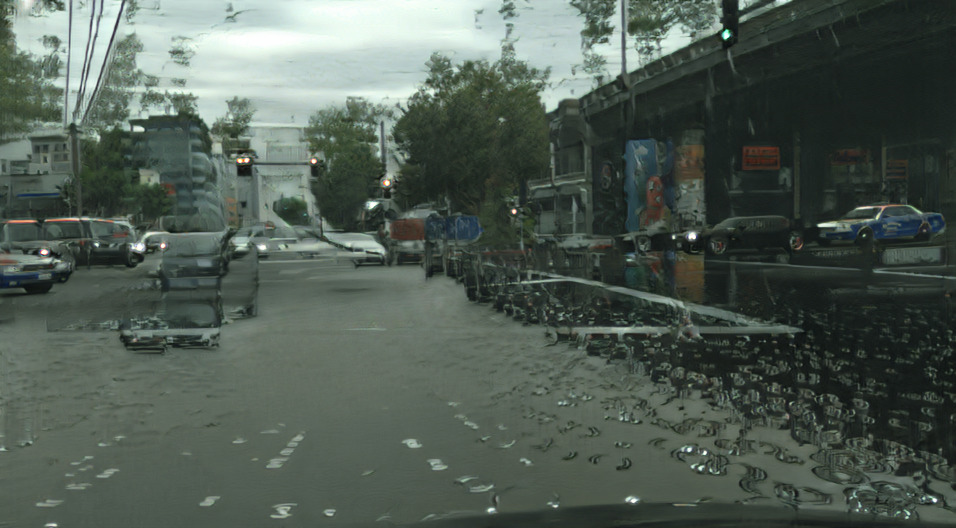}}\hfill
		{\includegraphics[width=0.165\textwidth]{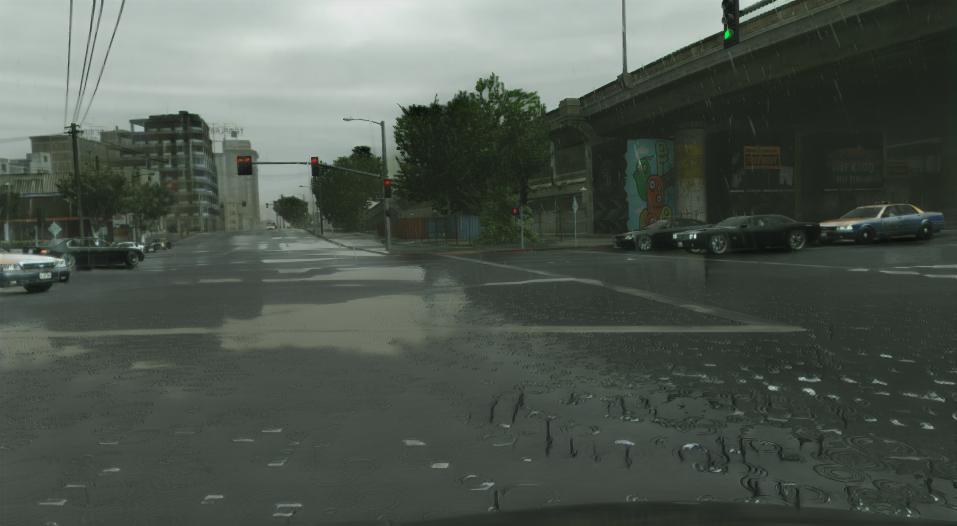}}\hfill\\\vspace{1pt}
		{\includegraphics[width=0.165\textwidth]{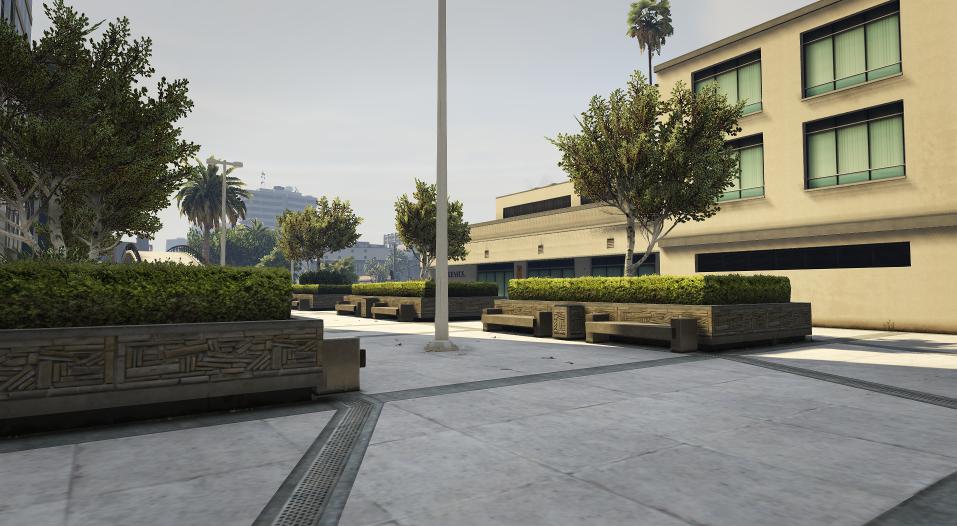}}\hfill
		{\includegraphics[width=0.165\textwidth]{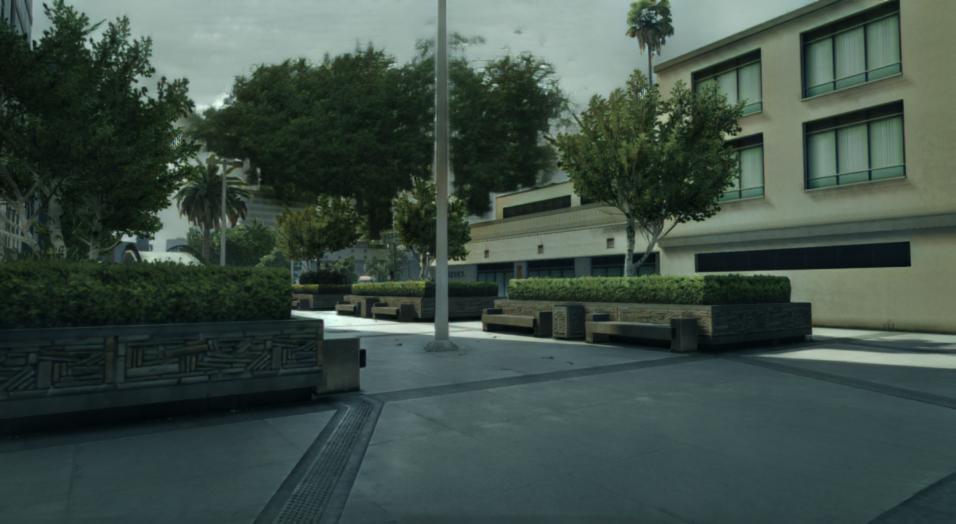}}\hfill
		{\includegraphics[width=0.165\textwidth]{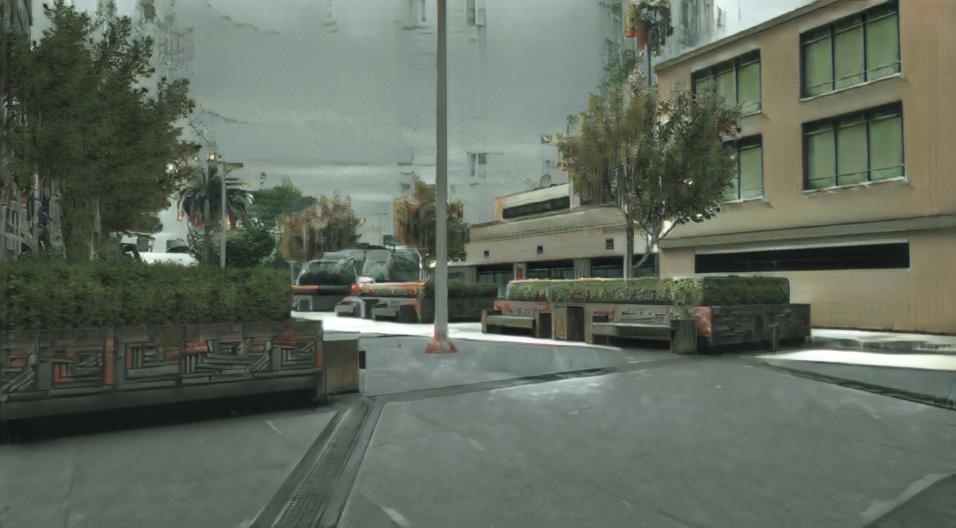}}\hfill
		{\includegraphics[width=0.165\textwidth]{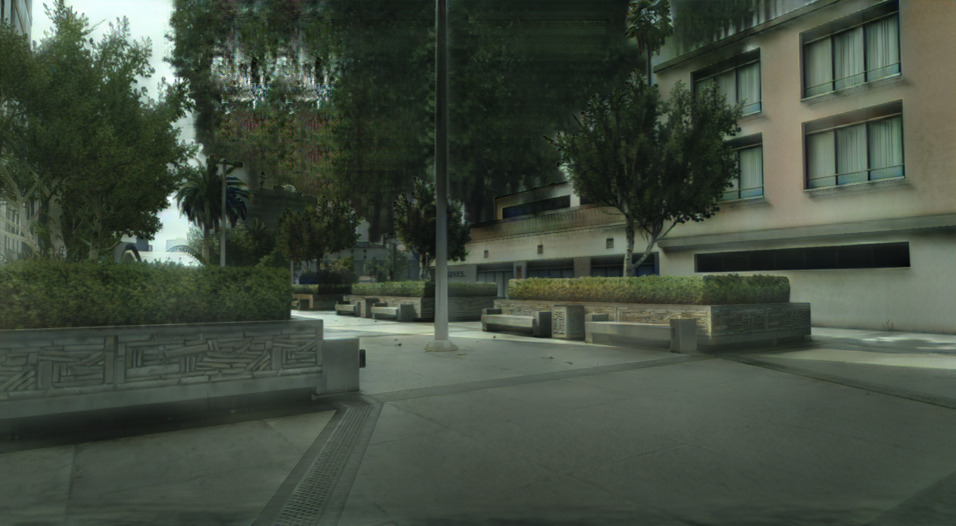}}\hfill
		{\includegraphics[width=0.165\textwidth]{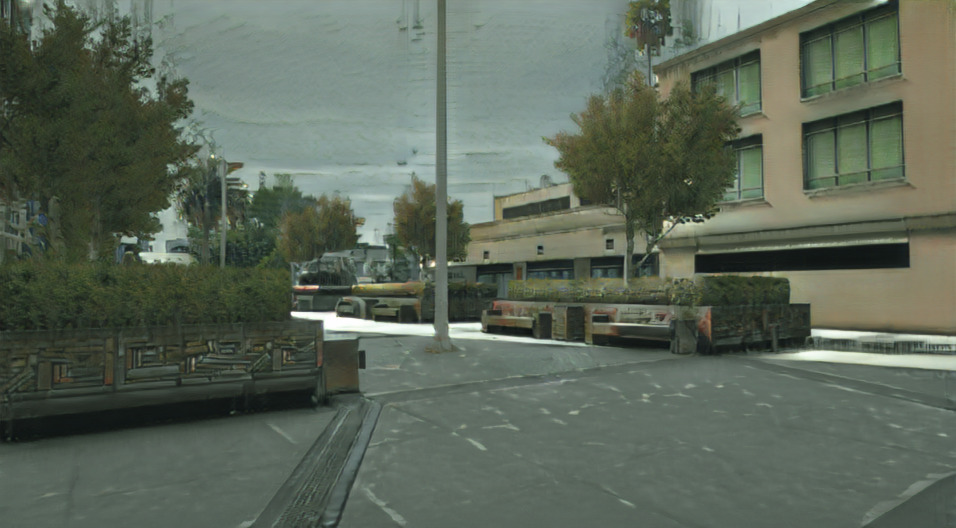}}\hfill
		{\includegraphics[width=0.165\textwidth]{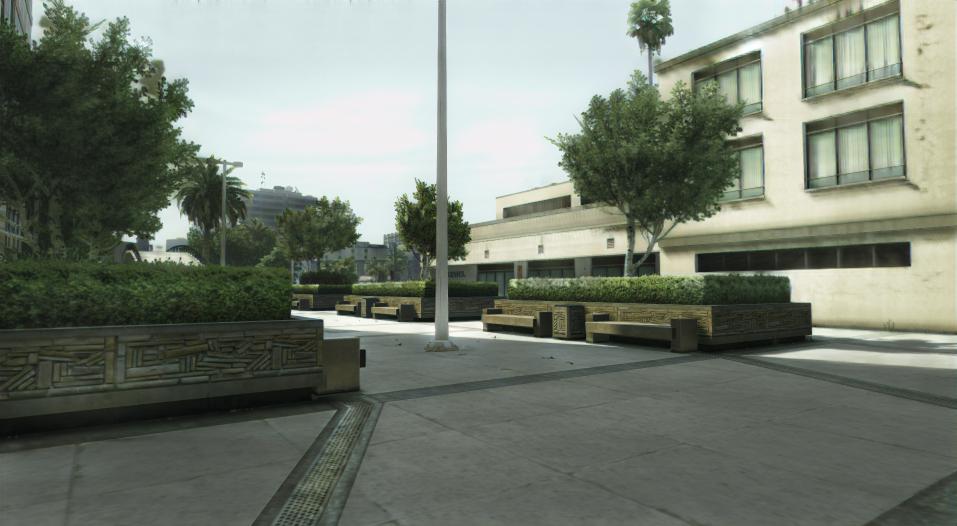}}\hfill\\\vspace{1pt}

		{\scriptsize Viper$\rightarrow$Cityscapes} \hfill\\\vspace{1pt}		
		{\includegraphics[width=0.165\textwidth]{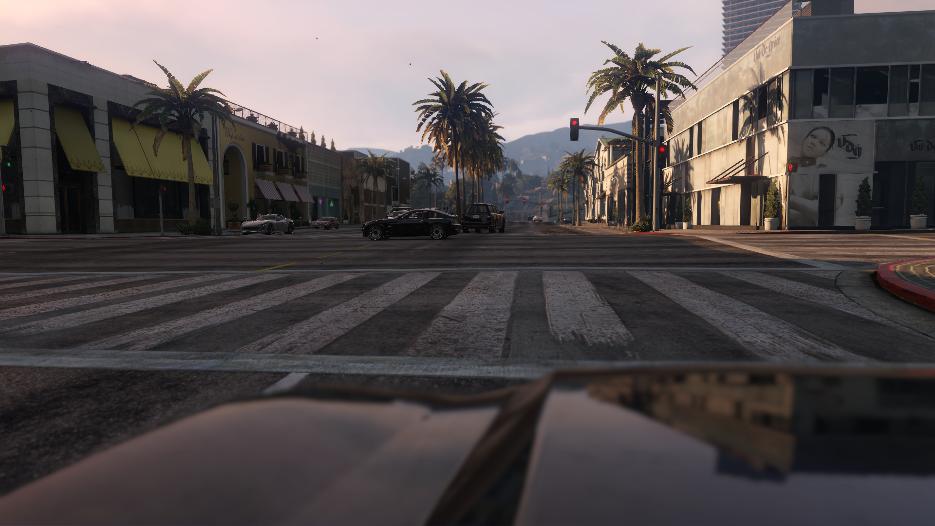}}\hfill
		{\includegraphics[width=0.165\textwidth]{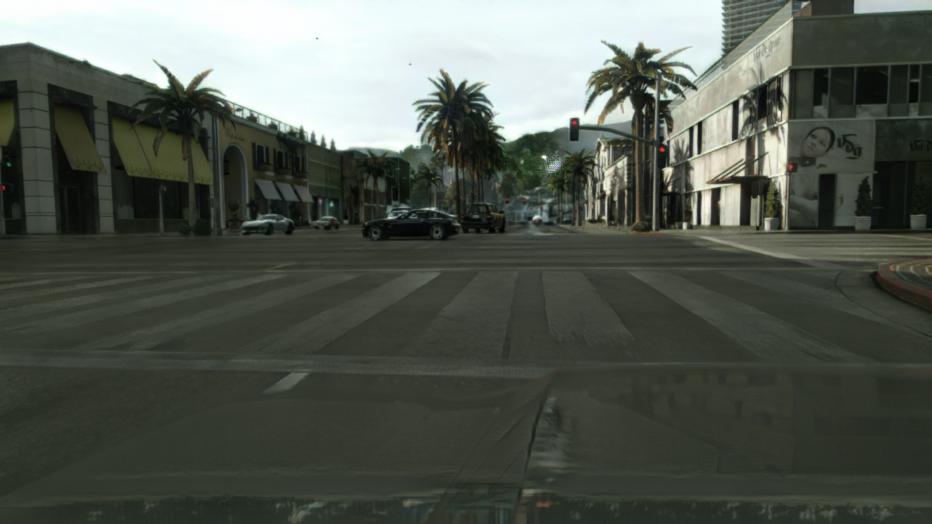}}\hfill
		{\includegraphics[width=0.165\textwidth]{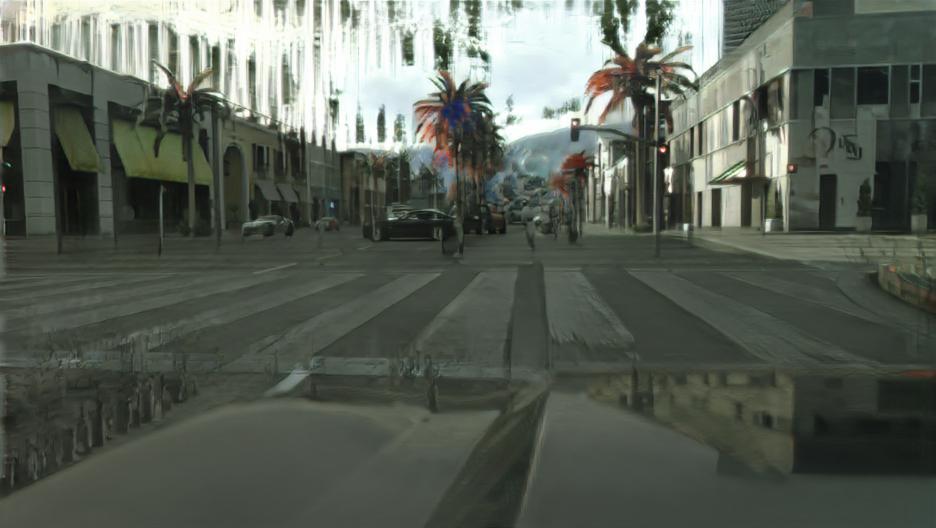}}\hfill
		{\includegraphics[width=0.165\textwidth]{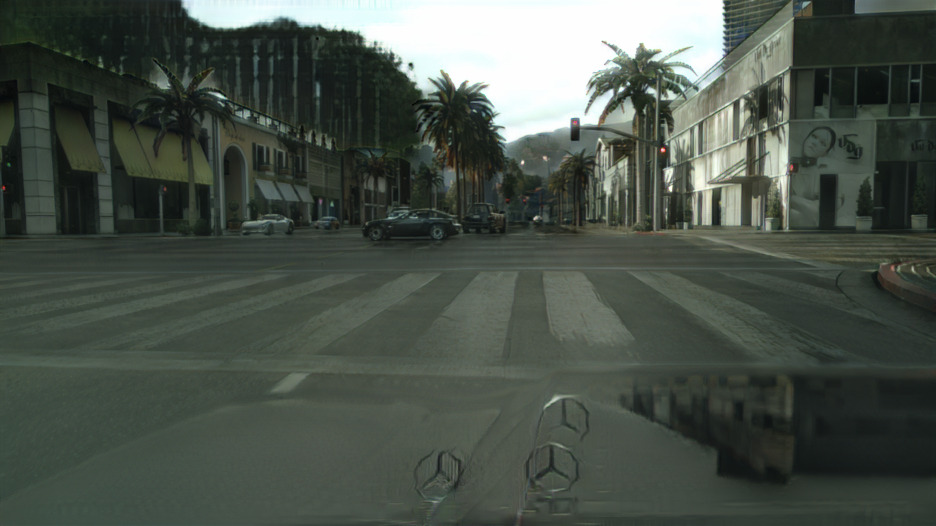}}\hfill
		{\includegraphics[width=0.165\textwidth]{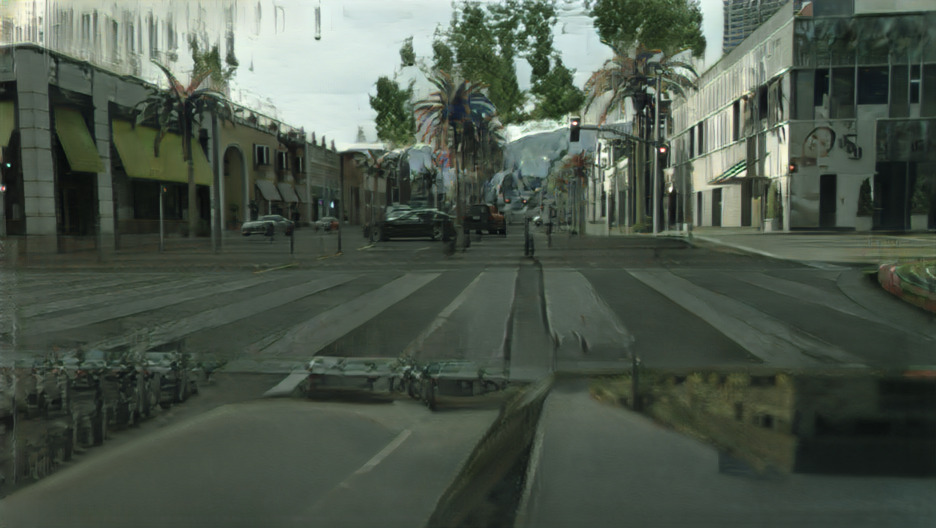}}\hfill
		{\includegraphics[width=0.165\textwidth]{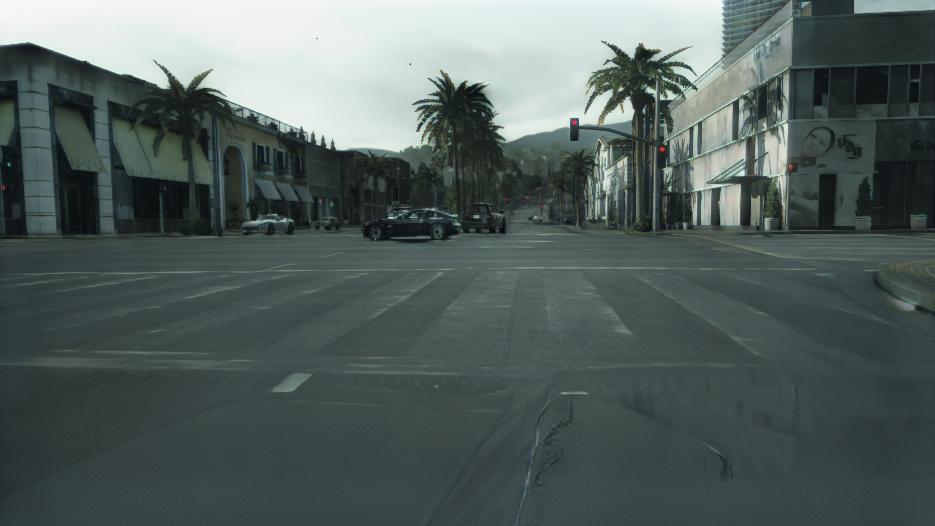}}\hfill\\\vspace{1pt}
		{\includegraphics[width=0.165\textwidth]{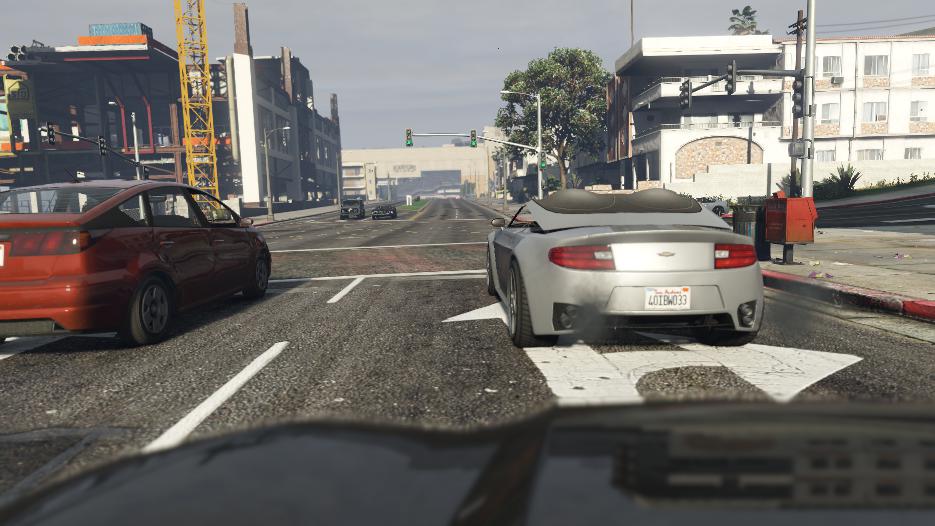}}\hfill
		{\includegraphics[width=0.165\textwidth]{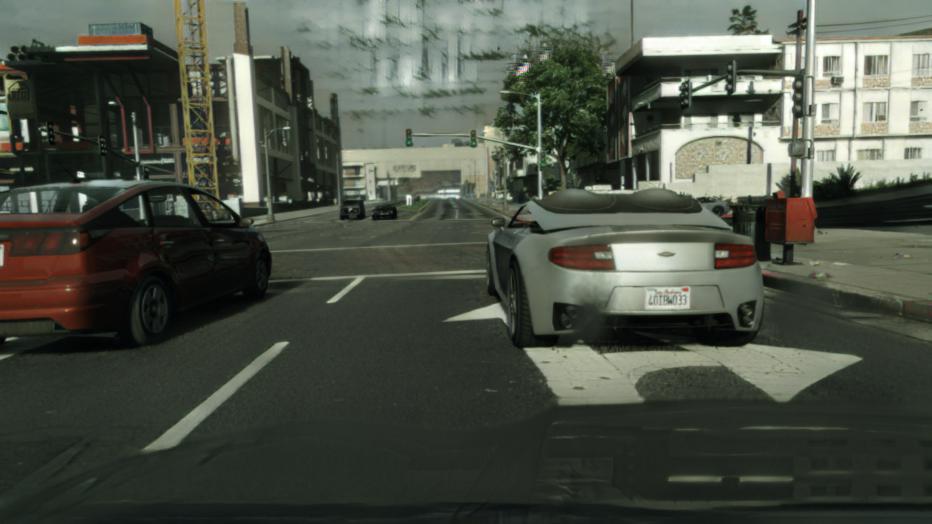}}\hfill
		{\includegraphics[width=0.165\textwidth]{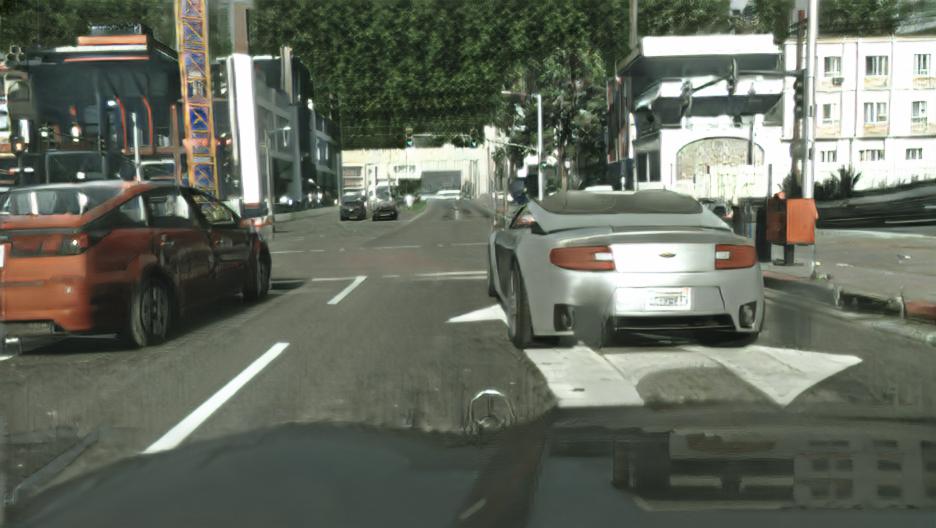}}\hfill
		{\includegraphics[width=0.165\textwidth]{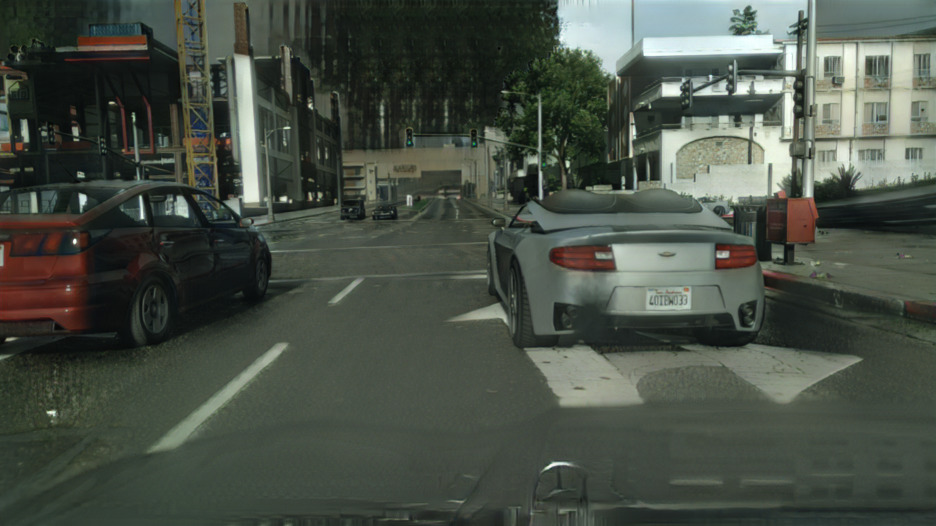}}\hfill
		{\includegraphics[width=0.165\textwidth]{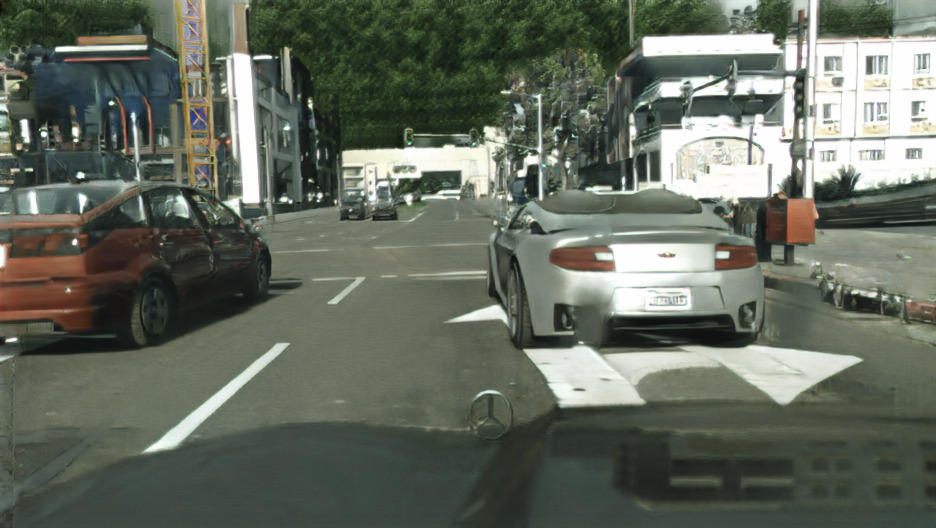}}\hfill
		{\includegraphics[width=0.165\textwidth]{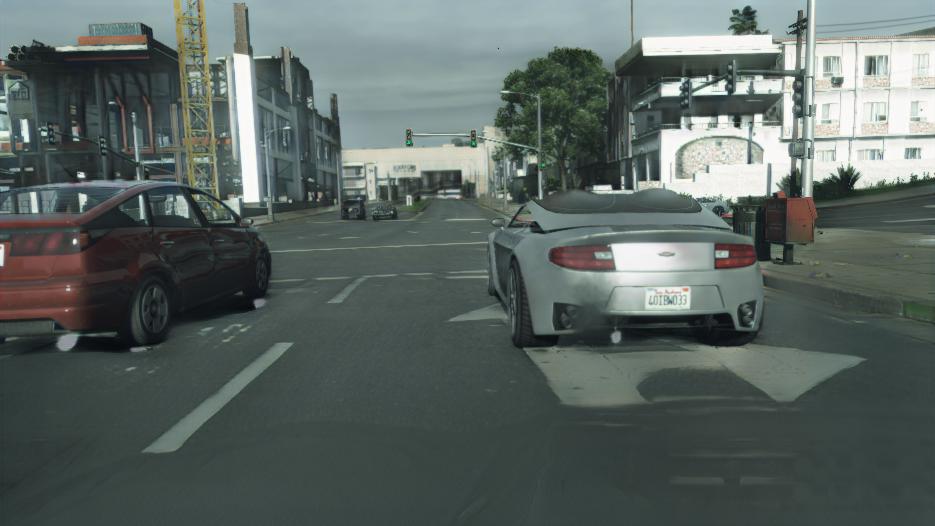}}\hfill\\\vspace{1pt}
		{\includegraphics[width=0.165\textwidth]{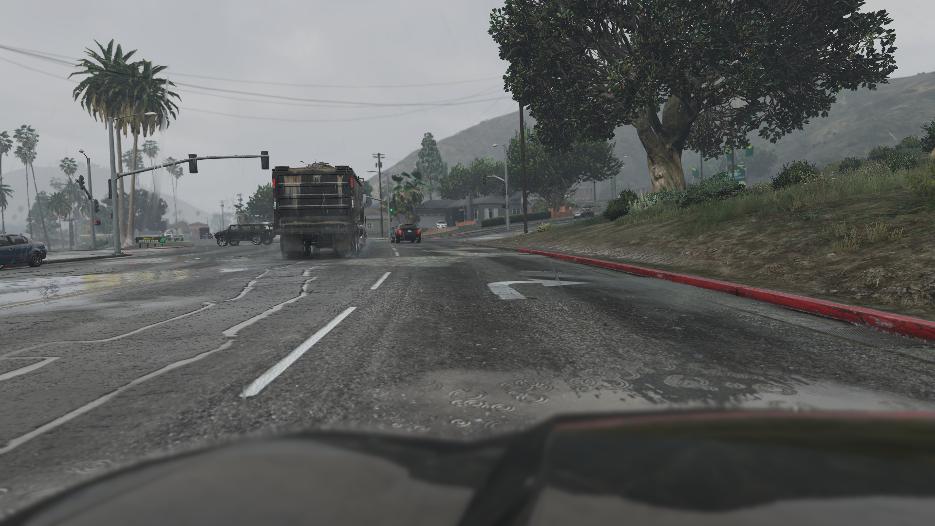}}\hfill
		{\includegraphics[width=0.165\textwidth]{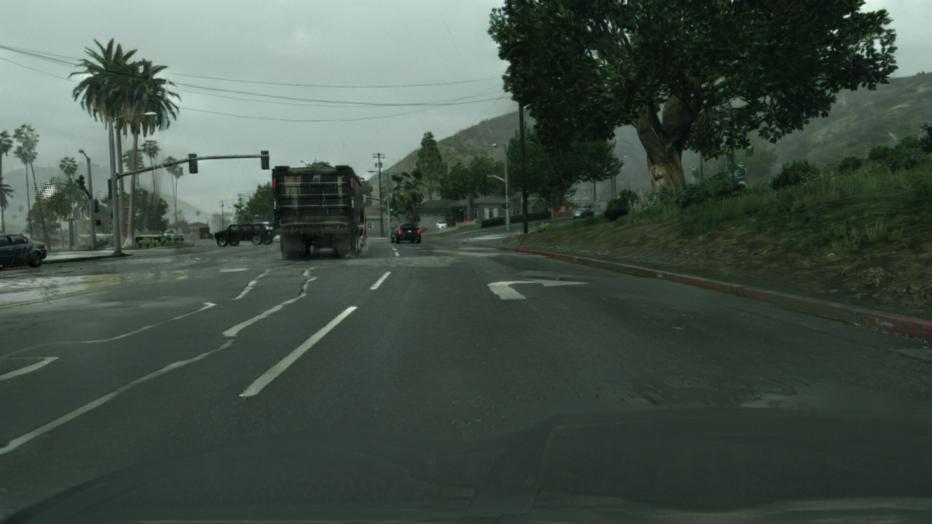}}\hfill
		{\includegraphics[width=0.165\textwidth]{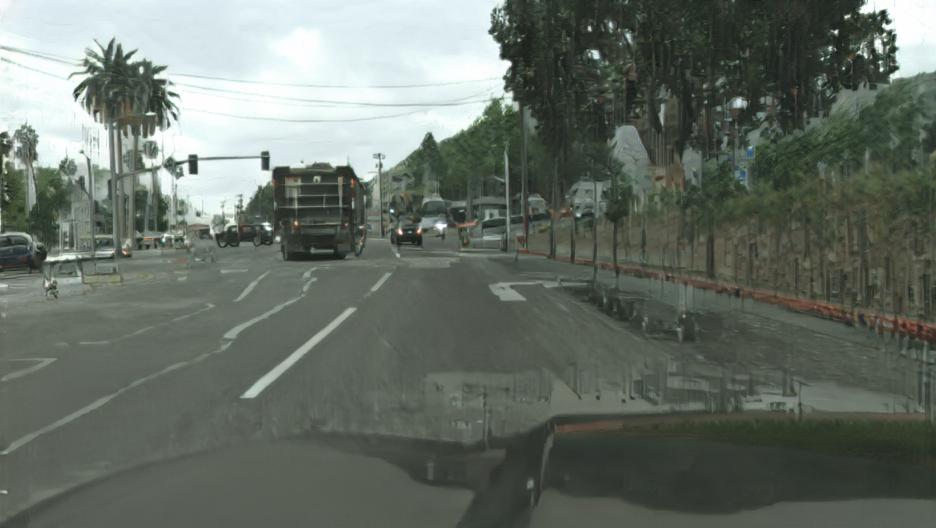}}\hfill
		{\includegraphics[width=0.165\textwidth]{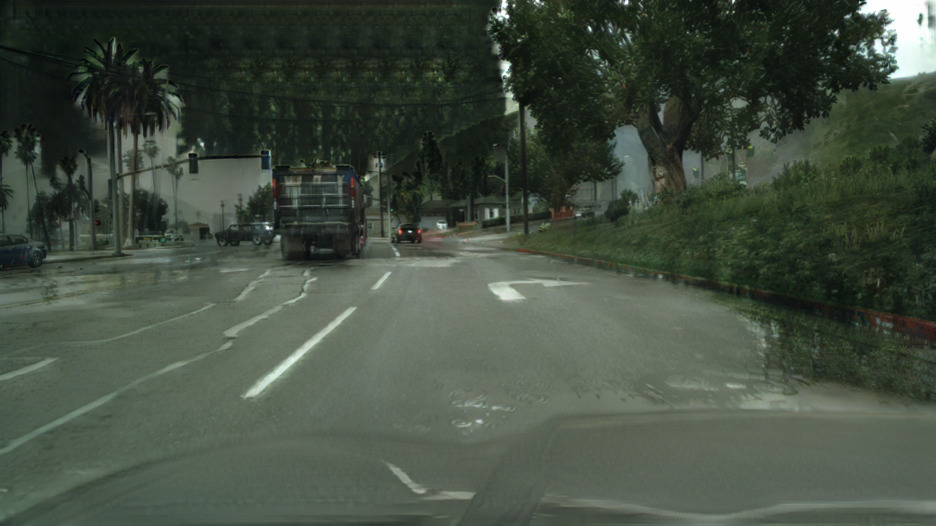}}\hfill
		{\includegraphics[width=0.165\textwidth]{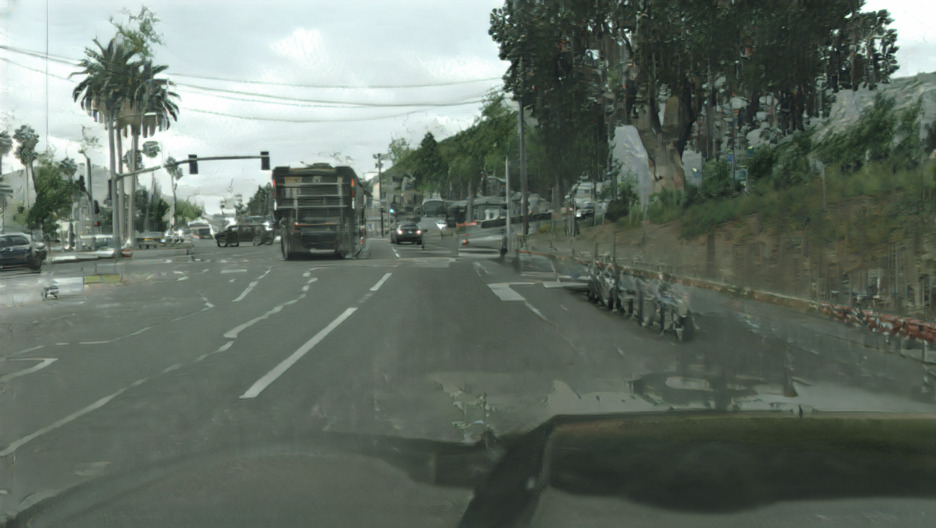}}\hfill
		{\includegraphics[width=0.165\textwidth]{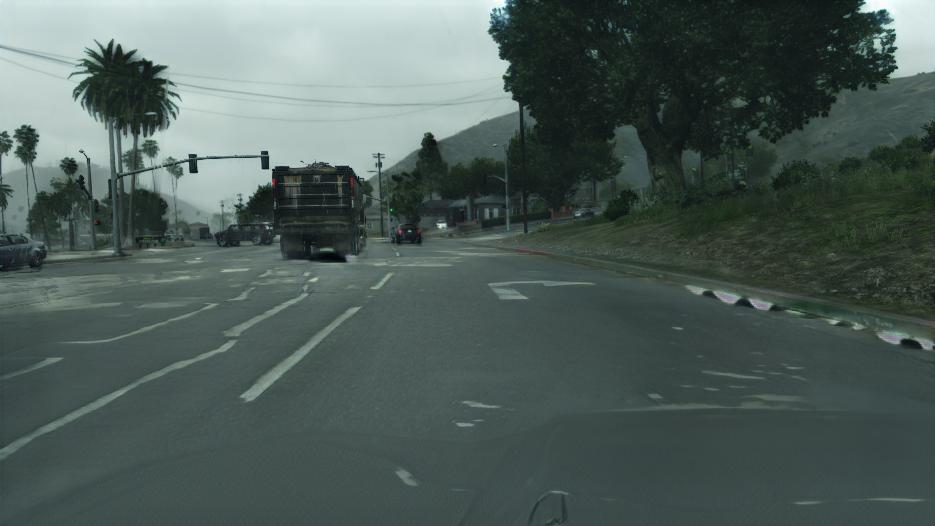}}\hfill\\\vspace{1pt}
		
		{\scriptsize Day$\rightarrow$Night} \hfill\\\vspace{1pt}
		{\includegraphics[width=0.165\textwidth]{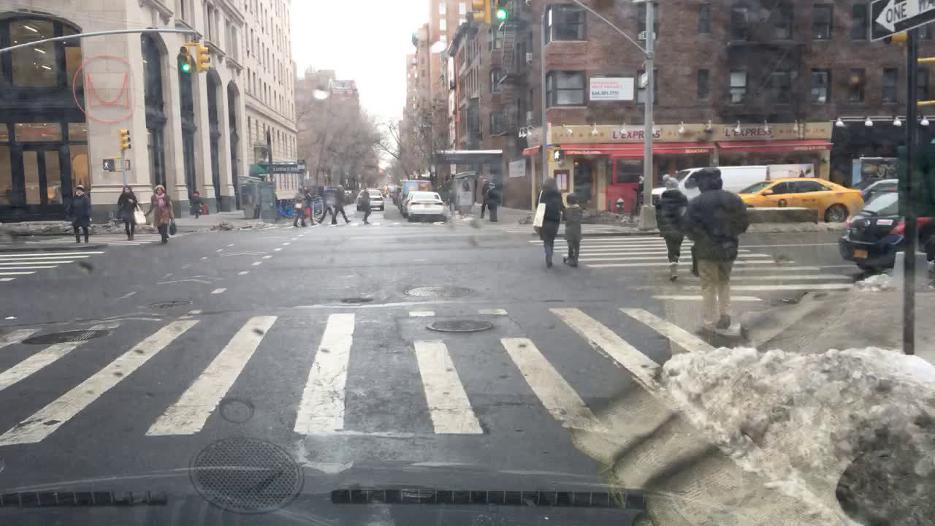}}\hfill
		{\includegraphics[width=0.165\textwidth]{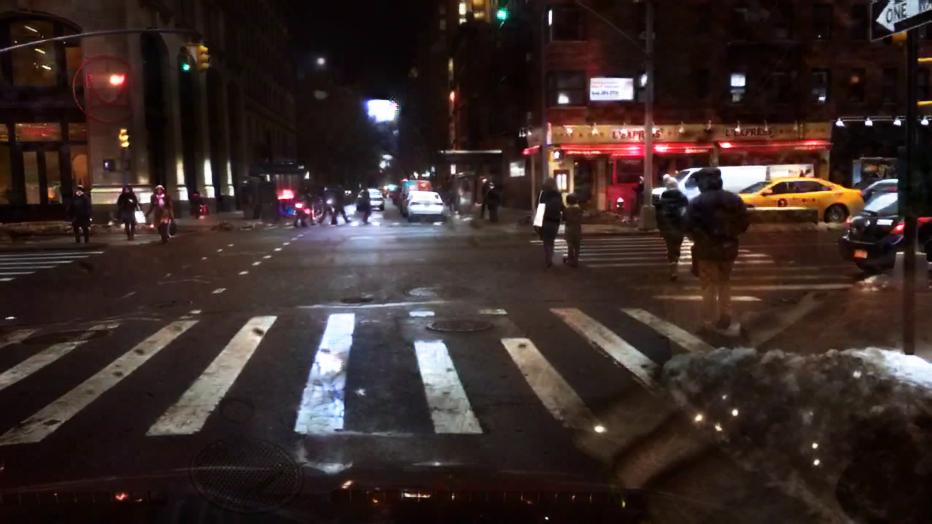}}\hfill
		{\includegraphics[width=0.165\textwidth]{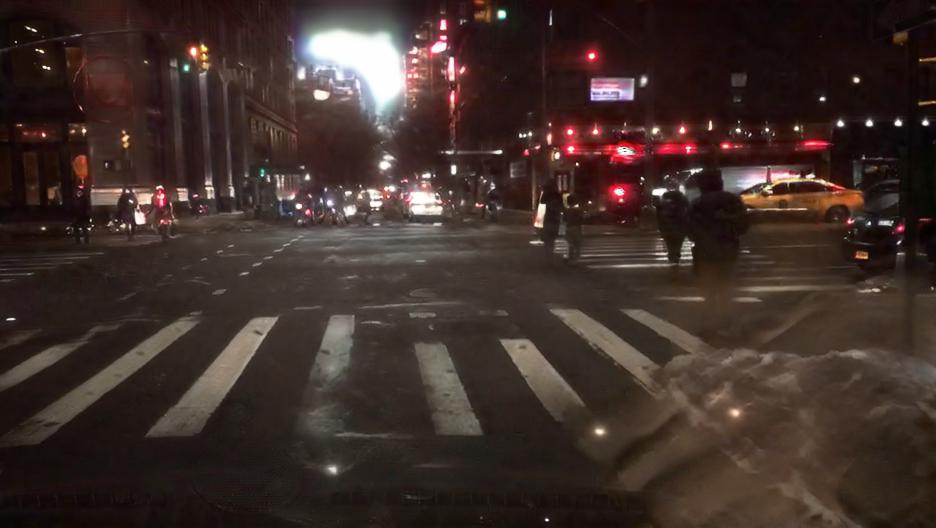}}\hfill
		{\includegraphics[width=0.165\textwidth]{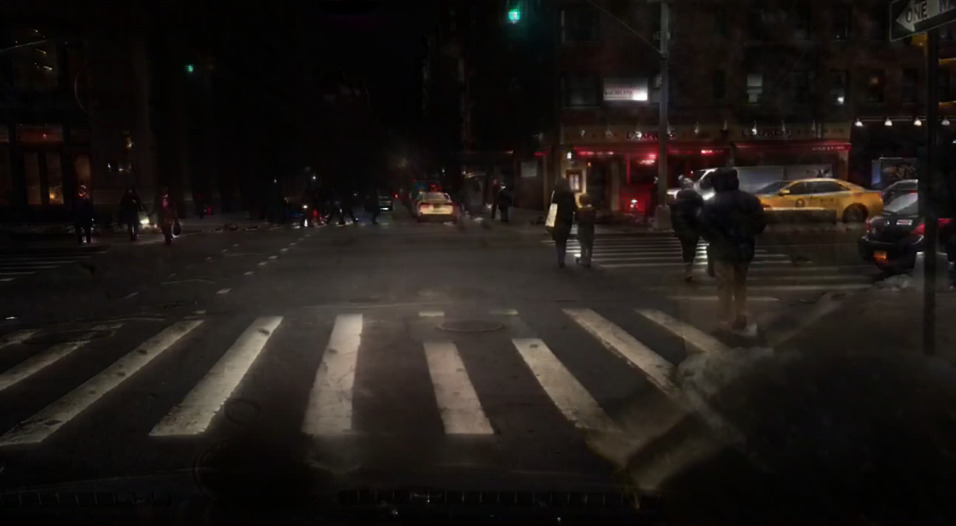}}\hfill
		{\includegraphics[width=0.165\textwidth]{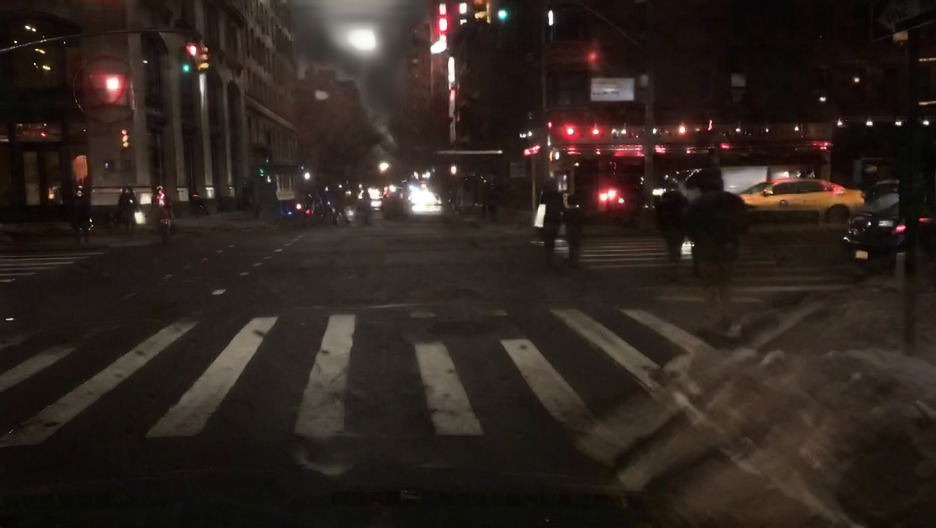}}\hfill
		{\includegraphics[width=0.165\textwidth]{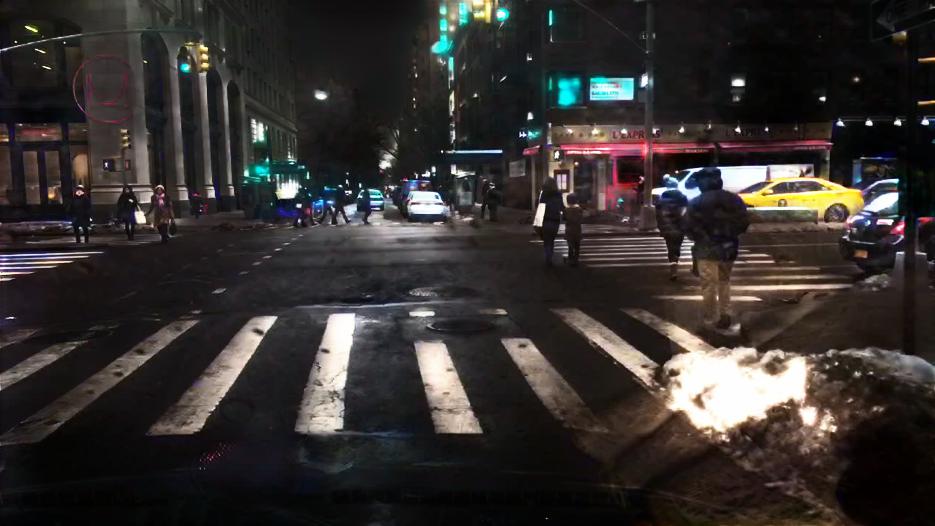}}\hfill\\\vspace{1pt}
		{\includegraphics[width=0.165\textwidth]{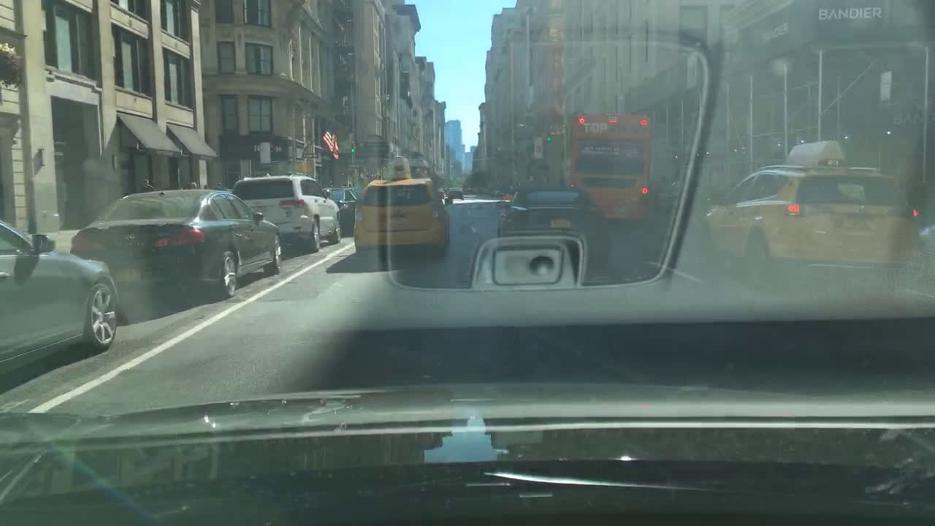}}\hfill
		{\includegraphics[width=0.165\textwidth]{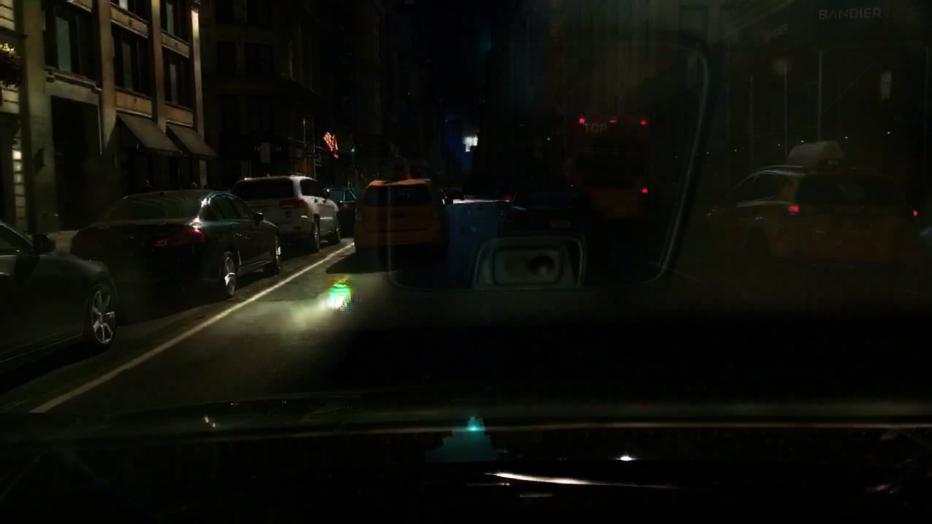}}\hfill
		{\includegraphics[width=0.165\textwidth]{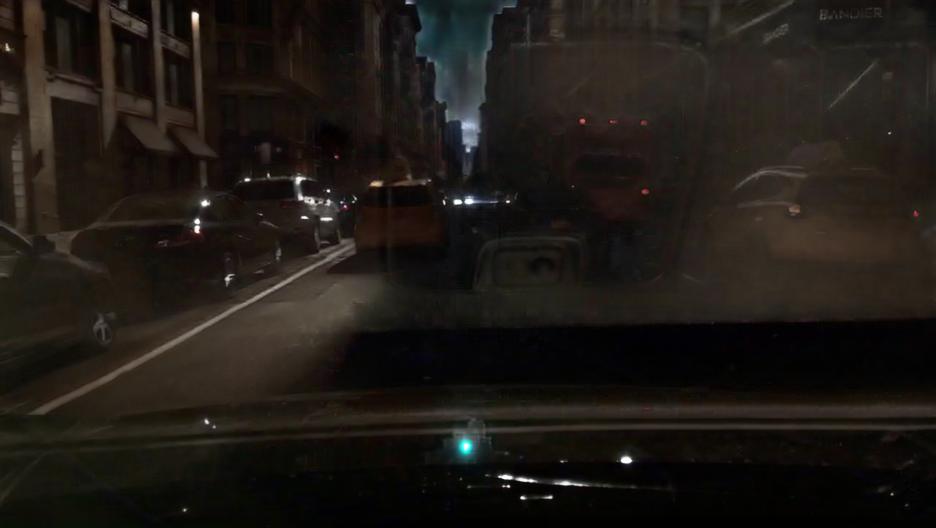}}\hfill
		{\includegraphics[width=0.165\textwidth]{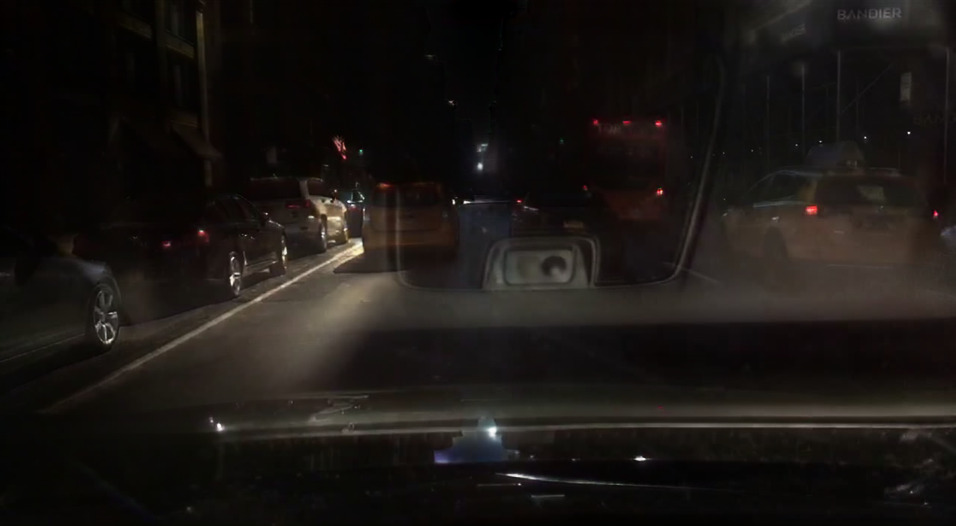}}\hfill
		{\includegraphics[width=0.165\textwidth]{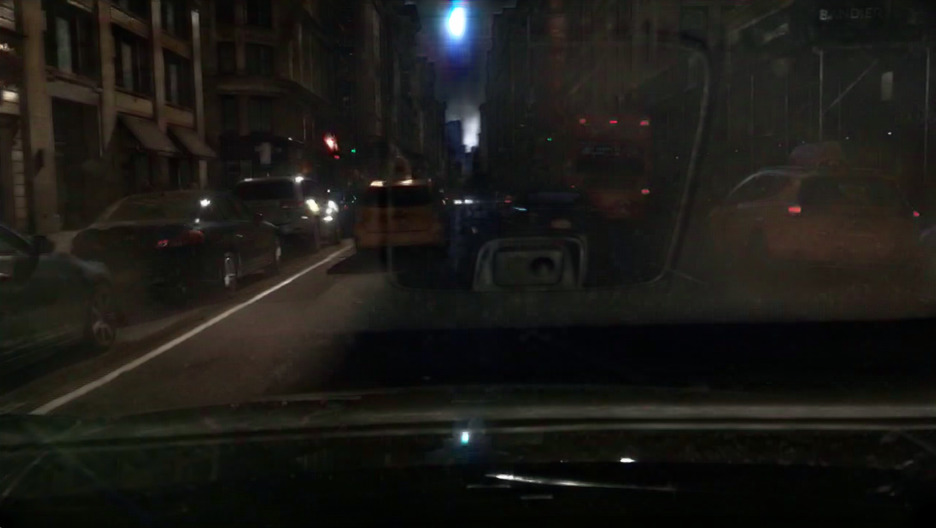}}\hfill
		{\includegraphics[width=0.165\textwidth]{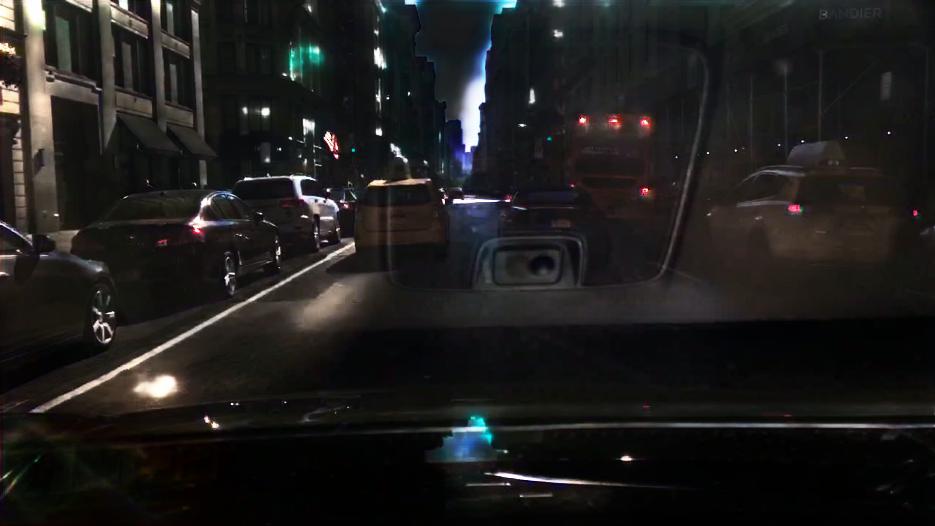}}\hfill\\\vspace{1pt}
		{\includegraphics[width=0.165\textwidth]{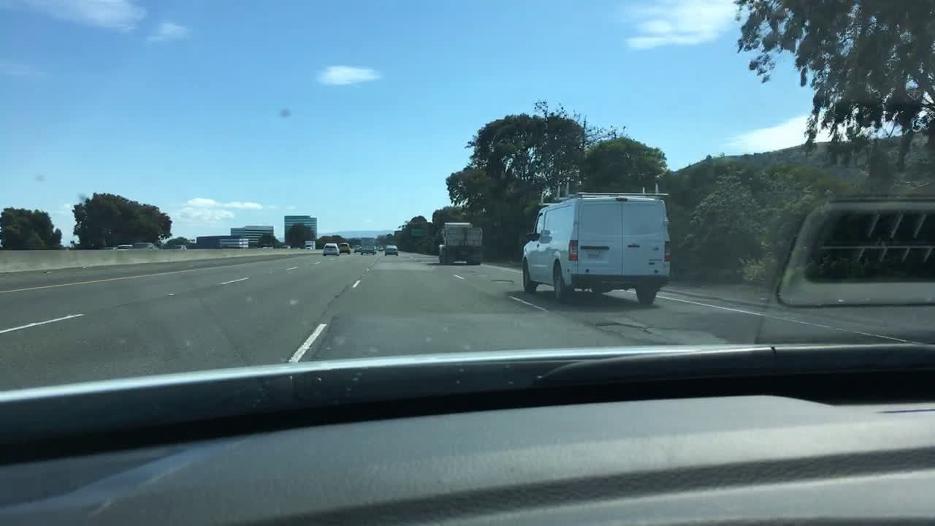}}\hfill
		{\includegraphics[width=0.165\textwidth]{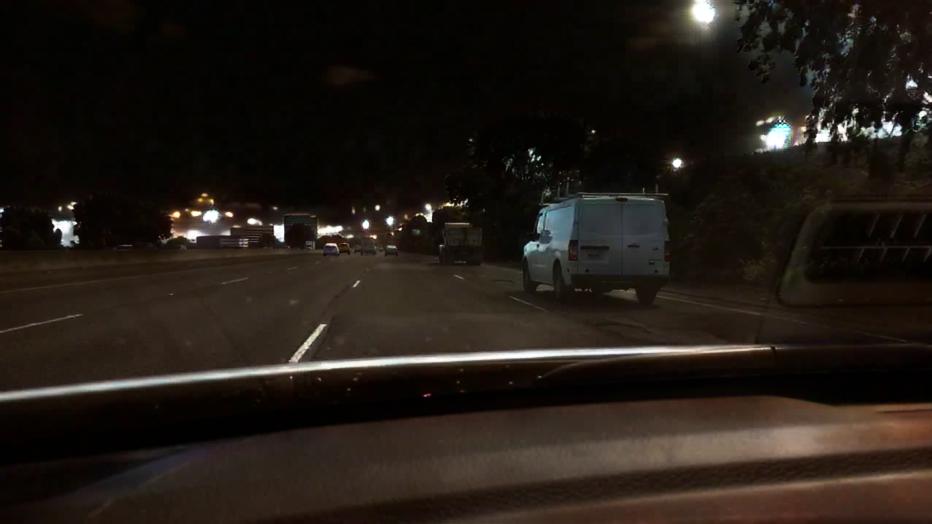}}\hfill
		{\includegraphics[width=0.165\textwidth]{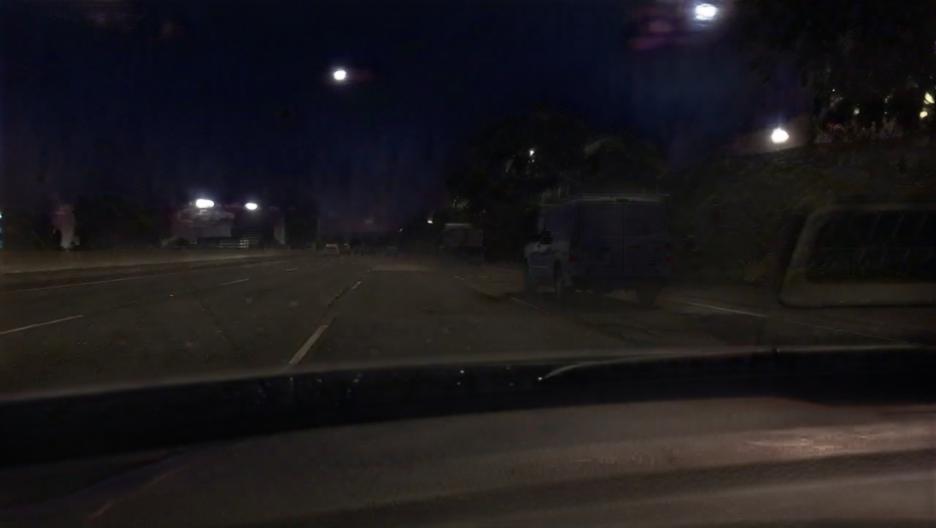}}\hfill
		{\includegraphics[width=0.165\textwidth]{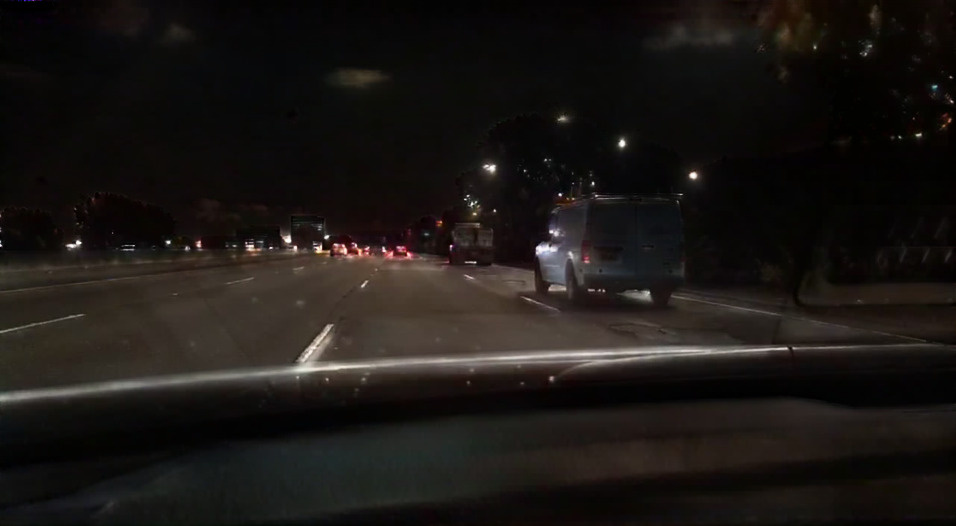}}\hfill
		{\includegraphics[width=0.165\textwidth]{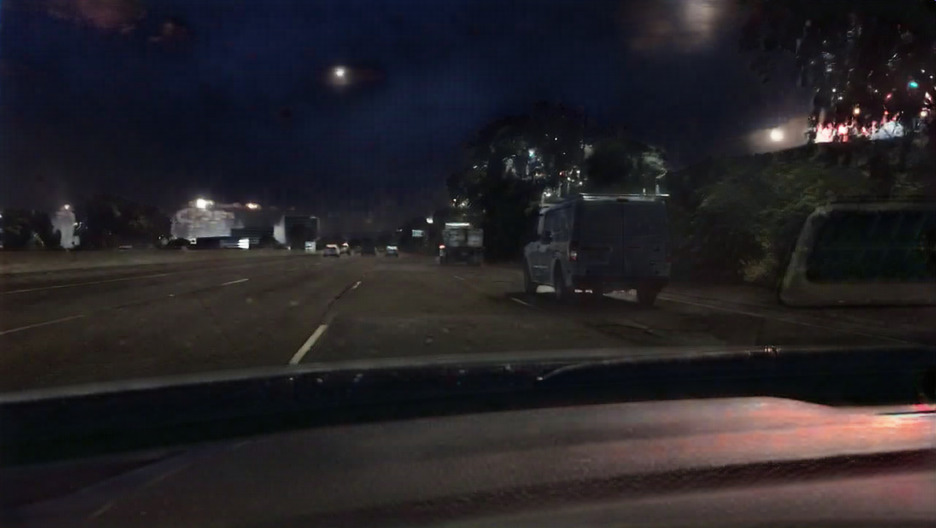}}\hfill
		{\includegraphics[width=0.165\textwidth]{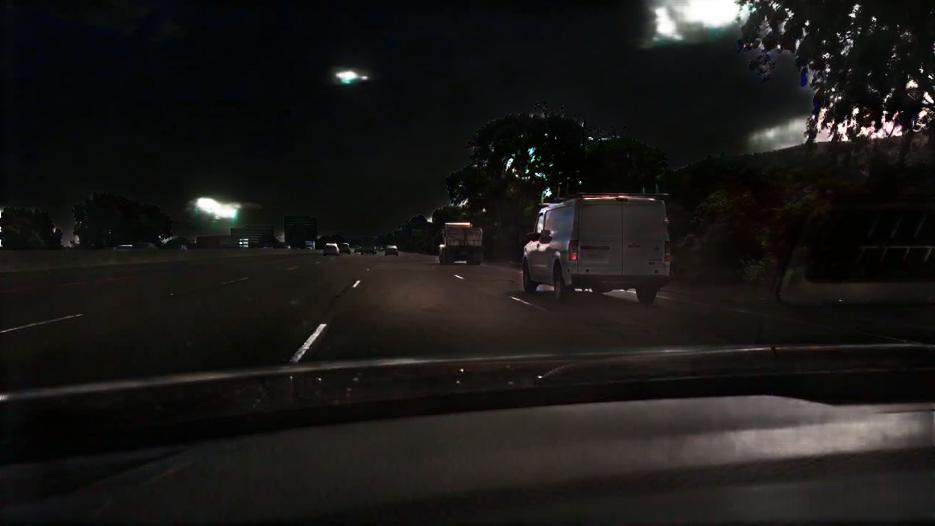}}\hfill\\\vspace{1pt}	
		
		{\scriptsize Clear$\rightarrow$Snowy} \hfill\\\vspace{1pt}
		{\includegraphics[width=0.165\textwidth]{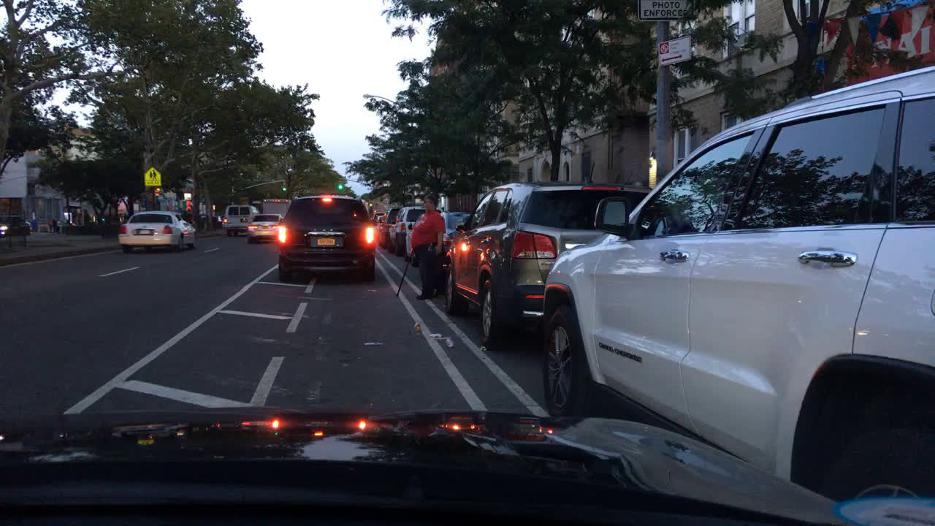}}\hfill
		{\includegraphics[width=0.165\textwidth]{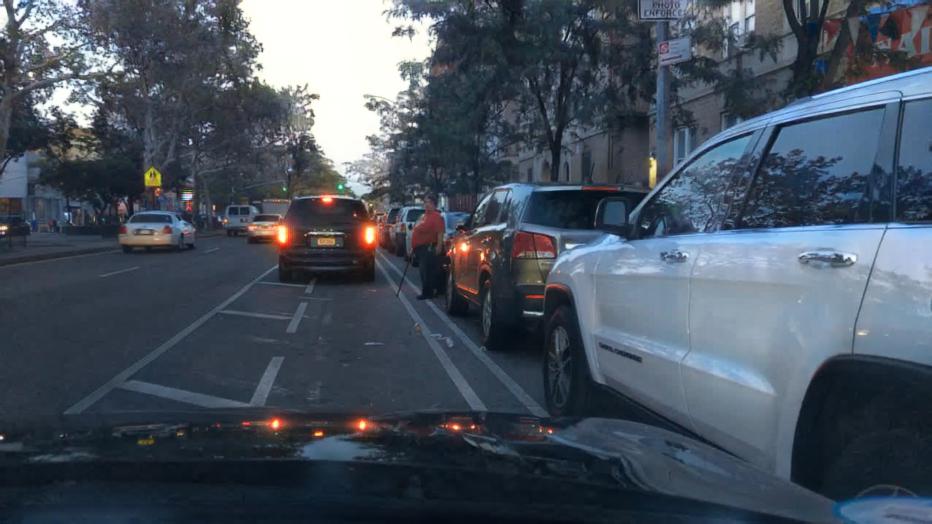}}\hfill
		{\includegraphics[width=0.165\textwidth]{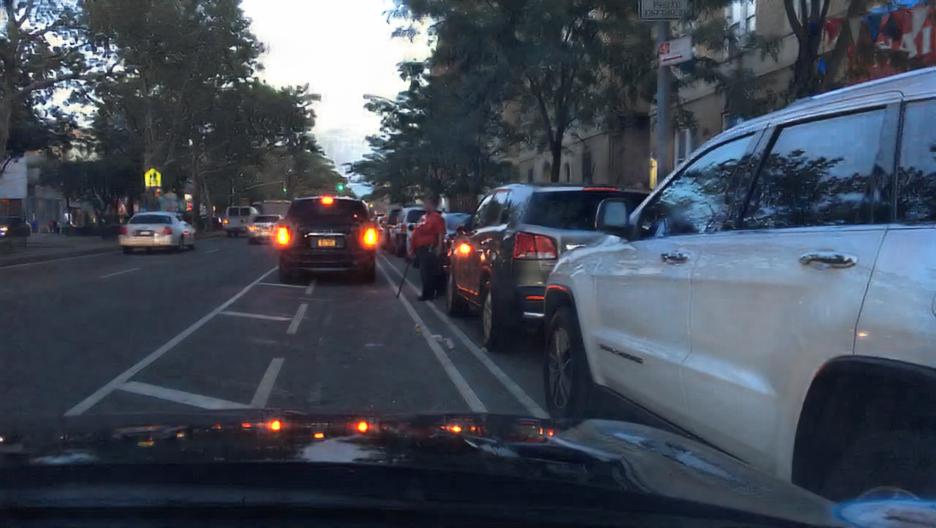}}\hfill
		{\includegraphics[width=0.165\textwidth]{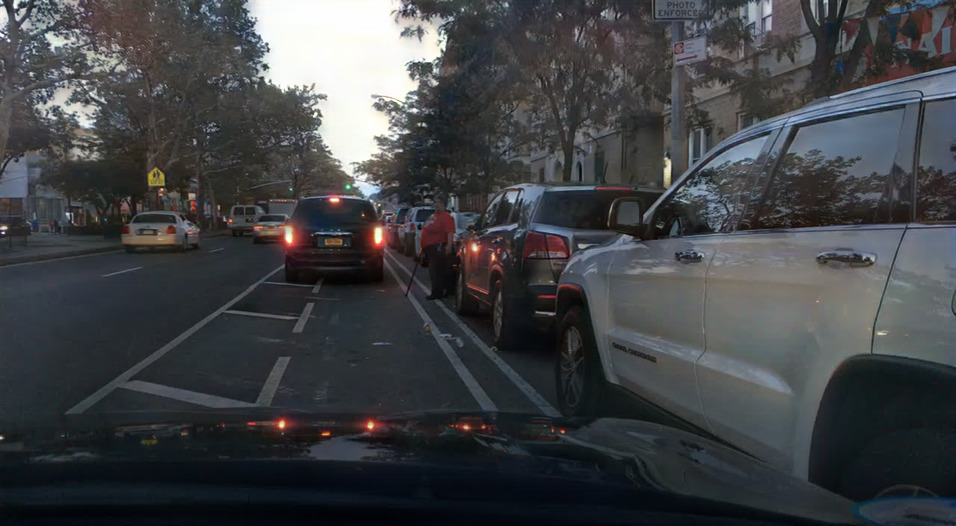}}\hfill
		{\includegraphics[width=0.165\textwidth]{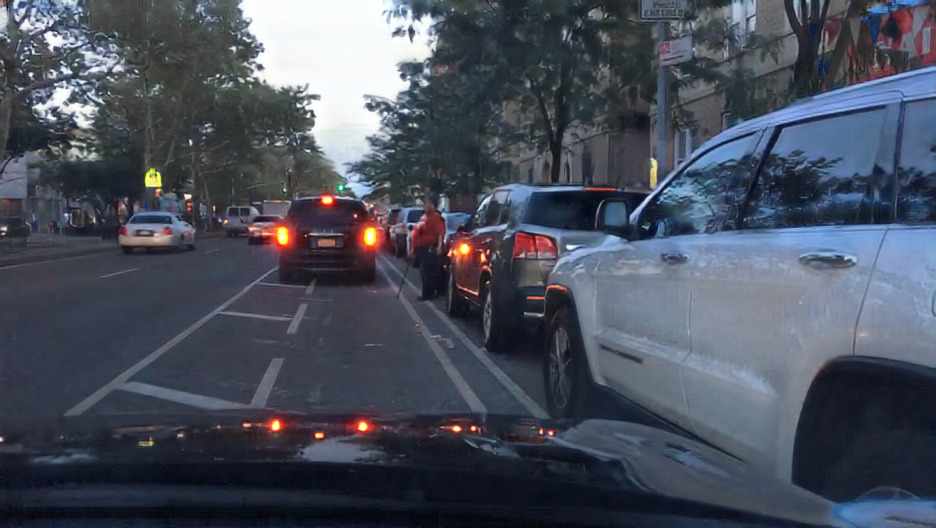}}\hfill
		{\includegraphics[width=0.165\textwidth]{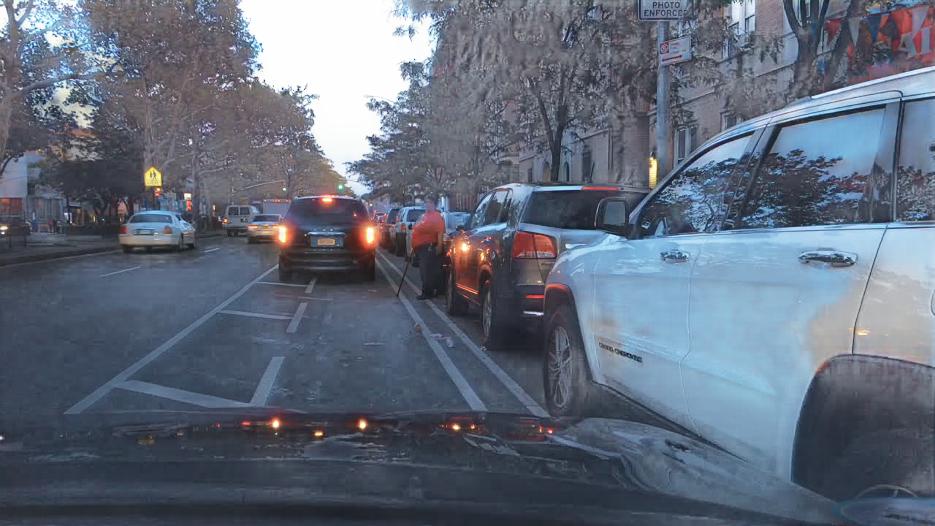}}\hfill\\\vspace{1pt}
		{\includegraphics[width=0.165\textwidth]{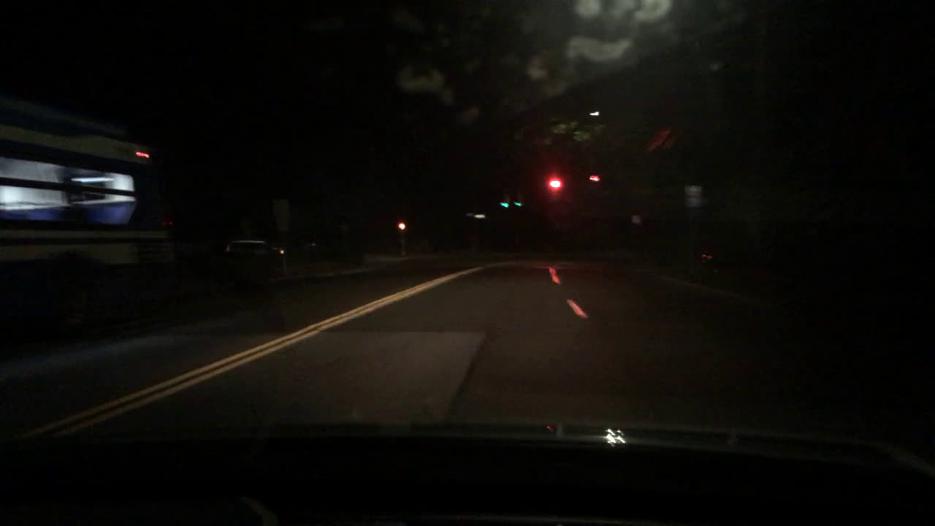}}\hfill
		{\includegraphics[width=0.165\textwidth]{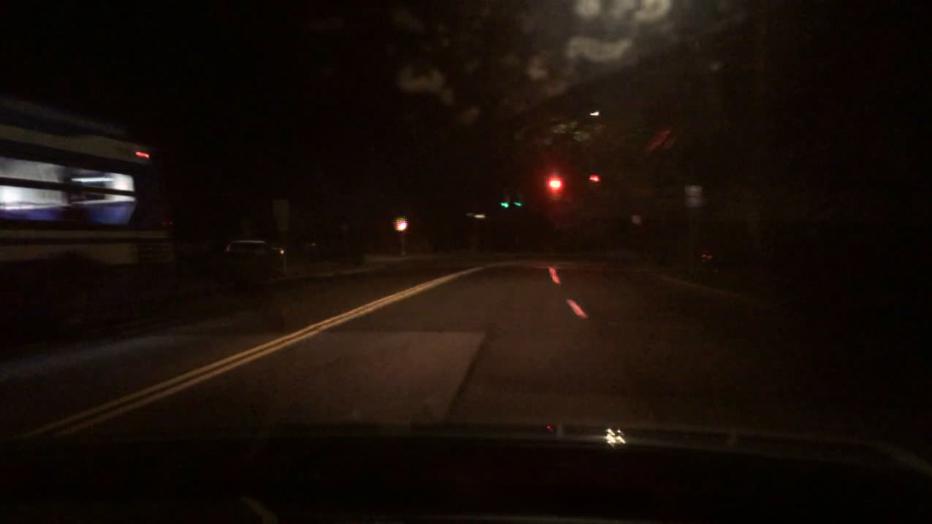}}\hfill
		{\includegraphics[width=0.165\textwidth]{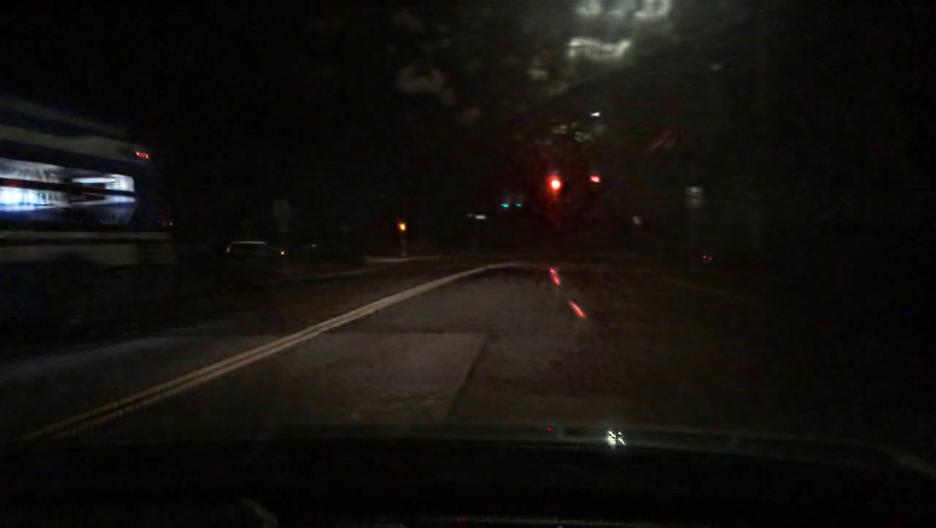}}\hfill
		{\includegraphics[width=0.165\textwidth]{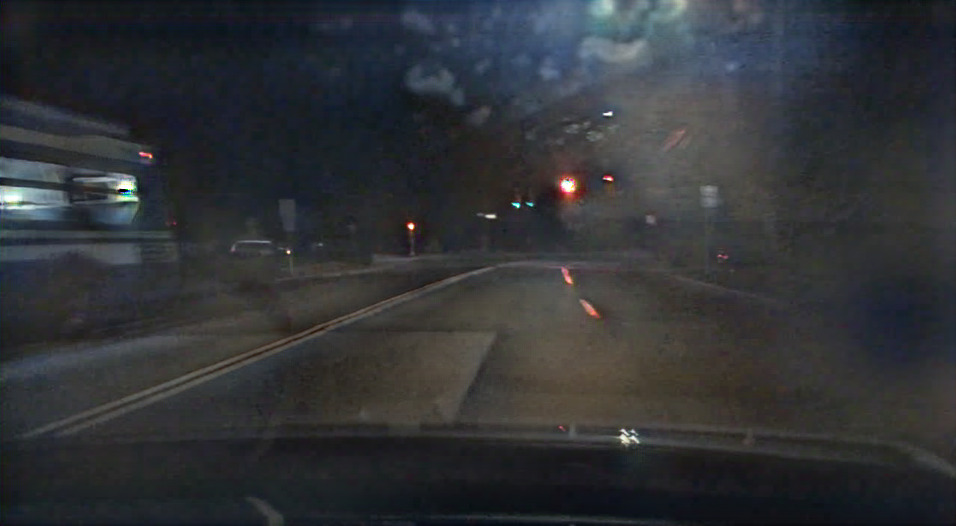}}\hfill
		{\includegraphics[width=0.165\textwidth]{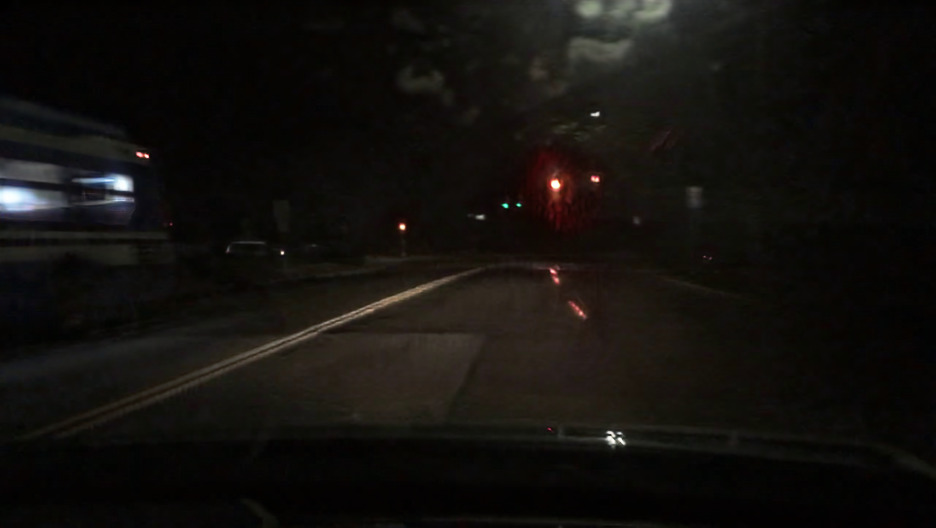}}\hfill
		{\includegraphics[width=0.165\textwidth]{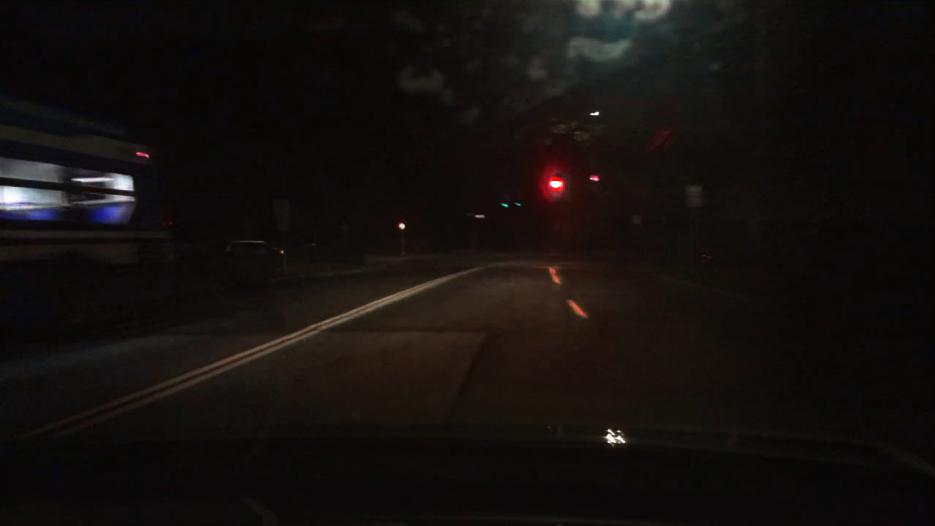}}\hfill\\\vspace{1pt}
		\vspace{-5pt}
		\subfigure[Input]
		{\includegraphics[width=0.165\textwidth]{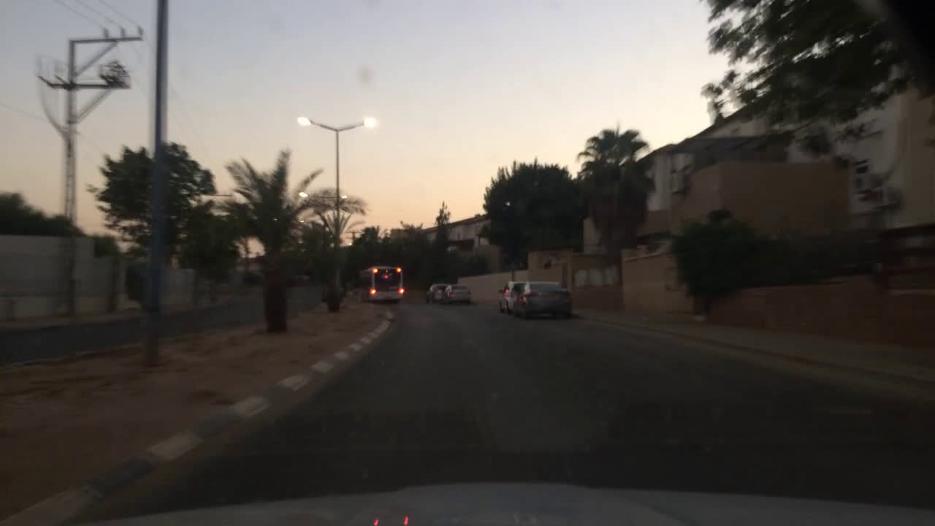}}\hfill
		\subfigure[MUNIT]
		{\includegraphics[width=0.165\textwidth]{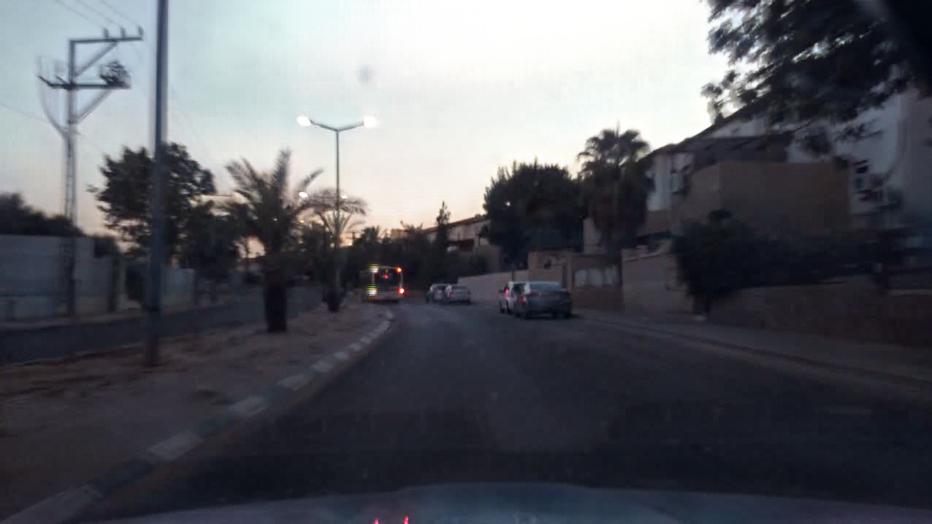}}\hfill
		\subfigure[CUT]
		{\includegraphics[width=0.165\textwidth]{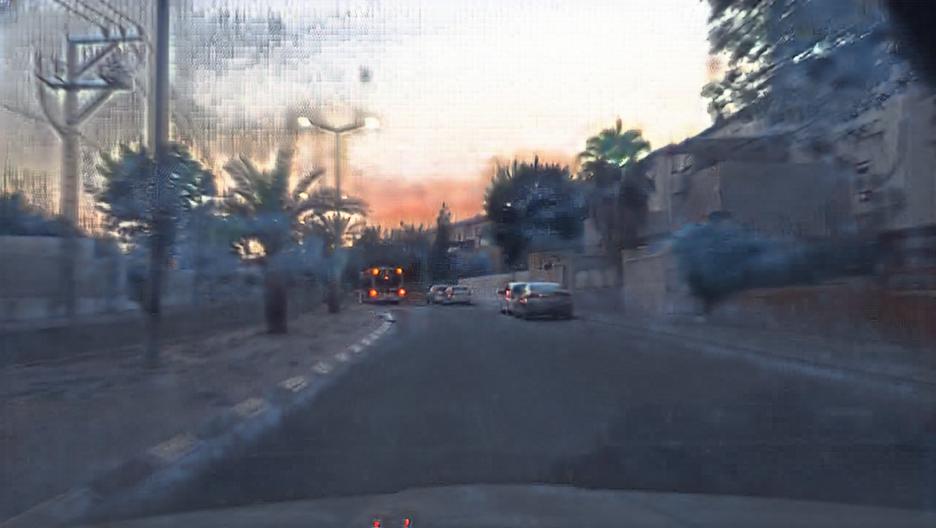}}\hfill
		\subfigure[TSIT]
		{\includegraphics[width=0.165\textwidth]{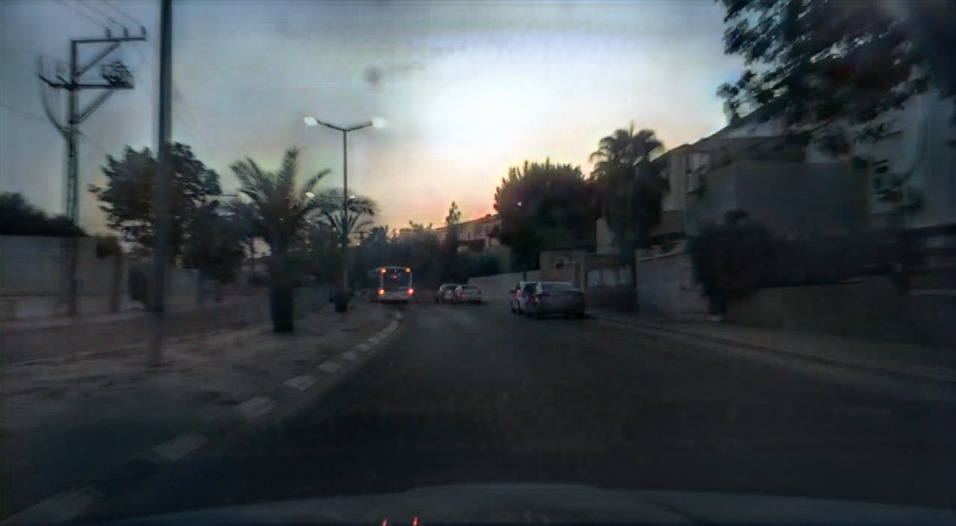}}\hfill
		\subfigure[QS-Attn]
		{\includegraphics[width=0.165\textwidth]{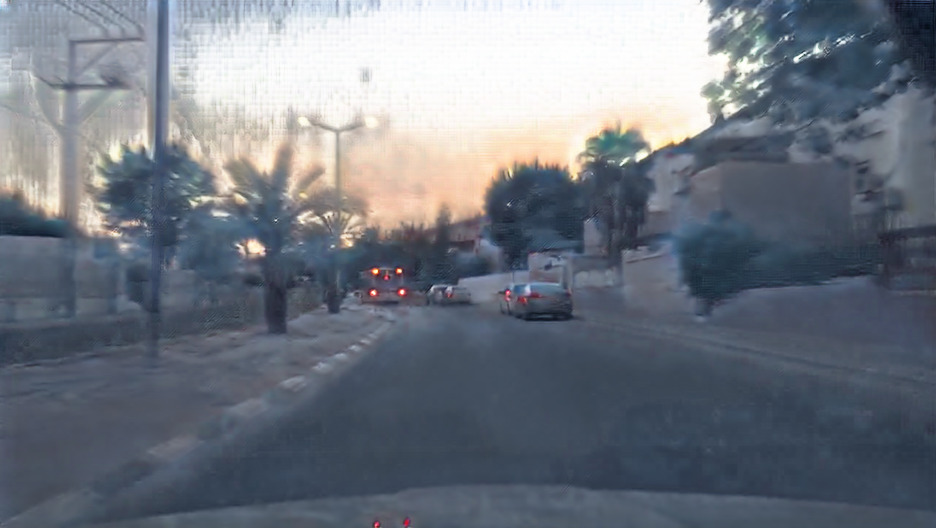}}\hfill
		\subfigure[FeaMGAN (ours)]
		{\includegraphics[width=0.165\textwidth]{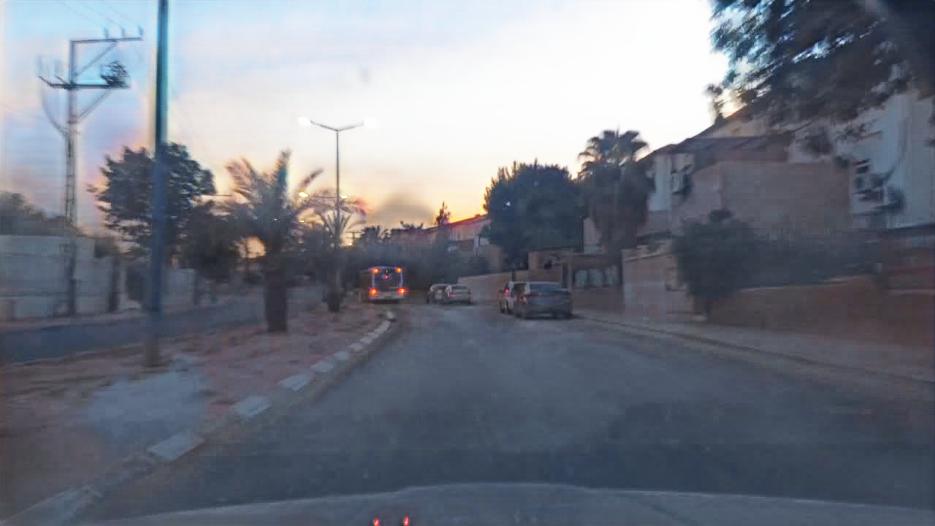}}\hfill
	\end{center}
	\vspace{-8pt}
	\caption{\textbf{Qualitative comparison to prior work.} Results are randomly sampled from the best model.}
	\label{fig:qualitative_comparison_additional_random}
\end{figure*}

\begin{figure*}[h] 
	\renewcommand{\thesubfigure}{}
	\begin{center}
		{\includegraphics[width=0.198\textwidth]{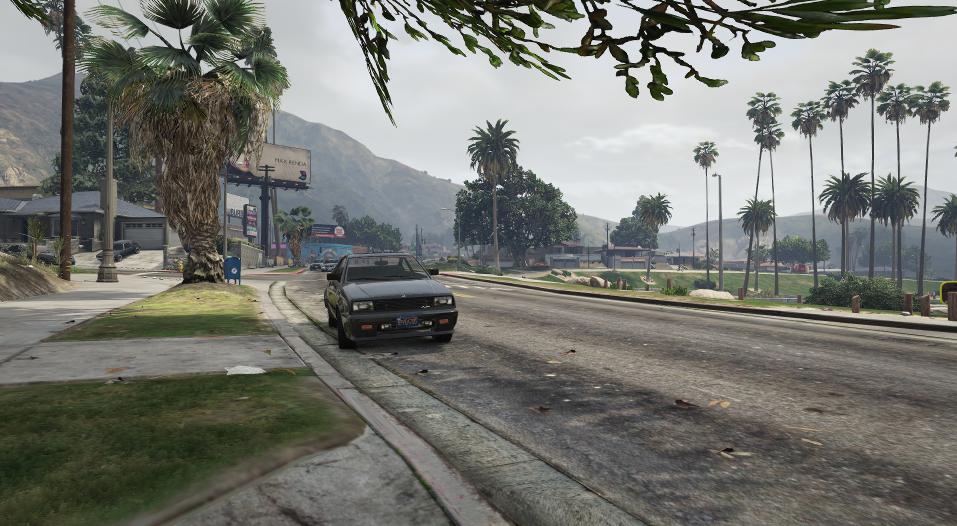}}\hfill
		{\includegraphics[width=0.198\textwidth]{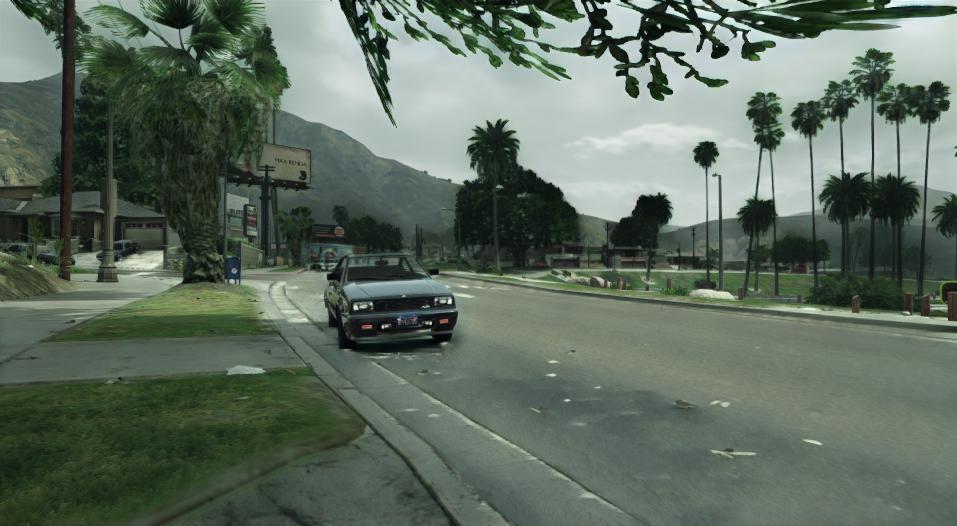}}\hfill
		{\includegraphics[width=0.198\textwidth]{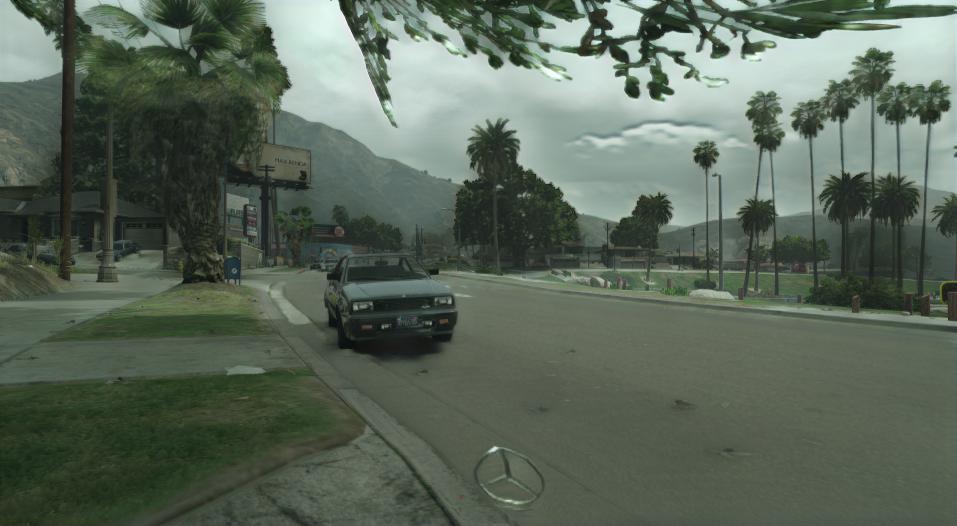}}\hfill
		{\includegraphics[width=0.198\textwidth]{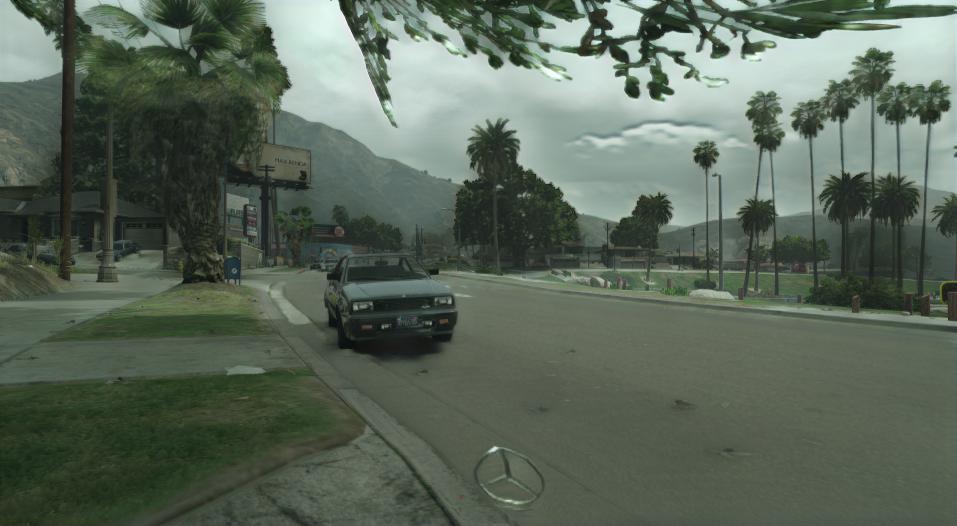}}\hfill
		{\includegraphics[width=0.198\textwidth]{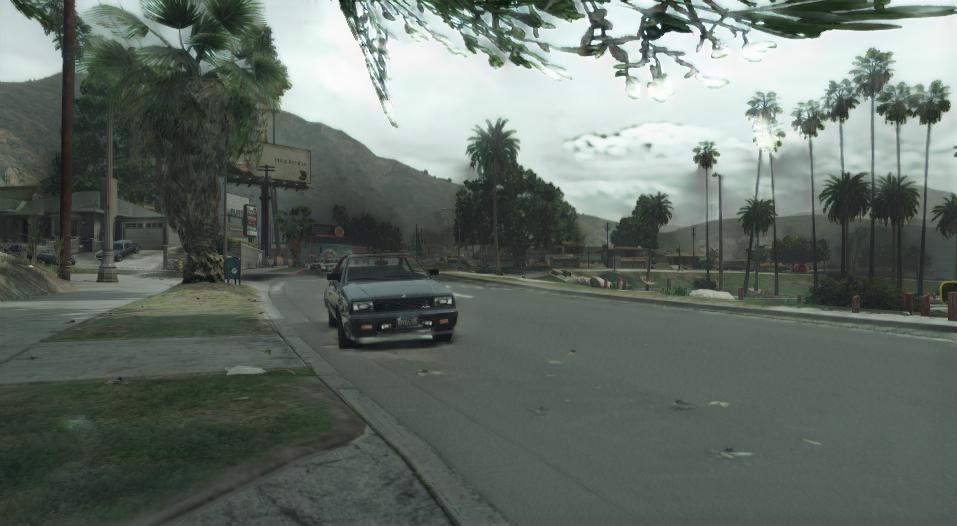}}\hfill \\ \vspace{1pt}
		{\includegraphics[width=0.198\textwidth]{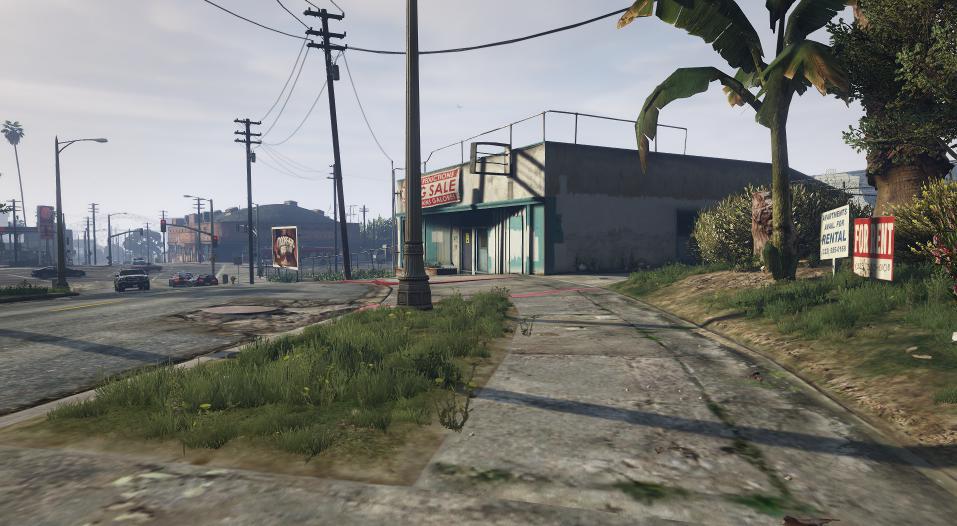}}\hfill
		{\includegraphics[width=0.198\textwidth]{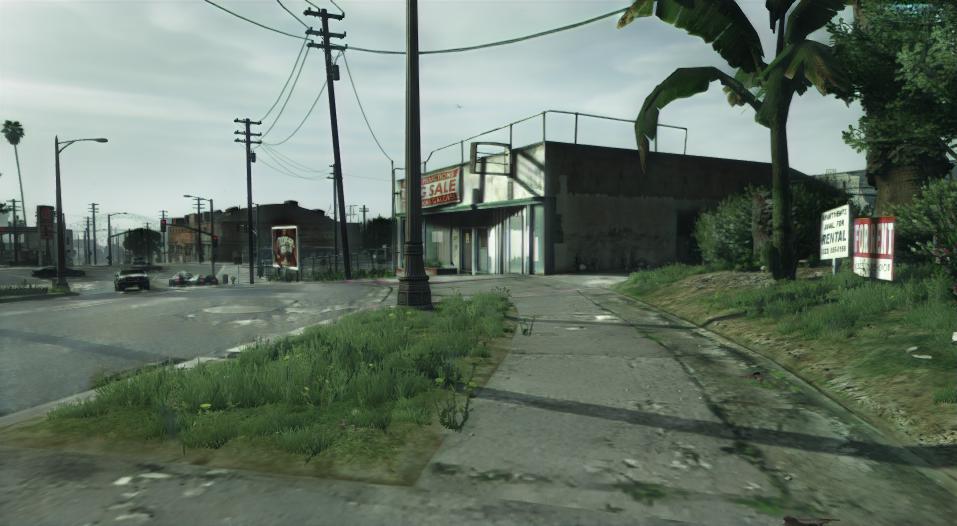}}\hfill
		{\includegraphics[width=0.198\textwidth]{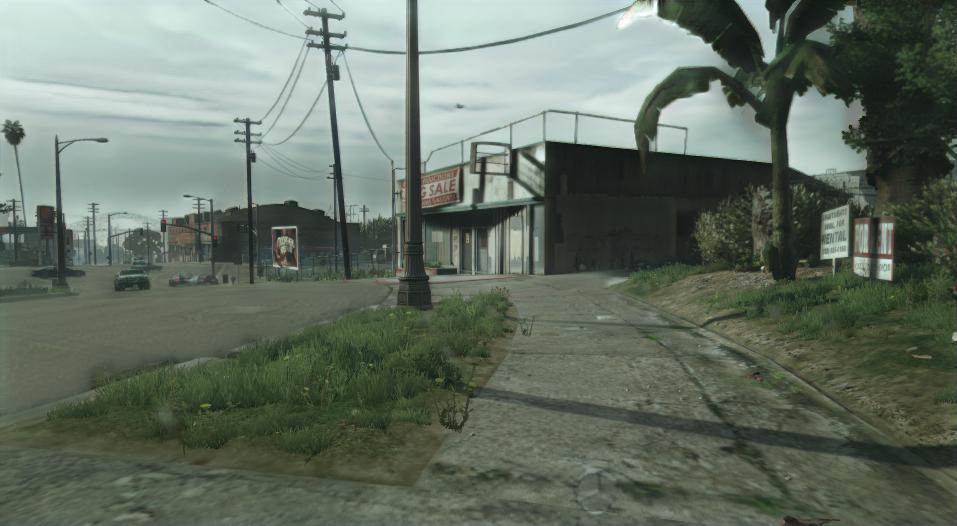}}\hfill
		{\includegraphics[width=0.198\textwidth]{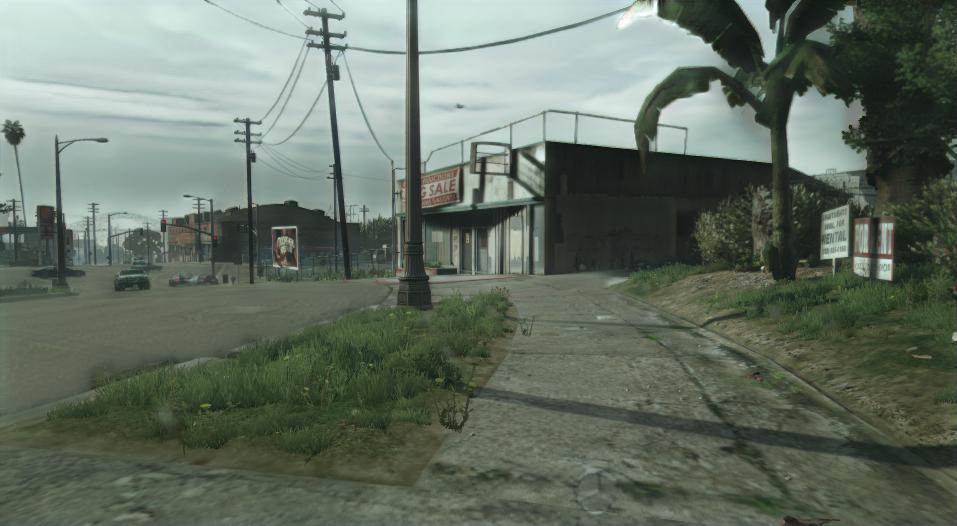}}\hfill
		{\includegraphics[width=0.198\textwidth]{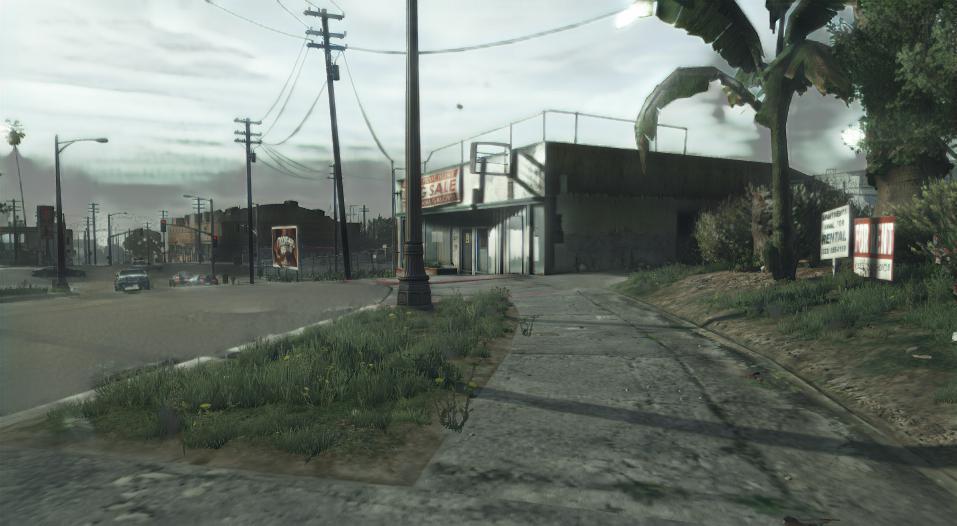}}\hfill \\ \vspace{1pt}
		{\includegraphics[width=0.198\textwidth]{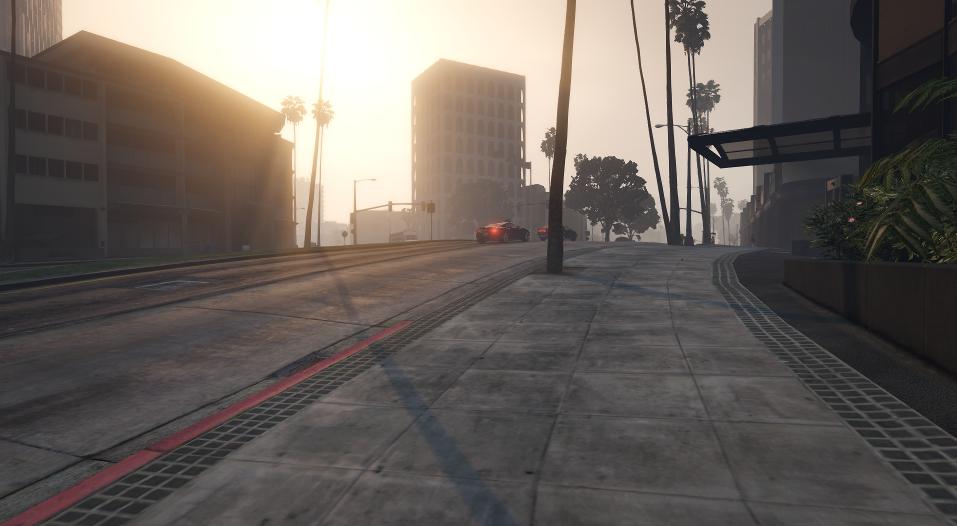}}\hfill
		{\includegraphics[width=0.198\textwidth]{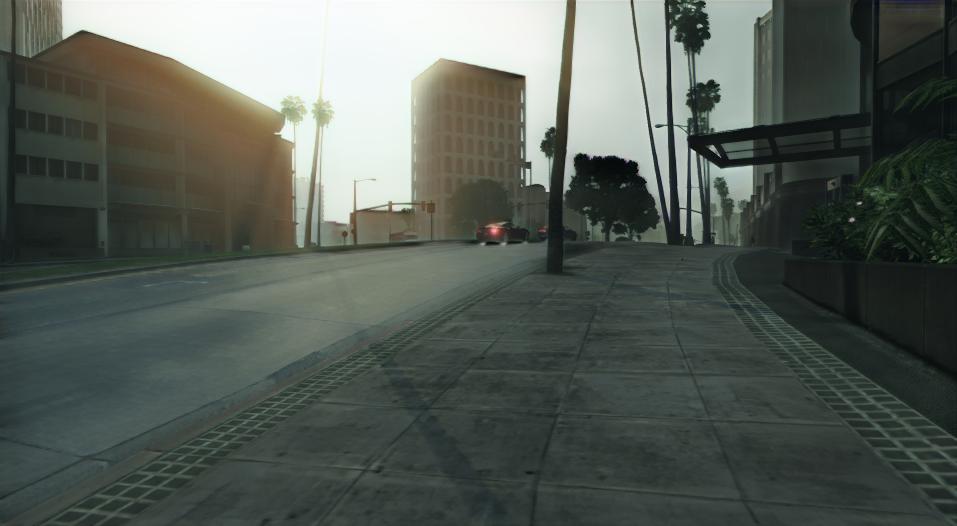}}\hfill
		{\includegraphics[width=0.198\textwidth]{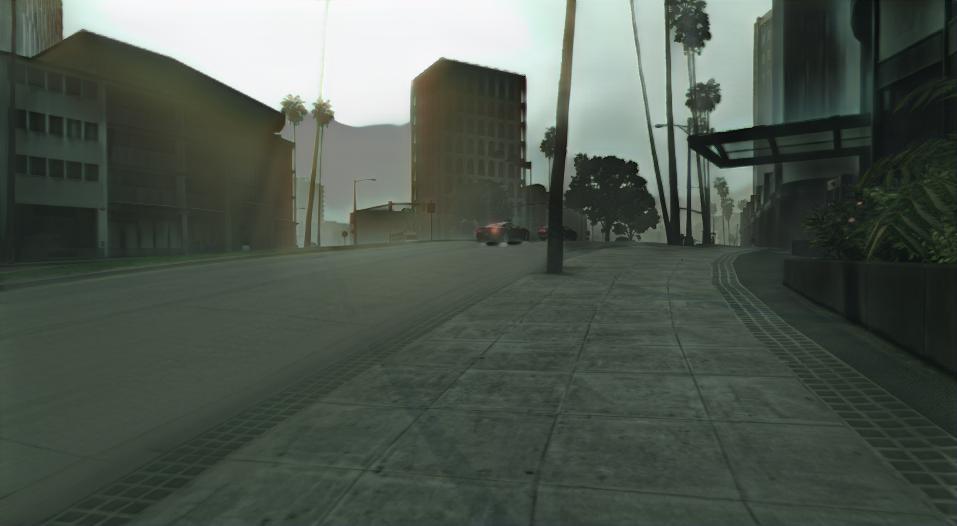}}\hfill
		{\includegraphics[width=0.198\textwidth]{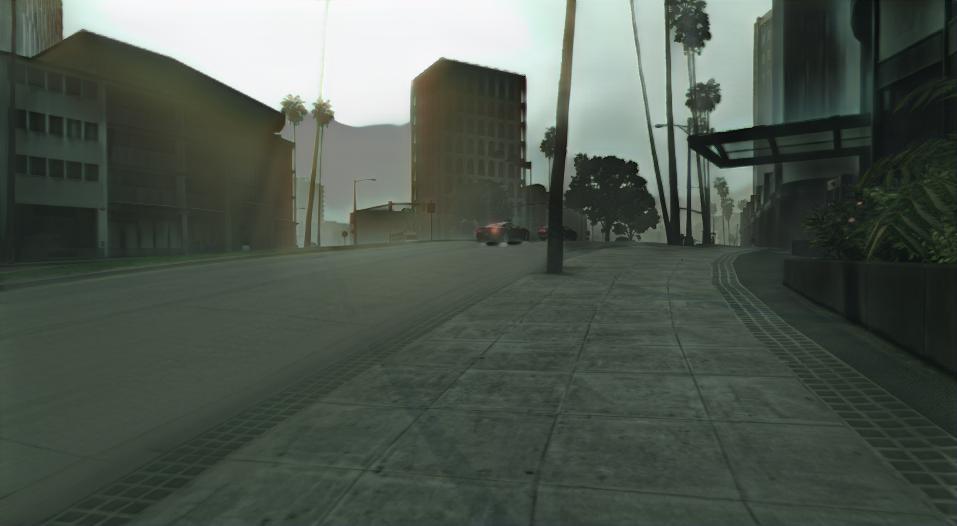}}\hfill
		{\includegraphics[width=0.198\textwidth]{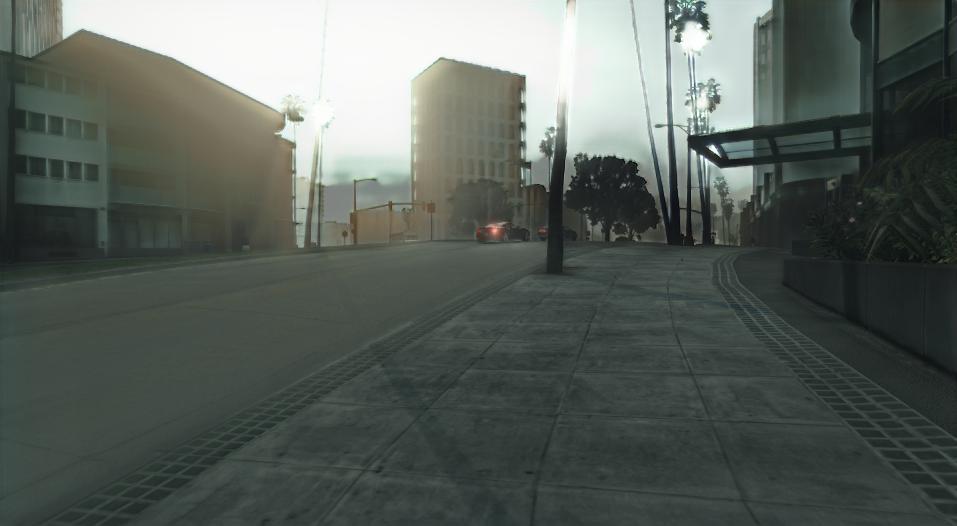}}\hfill \\ \vspace{1pt}
		{\includegraphics[width=0.198\textwidth]{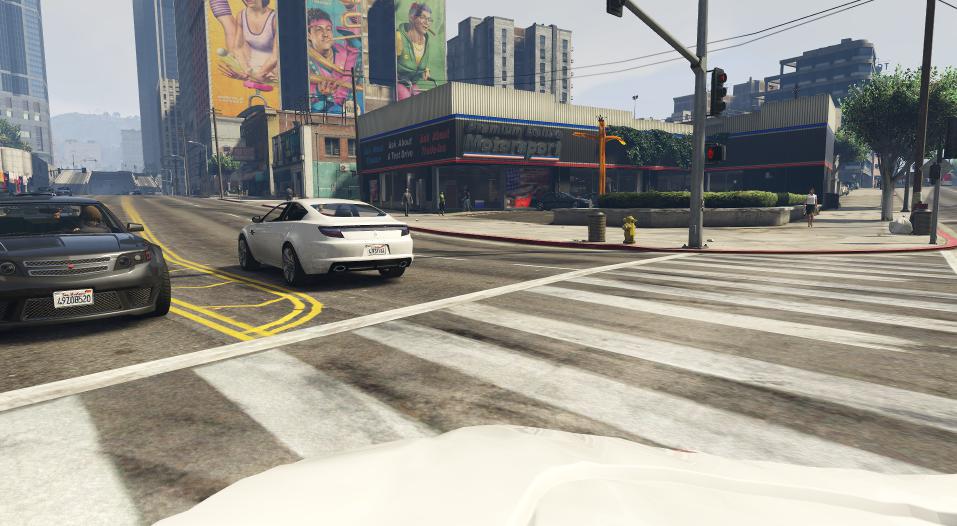}}\hfill
		{\includegraphics[width=0.198\textwidth]{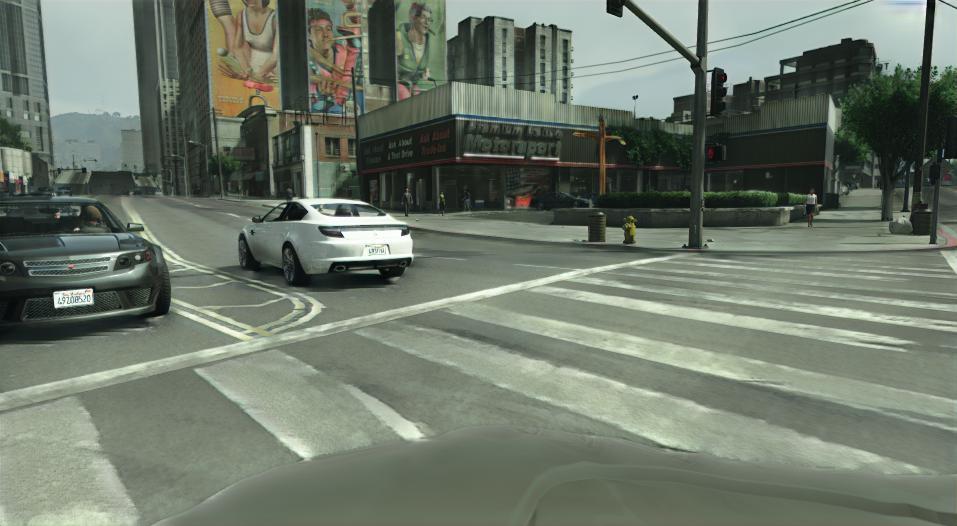}}\hfill
		{\includegraphics[width=0.198\textwidth]{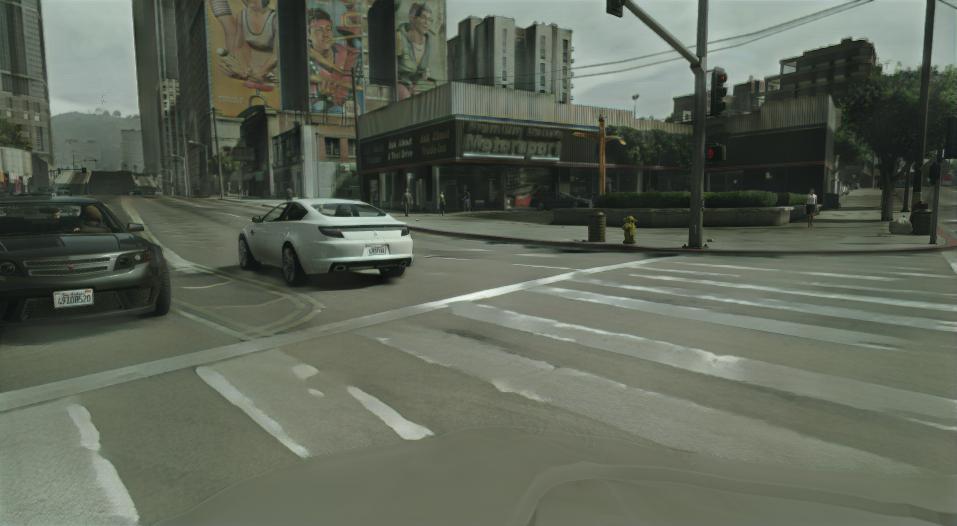}}\hfill
		{\includegraphics[width=0.198\textwidth]{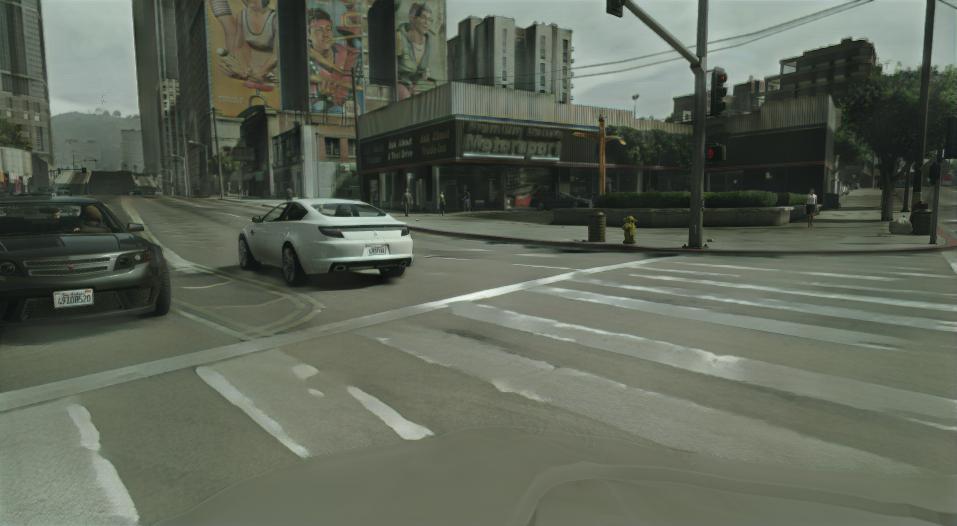}}\hfill
		{\includegraphics[width=0.198\textwidth]{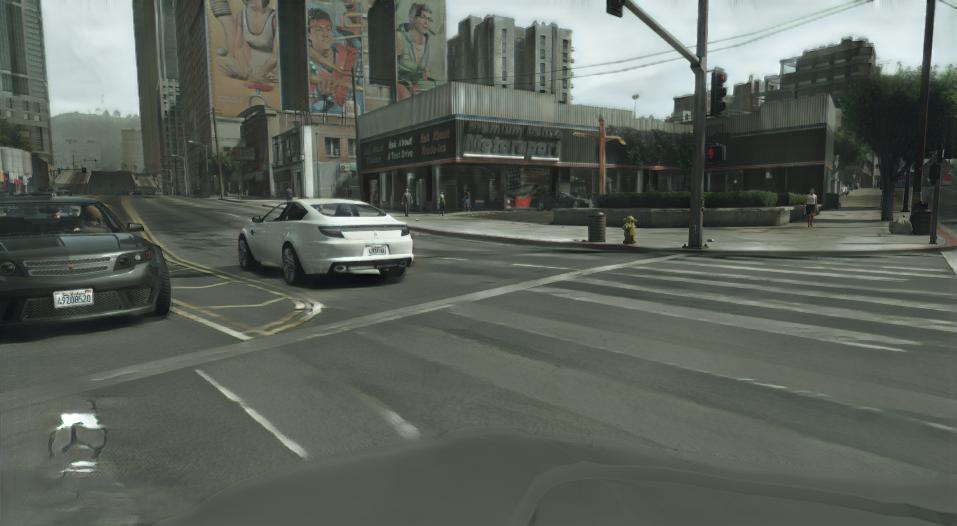}}\hfill \\ \vspace{1pt}
		{\includegraphics[width=0.198\textwidth]{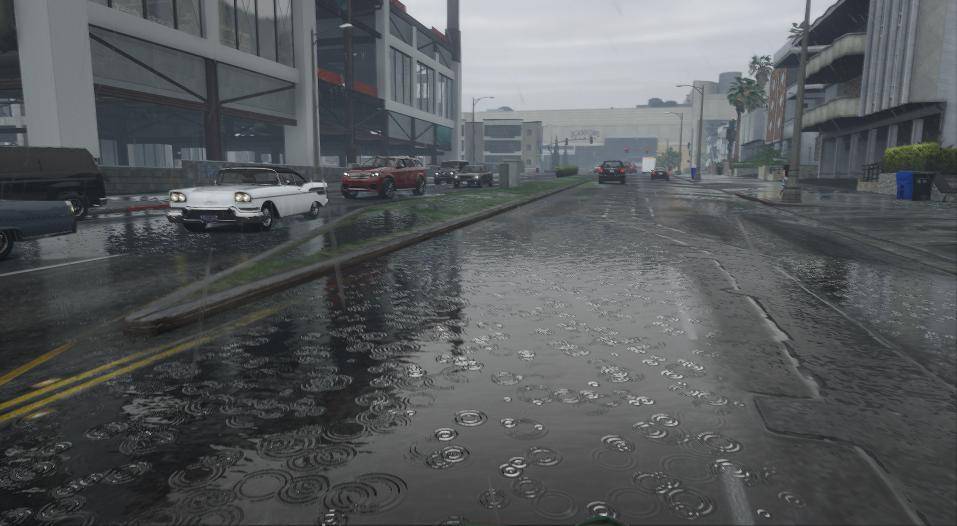}}\hfill
		{\includegraphics[width=0.198\textwidth]{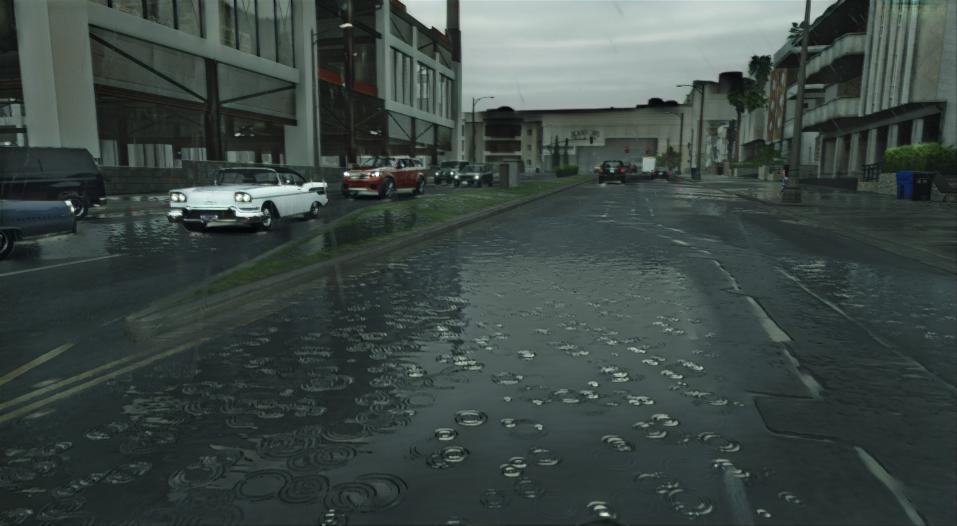}}\hfill
		{\includegraphics[width=0.198\textwidth]{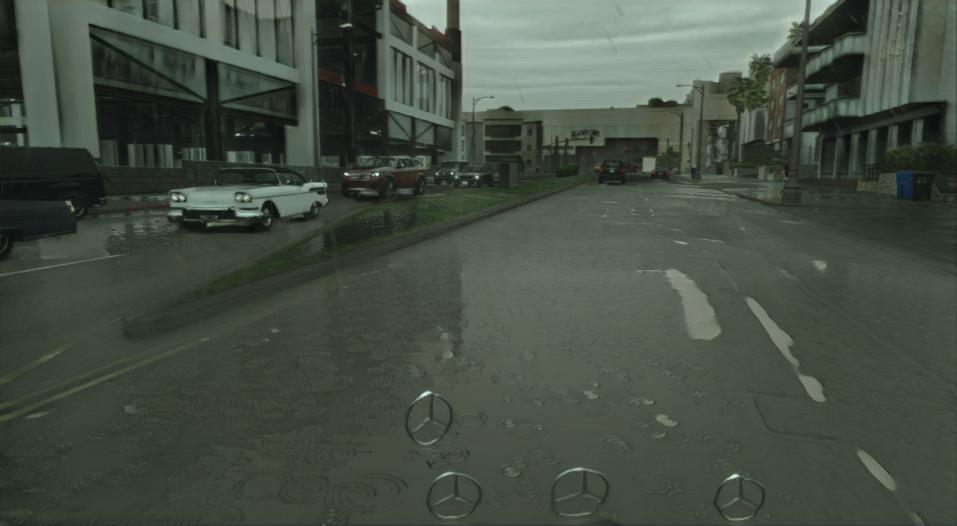}}\hfill
		{\includegraphics[width=0.198\textwidth]{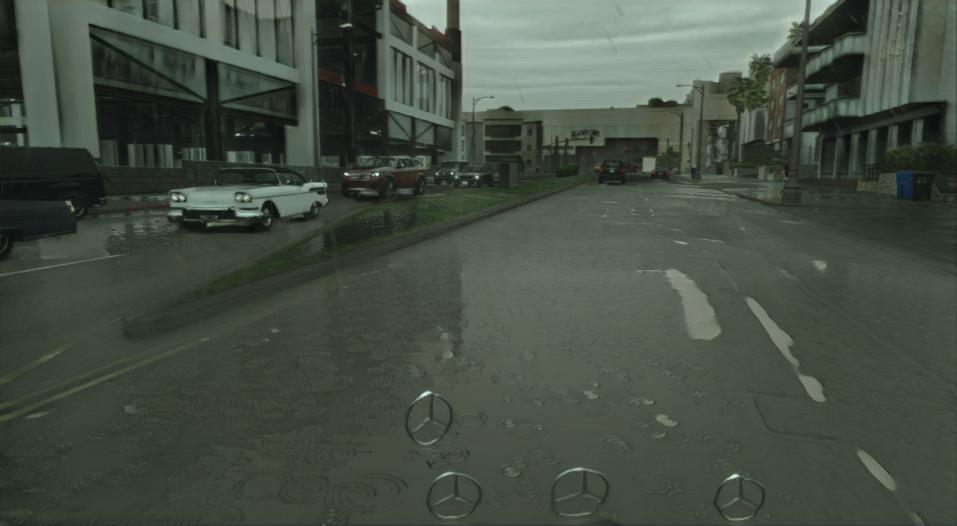}}\hfill
		{\includegraphics[width=0.198\textwidth]{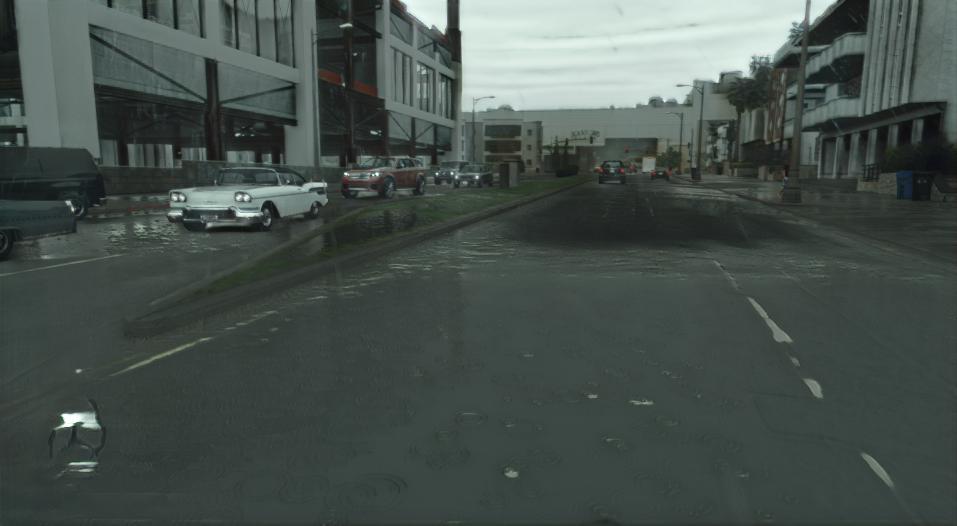}}\hfill \\ \vspace{1pt}
		{\includegraphics[width=0.198\textwidth]{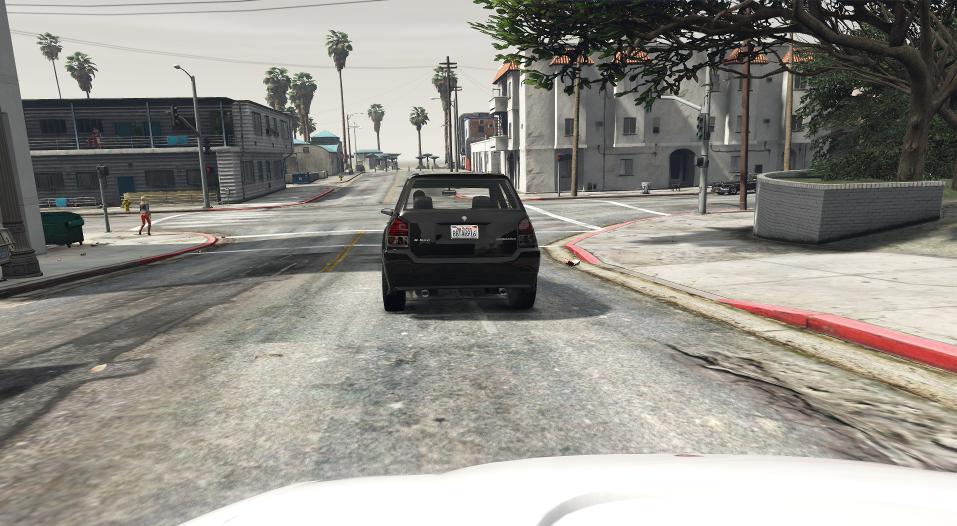}}\hfill
		{\includegraphics[width=0.198\textwidth]{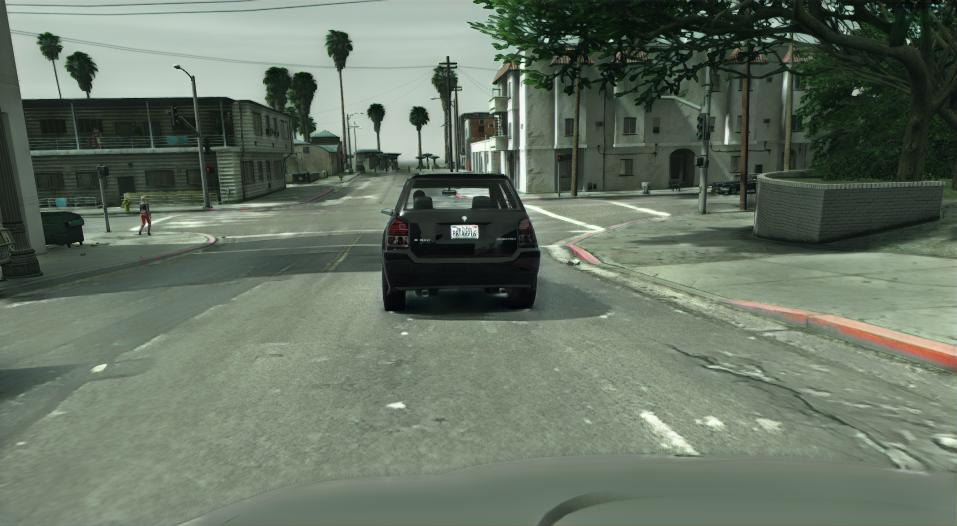}}\hfill
		{\includegraphics[width=0.198\textwidth]{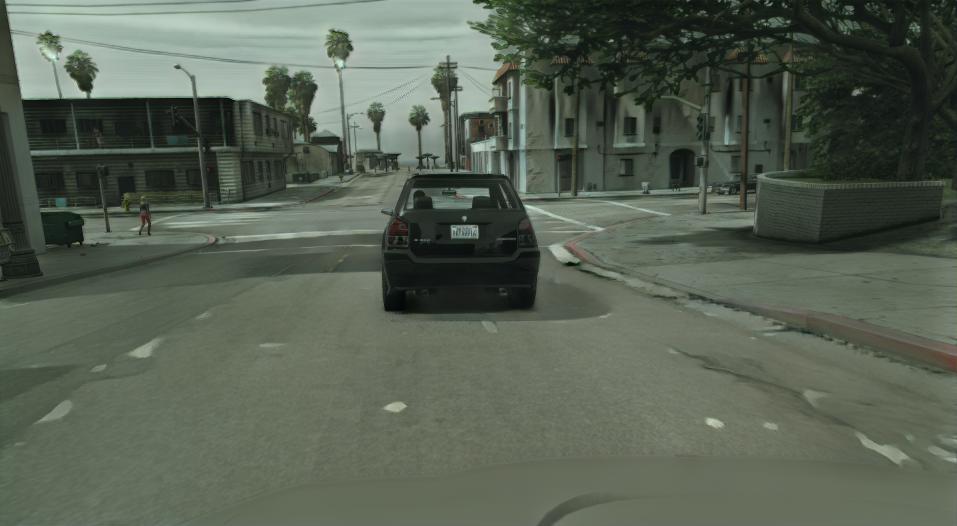}}\hfill
		{\includegraphics[width=0.198\textwidth]{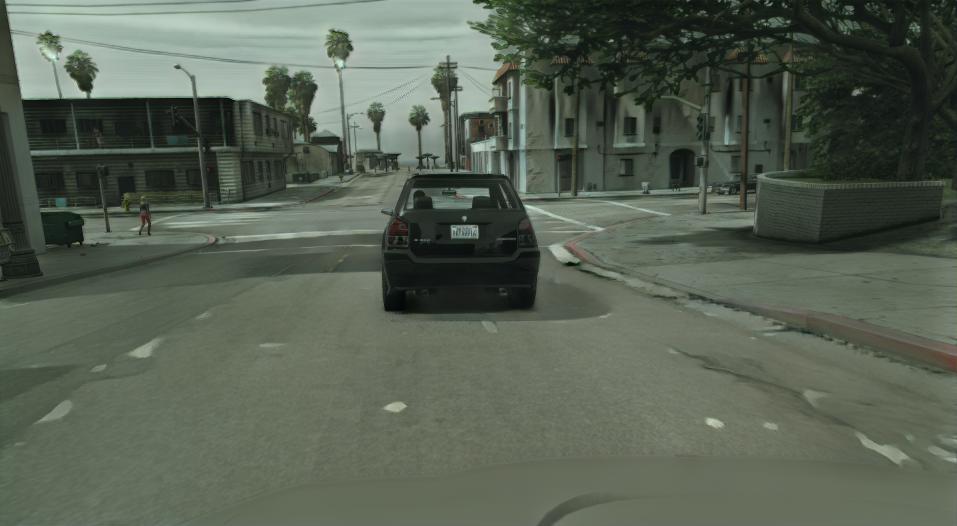}}\hfill
		{\includegraphics[width=0.198\textwidth]{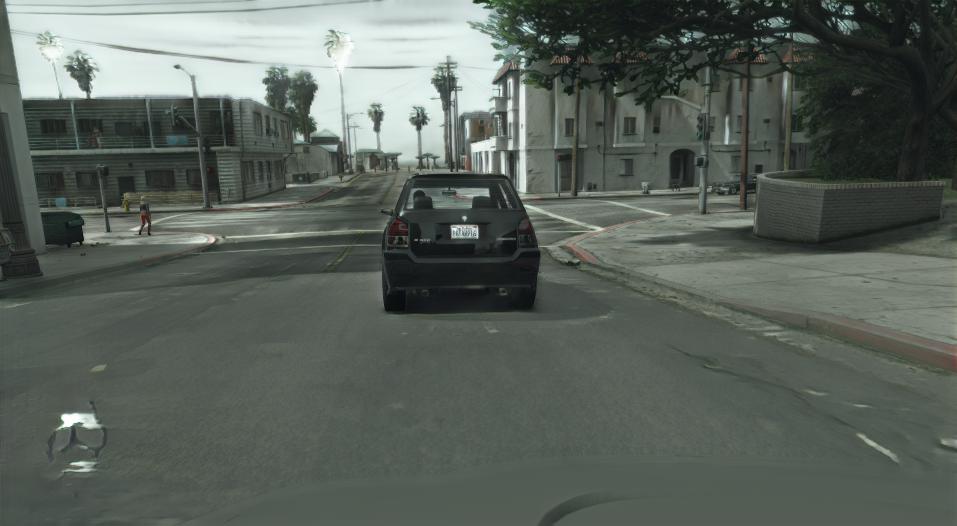}}\hfill \\ \vspace{1pt}
		{\includegraphics[width=0.198\textwidth]{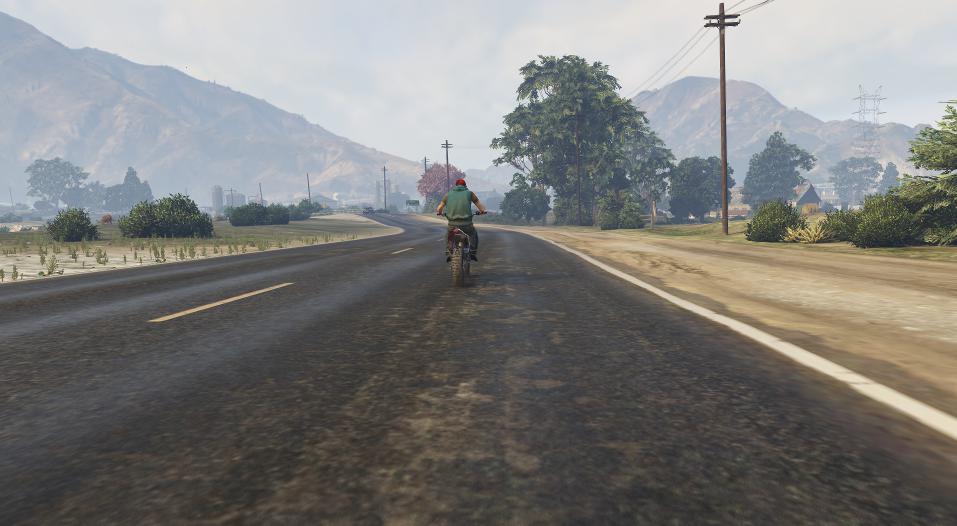}}\hfill
		{\includegraphics[width=0.198\textwidth]{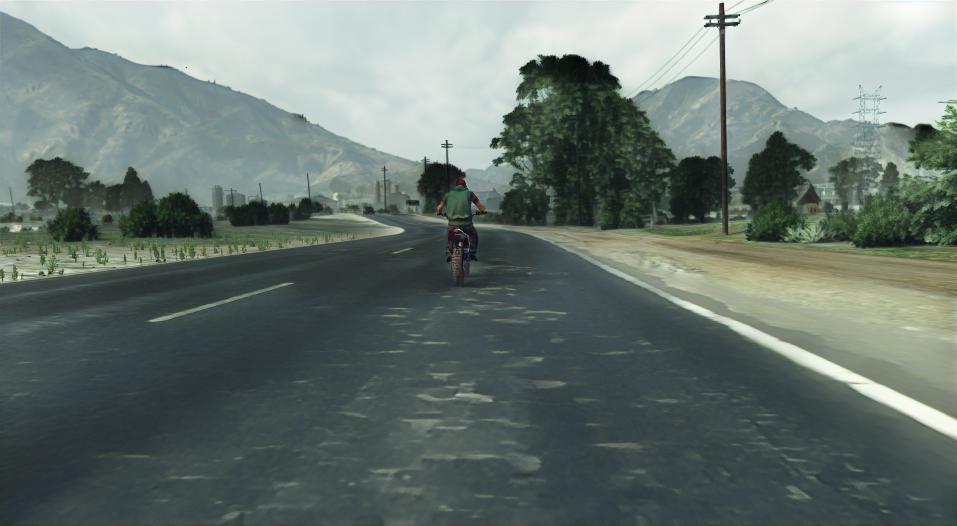}}\hfill
		{\includegraphics[width=0.198\textwidth]{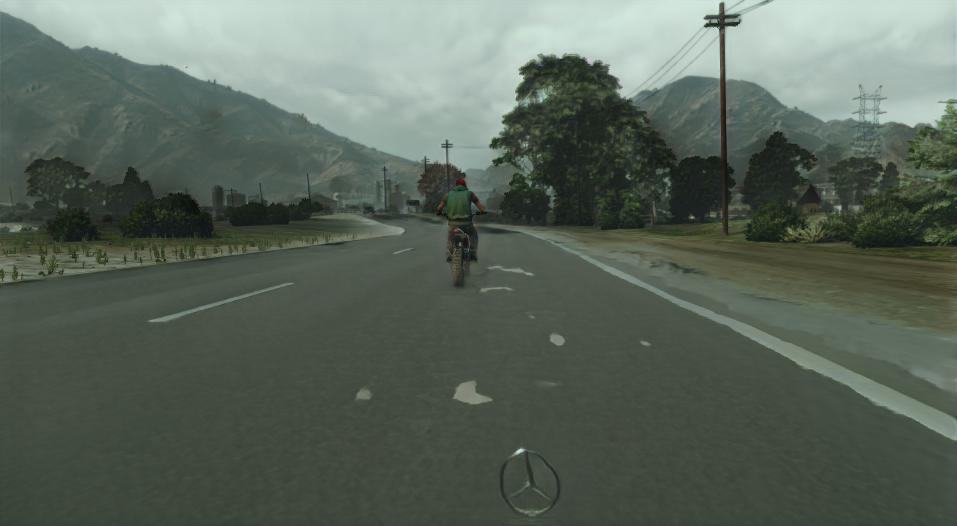}}\hfill
		{\includegraphics[width=0.198\textwidth]{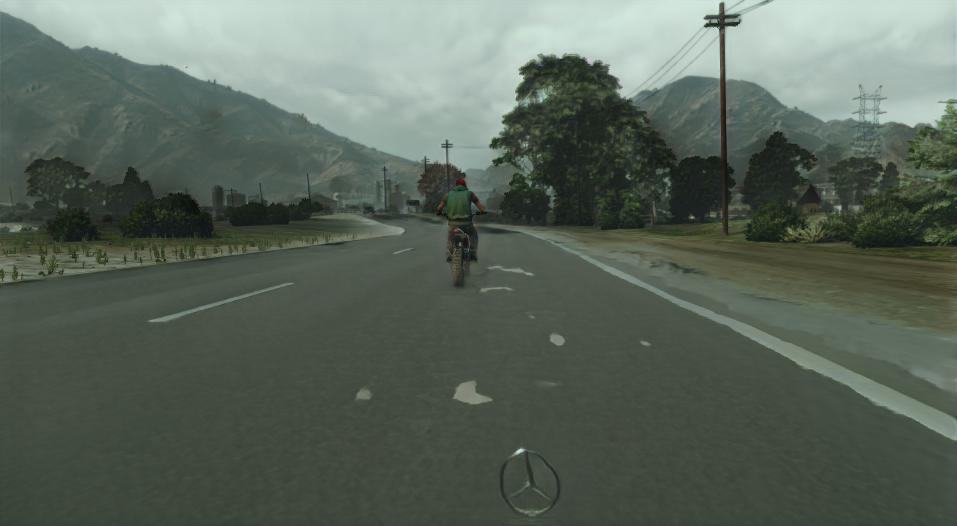}}\hfill
		{\includegraphics[width=0.198\textwidth]{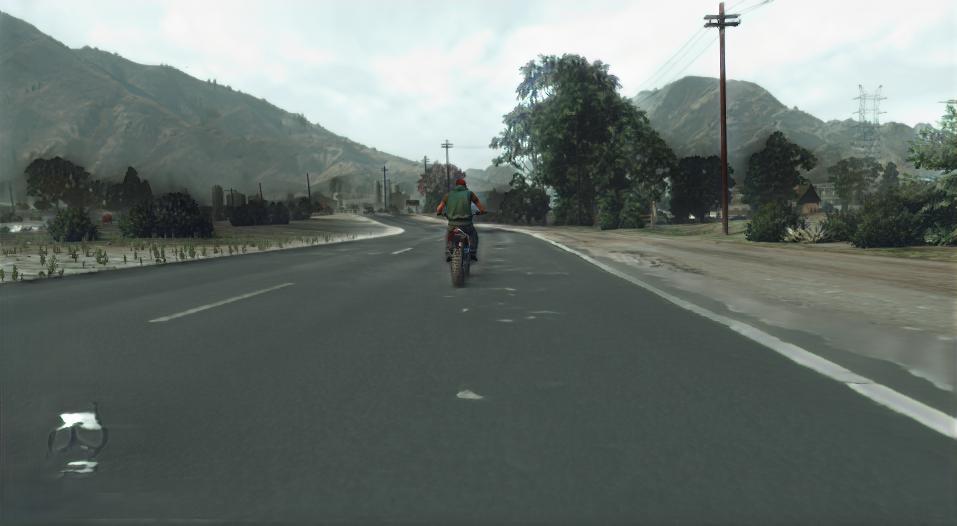}}\hfill \\ \vspace{1pt}
		{\includegraphics[width=0.198\textwidth]{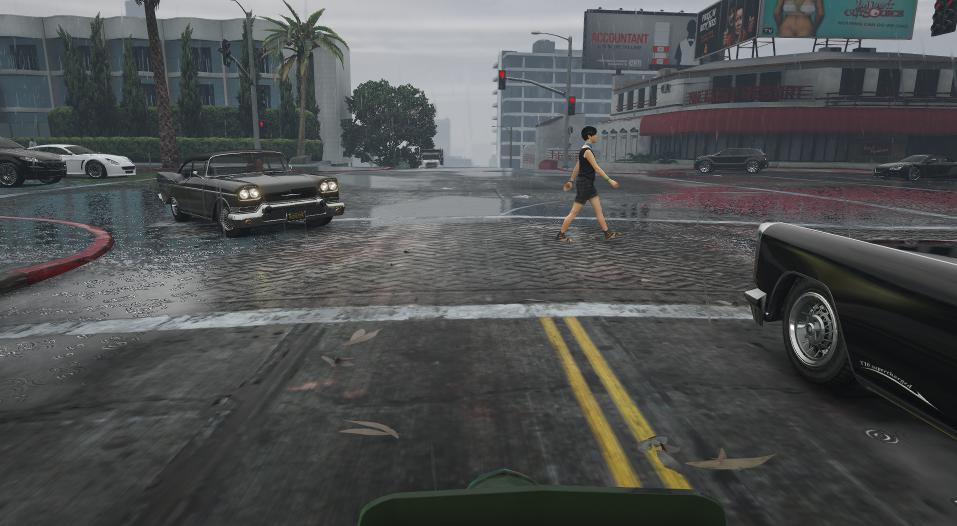}}\hfill
		{\includegraphics[width=0.198\textwidth]{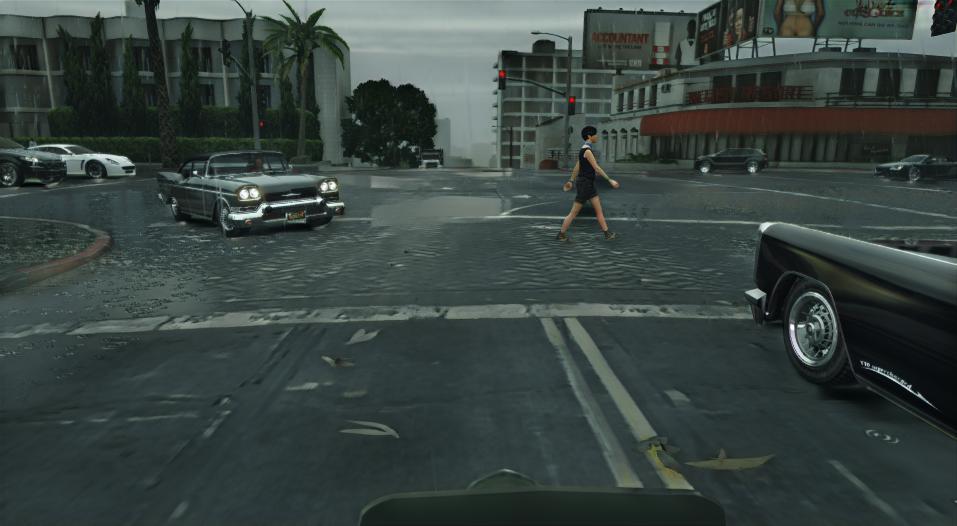}}\hfill
		{\includegraphics[width=0.198\textwidth]{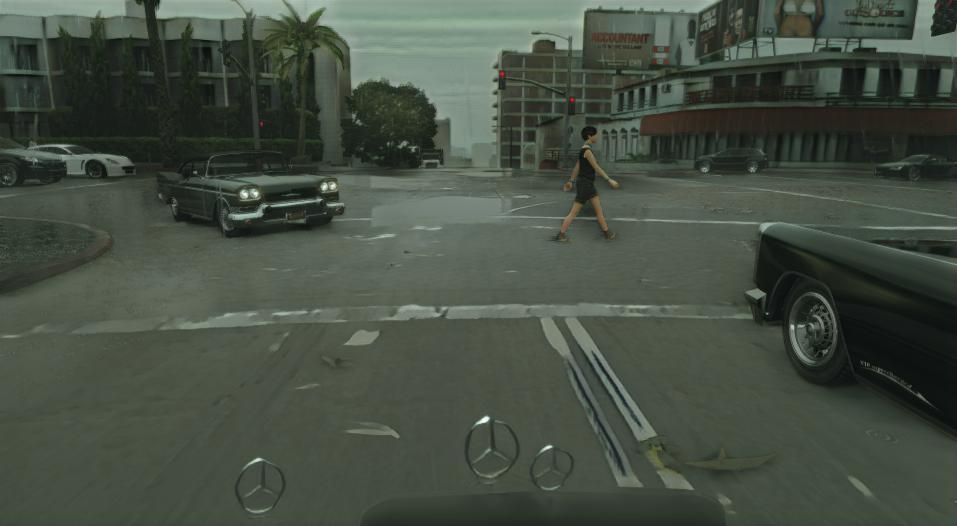}}\hfill
		{\includegraphics[width=0.198\textwidth]{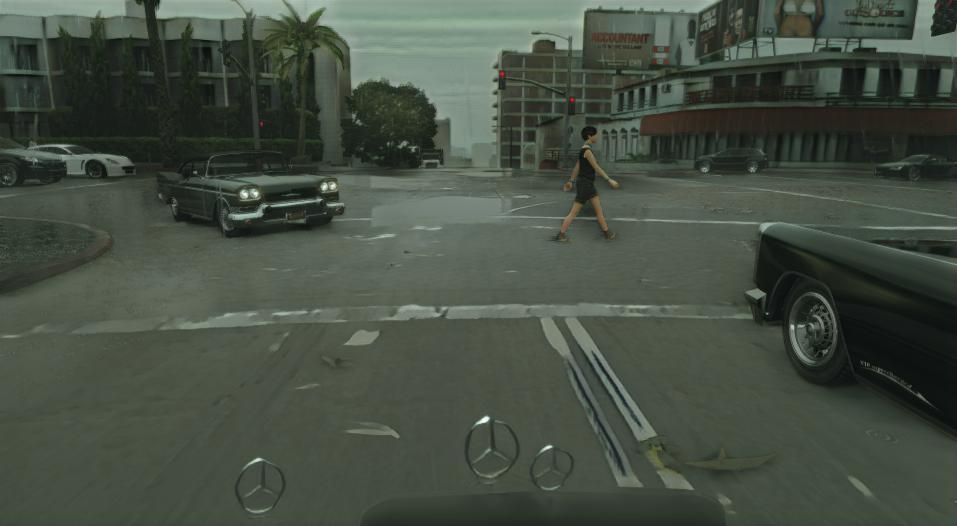}}\hfill
		{\includegraphics[width=0.198\textwidth]{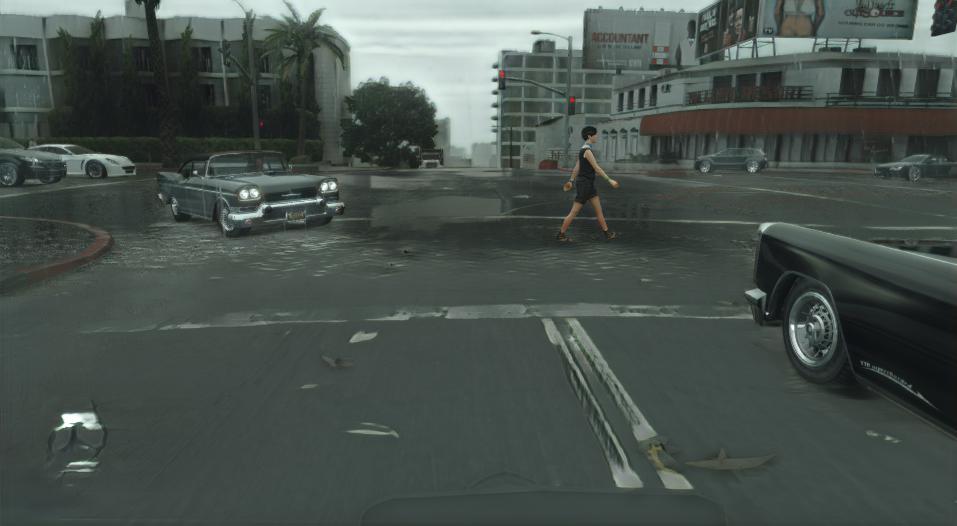}}\hfill \\ \vspace{1pt}
		{\includegraphics[width=0.198\textwidth]{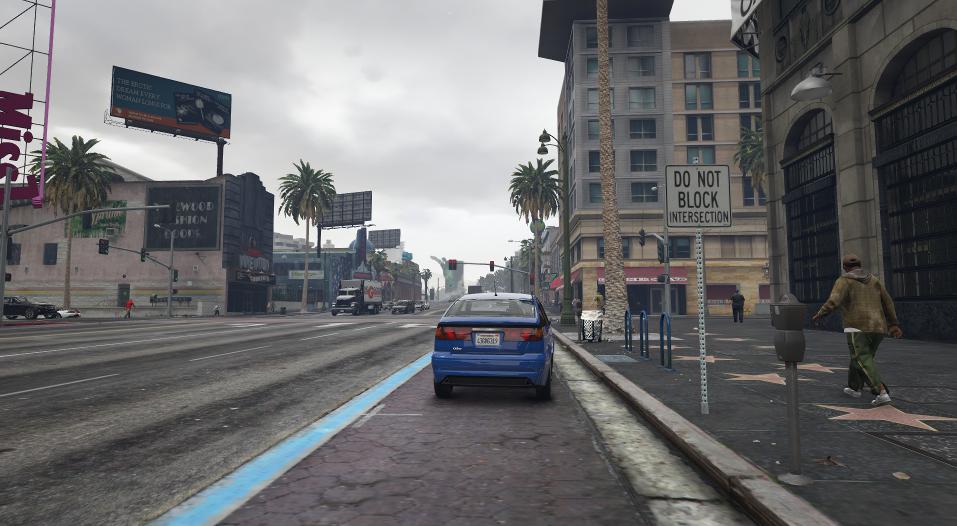}}\hfill
		{\includegraphics[width=0.198\textwidth]{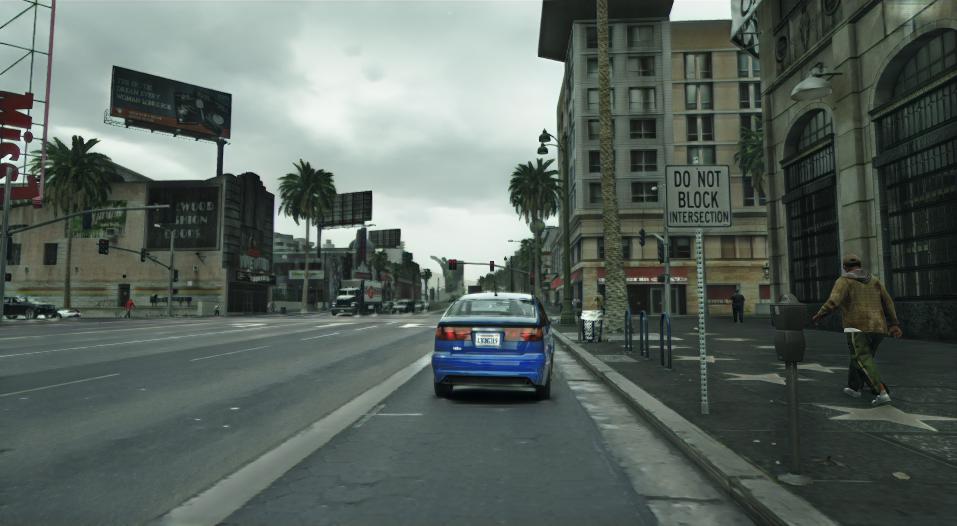}}\hfill
		{\includegraphics[width=0.198\textwidth]{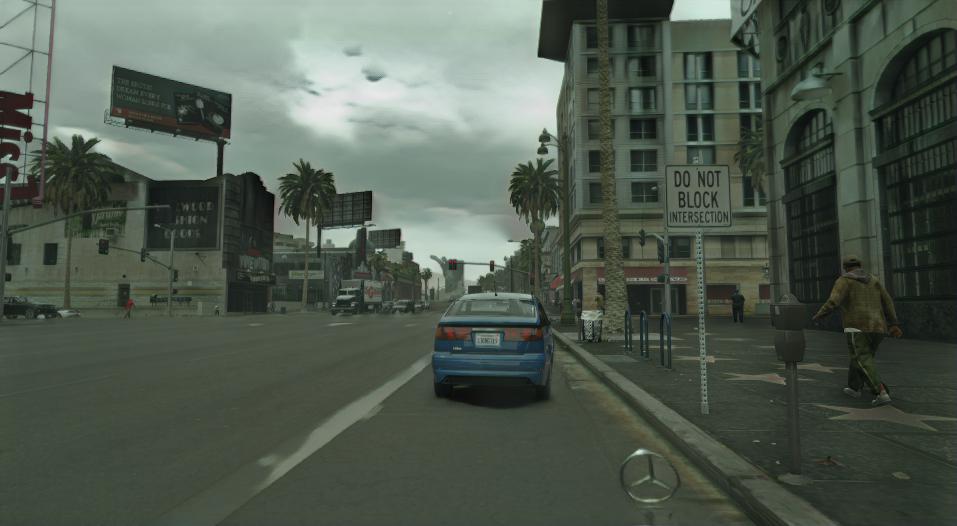}}\hfill
		{\includegraphics[width=0.198\textwidth]{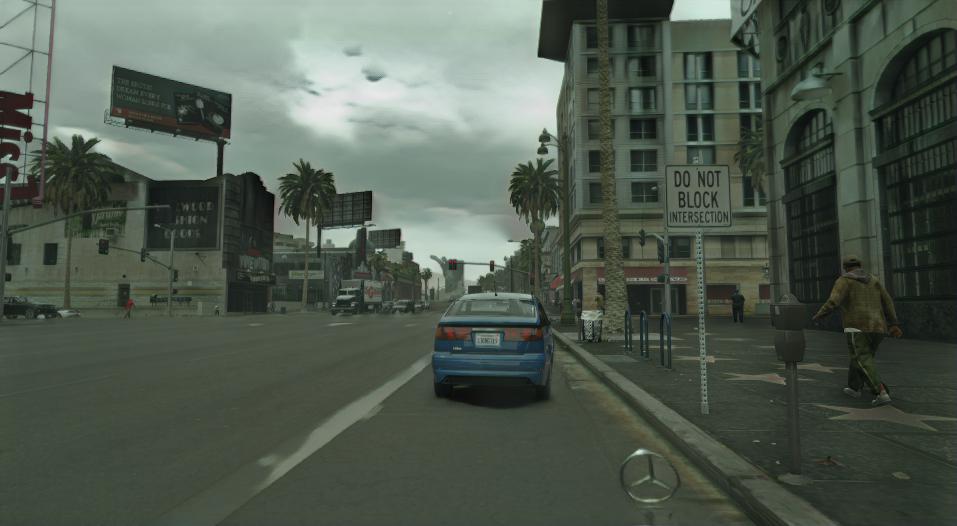}}\hfill
		{\includegraphics[width=0.198\textwidth]{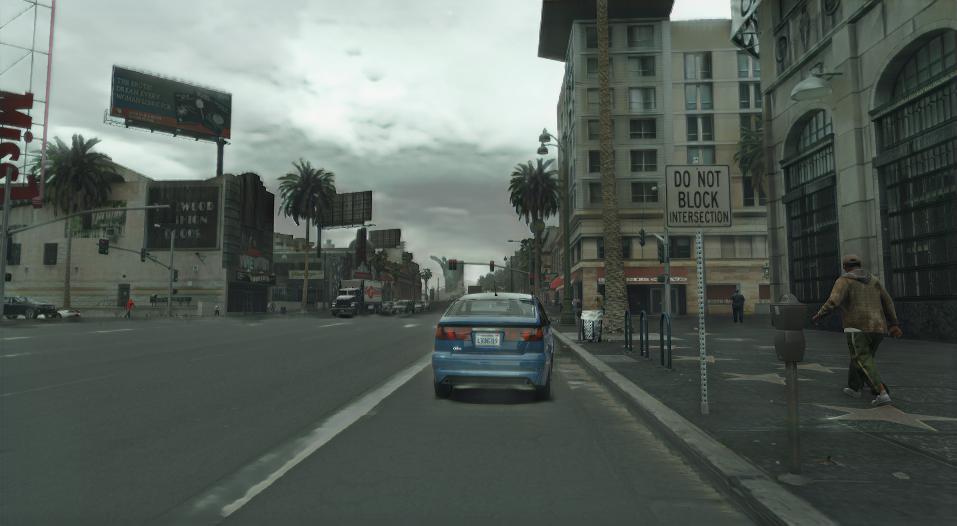}}\hfill \\ \vspace{1pt}
		{\includegraphics[width=0.198\textwidth]{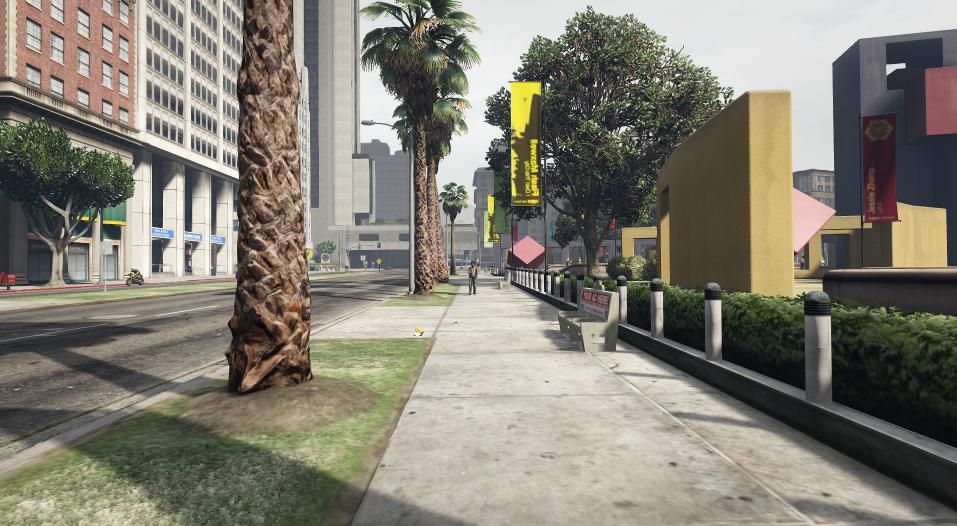}}\hfill
		{\includegraphics[width=0.198\textwidth]{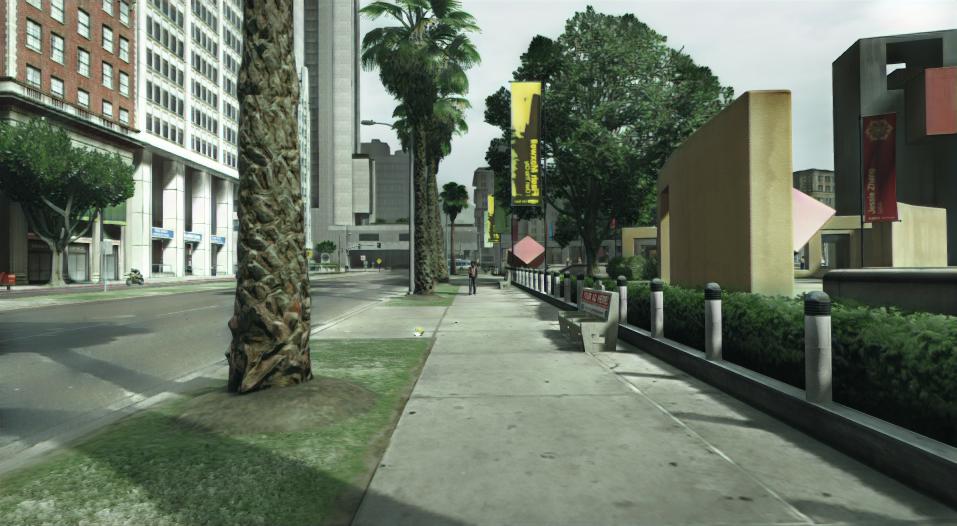}}\hfill
		{\includegraphics[width=0.198\textwidth]{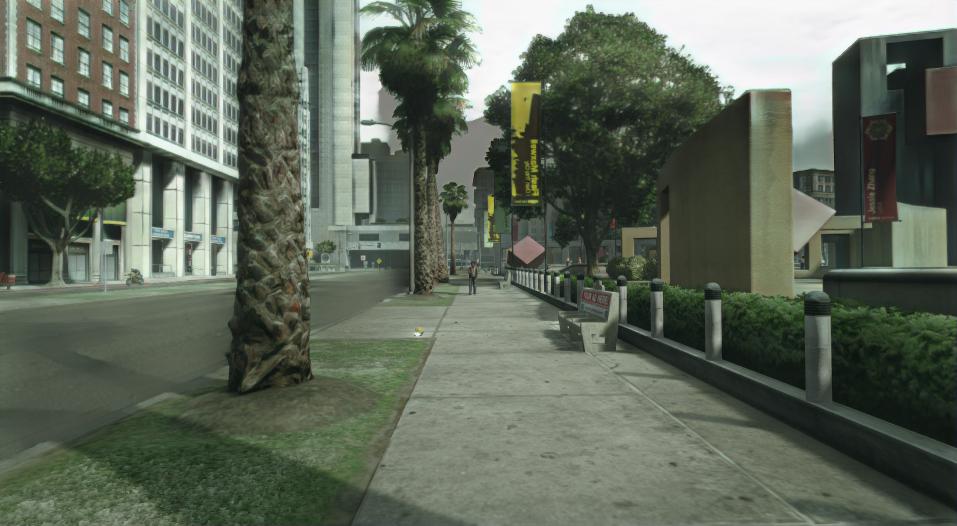}}\hfill
		{\includegraphics[width=0.198\textwidth]{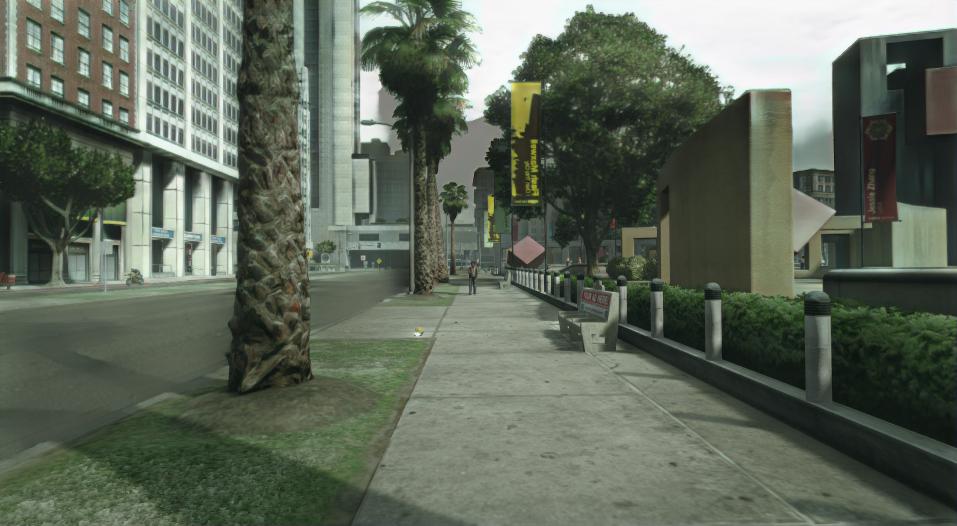}}\hfill
		{\includegraphics[width=0.198\textwidth]{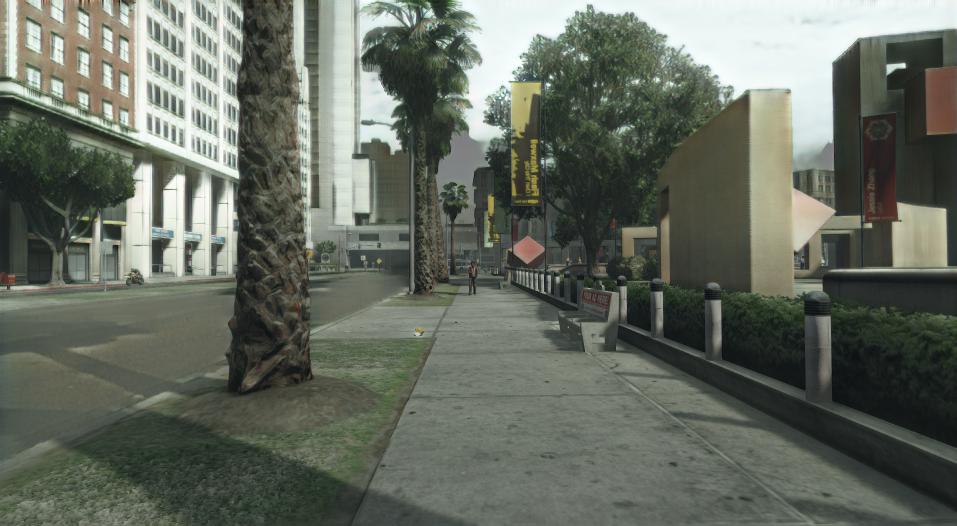}}\hfill \\
		\vspace{-4pt}
		\subfigure[Input]
		{\includegraphics[width=0.198\textwidth]{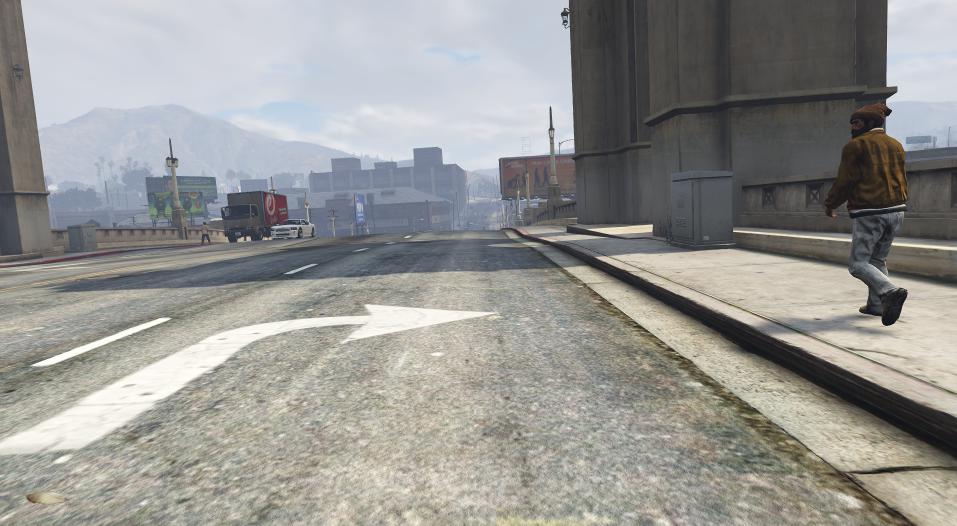}}\hfill
		\subfigure[252$\times$252]
		{\includegraphics[width=0.198\textwidth]{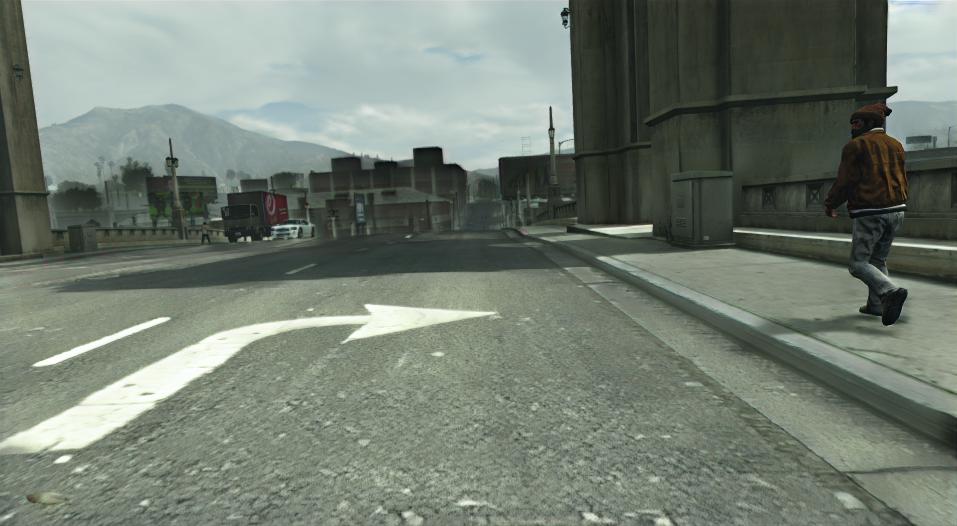}}\hfill
		\subfigure[352$\times$352]
		{\includegraphics[width=0.198\textwidth]{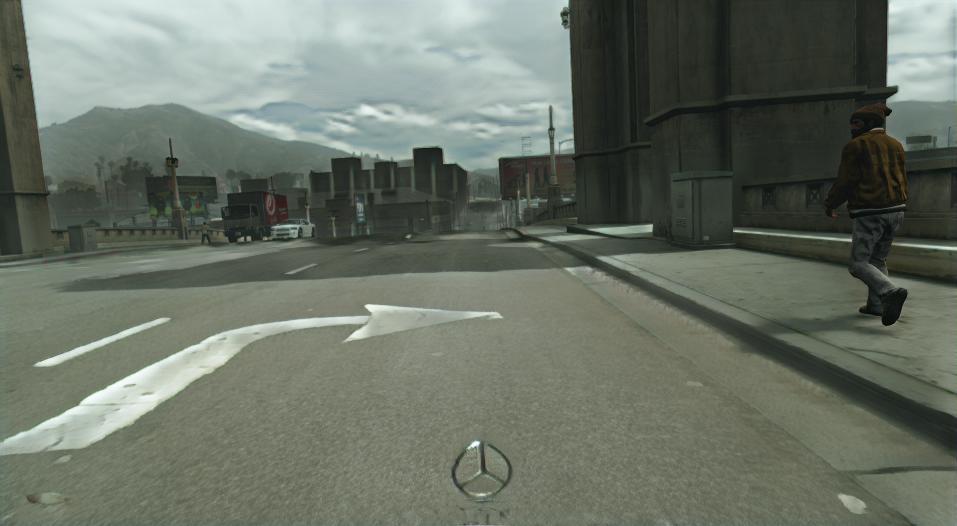}}\hfill
		\subfigure[464$\times$464]
		{\includegraphics[width=0.198\textwidth]{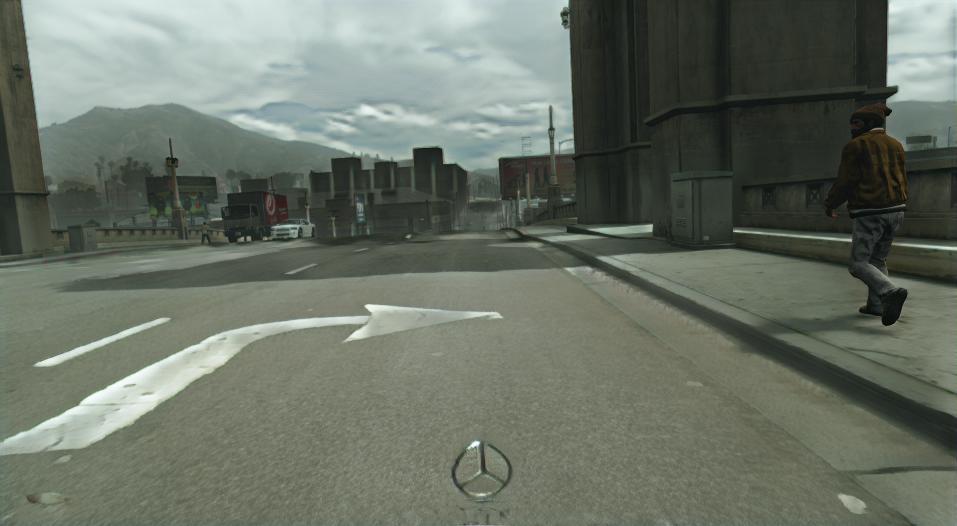}}\hfill
		\subfigure[512$\times$512]
		{\includegraphics[width=0.198\textwidth]{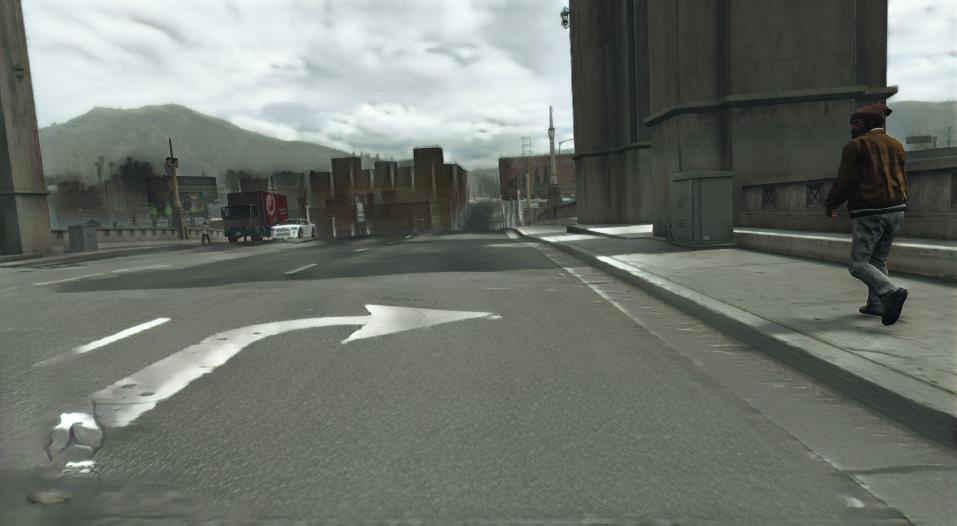}}\hfill 
	\end{center}
	\vspace{-8pt}
	\caption{\textbf{Qualitative ablation of crop sizes.} For each crop size, results are randomly sampled from the best model.}
	\label{fig:qualitative_ablation_crop_size_additional_random}
\end{figure*}

\begin{figure*}[h] 
	\renewcommand{\thesubfigure}{}
	\begin{center}
		{\includegraphics[width=0.198\textwidth]{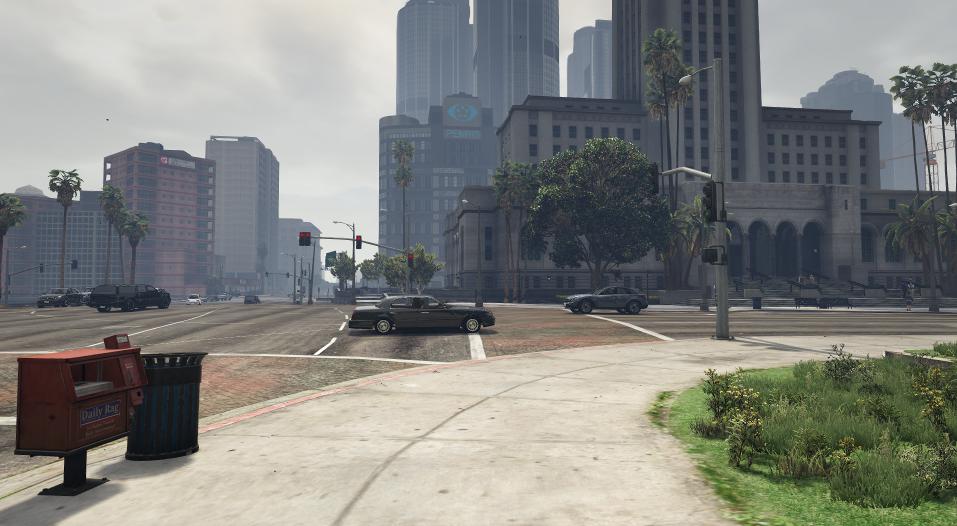}}\hfill
		{\includegraphics[width=0.198\textwidth]{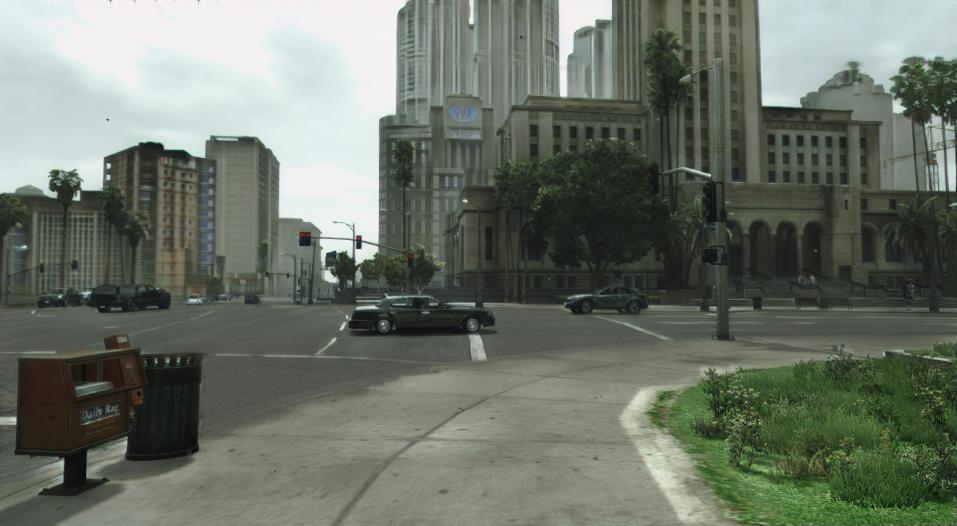}}\hfill
		{\includegraphics[width=0.198\textwidth]{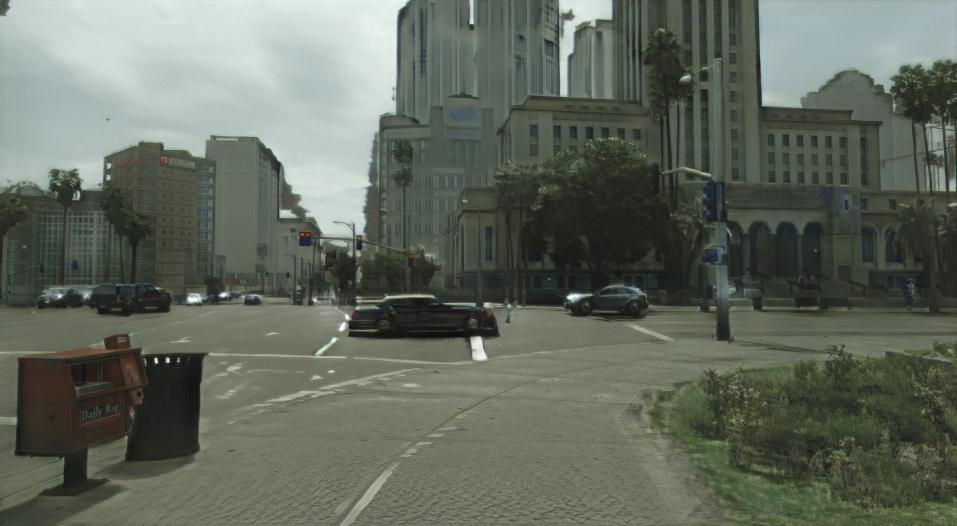}}\hfill
		{\includegraphics[width=0.198\textwidth]{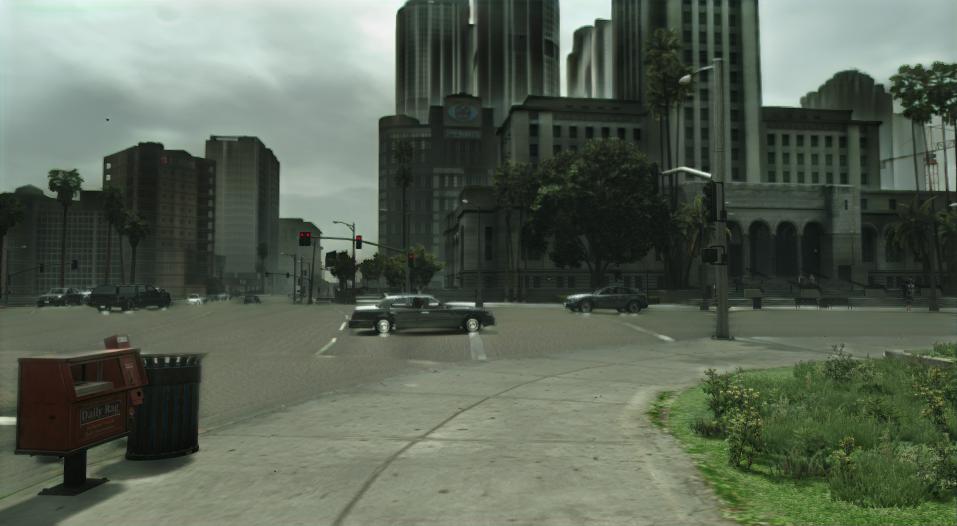}}\hfill
		{\includegraphics[width=0.198\textwidth]{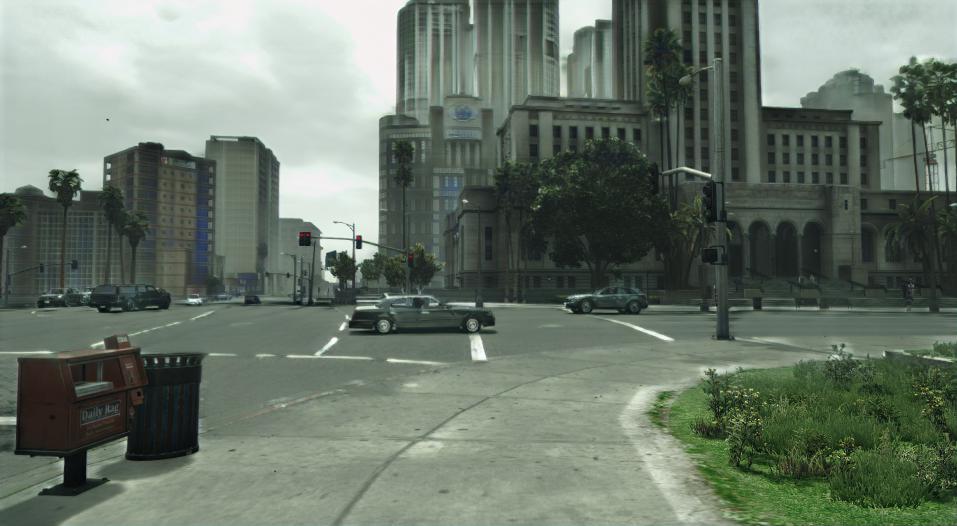}}\hfill \\ \vspace{1pt}
		{\includegraphics[width=0.198\textwidth]{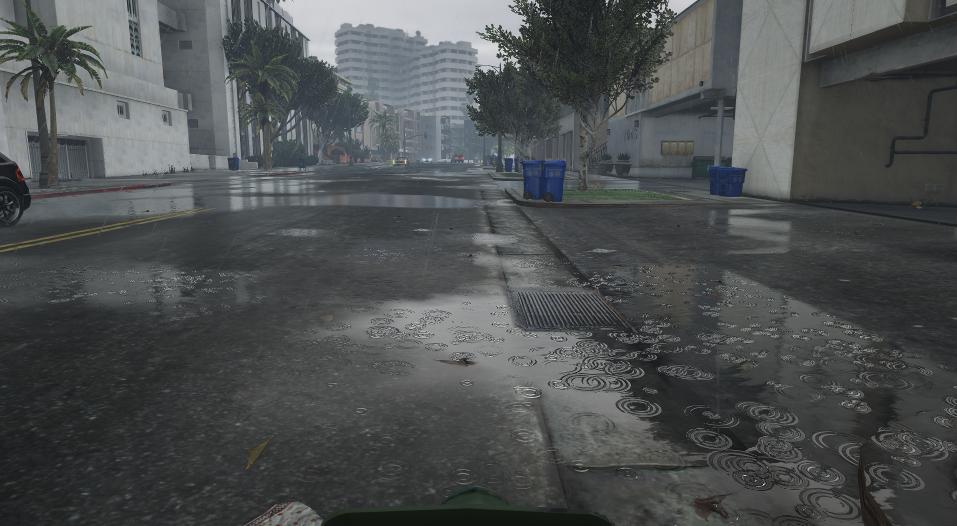}}\hfill
		{\includegraphics[width=0.198\textwidth]{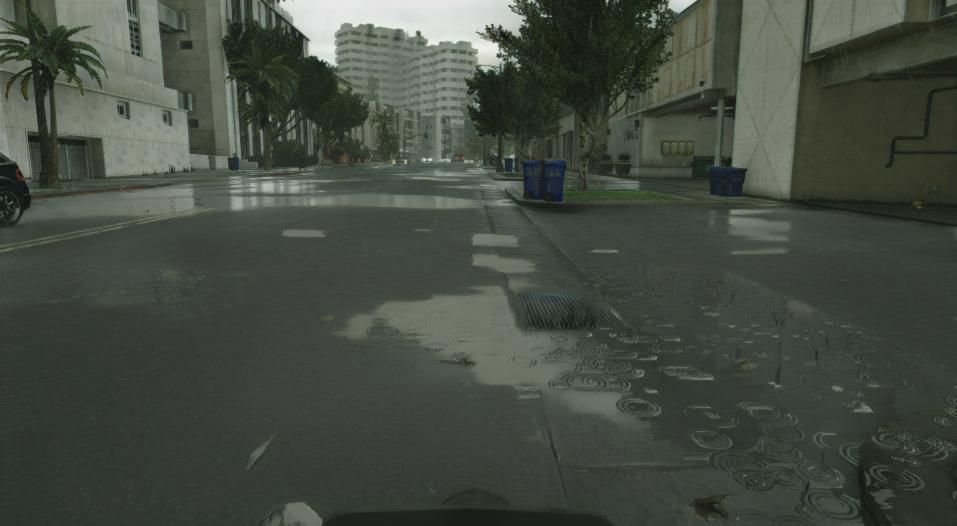}}\hfill
		{\includegraphics[width=0.198\textwidth]{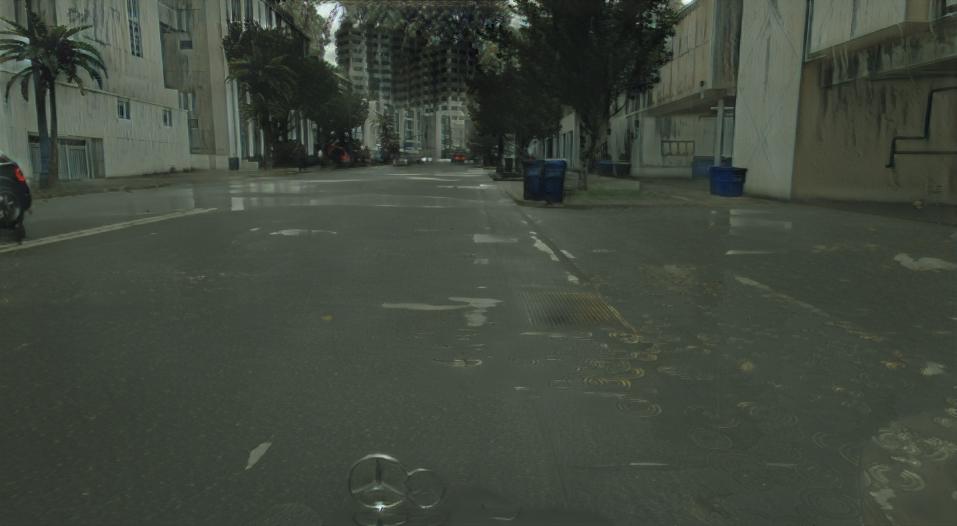}}\hfill
		{\includegraphics[width=0.198\textwidth]{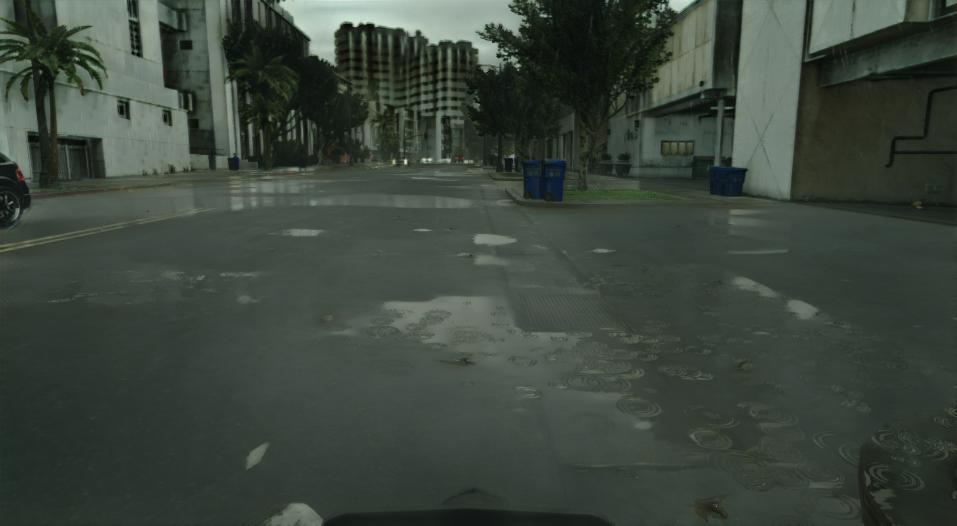}}\hfill
		{\includegraphics[width=0.198\textwidth]{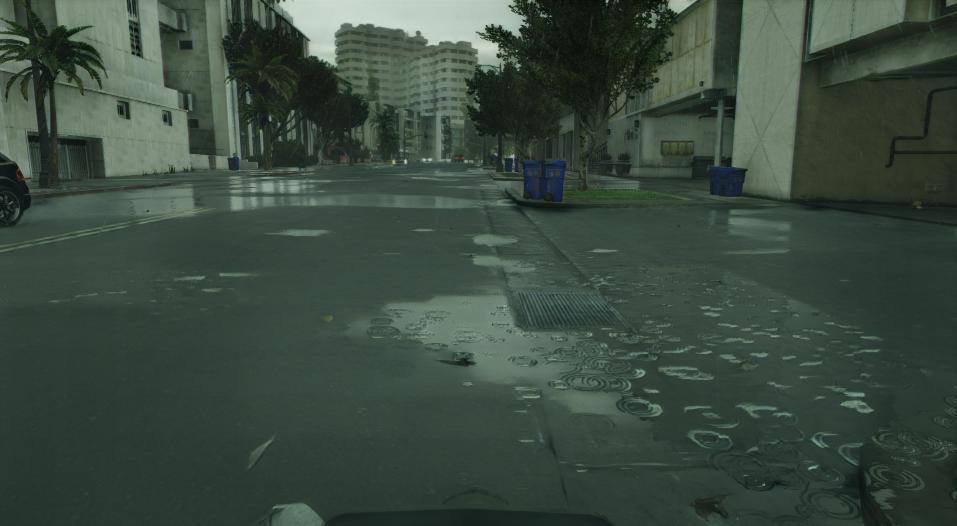}}\hfill \\ \vspace{1pt}
		{\includegraphics[width=0.198\textwidth]{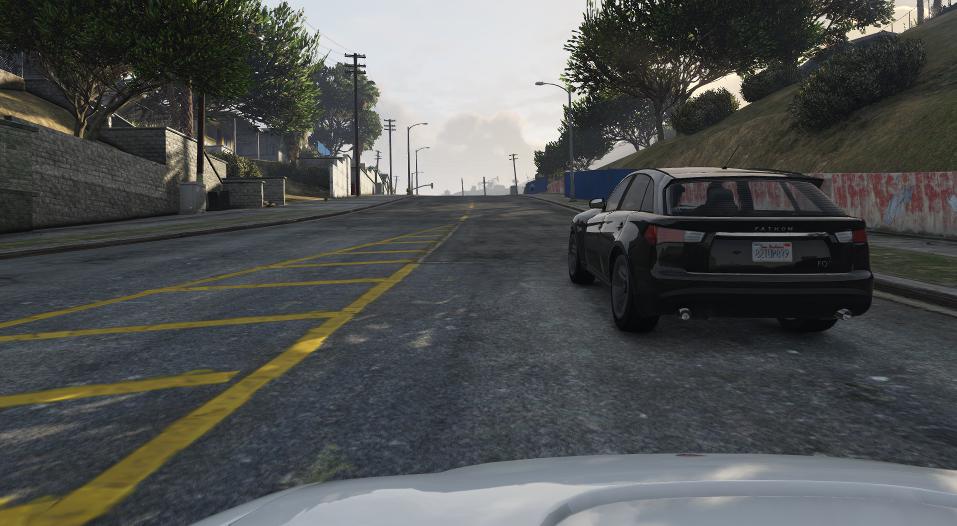}}\hfill
		{\includegraphics[width=0.198\textwidth]{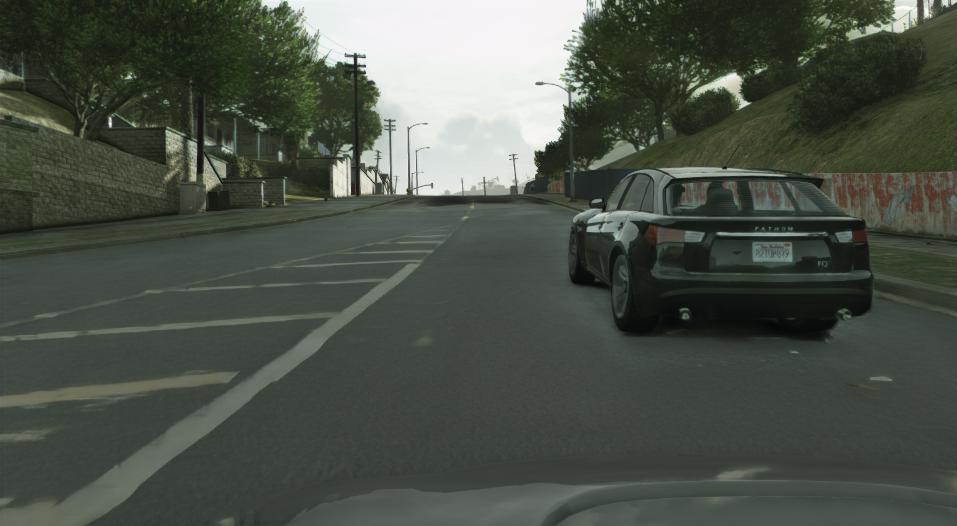}}\hfill
		{\includegraphics[width=0.198\textwidth]{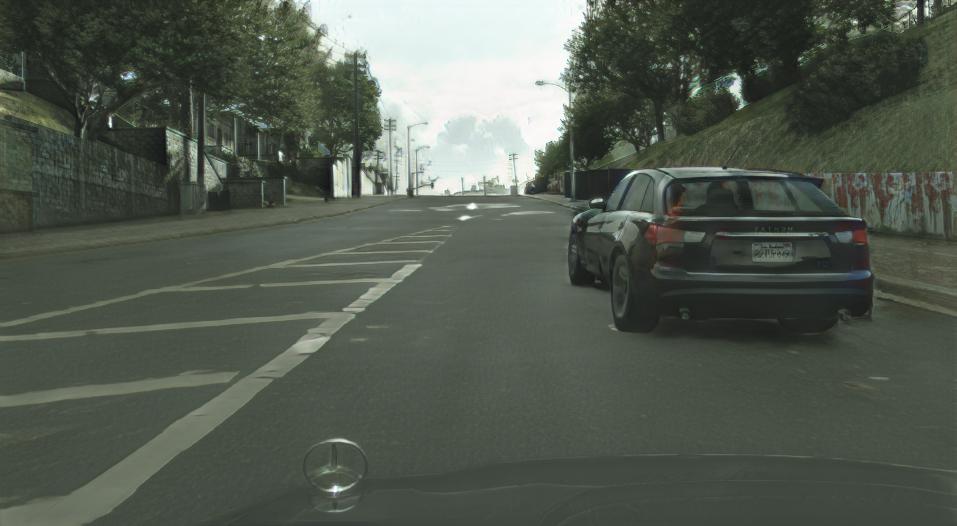}}\hfill
		{\includegraphics[width=0.198\textwidth]{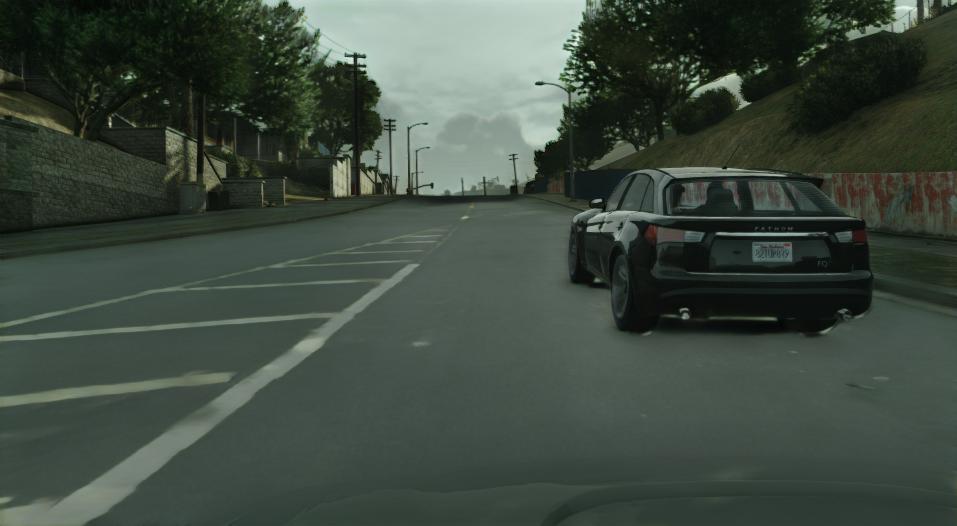}}\hfill
		{\includegraphics[width=0.198\textwidth]{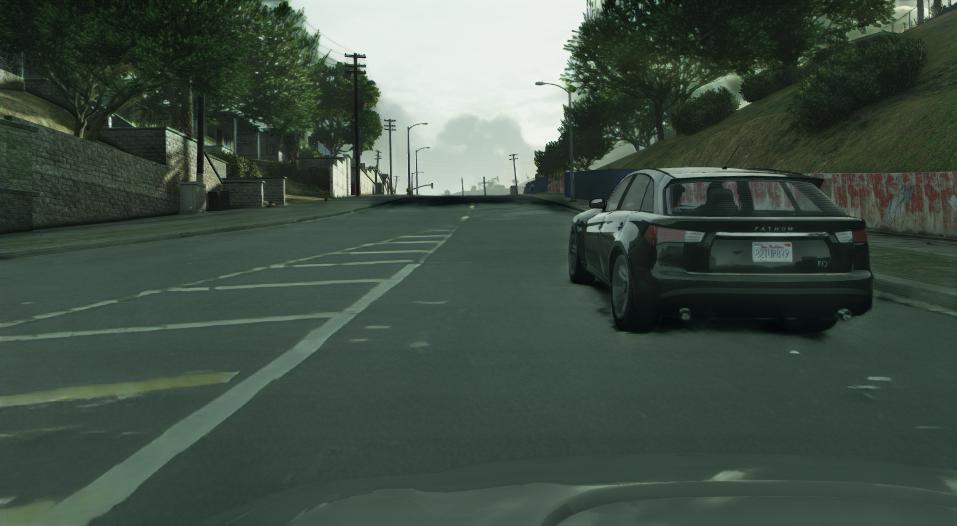}}\hfill \\ \vspace{1pt}
		{\includegraphics[width=0.198\textwidth]{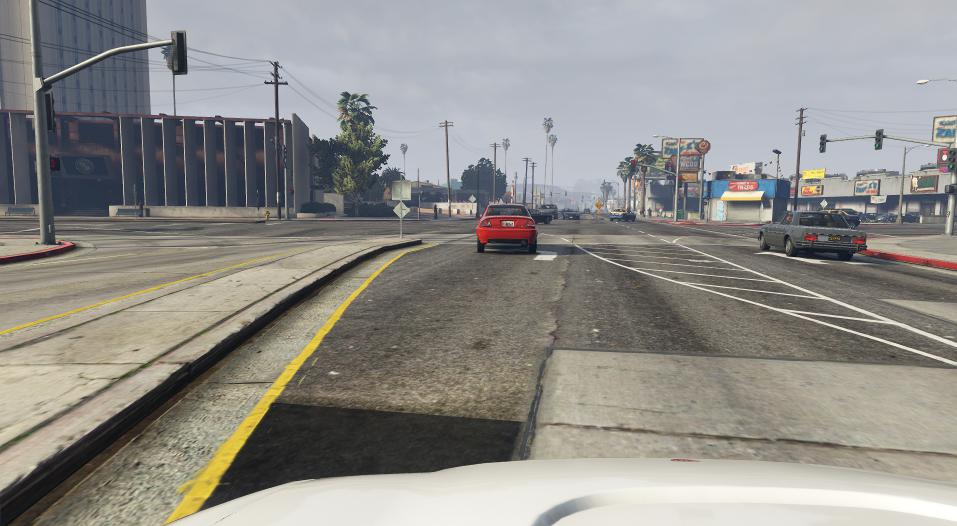}}\hfill
		{\includegraphics[width=0.198\textwidth]{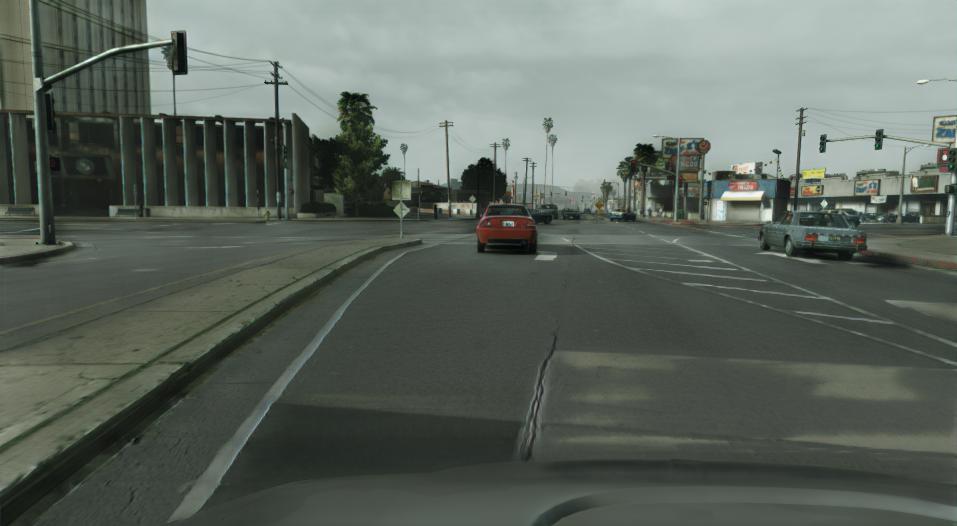}}\hfill
		{\includegraphics[width=0.198\textwidth]{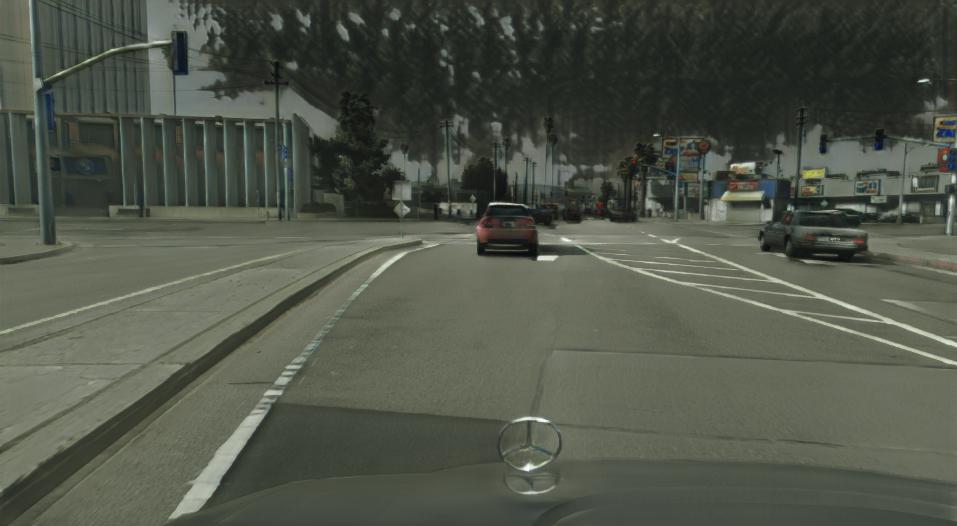}}\hfill
		{\includegraphics[width=0.198\textwidth]{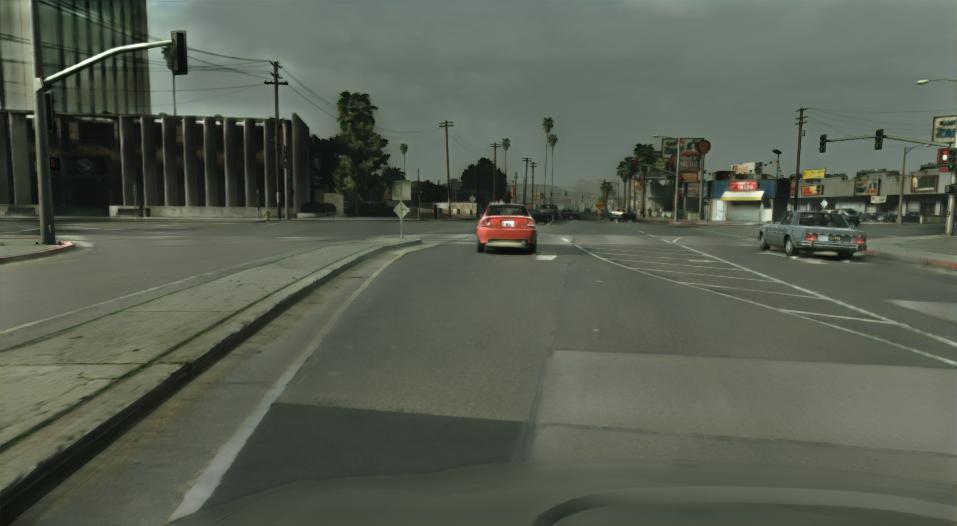}}\hfill
		{\includegraphics[width=0.198\textwidth]{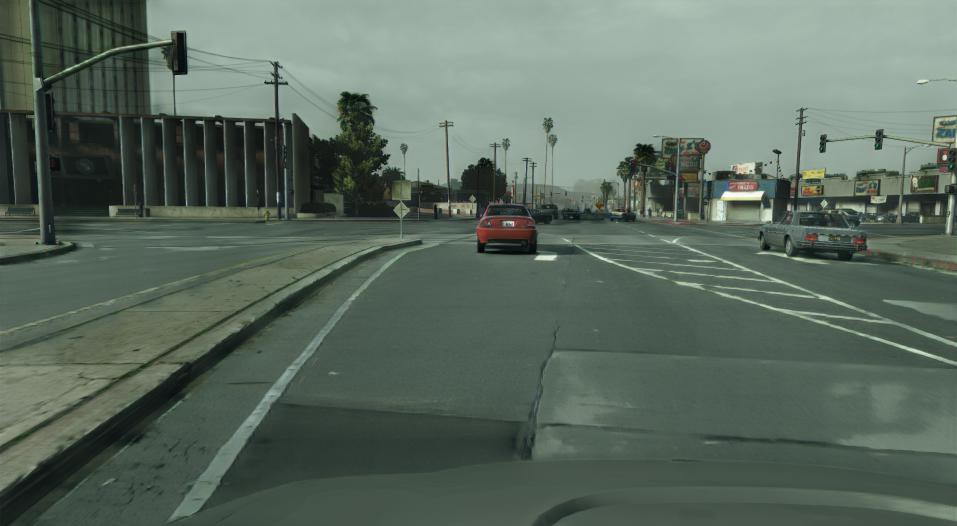}}\hfill \\ \vspace{1pt}
		{\includegraphics[width=0.198\textwidth]{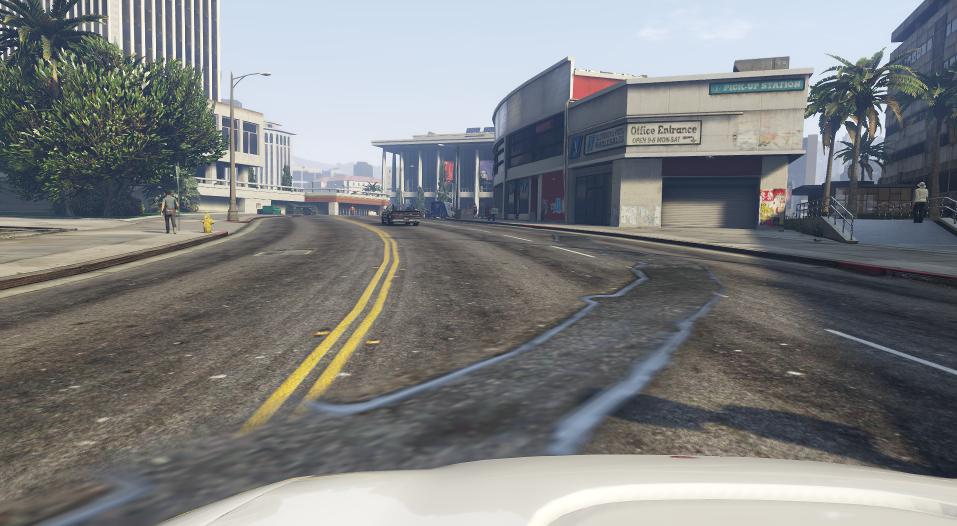}}\hfill
		{\includegraphics[width=0.198\textwidth]{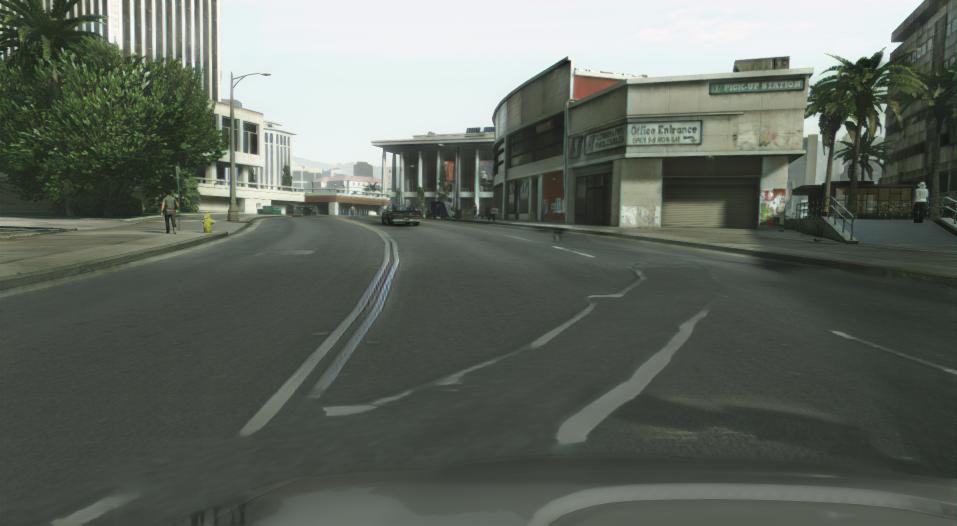}}\hfill
		{\includegraphics[width=0.198\textwidth]{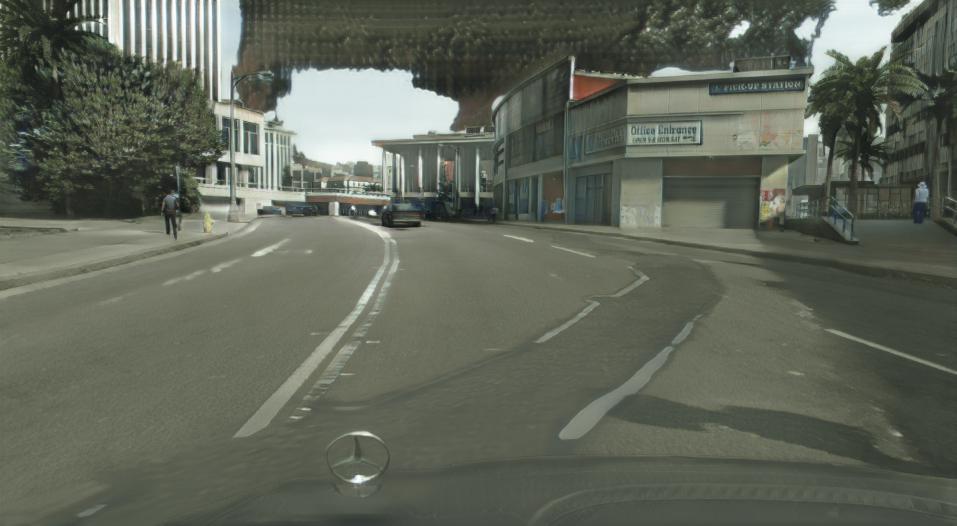}}\hfill
		{\includegraphics[width=0.198\textwidth]{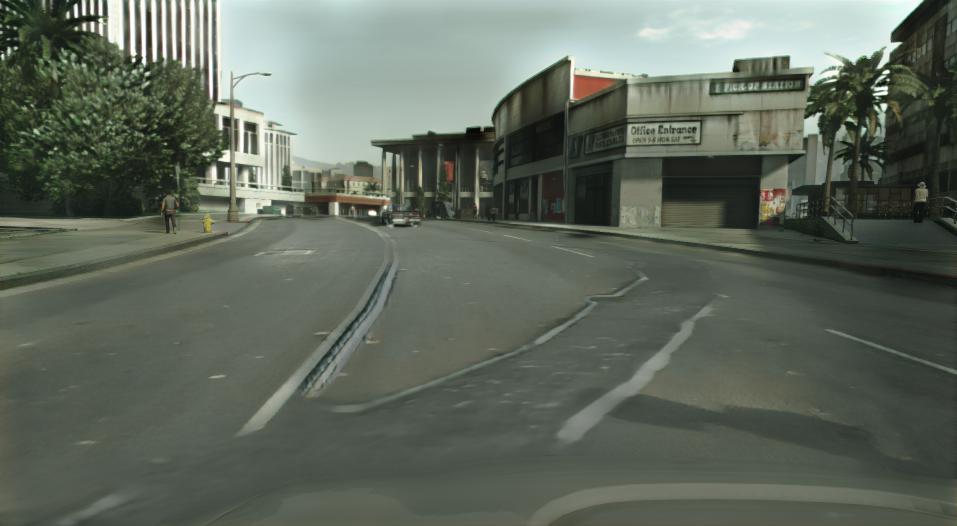}}\hfill
		{\includegraphics[width=0.198\textwidth]{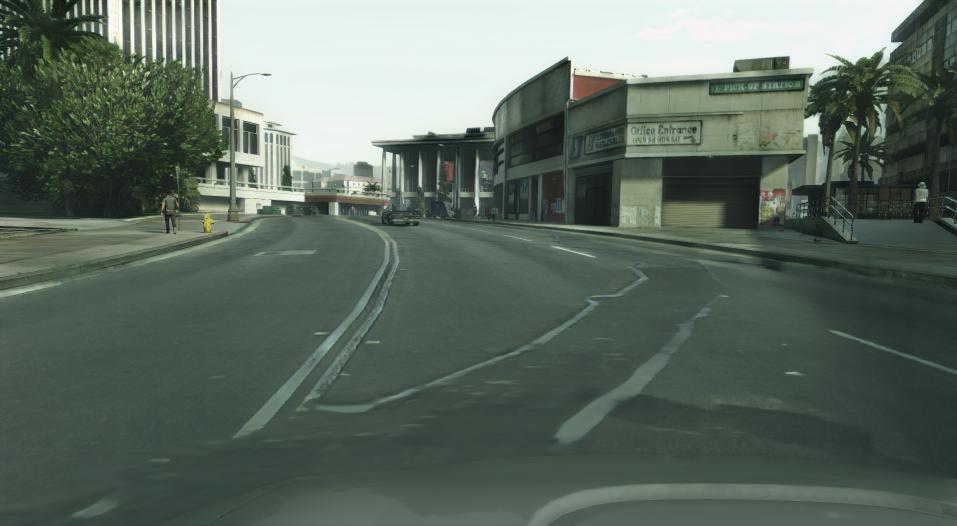}}\hfill \\ \vspace{1pt}
		{\includegraphics[width=0.198\textwidth]{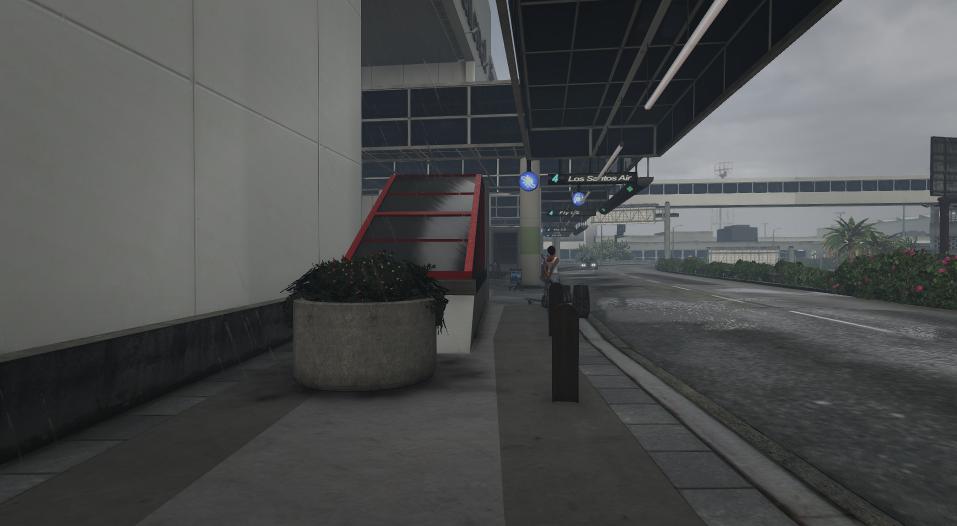}}\hfill
		{\includegraphics[width=0.198\textwidth]{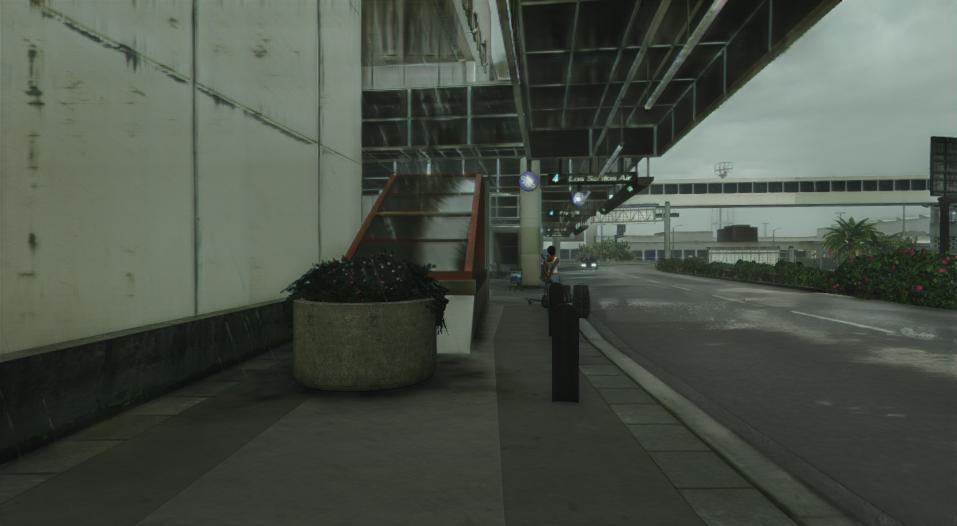}}\hfill
		{\includegraphics[width=0.198\textwidth]{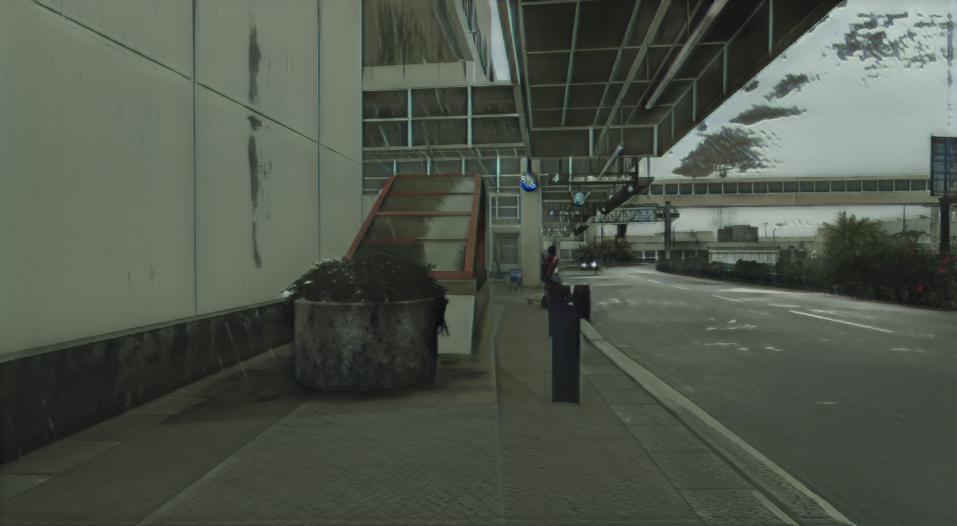}}\hfill
		{\includegraphics[width=0.198\textwidth]{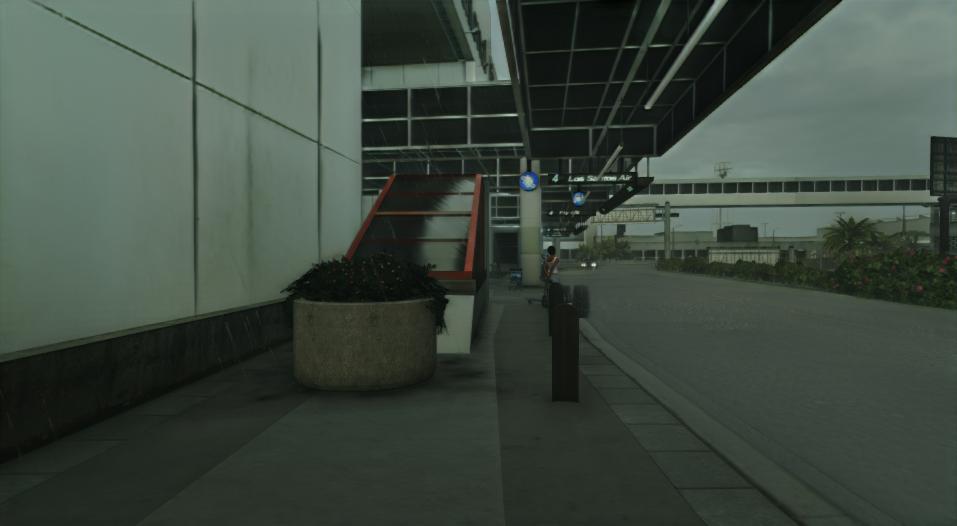}}\hfill
		{\includegraphics[width=0.198\textwidth]{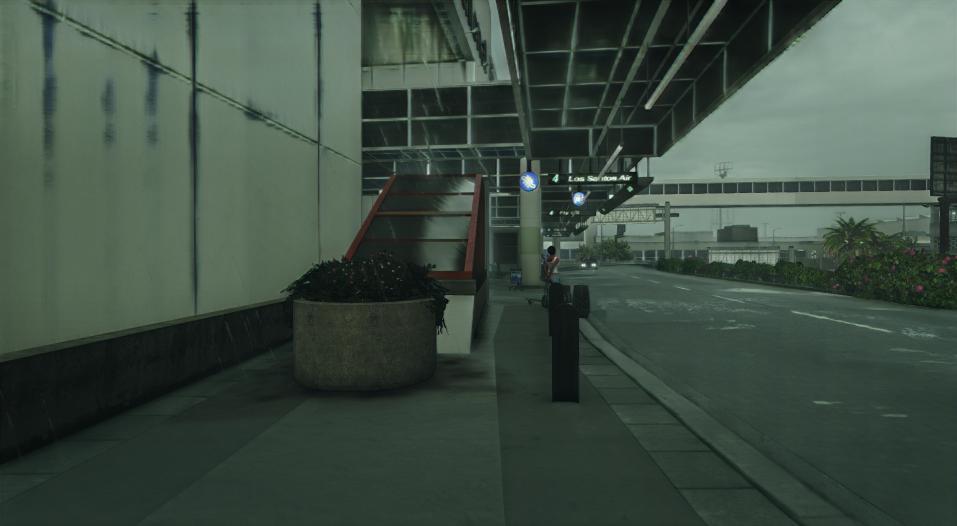}}\hfill \\ \vspace{1pt}
		{\includegraphics[width=0.198\textwidth]{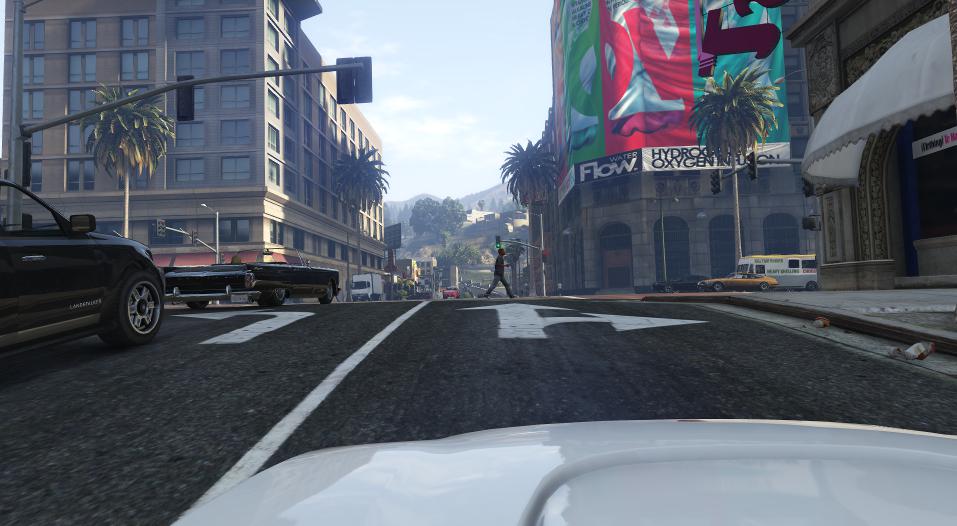}}\hfill
		{\includegraphics[width=0.198\textwidth]{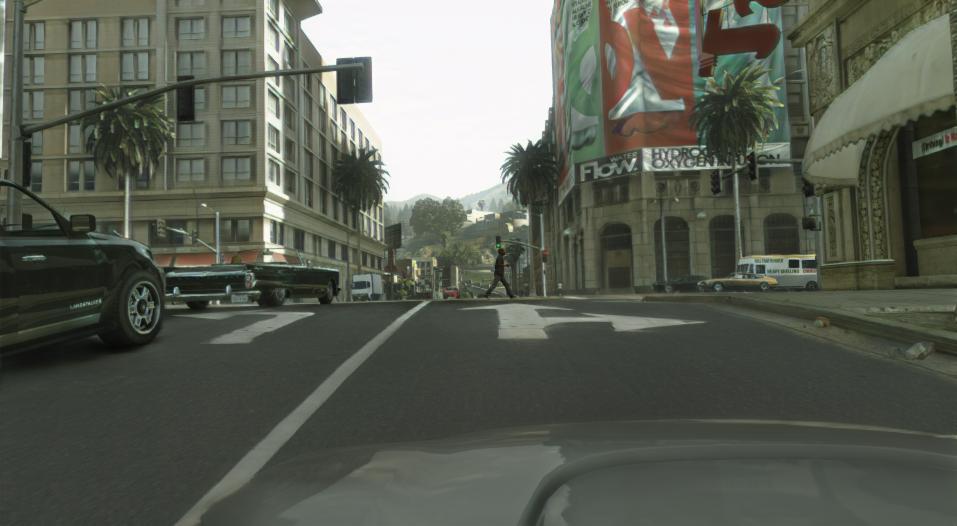}}\hfill
		{\includegraphics[width=0.198\textwidth]{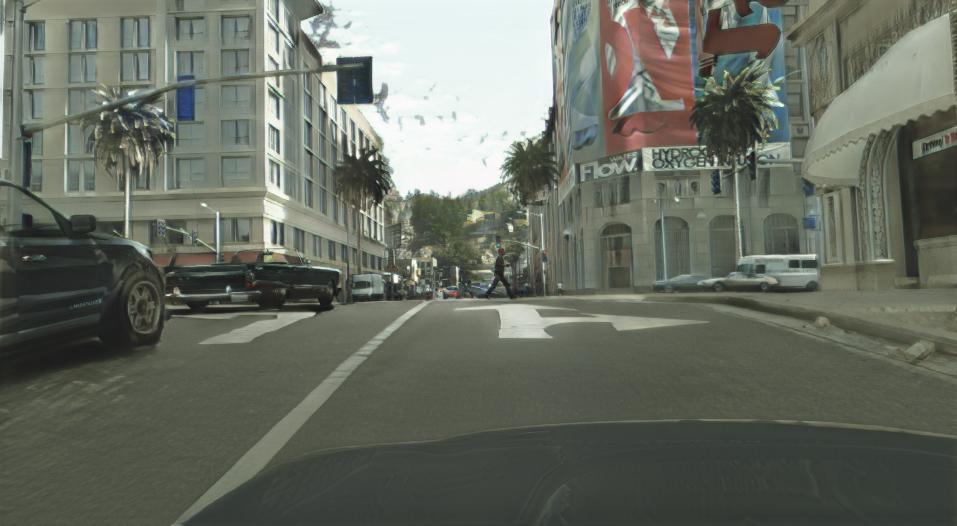}}\hfill
		{\includegraphics[width=0.198\textwidth]{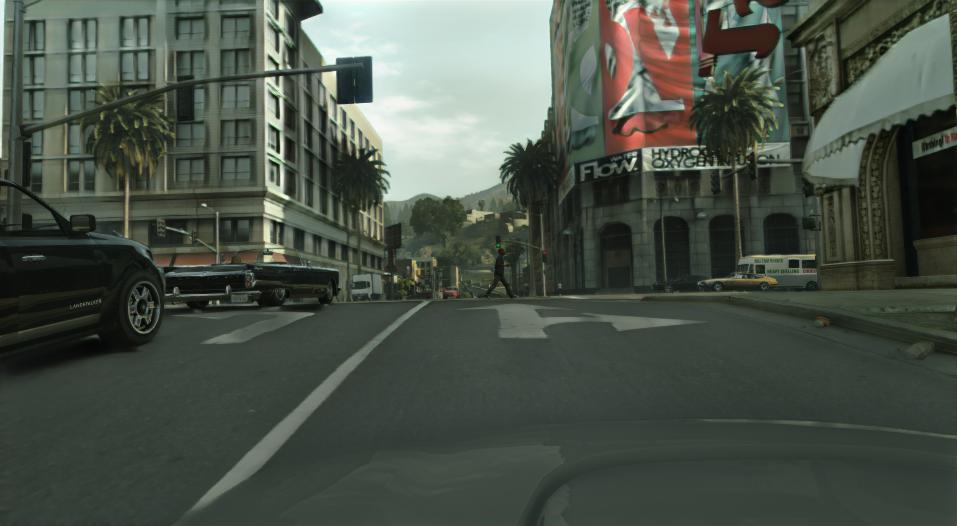}}\hfill
		{\includegraphics[width=0.198\textwidth]{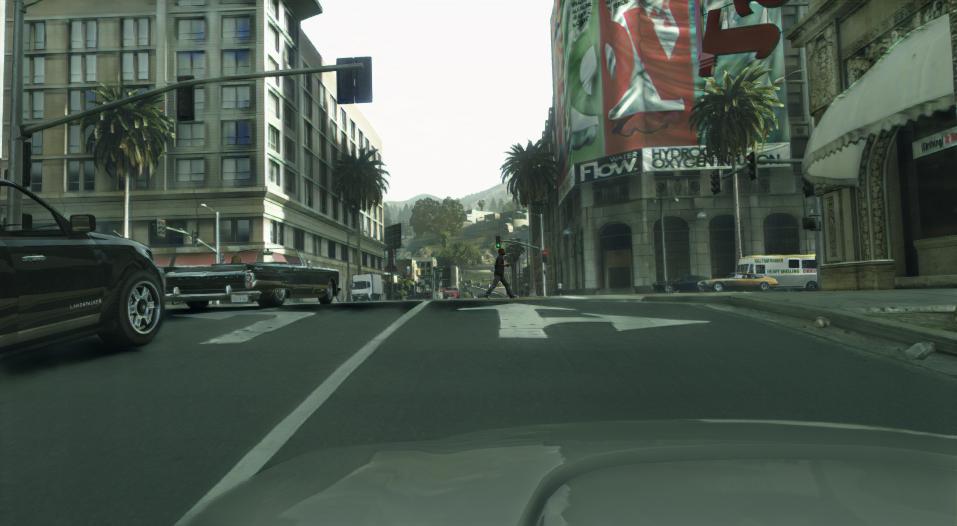}}\hfill \\ \vspace{1pt}
		{\includegraphics[width=0.198\textwidth]{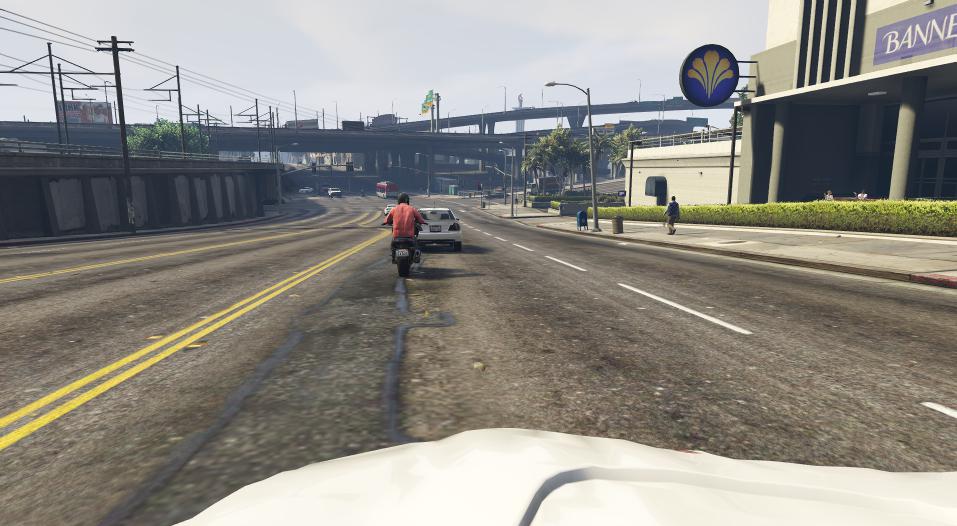}}\hfill
		{\includegraphics[width=0.198\textwidth]{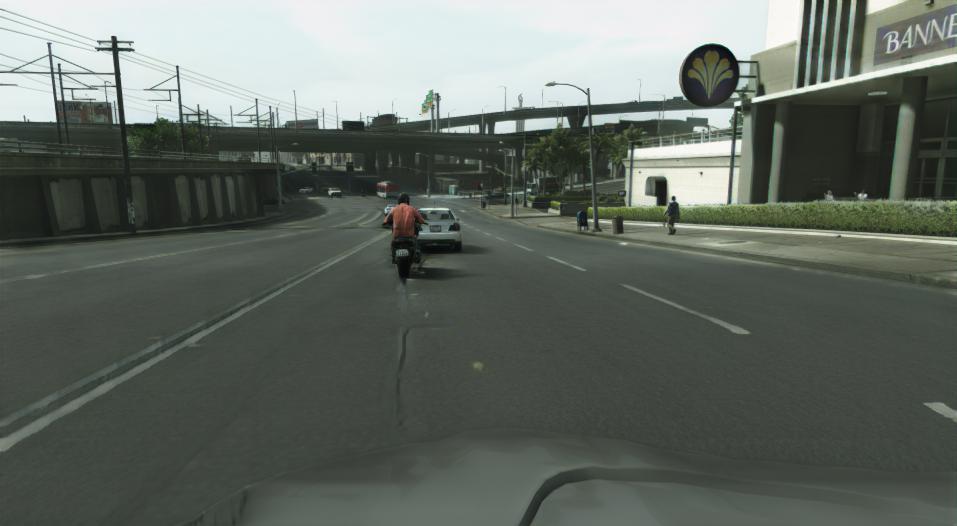}}\hfill
		{\includegraphics[width=0.198\textwidth]{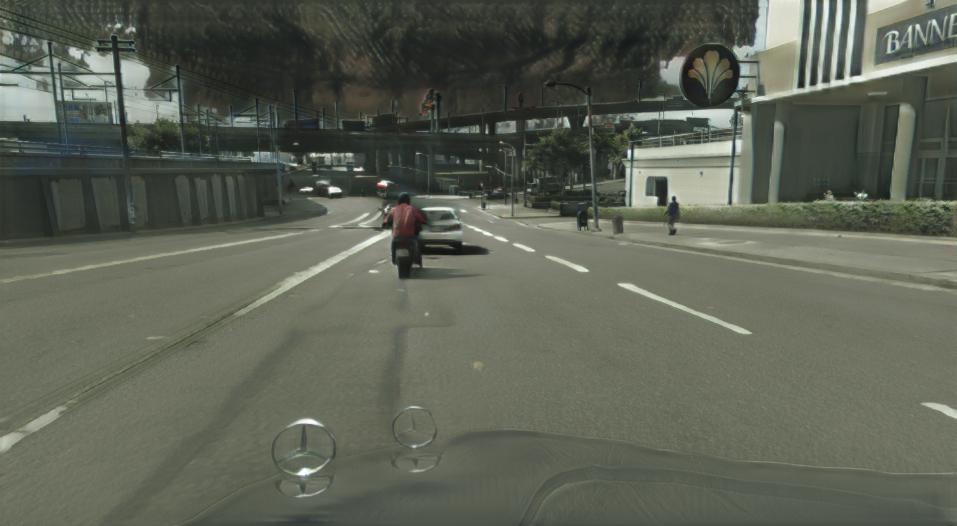}}\hfill
		{\includegraphics[width=0.198\textwidth]{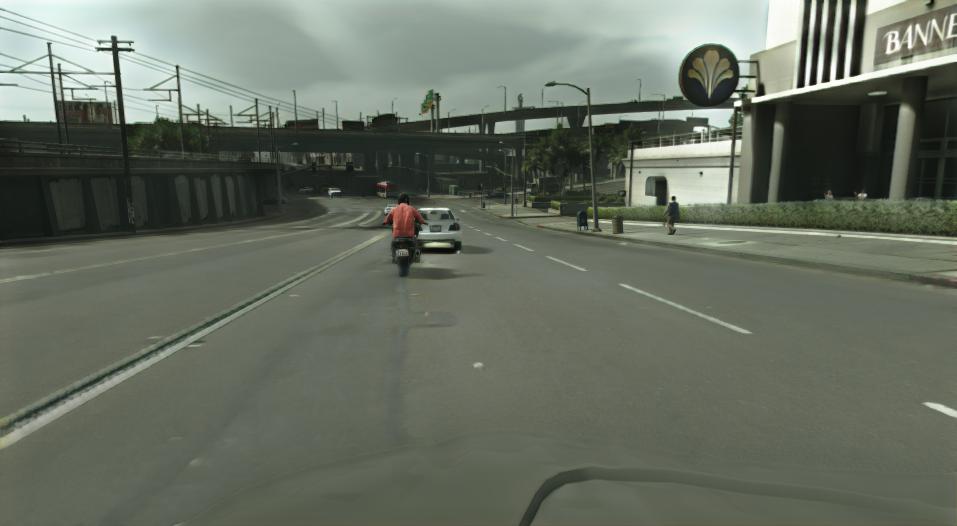}}\hfill
		{\includegraphics[width=0.198\textwidth]{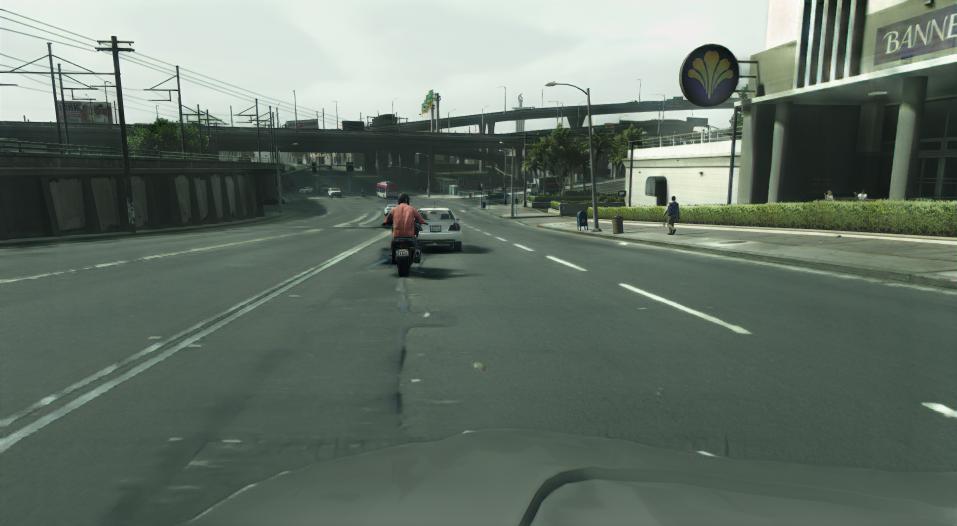}}\hfill \\ \vspace{1pt}
		{\includegraphics[width=0.198\textwidth]{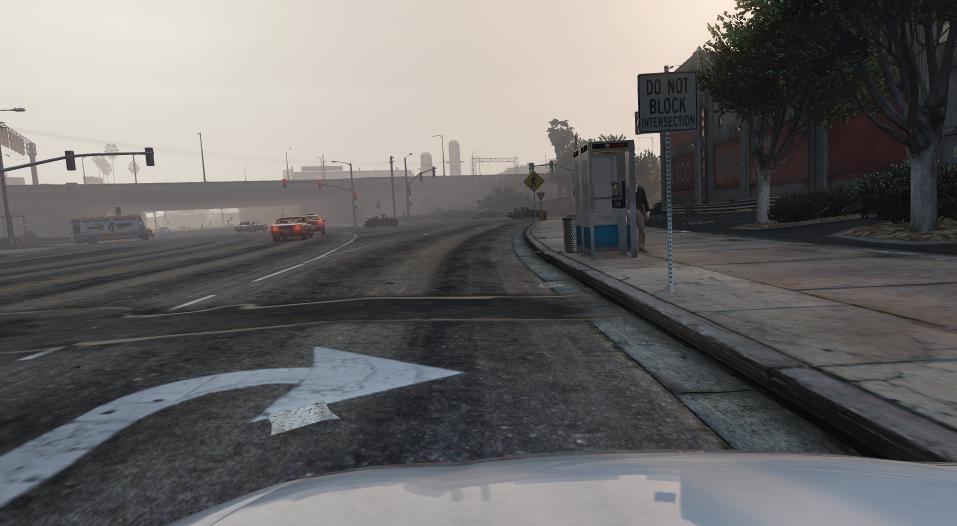}}\hfill
		{\includegraphics[width=0.198\textwidth]{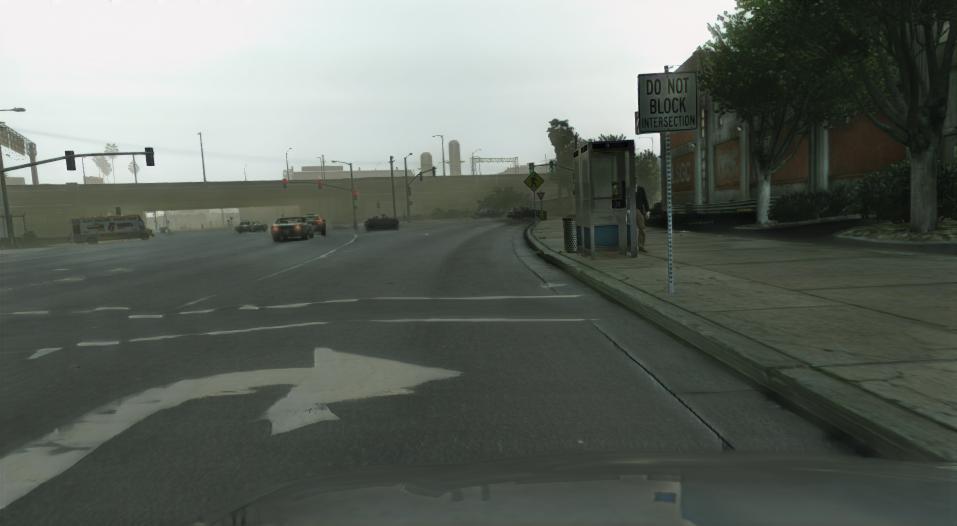}}\hfill
		{\includegraphics[width=0.198\textwidth]{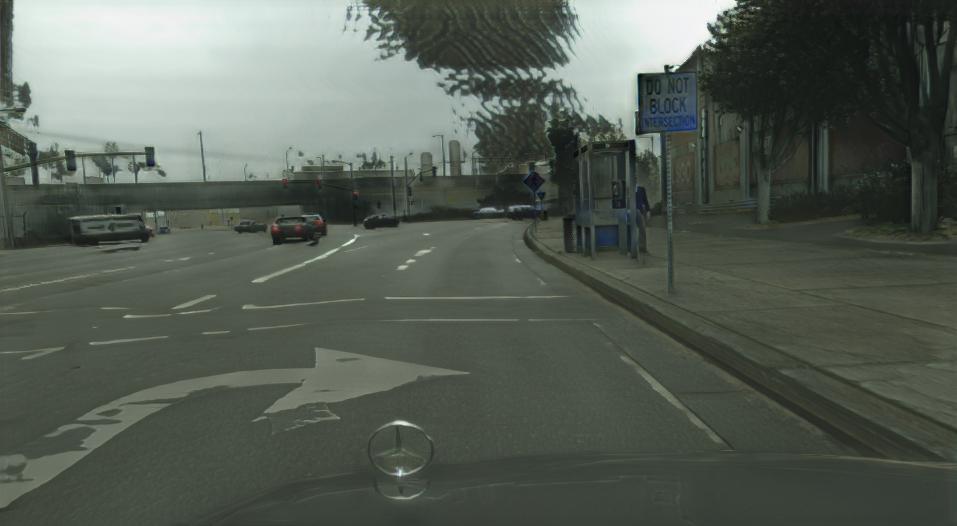}}\hfill
		{\includegraphics[width=0.198\textwidth]{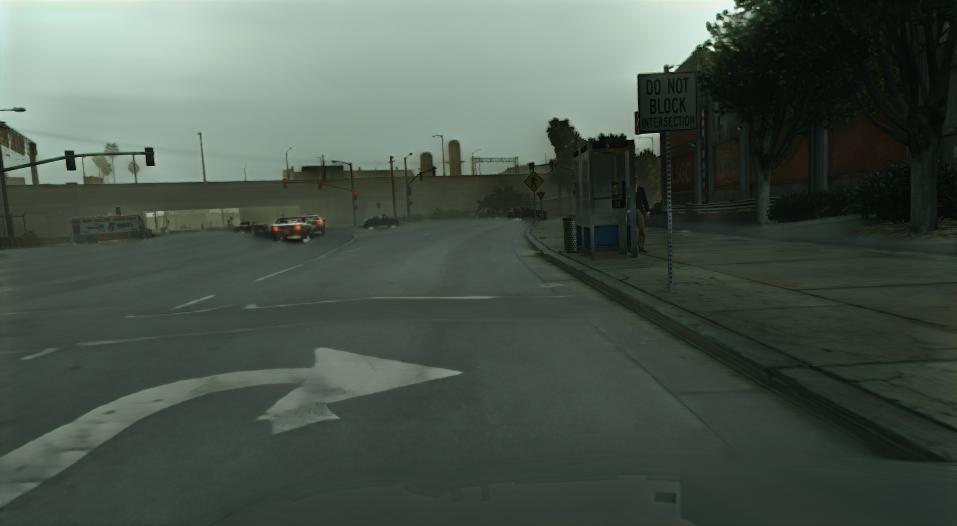}}\hfill
		{\includegraphics[width=0.198\textwidth]{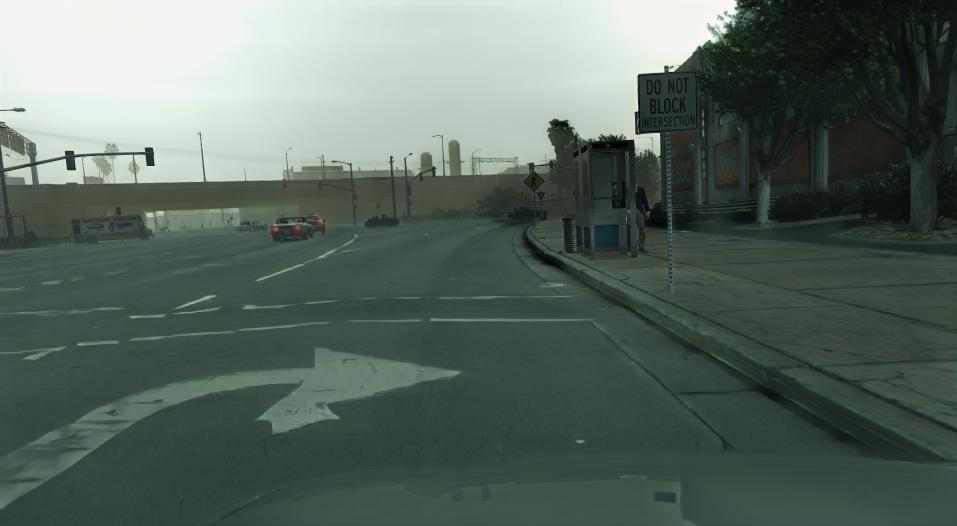}}\hfill \\ \vspace{1pt}
		{\includegraphics[width=0.198\textwidth]{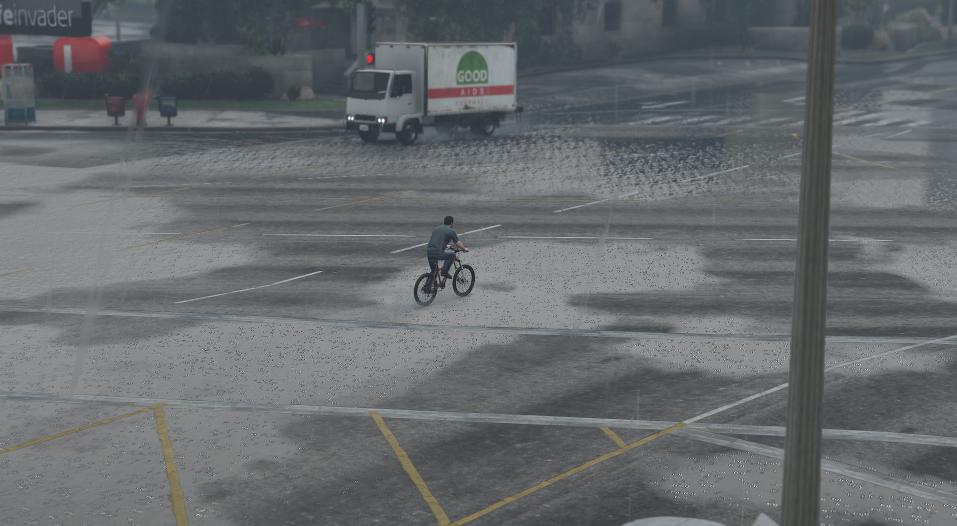}}\hfill
		{\includegraphics[width=0.198\textwidth]{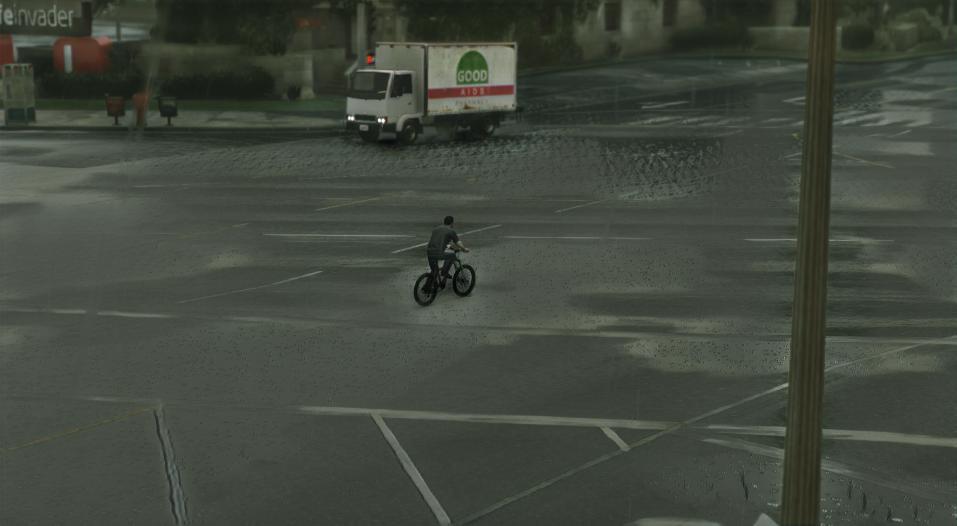}}\hfill
		{\includegraphics[width=0.198\textwidth]{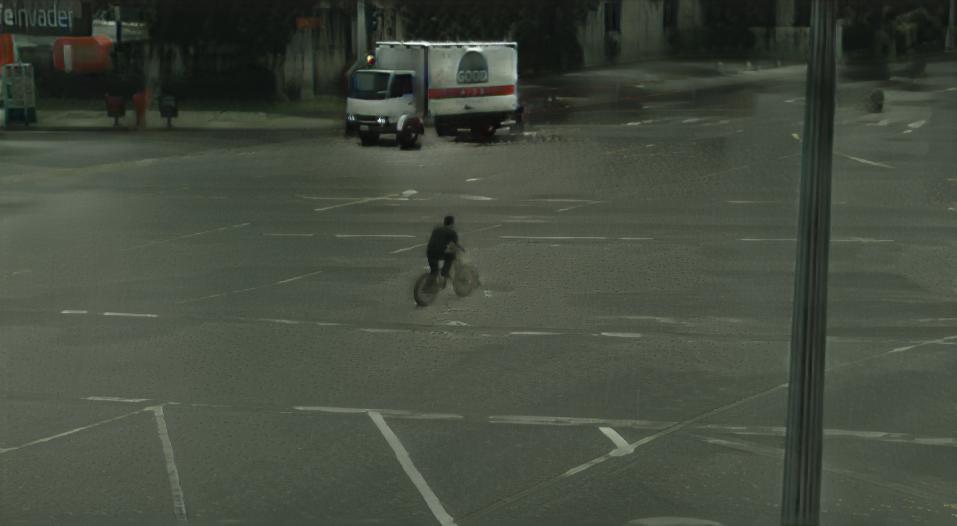}}\hfill
		{\includegraphics[width=0.198\textwidth]{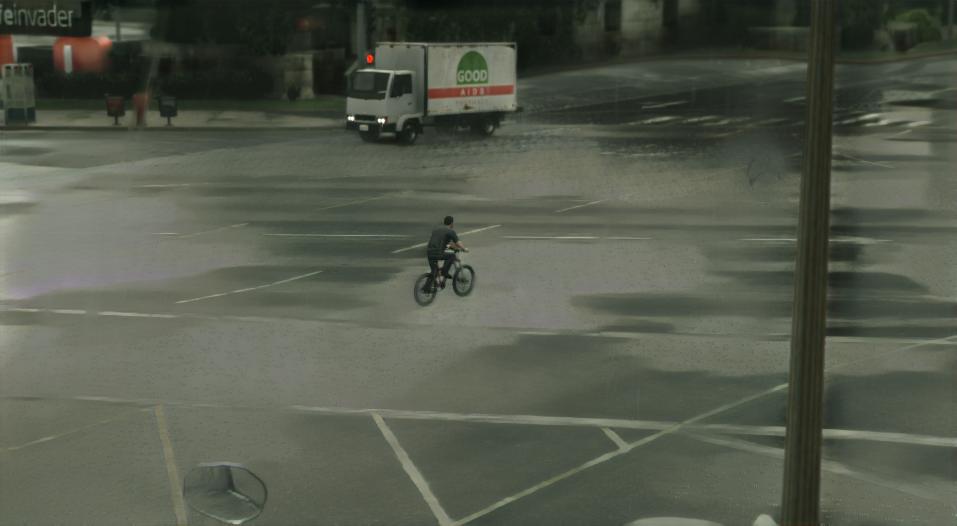}}\hfill
		{\includegraphics[width=0.198\textwidth]{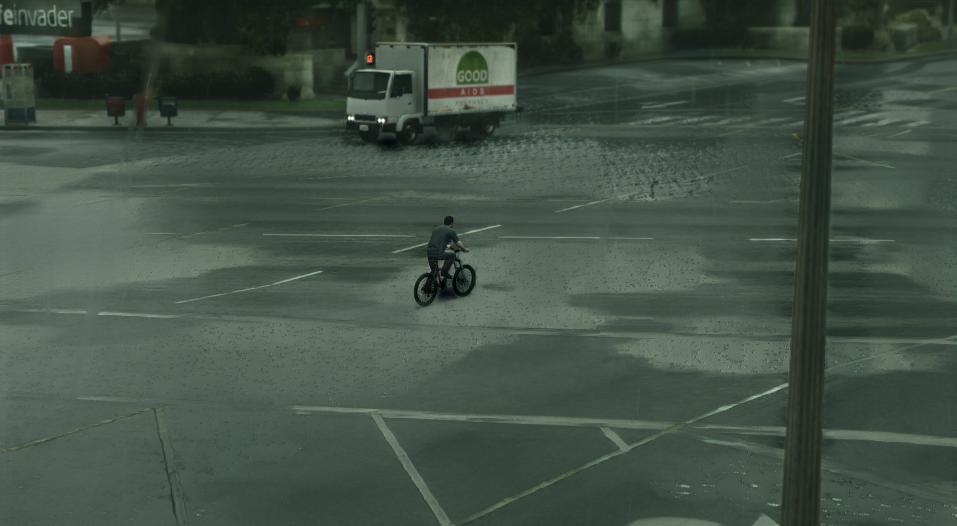}}\hfill \\ 
		\vspace{-4pt}
		\subfigure[Input]
		{\includegraphics[width=0.198\textwidth]{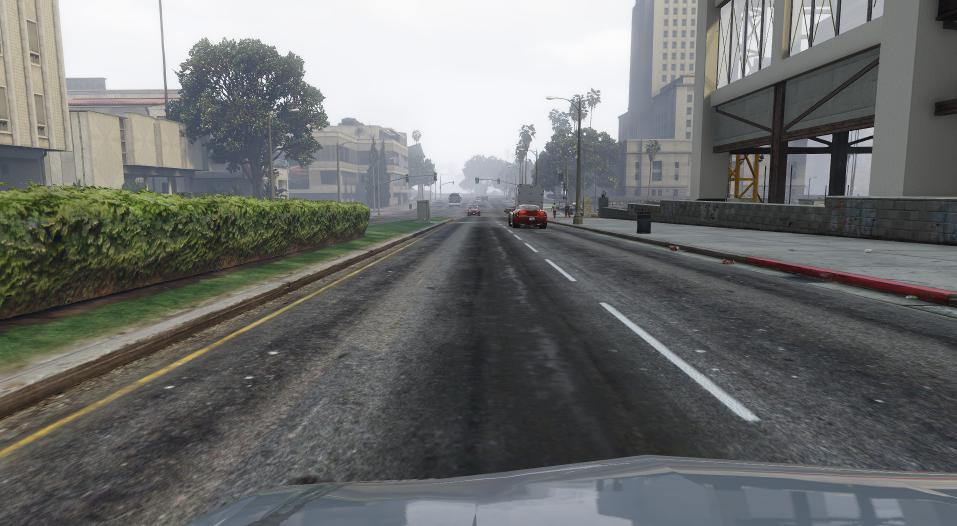}}\hfill
		\subfigure[Full]
		{\includegraphics[width=0.198\textwidth]{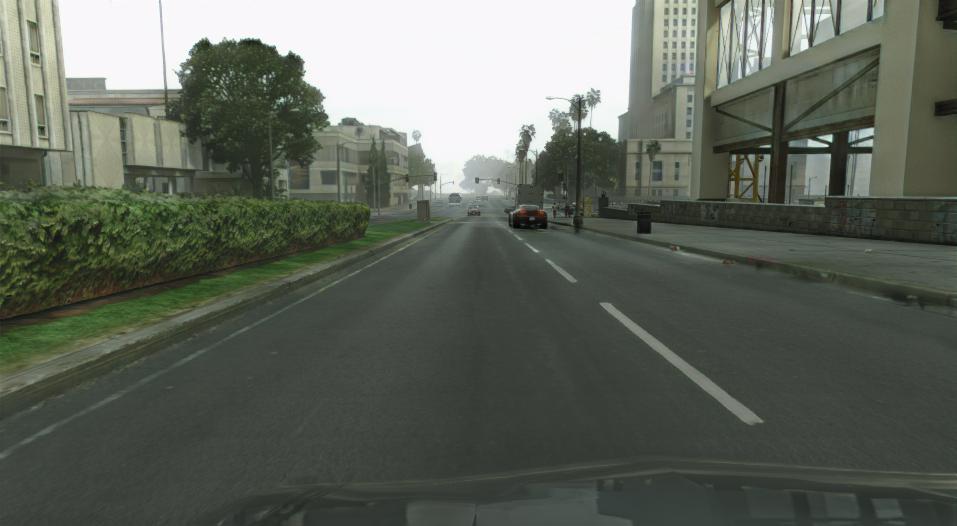}}\hfill
		\subfigure[w/o Dis. Mask]
		{\includegraphics[width=0.198\textwidth]{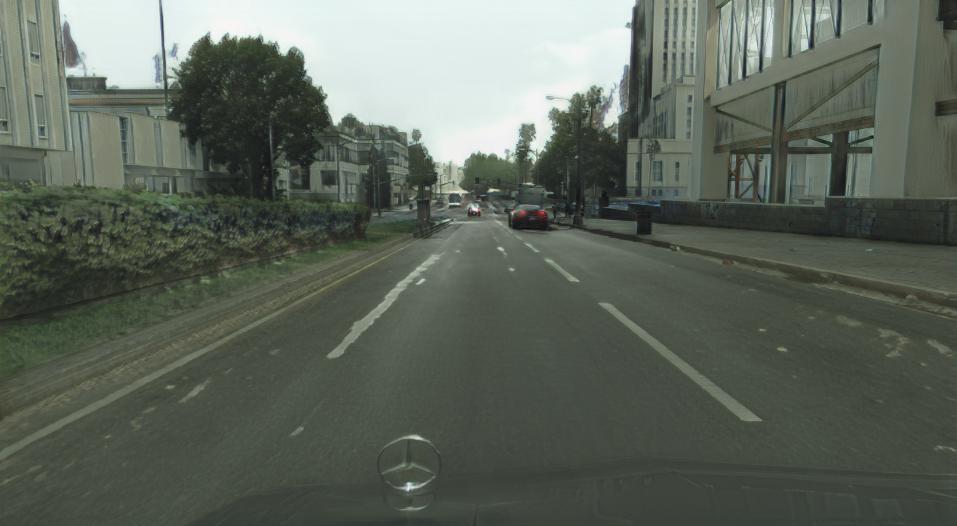}}\hfill
		\subfigure[w/o Local Dis.]
		{\includegraphics[width=0.198\textwidth]{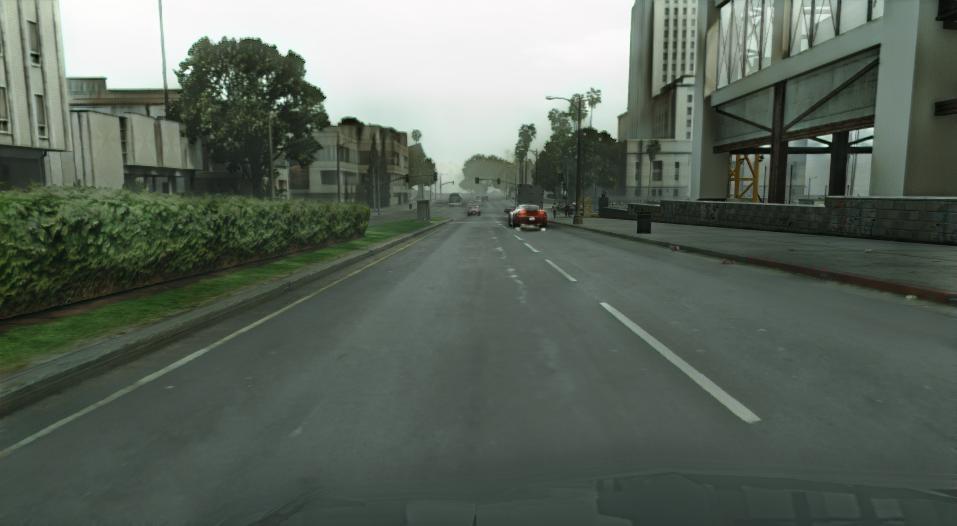}}\hfill
		\subfigure[w/ FADE w/o FATE]
		{\includegraphics[width=0.198\textwidth]{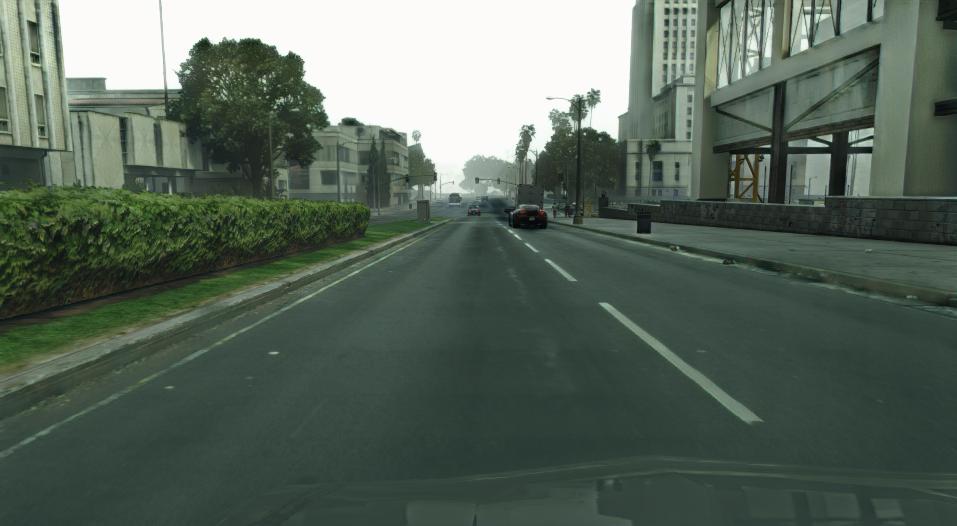}}\hfill 
	\end{center}
	\vspace{-8pt}
	\caption{\textbf{Qualitative ablations.} Results are randomly sampled from the best model.}
	\label{fig:qualitative_ablations_additional_random}
\end{figure*}
\begin{sidewaystable*}[h]\scriptsize
	\caption{\textbf{Extended quantitative comparison to prior work.} Models were trained using their official implementations. Results are reported as the average across five runs.}
	\setlength{\tabcolsep}{1.5pt}
	\begin{center}
		\begin{tabular}{lcccccccccccccc}
			\toprule
			\multirow{2}{*}{Method}&\multirow{2}{*}{FID}&\multirow{2}{*}{KID} &\multirow{2}{*}{sKVD}&\multicolumn{11}{c}{cKVD}\\
			\cmidrule(lr){5-15}
			&    &  &  & AVG& 		sky& 	ground&	road&	terrain&	vegetation&	building&	roadside-obj.&	person&	vehicle&	rest	\\
			\midrule
			\textbf{PFD$\rightarrow$Cityscapes} &   &  &  &  &  &  &   &  &  &  &    &  &  &    \\	
															
			Color Transfer& $91.01^{+0.05}_{-0.03}$ &	$94.82^{+0.14}_{-0.11}$ & $18.16^{+0.20}_{-0.18}$ & $50.87^{+1.18}_{-1.36}$ &$58.05^{+2.28}_{-1.24}$ & $16.66^{+0.19}_{-0.26}$ & $16.38^{+0.14}_{-0.09}$ & $26.91^{+1.23}_{-1.29}$ & $28.18^{+0.44}_{-0.26}$ & $32.60^{+0.33}_{-0.45}$ & $58.36^{+5.56}_{-7.85}$ & $125.37^{+11.20}_{-7.66}$ & $55.12^{+1.38}_{-1.86}$ & $91.11^{+0.39}_{-0.84}$ \\

			MUNIT & $40.36^{+1.22}_{-1.18}$ &$29.98^{+1.43}_{-1.14}$ & $14.99^{+0.06}_{-0.05}$ & $43.24^{+0.67}_{-0.71}$ &$37.92^{+0.85}_{-0.55}$ & $13.23^{+0.29}_{-0.23}$ & $14.33^{+0.16}_{-0.12}$ & $22.70^{+0.81}_{-0.75}$ & $24.97^{+0.37}_{-0.35}$ & $27.52^{+0.10}_{-0.15}$ & $58.24^{+3.37}_{-3.87}$ & $108.61^{+5.56}_{-4.30}$ & $45.35^{+0.32}_{-0.24}$ & $79.54^{+0.25}_{-0.49}$ \\

			CUT & $49.55^{+5.19}_{-3.63}$ &	$44.25^{+7.57}_{-4.99}$ & $16.85^{+1.28}_{-1.82}$ & $\mathbf{37.53}^{+1.09}_{-1.13}$ &$\mathbf{28.76}^{+0.98}_{-0.39}$ & $\mathbf{11.17}^{+1.26}_{-0.84}$ & $\mathbf{13.92}^{+0.58}_{-0.80}$ & $13.49^{+1.21}_{-1.54}$ & $24.20^{+1.23}_{-1.39}$ & $24.69^{+0.92}_{-0.86}$ & $57.45^{+4.33}_{-2.22}$ & $\mathbf{90.52}^{+3.33}_{-6.67}$ & $40.92^{+2.77}_{-3.08}$ & $70.20^{+3.10}_{-0.99}$ \\
				
			TSIT & $\mathbf{38.70}^{+1.59}_{-1.16}$ &	$\mathbf{28.70}^{+1.81}_{-1.27}$ & $\mathbf{10.80}^{+0.57}_{-0.29}$ & $42.35^{+0.73}_{-0.89}$ &$40.13^{+1.58}_{-1.28}$ & $13.74^{+0.39}_{-1.14}$ & $14.09^{+0.27}_{-0.78}$ & $23.48^{+0.84}_{-1.17}$ & $23.74^{+0.36}_{-0.47}$ & $25.76^{+0.16}_{-0.22}$ & $\mathbf{51.95}^{+2.31}_{-2.72}$ & $107.98^{+4.67}_{-5.61}$ & $43.49^{+1.32}_{-1.01}$ & $79.13^{+2.72}_{-2.05}$ \\
						
			QS-Attn& $49.42^{+5.71}_{-6.93}$ &	$42.87	^{+7.34}_{-10.03}$ & $14.01^{+0.34}_{-0.38}$ & $38.57^{+2.85}_{-1.68}$ &$29.50^{+2.40}_{-1.88}$ & $11.69^{+1.51}_{-0.69}$ & $\mathbf{13.92}^{+0.68}_{-0.76}$ & $\mathbf{13.22}^{+2.40}_{-1.23}$ & $23.99	^{+1.34}_{-1.72}$ & $\mathbf{23.36}^{+1.32}_{-1.56}$ & $57.88^{+2.89}_{-1.88}$ & $100.32^{+14.23}_{-6.18}$ & $\mathbf{40.77}^{+3.32}_{-2.01}$ & $71.06^{+5.17}_{-3.29}$ \\
			\midrule

			FeaMGan-S (ours)& $45.16^{+5.24}_{-3.23}$ &	$34.93^{+6.92}_{-3.48}$ & $13.87^{+0.66}_{-0.89}$ & $40.50^{+2.83}_{-1.96}$ &$40.57^{+5.56}_{-4.83}$ & $13.32^{+2.35}_{-1.48}$ & $16.00^{+1.51}_{-2.76}$ & $24.55^{+2.88}_{-3.76}$ & $\mathbf{20.82}^{+5.94}_{-4.03}$ & $27.54^{+1.07}_{-0.34}$ & $63.09^{+8.99}_{-4.97}$ & $102.53^{+1.58}_{-0.79}$ & $42.58^{+3.36}_{-4.07}$ & $\mathbf{53.99}^{+5.41}_{-5.45}$ \\
					
			FeaMGan (ours)& $46.12^{+4.60}_{-5.80}$ &	$36.56^{+6.70}_{-7.96}$ & $13.69^{+1.13}_{-1.15}$ & $ 41.19^{+2.89}_{-2.81}$ & $42.69^{+4.00}_{-5.01}$ & $14.97^{+3.86}_{-3.23}$ & $17.35^{+5.09}_{-4.28}$ & $26.51^{+5.70}_{-3.27}$ & $\mathbf{20.25}^{+2.88}_{-2.56}$ & $26.34^{+1.36}_{-0.77}$ & $64.64^{+4.94}_{-5.07}$ & $102.23^{+10.58}_{-10.91}$ & ${42.38}^{+2.91}_{-3.01}$ & $\mathbf{54.52}^{+5.00}_{-3.09}$ \\		
			\midrule	
			\midrule
			\textbf{Viper$\rightarrow$Cityscapes} &   &  &  &  &  &  &   &  &  &  &    &  &  &    \\	
		
			Color Transfer& $89.30^{+0.06}_{-0.05}$ &	$83.51^{+0.10}_{-0.10}$ & $20.20^{+0.32}_{-0.24}$ & $51.23^{+0.75}_{-0.75}$ &$65.74^{+1.33}_{-2.59}$ & $19.98^{+1.08}_{-0.44}$ & $16.87^{+0.16}_{-0.14}$ & $26.65^{+2.69}_{-2.39}$ & $28.79^{+0.45}_{-0.30}$ & $36.21^{+0.14}_{-0.14}$ & $41.97^{+1.52}_{-3.65}$ & $139.26^{+10.03}_{-10.10}$ & $57.10^{+1.04}_{-0.74}$ & $79.73^{+0.75}_{-0.82}$ \\
									
			MUNIT & $47.96^{+0.52}_{-1.11}$ &	$30.35^{+0.68}_{-1.29}$ & $14.14^{+0.10}_{-0.09}$ & $59.62^{+1,87}_{-2.34}$ &$46.44^{+2.83}_{-2.02}$ & $15.85^{+0.87}_{-0.71}$ & $14.11^{+0.31}_{-0.11}$ & $32.69^{+2.23}_{-3.94}$ & $25.75^{+0.27}_{-0.36}$ & $25.76^{+0.24}_{-0.24}$ & $\mathbf{39.99}^{+1.07}_{-1.26}$ & $274.68^{+15.46}_{-23.29}$ & $46.64^{+1.75}_{-1.70}$ & $74.33^{+0.40}_{-0.79}$ \\
			
			CUT & $60.35^{+6.50}_{-8.13}$ &	$49.48^{+7.19}_{-10.15}$ & $16.80^{+1.04}_{-1.11}$ & $51.02^{+3.71}_{-4.32}$ &$\mathbf{34.79}^{+6.87}_{-2.98}$ & $14.88^{+0.79}_{-0.76}$ & $16.80^{+2.50}_{-1.68}$ & $\mathbf{22.40}^{+2.41}_{-2.33}$ & $22.91^{+1.81}_{-0.84}$ & $\mathbf{23.34}^{+1.10}_{-1.04}$ & $45.00^{+5.23}_{-2.68}$ & $224.47^{+29.25}_{-28.29}$ & $42.29^{+2.56}_{-2.55}$ & $63.36^{+1.72}_{-3.17}$ \\

			TSIT & $\mathbf{45.26}^{+1.92}_{-1.39}$ &	$\mathbf{28.40}^{+2.55}_{-2.16}$ & $\mathbf{8.47}^{+0.25}_{-0.26}$ & $50.03^{+3.06}_{-2.12}$ &$46.25^{+0.69}_{-0.93}$ & $\mathbf{14.46}^{+2.11}_{-1.95}$ & $\mathbf{12.28}^{+0.98}_{-0.97}$ & $31.95^{+5.17}_{-4.96}$ & $24.86^{+1.50}_{-1.27}$ & $24.91^{+1.26}_{-1.43}$ & $45.19^{+2.35}_{-2.10}$ & $184.05^{+18.06}_{-10.62}$ & $\mathbf{44.59}^{+2.46}_{-1.55}$ & $71.72^{+3.90}_{-3.72}$ \\
			
			QS-Attn& $55.62^{+12.05	}_{-9.66}$ &	$39.31^{+11.87}_{-9.92}$ & $12.99^{+1.87}_{-1.60}$ & $63.22^{+17.47}_{-13.74}$ &$36.44^{+15.38}_{-4.97}$ & $16.04^{+1.56}_{-1.26}$ & $15.25^{+1.10}_{-2.49}$ & $25.20^{+4.27}_{-2.30}$ & $26.09^{+1.88}_{-2.02}$ & $24.24^{+1.26}_{-0.89}$ & $46.54^{+1.63}_{-1.84}$ & $326.60^{+171.61}_{-128.90}$ & $46.44^{+3.78}_{-5.51}$ & $69.33^{+5.26}_{-5.06}$ \\
			\midrule	

			FeaMGan-S (ours)& $52.79^{+2.50}_{-2.79}$ &  $35.92^{+3.88}_{-3.18}$ &	$14,34^{+0.65}_{-0.73}$ &  $\mathbf{45.38}^{+1.53}_{-1.63}$ &$56.75^{+5.13}_{-8.76}$ & $18.51^{+1.49}_{-1.08}$ & $16.68^{+1.90}_{-3.18}$ & $42.85^{+1.59}_{-1.97}$ & $\mathbf{22.70}^{+1.41}_{-1.40}$ & $26.82^{+0.74}_{-1.12}$ & $45.27^{+1.37}_{-1.11}$ & $\mathbf{130.76}^{+6.27}_{-11.95}$ & $45.25^{+1.49}_{-3.29}$ & $\mathbf{48.19}^{+2.45}_{-1.49}$ \\
	
			FeaMGan (ours)& $51.56^{+1.97}_{-3.56}$ &	$34.63^{+3.32}_{-5.48}$ & $14.01^{+0.58}_{-0.73}$ & $\mathbf{47.21}^{+1.29}_{-1.10}$ &$58.87^{+3.48}_{-1.62}$ & $21.20^{+0.72}_{-0.93}$ & $18.03^{+1.38}_{-0.62}$ & $50.01^{+7.63}_{-3.72}$ & $23.55^{+3.08}_{-2.42}$ & $26.67^{+0.46}_{-0.66}$ & $45.55^{+1.11}_{-1.79}$ & $\mathbf{132.77}^{+2.96}_{-3.46}$ & $45.13^{+1.72}_{-1.51}$ & $\mathbf{50.32}^{+1.55}_{-1.24}$ \\
			\midrule	
			\midrule
			\textbf{Day$\rightarrow$Night} &   &  &  &  &  &  &   &  &  &  &    &  &  &    \\	
	
			Color Transfer& $125.90^{+0.13}_{-0.10}$ &	$140.60^{+0.10}_{-0.10}$ & $32.58^{+0.32}_{-0.52}$ & $56.52^{+1.76}_{-1.26}$ &$47.62^{+0.50}_{-0.78}$ & $27.41^{+1.37}_{-1.27}$ & $15.89^{+0.46}_{-0.23}$ & $\mathbf{32.60}^{+1.76}_{-2.27}$ & $44.24^{+0.30}_{-0.25}$ & $32.61^{+0.68}_{-1.07}$ & $128.57^{+11.25}_{-13.18}$ & $108.52^{+8.17}_{-6.36}$ & $25.65^{+0.43}_{-0.37}$ & $102.06^{+0.39}_{-0.31}$ \\
			
			MUNIT & $42.53^{+1.65}_{-1.27}$ &	$31.83^{+1.73}_{-0.98}$ & $15.02^{+0.64}_{-0.65}$ & $50.83^{+1.25}_{-0.88}$ &$\mathbf{29.25}^{+0.23}_{-0.29}$ & $28.00^{+0.72}_{-0.50}$ & $13.49^{+0.30}_{-0.16}$ & $36.57^{+1.16}_{-0.53}$ & $44.86^{+0.31}_{-0.35}$ & $\mathbf{24.96}^{+0.69}_{-0.59}$ & $115.00^{+6.94}_{-5.17}$ & $101.70^{+4.53}_{-4.06}$ & $19.66^{+0.34}_{-0.33}$ & $94.82^{+1.31}_{-1.37}$ \\
			
			CUT & $\mathbf{34.36}^{+3.71}_{-6.12}$ &	$\mathbf{20.54}^{+4.81}_{-7.05}$ & $10.16^{+1.98}_{-1.14}$ & $53.55^{+3.05}_{-3.22}$ &$31.89^{+0.92}_{-2.06}$ & $27.44^{+1.58}_{-1.46}$ & $\mathbf{13.14}^{+0.64}_{-0.73}$ & $40.93^{+6.70}_{-8.59}$ & $49.79^{+3.41}_{-2.78}$ & $25.52^{+1.69}_{-1.71}$ & $104.26^{+7.21}_{-10.28}$ & $122.50^{+11.90}_{-10.06}$ & $27.30^{+2.60}_{-2.87}$ & $92.76^{+0.52}_{-1.51}$ \\
		
			TSIT & $54.979^{6.83}_{-7.99}$ &	$33.21^{+5.26}_{-6.21}$ & $12.71^{+5.77}_{-3.49}$ & $57.91^{+2.92}_{-2.28}$ &$36.27^{+2.32}_{-1.39}$ & $31.56^{+1.18}_{-2.21}$ & $16.93^{+2.20}_{-1.07}$ & $45.23^{+9.74}_{-4.86}$ & $54.82^{+4.55}_{-4.89}$ & $29.09^{+3.19}_{-1.65}$ & $143.47^{+14.01}_{-10.47}$ & $99.30^{+3.07}_{-5.53}$ & $27.43^{+2.98}_{-4.22}$ & $94.98^{+2.46}_{-2.60}$ \\
			
			QS-Attn& $46.68^{+2.73}_{-2.03}$ &	$21.47^{+3.94}_{-2.55}$ & $\mathbf{7.58}^{+1.27}_{-1.77}$ & $52.02^{+4.14}_{-3.29}$ &$31.62^{+1.73}_{-1.66}$ & $\mathbf{26.73}^{+2.64}_{-3.41}$ & $13.26^{+0.99}_{-0.92}$ & $38.25^{+5.42}_{-3.84}$ & $47.26^{+3.31}_{-3.13}$ & $25.42^{+2.05}_{-1.80}$ & $\mathbf{100.84}^{+11.90}_{-7.85}$ & $123.79^{+18.23}_{-14.72}$ & $26.67^{+4.28}_{-3.72}$ & $86.39^{+5.62}_{-4.37}$ \\
			\midrule

			FeaMGan-S (ours)& $70.40^{+15.29}_{-4.76}$ &	$51.30^{+21.06}_{-6.09}$ & $14.68^{+3.45}_{-1.84}$ & $\mathbf{46.66}^{+2.63}_{-2.20}$ &$30.35^{+1.05}_{-0.84}$ & $35.47^{+3.71}_{-3.90}$ & $17.26^{+1.46}_{-1.08}$ & $47.29^{+10.32}_{-8.77}$ & $\mathbf{27.12}^{+1.14}_{-1.10}$ & $25.22^{+1.36}_{-1.38}$ & $116.25^{+4.47}_{-4.70}$ & $\mathbf{70.75}^{+4.87}_{-5.91}$ & $\mathbf{19.29}^{+0.52}_{-0.87}$ & $\mathbf{77.60}^{+4.37}_{-2.09}$ \\

			FeaMGan (ours)& $66.39^{+6.39}_{-8.43}$ & $46.96^{+10.07}_{-10.41}$ & $13.14^{+3.34}_{-2.01}$ & $\mathbf{46.88}^{+2.52}_{-2.83}$ & $29.72^{+2.56}_{-1.42}$ & $35.94^{+1.75}_{-1.77}$ &	$17.48^{+0.98}_{-1.44}$ & $49.78^{+7.45}_{-7.78}$ & $28.78^{+1.94}_{-3.03}$ &$25.65^{+0.71}_{-0.61}$ &  $115.66^{+2.72}_{-5.03}$ & $\mathbf{70.94}^{+9.44}_{-11.48}$ & $\mathbf{19.23}^{+0.64}_{-0.78}$ & $\mathbf{75.57}^{+4.86}_{-3.73}$ \\
			\midrule	
			\midrule
			\textbf{Clear$\rightarrow$Snowy} &   &  &  &  &  &  &   &  &  &  &    &  &  &    \\
							
			Color Transfer& $46.85^{+0.12}_{-0.38}$ &	$19.44^{+0.43}_{-1.43}$ & $14.91^{+0.86}_{-2.97}$ & $42.89^{+13.81}_{-3.96}$ &$25.78^{+1.19}_{-2.03}$ & $22.99^{+3.22}_{-2.15}$ & $16.01^{+0.37}_{-0.35}$ & $21.54^{+1.72}_{-1.52}$ & $41.13^{+7.40}_{-2.60}$ & $24.20^{+1.99}_{-0.78}$ & $\mathbf{57.67}^{+18.52}_{-11.79}$ & $128.26^{+94.11}_{-29.48}$ & $25.95^{+7.36}_{-2.36}$ & $65.39^{+5.38}_{-1.91}$ \\
			
			MUNIT & $\mathbf{44.74}^{+1.23}_{-0.79}$ &	$17.48^{+0.59}_{-0.86}$ & $11.65^{+0.34}_{-0.22}$ & $48.10^{+0.49}_{-0.73}$ &$28.47^{+0.78}_{-0.73}$ & $25.64^{+0.30}_{-0.46}$ & $15.27^{+0.21}_{-0.13}$ & $26.21^{+0.60}_{-0.48}$ & $40.31^{+0.58}_{-0.37}$ & $24.26^{+0.13}_{-0.09}$ & $101.98^{+1.73}_{-4.77}$ & $116.14^{+3.89}_{-4.34}$ & $21.63^{+0.41}_{-0.18}$ & $81.08^{+0.31}_{-0.63}$ \\
				
			CUT & $46.03^{+1.08}_{-0.85}$ &	$15.70^{+0.77}_{-0.94}$ & $14.71^{+1.15}_{-0.94}$ & $43.91^{+0.97}_{-0.78}$ &$26.74^{+0.50}_{-0.63}$ & $\mathbf{21.96}^{+0.75}_{-0.96}$ & $\mathbf{13.15}^{+0.31}_{-0.50}$ & $21.49^{+1.02}_{-1.08}$ & $35.20^{+0.47}_{-0.28}$ & $25.31^{+0.50}_{-0.39}$ & $76.67^{+4.18}_{-6.21}$ & $119.13^{+16.19}_{-10.00}$ & $23.91^{+0.59}_{-1.18}$ & $75.51^{+1.17}_{-0.60}$ \\
		
			TSIT & $79.29^{+5.08}_{-6.69}$ &	$40.02^{+7.17}_{-7.14}$ & $12.97^{+0.37}_{-0.51}$ & $41.52^{+3.31}_{-2.59}$ &$28.02^{2.81}_{-1.34}$ & $22.72^{+2.06}_{-1.40}$ & $14.32^{+0.64}_{-0.48}$ & $\mathbf{18.92}^{+2.42}_{-2.13}$ & $34.54^{+2.08}_{-2.52}$ & $23.02^{+0.69}_{-0.66}$ & $72.13^{+7.37}_{-4.90}$ & $104.05^{+12.49}_{-12.64}$ & $21.64^{+2.06}_{-1.56}$ & $75.84^{+4.73}_{-5.53}$ \\
				
			QS-Attn& $60.91^{+0.79}_{-1.02}$ &	$18.85^{+1.05}_{-1.36}$ & $14.19^{+1.70}_{-1.01}$ & $44.00^{+1.95}_{-1.80}$ &$25.60^{+0.50}_{-0.75}$ & $22.04^{+1.20}_{-1.41}$ & $13.24^{+0.17}_{-0.16}$ & $22.71^{+0.46}_{-0.71}$ & $36.02^{+2.20}_{-1.40}$ & $26.45^{+1.42}_{-1.75}$ & $78.58^{+4.70}_{-4.53}$ & $114.07^{+13.76}_{-8.81}$ & $25.17^{+3.07}_{-1.42}$ & $76.08^{+1.84}_{-2.29}$ \\
			\midrule
			
			FeaMGan-S (ours)& $57.93^{+1.37}_{-1.68}$ &	$16.24^{+1.19}_{-0.59}$ & $11.88^{+0.55}_{-0.52}$ & $\mathbf{38.28}^{+2.86}_{-2.83}$ &$\mathbf{22.69}^{+0.67}_{-1.34}$ & $25.71^{+1.14}_{-1.76}$ & $15.82^{+0.82}_{-1.27}$ & $37.47^{+3.98}_{-4.13}$ & $\mathbf{25.94}^{+1.50}_{-2.69}$ & $\mathbf{21.80}^{+0.53}_{-0.61}$ & $75.32^{+4.53}_{-6.80}$ & $\mathbf{81.47}^{+22.52}_{-14.46}$ & $\mathbf{19.10}^{+0.55}_{-0.30}$ & $\mathbf{57.46}^{+2.18}_{-3.98}$ \\

			FeaMGan (ours)& $56.78^{+0.81}_{-0.32}$ &	$\mathbf{14.77}^{+0.98}_{-1.59}$ & $\mathbf{11.36}^{+0.21}_{-0.43}$ & $41.72^{+2.61}_{-1.76}$ &$\mathbf{22.71}^{+1.17}_{-1.09}$ & $26.64^{+3.01}_{-2.07}$ & $16.19^{+1.74}_{-1.13}$ & $38.00^{+3.46}_{-2.77}$ & $\mathbf{27.78}^{+2.95}_{-2.21}$ & $\mathbf{21.41}^{+0.61}_{-0.69}$ & $79.35^{+3.72}_{-1.97}$ & $105.08^{+35.80}_{-25.18}$ & $\mathbf{19.50}^{+0.67}_{-1.04}$ & $\mathbf{60.59}^{+1.99}_{-1.90}$ \\
			\bottomrule
		\end{tabular}
	\end{center}
	\label{tab:quantitative_comparison_extended}
\end{sidewaystable*}

\begin{sidewaystable*}[h]\scriptsize
	\caption{\textbf{Extended quantitative evaluation for ablation study.} Results are reported as the average across five runs.}
	\setlength{\tabcolsep}{1.5pt}
	\begin{center}
		\begin{tabular}{lcccccccccccccc}
			\toprule
			\multirow{2}{*}{Method}&\multirow{2}{*}{FID}&\multirow{2}{*}{KID} &\multirow{2}{*}{sKVD}&\multicolumn{11}{c}{cKVD}\\
			\cmidrule(lr){5-15}
			&    &  &  & AVG& 		sky& 	ground&	road&	terrain&	vegetation&	building&	roadside-obj.&	person&	vehicle&	rest	\\
			\midrule
			FeaMGan (Full)& $46.11^{+4.60}_{-5.80}$ &	$36.66^{+6.70}_{-7.96}$ & $13.69^{+1.13}_{-1.15}$ & $ 41.19^{+2.89}_{-2.81}$ & $42.69^{+4.00}_{-5.01}$ & $14.97^{+3.86}_{-3.23}$ & $17.35^{+5.09}_{-4.28}$ & $26.51^{+5.70}_{-3.27}$ & $\mathbf{20.25}^{+2.88}_{-2.56}$ & $26.34^{+1.36}_{-0.77}$ & $64.64^{+4.94}_{-5.07}$ & $102.23^{+10.58}_{-10.91}$ & $\mathbf{42.38}^{+2.91}_{-3.01}$ & $54.52^{+5.00}_{-3.09}$ \\
						
			w/o Dis. Mask & $ \mathbf{37.10}^{+3,03}_{-5,46}$ & $ \mathbf{25.88}^{+3.60}_{-8.65}$ & $ 14.73^{+1.27}_{-1.66}$ & $\mathbf{39.65}^{+4.73}_{-4.16}$ & $\mathbf{26.70}^{+4.09}_{-5.99}$ & $15.81^{+5.28}_{-3.67}$ & $16.65^{+4.43}_{-2.85}$ & $31.02^{+11.19}_{-8.23}$ & $22.97^{+4.15}_{-6.89}$ & $\mathbf{25.39}^{+2.75}_{-1.24}$ & $67.01^{+6.63}_{-7.99}$ & $\mathbf{93.78}^{+10.71}_{-7.90}$ & $44.23^{+2.88}_{-4.95}$ & $\mathbf{52.91}^{+5.34}_{-5.61}$\\
			
			w/ FADE w/o FATE & $45.46^{+2.68}_{-2.65}$ & $ 35.73^{+4.44}_{-3.53}$ & $ \mathbf{13.17}^{+0.54}_{-0.60}$ & $40.90^{+4.40}_{-2.04}$ & $41.49^{+8.66}_{-8.86}$ & $13.78^{+3.22}_{-1.70}$ & $16.78^{+4.99}_{-2.30}$ & $25.30^{+2.41}_{-2.12}$ & $20.58^{+1.62}_{-1.88}$ & $27.21^{+1.93}_{-0.68}$ & $\mathbf{63.12}^{+3.77}_{-5.25}$ & $104.43^{+13.22}_{-7.73}$ & $42.44^{+1.27}_{-0.91}$ & $53.83^{+5.46}_{-1.96}$\\
			
			w/ Random Crop & $47.88^{+5.10}_{-3.82}$ & $38.48^{+7.27}_{-4.14}$ & $	13.37^{+0.61}_{-1.14}$ & $ 40.18^{+1.55}_{-1.66}$ & $39.88^{+3.92}_{-7.65}$ & $\mathbf{12.90}^{+0.94}_{-0.68}$ & $\mathbf{14.65}^{+1.62}_{-2.00}$ & $\mathbf{25.09}^{+2.01}_{-2.93}$ & $21.89^{+5.21}_{-3.36}$ & $27.32^{+2.05}_{-1.87}$ & $64.32^{+2.86}_{-1.79}$ & $98.81^{+7.63}_{-5.61}$ & $43.08^{+3.81}_{-2.90}$ & $53.86^{+1.66}_{-1.56}$ \\
						
			w/ VGG Crop& $51.23^{+5.12}_{-2.11}$ & $	42.46^{+6.65}_{-3.42}$ & $	13.56^{+0.78}_{-0.81}$ & $ 40.62^{+2.22}_{-1.99}$ & $40.32^{+4.36}_{-5.72}$ & $13.3^{+2.06}_{-1.83}$ & $15.67^{+2.15}_{-2.14}$ & $26.47^{+4.07}_{-1.25}$ & $21.09^{+1.61}_{-2.41}$ & $27.28^{+0.98}_{-1.46}$ & $65.23^{+8.64}_{-3.75}$ & $99.61^{+9.27}_{-7.66}$ & $43.19^{+2.16}_{-3.12}$ & $53.94^{+3.81}_{-5.04}$ \\
			\midrule			
			w/o Local Dis. &   &  &  &  &  &  &   &  &  &  &    &  &  &    \\			
			- w/ 256$\times$256 Crop& $48.57^{+5.16}_{-2.30}$ & $ 38.89^{+6.82}_{-3.47}$ & $ \mathbf{12.89}^{+1.32}_{-0.90}$ & $ 41.26^{+3.00}_{-2.64}$ & $	42.31^{+2.92}_{-3.88}$ & $	13.57^{+1.65}_{-2.15}$ & $ 15.98^{+1.44}_{-1.58}$ & $	\mathbf{25.28}^{+4.26}_{-2.75}$ & $	22.18^{+8.57}_{-4.62}$ & $ \mathbf{26.56}^{+2.06}_{-1.46}$ & $\mathbf{61.13}^{+2.92}_{-2.21}$ & $	107.48^{+6.40}_{-11.62}$ & $ 42.44^{+4.20}_{-2.30}$ & $	55.62^{+5.79}_{-4.61}$\\
			
			- w/ 352$\times$352 Crop& $47.26^{+3.44}_{-2.31}$ & $ 37.75^{+5.30}_{-3.23}$ & $ 14.38^{+0.76}_{-1.00}$ & $ 39.30^{+2.59}_{-1.40}$ & $34.44^{+7.70}_{-5.20}$ & $\mathbf{13.09}^{+1.62}_{-1.17}$ & $15.84^{+1.79}_{-1.04}$ & $ 25.83^{+1.56}_{-2.01}$ & $21.50^{+3.84}_{-1.63}$ & $27.20^{+1.47}_{-1.24}$ & $61.24^{+3.62}_{-5.82}$ & $98.25^{+4.82}_{-3.34}$ & $42.24^{+2.34}_{-1.79}$ & $53.38^{+3.71}_{-2.18}$\\
						
			- w/ 464$\times$464 Crop& $\mathbf{46.61}^{+2.57}_{-3.25}$ & $ \mathbf{37.25}^{+3.17}_{-4.48}$ & $ 15.04^{+0.79}_{-0.45}$ & $\mathbf{38.62}^{+1.52}_{-1.68}$ & $\mathbf{31.60}^{+4.89}_{-3.80}$ & $13.13^{+1.08}_{-0.99}$ & $\mathbf{15.38}^{+1.77}_{-1.93}$ & $27.06^{+1.37}_{-2.76}$ & $22.23^{+3.99}_{-3.40}$ & $29.67^{+0.67}_{-1.10}$ & $63.38^{+2.10}_{-3.73}$ & $\mathbf{87.51}^{+3.83}_{-4.99}$ & $44.41^{+2.76}_{-2.76}$ & $51.77^{+2.19}_{-3.06}$\\
			
			- w/ 512$\times$512 Crop& $55.89^{+19.35}_{-12.54}$ & $ 49.12^{+26.93}_{-16.19}$ & $ 15.94^{+4.24}_{-2.24}$ & $ 39.35^{+4.47}_{-3.91}$ & $36.48^{+6.33}_{-12.08}$ & $14.68^{+2.82}_{-2.30}$ & $ 16.06^{+3.25}_{-3.12}$ & $26.87^{+8.35}_{-4.31}$ & $\mathbf{19.61}^{+4.76}_{-4.21}$ & $27.37^{+3.00}_{-2.34}$ & $62.40^{+3.93}_{-4.27}$ & $98.90^{+8.87}_{-5.05}$ & $\mathbf{40.32}^{+3.12}_{-4.21}$ & $\mathbf{50.86}^{+5.50}_{-7.19}$\\	
			\bottomrule
		\end{tabular}
	\end{center}
	\label{tab:quantitative_ablation_extended}
\end{sidewaystable*}
}


\end{document}